\documentclass{phdthesis}
\usepackage[english]{babel}
\usepackage[utf8]{inputenc}
\usepackage{graphicx}
\usepackage{amsmath,amssymb}
\usepackage{color}
\usepackage{graphicx}
\usepackage{subfigure}
\usepackage{booktabs}
\usepackage{graphicx,subfigure}
\usepackage{float}

\newcommand{\specialcell}[2][c]{%
  \begin{tabular}[#1]{@{}c@{}}#2\end{tabular}}
\usepackage{algorithm,algpseudocode}
\newcommand{\etal}{\textit{et al.}}
\usepackage{amssymb}
\usepackage{multirow}
\usepackage{colorspace}

\usepackage{algorithm}
\usepackage{tikz}
\usepackage{comment}
\usepackage{color}
\usepackage[accsupp]{axessibility}
\usepackage{bbm}
\usepackage{nicefrac}
\usepackage{bm}
\newcommand{\tablestyle}[2]{\setlength{\tabcolsep}{#1}\renewcommand{\arraystretch}{#2}\centering\footnotesize}
\usepackage{xcolor}
\usepackage{booktabs}
\usepackage[normalem]{ulem}
\usepackage{multirow}
\usepackage{array}
\usepackage{tabularx}
\usepackage{xspace}
\usepackage{tikz}
\usetikzlibrary{arrows,shapes,calc,matrix,fit,backgrounds}
\usepackage{pgfplots}
\usepackage{pgfplotstable}
\pgfplotsset{compat=1.9}
\usepgfplotslibrary{fillbetween}

\usepackage{xstring}

\usepgfplotslibrary{external}
\IfBeginWith*{\jobname}{fig/extern/}{\finalcopy}{}

\tikzset{every mark/.append style={solid}}
\pgfplotsset{%smooth,
	grid=both, width=\columnwidth, try min ticks=5,
	every axis/.append style={font=\scriptsize},
	every axis plot/.append style={thick,mark=none,mark size=1.2,tension=0.18},
	legend cell align=left, legend style={fill opacity=0.8},
}

\pgfplotsset{
	dash/.style={mark=o,dashed,opacity=0.7},
	dott/.style={mark=o,dotted,opacity=0.7},
}

\definespotcolor{mygreen}{PANTONE 7716 C}{.83, 0, .40, .11}
\definecolor{amber}{rgb}{1.0, 0.75, 0.0}
\makeatother

\usepackage{makeidx}
\usepackage[toc]{glossaries}
\usepackage{xspace}

\makeglossaries
\DeclareMathOperator{\similarity}{sim}

\bibliography{egbib}

\title{Neural Network Training and Non-Differentiable Objective Functions}
\author{Yash Patel}

\supervisor{Prof. Ing. Jiří Matas, Ph.D.}
\supervisorAffiliation{
	Czech Technical University in Prague\\
	Faculty of Electrical Engineering\\
	Department of Cybernetics\\
        Karlovo náměstí 13\\
	121 35 Prague 2\\
	Czech Republic}

\placeyear{Prague, May 2023}
\date{May 2023}

\begin{document}

%% Uvodni stranky
\maketitle
\chapter*{Abstract}

Many important computer vision tasks are naturally formulated to have a non-differentiable objective. Therefore, the standard, dominant training procedure of a neural network is not applicable since back-propagation requires the gradients of the objective with respect to the output of the model. Most deep learning methods side-step the problem sub-optimally by using a proxy loss for training, which was originally designed for another task and is not tailored to the specifics of the objective. The proxy loss functions may or may not align well with the original non-differentiable objective. An appropriate proxy has to be designed for a novel task, which may not be feasible for a non-specialist. This thesis makes four main contributions toward bridging the gap between the non-differentiable objective and the training loss function. Throughout the thesis, we refer to a loss function as a surrogate loss if it is a differentiable approximation of the non-differentiable objective. Note that we use the terms objective and evaluation metric interchangeably. 

First, we propose an approach for learning a differentiable surrogate of a decomposable and non-differentiable evaluation metric. The surrogate is learned jointly with the task-specific model in an alternating manner. The approach is validated on two practical tasks of scene text recognition and detection, where the surrogate learns an approximation of edit distance and intersection-over-union, respectively. In a post-tuning setup, where a model trained with the proxy loss is trained further with the learned surrogate on the same data, the proposed method shows a relative improvement of up to $39$\% on the total edit distance for scene text recognition and $4.25$\% on $F_{1}$ score for scene text detection.

Second, an improved version of training with the learned surrogate where the training samples that are hard for the surrogate are filtered out. This approach is validated for scene text recognition. It outperforms our previous approach and attains an average improvement of $11.2\%$ on total edit distance and an error reduction of $9.5\%$ on accuracy on several popular benchmarks. Note that the two proposed methods for learning a surrogate and training with the surrogate do not make any assumptions about the task at hand and can be potentially extended to novel tasks. 

Third, for recall@k, a non-decomposable and non-differentiable evaluation metric, we propose a hand-crafted surrogate that involves designing differentiable versions of sorting and counting operations. An efficient mixup technique for metric learning is also proposed that mixes the similarity scores instead of the embedding vectors. The proposed surrogate attains state-of-the-art results on several metric learning and instance-level search benchmarks when combined with training on large batches. Further, when combined with the kNN classifier, it also serves as an effective tool for fine-grained recognition, where it outperforms direct classification methods.

Fourth, we propose a loss function termed Extended SupCon that jointly trains the classifier and backbone parameters for supervised contrastive classification. The proposed approach benefits from the robustness of contrastive learning and maintains the probabilistic interpretation like a soft-max prediction. Empirical results show the efficacy of our approach under challenging settings such as class imbalance, label corruption, and training with little labeled data.

Overall the contributions of this thesis make the training of neural networks more scalable -- to new tasks in a nearly labor-free manner when the evaluation metric is decomposable, which will help researchers with novel tasks. For non-decomposable evaluation metrics, the differentiable components developed for the recall@k surrogate, such as sorting and counting, can also be used for creating new surrogates.

Automatic translations of the abstract to the Czech language by Google Translate and ChatGPT are included in the appendix.

\newpage
\chapter*{Acknowledgements}
I extend gratitude to my advisor, Prof. Jiří Matas for attracting my interest towards doing a Ph.D. I would not be able to pursue my Ph.D. degree without his support, ideas and motivation. I would like to thank the institute, Czech Technical University in Prague, for providing a nourishing research platform. 

Many thanks to my parents, Mr. Komal Singh Patel and Mrs. Rajni Patel, and my partner Kristína Cinová for their unconditional love and support, without which my research journey would not have been possible. Thanks also to my sisters Dr. Shivani Bhatnager and Mrs. Priyanka Singh along with my niece Kyra Bhatnager and nephew Amartya Singh for making me explain machine learning to them, it always helps to put my work into different perspectives. Last but not the least, I would like to thank my friends and colleagues at the Visual Recognition Group for feedback, support and engaging research discussions. 

This research was supported by Research Center for Informatics (project CZ.02 .1.01/ 0.0/0.0/16\_019/0000765 funded by OP VVV), by the Grant Agency of the Czech Technical University in Prague, grant No. SGS23/ 173/ OHK3/ 3T/ 13, by Project StratDL in the realm of COMET K1 center Software Competence Center Hagenberg, and Amazon Research Award.

\tableofcontents
\mainmatter

\chapter{Introduction}

Training deep networks by gradient descent on the user-defined objective is not possible when the objective is non-differentiable. Deep learning methods use a proxy loss function as a workaround. A proxy loss is a differentiable function previously designed and used for another task but is not tailored to the specifics of the user-defined objective. The use of proxy loss can empirically lead to a reasonable performance, but it may not align well with the user-defined objective, leading to sub-optimal performance. Often, the goal is to perform well on standard benchmarks where the performance is measured using an evaluation metric. In this thesis, we assume that the evaluation metric captures the user-defined objective, and thus the terms evaluation metric and objective function are interchangeably used. 

Examples of such a mismatch between the loss and the objective exist in object detection~\cite{yjw+16}, where the evaluation metric is intersection-over-union. Still, many popular approaches use $L_{n}$-norm as a proxy loss~\cite{fasterrcnn,redmon2016you}. Another example is scene text recognition~\cite{phm+20,pm+21} where the evaluation metric is edit distance but per-character cross-entropy and CTC~\cite{graves2006connectionist} are commonly used as a proxy. Similarly, in image retrieval~\cite{bxk+20,pgr+19} where the benchmarks use recall@k or average precision, many variants of a proxy triplet~\cite{skp+15} and margin~\cite{wms+17} losses have been studied. The thesis deals with training neural networks using learned or hand-crafted differentiable approximations of the test-time objective function, explicitly focusing on the cases when the evaluation metric is non-differentiable. 

\section{Background}
Supervised deep learning requires four main components: an annotated dataset, a model, a loss function, and an optimizer. The data is collected in a task-specific manner and contains the annotations required to train for a task, {\em e.g.}, semantic class labels for image classification~\cite{deng2009imagenet}, bounding boxes with semantic labels for object detection~\cite{linCOCO}, per-pixel semantic labels for segmentation~\cite{zhou2019semantic}, etc. The model typically consists of two components: a backbone and a prediction module. The backbone of the model is designed to transform the input data to a discriminative representation, {\em e.g.}, ViT~\cite{dbk+21} for images, BERT ~\cite{devlin2018bert} for text, ALBEF~\cite{li2021align} for joint vision-language modeling, etc. The current trend is to use a backbone pre-trained on large scale datasets either in a supervised or self-supervised manner~\cite{radford2021learning,singh2022revisiting}. The prediction module on top of these backbones transforms the representation to the required task-specific prediction. The loss function compares the model's prediction with the ground truth. Thus the design of the loss function needs to take into account the task-specific predictions and ground truth, {\em e.g.}, cross-entropy loss for image classification~\cite{he2016deep}, smooth-$L_{1}$ to regress the bounding box coordinates for object detection~\cite{fasterrcnn}, per-pixel cross-entropy loss for semantic segmentation~\cite{jadon2020survey}, etc. The exact choice of the optimizer is often a hyper-parameter which is empirically determined along with the learning rate, learning rate schedule, and weight decay. Popular choices are Adam~\cite{kingma2014adam}, AdamW~\cite{lh+19}, RMSProp~\cite{ruder2016overview}, Adagrad~\cite{duchi2011adaptive}, Adadelta~\cite{zeiler2012adadelta}, SGD, etc. 

Once these components are chosen, the model is trained to minimize the expectation of a loss on the training data. During feed-forward, the data is fed to the model to obtain the predictions, which are then compared against the ground-truth annotations by the loss function. During back-propagating, the chain rule is employed to obtain the gradients of the loss with respect to the model weights. The optimizer then updates these weights based on the obtained gradients and the learning rate. For the chain rule in backpropagation to work, every module in the model and the loss function are required to be differentiable. The model is usually trained until the loss on the validation set keeps decreasing, {\em i.e.}, until the model starts over-fitting on the training data. 

Once the model is trained, it is evaluated on unseen test data, and the comparison between the predictions and the ground truth is now made using a test-time evaluation metric. The evaluation metric is designed to fulfill task-specific requirements and does not depend on the training process. 

With supervised deep learning, the performance on a wide range of computer vision tasks has been pushed to the levels of real-world practical use. The progress has been systematically made by improving each component of supervised deep learning, such as deeper and more powerful model architectures \cite{krizhevsky2012imagenet,he2016deep,simonyan2014very,dbk+21} and introduction of large-scale training datasets~\cite{deng2009imagenet,lin2014microsoft}. At first, the progress was expensive as designing architectures demanded detailed domain expertise, and creating new datasets is costly. Therefore, to decrease the human effort, there has been a substantial effort in automating the process of designing better task-specific architectures \cite{elsken2018neural,ryoo2019assemblenet,zoph2016neural} and employing self-supervised methods of learning to reduce the dependence on human-annotated data \cite{gidaris2018unsupervised,caron2018deep,gomez2017self,hfw+20,he2022masked}.

There are numerous reasons why a trained model may perform sub-optimally on a test set. Name a few; first, the trained model can over-fit on the training data and perform poorly on the test data sampled identically and independently from the same distribution. To remedy this issue, several regularization techniques have been investigated~\cite{srivastava2014dropout,ioffe2015batch,ba2016layer}. Over-fitting and under-fitting often happen due to improper choice of the hyper-parameters such as learning rate, schedule, and weight decay. Several approaches and recommendations have been proposed to mitigate these~\cite{bengio2012practical,smith2017learning,law2020grid,klein2019hyperparameter}. Second, the test set could be from a different distribution leading to a domain gap. There are several papers in the literature on domain adaption to make the models robust to the distribution shift ~\cite{tang2019multi,li2020d,huang2020improving,liu2021rethinking,liu2021self,zheng2015cross,long2015learning,hussain2016deep,ganin2016domain,ganin2015unsupervised}. Further, the use of backbone models pre-trained on very large scale datasets can lead to good generic representations, which can implicitly bridge the domain gap~\cite{radenovic2023filtering,radford2021learning,singh2022revisiting,mahajan2018exploring}. Third, the evaluation metric may not be known at the time of training leading to a wrong choice of the loss function. 

Another possibility is that the evaluation metric is known but is non-differentiable and thus can not be used during the training. Relatively little attention has been paid to the case where the test-time evaluation metric cannot be directly used as a loss function.

\section{Limitation of Proxy Loss Functions}
If the evaluation metric is known at the time of training and is differentiable, it can be directly used as a loss function for training. However, the evaluation metrics are known for many practical problems in computer vision but are non-differentiable. An example of a non-differentiable metric is edit distance, also known as Levenshtein distance, which is computed by counting unit operations of addition, deletion, and substitution necessary to transform one text string into another. Edit distance is a common choice for evaluating scene text recognition methods. This metric is non-differentiable as it has a discrete range and is often implemented using dynamic programming, which makes it infeasible to obtain the gradients. Another example is intersection-over-union between the two bounding boxes for object detection, which can be easily implemented in a differentiable manner for axis-aligned bounding boxes, but the implementation is inconvenient for rotated bounding boxes~\cite{8886046}. Another example is in lossy image compression, where the end user is human. Thus the optimal evaluation metric is a human's perception of similarity, which is complicated and unknown to be adequately expressed as a mathematical function. Another example includes recall at top-k, a popular metric for evaluating retrieval approaches on open-set datasets. Recall@k is the ratio of the number of positive samples in top-k ranks and the total number of positive samples. Recall at top-k is non-differentiable as it requires sorting and counting operations, both of which are non-differentiable. 

\begin{figure}
    \centering
    \includegraphics[width=\textwidth]{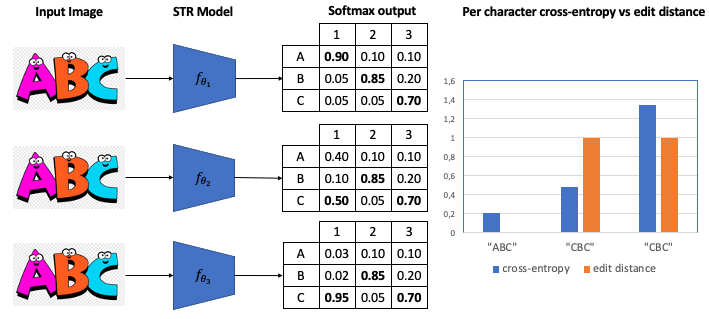}
    \caption{Left: Per-character softmax predictions are obtained using three different models for a cropped word image. The final predictions are obtained through argmax on these softmax predictions. Right: for the three predictions, a comparison between the mean of per-character cross-entropy loss and the edit distance is shown. The first prediction is correct and has an edit distance of zero. However, the cross-entropy loss still penalizes the model. The edit distance value for the second and the third predictions is the same. However, the value of cross-entropy is very different.}
    \label{fig:motivation_str}
\end{figure}

\begin{figure}
	\centering
	\includegraphics[width=\textwidth]{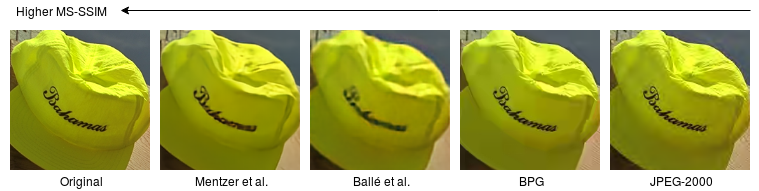}
	\caption{An example from the Kodak dataset \cite{kodak}. In order of MS-SSIM values: Mentzer ~\etal \cite{mentzer2018conditional} $>$ Ballé  ~\etal \cite{balle2016end} $>$ BPG \cite{bpg} $>$ JPEG-2000 \cite{skodras2001jpeg}. However, the order of performance based on $5$ human evaluations is: BPG \cite{bpg} $>$ Mentzer ~\etal \cite{mentzer2018conditional} $>$ JPEG-2000 \cite{skodras2001jpeg} $>$ Ballé  ~\etal \cite{balle2016end}. Visually the foreground and text in BPG are better in quality.}
	\label{fig:motivation_compression}
\end{figure}

\begin{figure}[t]
    \centering
    \includegraphics[width=\textwidth]{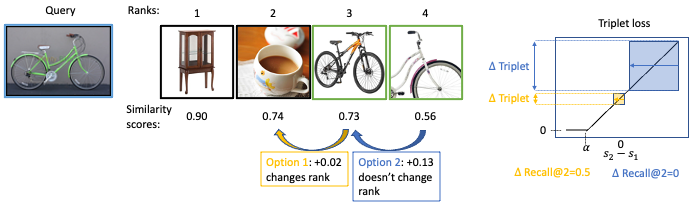}
    \caption{Similarity scores to the samples in the image collection are shown for a query image. Consider the change in similarity scores with two options: \textcolor{amber}{Option 1} is a small change in similarity that leads to a change in ranking, \textcolor{blue}{Option 2} is a bigger change in similarity that does not lead to any change in ranks. \textcolor{amber}{Option 1} leads to a change in the value of the evaluation metric, {\em i.e.}, recall@2 while the proxy loss, {\em i.e.}, triplet loss has a minor change in value. \textcolor{blue}{Option 2} changes the value of the triplet loss a lot, whereas the value of the evaluation metric does not change at all.}
    \label{fig:motivation_retrieval}
\end{figure}

Existing approaches for these tasks side-step the issue by using an alternative function as a loss function. In this thesis, we term this as a proxy loss function. A proxy loss function for a task is any function that can be used to attain a reasonable performance but is not tailored for the test-time objective. However, this function may not align well with the test-time evaluation metric, leading to a sub-optimal performance during inference. For the above-mentioned examples:
\begin{itemize}
\item \textbf{Scene Text Recognition}. Given an input image of a cropped word, scene text recognition is to predict the transcription of the word. An adequate evaluation metric for the task is the edit distance. Popular methods~\cite{baek2019wrong,litman2020scatter} use either per-character cross-entropy or CTC~\cite{graves2006connectionist} as the proxy loss function. Note that while the value of the edit distance decreases only when a correct unit operation is being made, the value of per-character cross-entropy may decrease when the probability of predicting the correct character increases. Further, when the prediction is correct, the per-character cross-entropy loss will continue to penalize the model until the correct predictions are very confident. Thus, the proxy loss function, in this case, is harsher on the model than the evaluation metric. The mismatch between the per-character cross-entropy loss and the edit distance is shown in Figure \ref{fig:motivation_str} through an example.
    \item \textbf{Scene Text Detection}. Given a natural scene image, the goal is to precisely localize all instances of text at a word level. The ground truth consists of rotated bounding boxes. Popular approaches use smooth-$L_{1}$ or $L_{2}$ distance for regressing bounding box coordinates. The use of these proxy losses does not have a strong correlation with IoU~\cite{rezatofighi2019generalized}. As noted by Yu {\em et al.}~\cite{yu2016unitbox}, IoU accounts for a bounding box as a whole, whereas regressing using an $L_{n}$ proxy loss treats each point independently.
    \item \textbf{Image compression}. The objective of compression approaches is to minimize the storage cost of images. In a lossy setting, this is achieved by removing information that is least noticeable to humans. Learning-based compression approaches follow a rate-distortion objective, which tries to minimize the storage requirements while keeping the distortion as low as possible. These approaches use peak-signal-to-noise ratio, mean squared error, or MS-SSIM as the proxy distortion loss function. Through extensive human evaluations, our work shows that PSNR or MS-SSIM do not correlate well with a human's perception of similarity. In fact, in a binary classification setup, where a human is asked to pick which of two images is closer to the original, these metrics are only slightly better than random predictions \cite{patel2019human,patel2019deep,patel2021saliency}. Figure \ref{fig:motivation_compression} shows this through a visual comparison of popular compression techniques.
    \item \textbf{Image Retrieval}. It is a task of ranking all database images according to the relevance to a query. The relevance could be at a semantic or at an instance level. Existing methods for this task use ranking proxy loss functions such as contrastive~\cite{hcl06}, triplet~\cite{skp+15}, and margin~\cite{wms+17} that pull the samples from the same class closer to one another and push the samples from different class away. As shown in Figure \ref{fig:motivation_retrieval}, the value of a proxy loss function changes with the change in the similarity scores. However, the evaluation metric recall@k for the task only depends on the rank of positive samples in the retrieved list. A small change in the value of the similarity score that changes the rank will cause a big change in the value of the evaluation metric, whereas it will not change the value of the proxy loss substantially~\cite{bxk+20}.
    \item \textbf{Supervised Contrastive Classification}. The goal is to learn representations that are useful for classification. The objective follows contrastive learning where an image, its views obtained via applying augmentations, and other images with the same semantic label are pulled closer to one another, and the samples from a different class are pushed apart. Contrastive supervised classification has shown to be superior and more robust than the use of standard soft-max cross-entropy loss function. With the goal of learning to classify, these approaches follow a two-step training procedure. First, the representations are learned via contrastive training, and then the model is fine-tuned with the cross entropy loss to perform classification. As shown by ~\cite{khosla2020supervised}, a simple approach to jointly train the representations and the classifier by combining contrastive loss and cross-entropy loss is sub-optimal.
\end{itemize}

The above-mentioned non-differentiable evaluation metrics can be categorized into decomposable and non-decomposable. An evaluation metric is decomposable if a per-point evaluation is available, {\em i.e.}, for a prediction of the model, there is a fixed ground truth, {\em e.g.}, edit distance, intersection-over-union, and human perception of similarity. If the per-point evaluation is not possible, {\em i.e.}, the metric is computed on a set of samples, the evaluation metric is termed non-decomposable, {\em e.g.}, recall@k, and average precision.

\section{Contributions}

The thesis focuses on bridging the gap between the training loss function and the test-time evaluation metric. The proposed solutions vary depending on the nature of the evaluation metric. For decomposable and non-differentiable evaluation metrics such as edit distance, intersection-over-union, and human perception of similarity, we propose to learn a differentiable surrogate of the evaluation metric and train the task-specific model with the learned surrogate~\cite{patel2020learning,patel2021feds,patel2021saliency}. For non-decomposable and non-differentiable evaluation metrics such as recall@k, we resort to a hand-crafted solution~\cite{patel2022recall}. However, the components involved in the developing recall@k surrogate, such as differentiable sorting and differentiable counting, are general and can be used for other evaluation metrics involving these operations as well.

The thesis makes the following contributions:

\begin{itemize}
    \item A technique for training a neural network by minimizing a surrogate loss that approximates the target evaluation metric, which is decomposable and non-differentiable. The surrogate is learned via a deep embedding where the Euclidean distance between the prediction and the ground truth corresponds to the value of the evaluation metric. The effectiveness of the proposed technique is demonstrated in a post-tuning setup, where a trained model is tuned using the learned surrogate. Without a significant computational overhead and any bells and whistles, improvements are demonstrated on challenging and practical tasks of scene-text recognition and detection. In the recognition task, the model is tuned using a surrogate approximating the edit distance metric and achieves up to $39\%$ relative improvement in the total edit distance. In the detection task, the surrogate approximates the intersection over union metric for rotated bounding boxes and yields up to $4.25\%$ relative improvement in the $F_{1}$ score. This work was published in~\cite{patel2020learning} and detailed in Chapter \ref{chapter:ls}.
    \item A procedure to robustly train a scene text recognition model using a learned surrogate of edit distance. The approach borrows from self-paced learning~\cite{kumar2010self} and filters out the training examples that are hard for the surrogate. The filtering is performed by judging the quality of the approximation using a ramp function, enabling end-to-end training. The experiments are conducted in a post-tuning setup, where a trained scene text recognition model is tuned using the learned surrogate of edit distance. The efficacy is demonstrated by improvements on various challenging scene text datasets such as IIIT-5K~\cite{mishra2012scene}, SVT~\cite{wang2011end}, ICDAR~\cite{lucas2003icdar,karatzas2013icdar,karatzas2015icdar}, SVTP~\cite{quy2013recognizing}, and CUTE~\cite{risnumawan2014robust}. The proposed method provides an average improvement of $11.2 \%$ on total edit distance and an error reduction of $9.5\%$ on accuracy. This work was published in~\cite{patel2021feds} and detailed in Chapter \ref{chapter:feds}.
    \item A differentiable surrogate of recall at top-k for learning visual representation models for retrieval (see Section \ref{sec:rs@k_method}). Since recall@k is a non-decomposable and non-differentiable function, our work relies on hand-crafting the solution. 
    \item An implementation for training with the proposed recall@k surrogate that side-steps the GPU memory constraints and can train up to a batch size of $16$k images on a single GPU (see Section \ref{sec:rs@k_implementation_details}).
    \item An efficient mixup technique that operates on pairwise scalar similarities and virtually increases the batch size further (see Section \ref{sec:rs@k_method}). The proposed mixup technique, in theory, is the same as the standard embedding mixup. However, in practice, it is computationally and memory efficient the mixed samples are virtual, and the technique only operates on pairwise similarity scores.
    \item With synergy between the above three components, our work attains state-of-the-art results on several metric learning benchmarks such as iNaturalist~\cite{vms+18}, Stanford Online Products~\cite{ohb16}, Stanford Cars~\cite{ksd+13}, and PUK VehicleID~\cite{ltw+16}. The same approach also attains state-of-the-art results, for instance-level search on revisited Oxford and Paris~\cite{rit+18}. This work was published in~\cite{patel2022recall} and presented in Chapter \ref{chapter:rs@k}.
    \item For fine-grained plant recognition, a model trained with the proposed approach and evaluated with kNN classification outperforms the classification approaches trained using the soft-max cross-entropy loss with performance-enhancing techniques such as class prior adaptation, heavy data augmentations, etc.
    \item A new version of the supervised contrastive training that jointly learns the classifier's parameters and the network's backbone. The proposed approach enjoys the robustness of contrastive training while still maintaining probabilistic interpretation like soft-max cross-entropy. The joint training of the backbone and the classifier eliminates the need for two-stage training. We empirically show that our proposed objective functions significantly improve over the standard cross entropy loss with more training stability and robustness in various challenging settings. This work was published in~\cite{aljundi2023contrastive} and presented in Chapter \ref{chapter:esupcon}.
    \item Additionally, we revisit a previously proposed contrastive-based objective function that approximates cross-entropy loss and present a simple extension to learn the classifier jointly (see Section \ref{sec:esupcon_jointlearning}).
\end{itemize}

An additional contribution of our research that is relevant but not included in the thesis is on image compression, where a new end-to-end trainable model for lossy image compression is proposed that includes several novel components. The method incorporates an adequate perceptual similarity metric, saliency in the images, and a hierarchical auto-regressive model. Our work demonstrates that the popularly used evaluation metrics such as MS-SSIM and PSNR are inadequate for judging the performance of image compression techniques as they do not align with the human perception of similarity. Alternatively, a new metric is proposed, which is learned on perceptual similarity data specific to image compression. The proposed compression model incorporates the salient regions and optimizes on the proposed perceptual similarity metric. The model not only generates images that are visually better but also gives superior performance for subsequent computer vision tasks such as object detection and segmentation when compared to existing engineered or learned compression techniques. Note that details of these contributions are excluded from the thesis as the work was done during an internship at AWS-AI. We refer the reader to the related publication~\cite{patel2021saliency} for more details.

\section{Structure of the Thesis}
The rest of the thesis is organized as follows. Chapter \ref{chapter:related_work} reviews existing methods on surrogate loss functions that attempt to train on non-differentiable objectives. Related work for specific tasks and loss functions are in each subsequent chapter. Chapter \ref{chapter:ls} presents LS, a method for learning surrogate loss functions for decomposable evaluation metrics. Chapter \ref{chapter:feds} presents FEDS, an improved approach for learning a surrogate loss function for edit distance via filtering. Chapter \ref{chapter:rs@k} presents a hand-crafted surrogate of recall@k, along with an efficient mixup technique. Chapter \ref{chapter:esupcon} presents ESupCon, a loss function for training classification models end-to-end via contrastive learning. The conclusions are made in Chapter \ref{chapter:conclusions}. For general curiosity, the Czech version of the abstract, translated by Google Translate and ChatGPT, is included in the appendix.

From an application point-of-view, Chapter \ref{chapter:ls} focuses on scene text recognition and detection, Chapter \ref{chapter:feds} on scene text recognition, Chapter \ref{chapter:rs@k} on metric learning, instance level search, and fine-grained recognition, Chapter \ref{chapter:esupcon} on image classification.

\newpage
\section{Publications}
This thesis builds on the results previously published in the following publications:
\begin{itemize}
    \item Learning Surrogates via Deep Embedding, \textbf{Yash Patel}, Tomas Hodan, Jiri Matas. \textit{European Conference on Computer Vision (ECCV) 2020} \cite{patel2020learning}.
    \item FEDS--Filtered Edit Distance Surrogate, \textbf{Yash Patel}, Jiri Matas. \textit{International Conference on Document Analysis and Recognition (ICDAR) 2021} \cite{patel2021feds}.
    \item Recall@k Surrogate Loss with Large Batches and Similarity Mixup, \textbf{Yash Patel}, Giorgos Tolias, Jiri Matas. \textit{IEEE/ CVF Conference on Computer Vision and Pattern Recognition (CVPR) 2022} \cite{patel2022recall}.
    \item Plant recognition by AI: Deep neural nets, transformers, and kNN in deep embeddings, Luk\'{a}\v{s} Picek, Milan \v{S}ulc, \textbf{Yash Patel} and Ji\v{r}\'{i} Matas. \textit{Frontiers in Plant Science 2022} \cite{picek2022plant}.
    \item Contrastive Classification and Representation Learning with Probabilistic Interpretation, Rahaf Aljundi, \textbf{Yash Patel}, Milan Sulc, Daniel Olmeda Reino, Nikolay Chumerin. \textit{Association for the Advancement of Artificial Intelligence (AAAI)} 2023 \cite{aljundi2023contrastive}.
    \end{itemize}

The following publications are related to the topic but were not included in the thesis, in order to keep the thesis more focused and easier to follow:
\begin{itemize}
    \item Saliency driven perceptual image compression, \textbf{Yash Patel}, Srikar Appalaraju, R. Manmatha. \textit{IEEE/ CVF Winter Conference on Applications of Computer Vision (WACV) 2021} \cite{patel2021saliency}.
    \item Neural Network-based Acoustic Vehicle Counting, Slobodan Djukanović, \textbf{Yash Patel}, Jiři Matas, Tuomas Virtanen. \textit{European Signal Processing Conference (EUSIPCO) 2021} \cite{djukanovic2021neural}. 
\end{itemize}

The following publications were published during the duration of the Ph.D. but are not included in the thesis because they are not directly related to the topic of the thesis:
\begin{itemize}
    \item ICDAR2019 Robust Reading Challenge on Multi-lingual Scene Text Detection and Recognition--RRC-MLT-2019, Nibal Nayef\textbf{*}, \textbf{Yash Patel}\textbf{*}, Michal Bušta, Pinaki Nath Chowdhury, Dimosthenis Karatzas, Wafa Khlif, Jiri Matas, Umapada Pal, Jean-Christophe Burie, Cheng-lin Liu, Jean-Marc Ogier (* indicates equal contribution). \textit{International Conference on Document Analysis and Recognition (ICDAR) 2019} \cite{nayef2019icdar2019}.
    \item Filtering, Distillation, and Hard Negatives for Vision-Language Pre-Training, Filip Radenovic, Abhimanyu Dubey, Abhishek Kadian, Todor Mihaylov, Simon Vandenhende, \textbf{Yash Patel}, Yi Wen, Vignesh Ramanathan, Dhruv Mahajan. \textit{IEEE/ CVF Conference on Computer Vision and Pattern Recognition (CVPR) 2023} \cite{radenovic2023filtering}.
    \item DocILE Benchmark for Document Information Localization and Extraction, Štěpán Šimsa, Milan Šulc, Michal Uřičář, \textbf{Yash Patel}, Ahmed Hamdi, Matěj Kocián, Matyáš Skalický, Jiří Matas, Antoine Doucet, Mickaël Coustaty, Dimosthenis Karatzas. \textit{International Conference on Document Analysis and Recognition (ICDAR) 2023} \cite{vsimsa2023docile_icdar}.
\end{itemize}

The following publications were not included as they are currently under review:
\begin{itemize}
    \item Generalized Differentiable RANSAC, Tong Wei, \textbf{Yash Patel}, Alexander Shekhovtsov, Jiri Matas, Daniel Barath. \textit{arXiv pre-print 2023} \cite{wei2023fully}.
    \item Self-Guided Semantic Alignment for Text Supervised Segmentation, \textbf{Yash Patel}, Yusheng Xie, Yi Zhu, Srikar Appalaraju, R. Manmatha. \textit{arXiv pre-print 2023} \cite{patel2023simcon}. 
\end{itemize}
\chapter{Related Work}
\label{chapter:related_work}

This chapter overviews some of the popular and general approaches for training with an approximation of the non-differentiable evaluation metrics. Specific related work to each task and the evaluation metric are provided in the subsequent chapters. 

Due to the non-differentiable nature of many practical evaluation metrics, there exists a large body of proxy loss functions that have been proposed as smooth metric relaxations. Examples include AUCPR loss~\cite{eban2017scalable}, pairwise AUCROC loss~\cite{rakotomamonjy2004optimizing}, Lovasz-Softmax loss for IoU metric~\cite{berman2018lovasz}, cost-sensitive classification for F-measure~\cite{puthiya2014optimizing}. 

Similar to the focus of our research on learning surrogates, there also have been efforts to learn the loss functions~\cite{xu2018autoloss,wu2018learning}. However, these loss-learning approaches are still based on metric relaxation schemes. Another set of approaches embeds the true evaluation metric as a correction term for optimization~\cite{hazan2010direct,song2016training}. These approaches are limited to the evaluation metrics that are available in a closed form and thus cannot be extended to the evaluation metrics such as edit distance. Our work on learning surrogates~\cite{patel2020learning,patel2021feds} follows a simpler approach of directly learning a metric space for approximating the target evaluation metric without making any assumptions about the underlying decomposable evaluation metric. 

An approach similar to our work is MetricOpt~\cite{huang2021metricopt}. This approach optimizes a model on arbitrary non-differentiable evaluation metrics such as misclassification rate and recall. This approach operates in a black-box setting where the computation details of the target metric are unknown and fine-tunes a pre-trained model using an approximation of the evaluation metric. Instead of fine-tuning the entire model, additional adapter parameters are introduced and fine-tuned using the approximation of the evaluation metric. Unlike our work, the approximation of the evaluation metric is learned using a straightforward regression task. Based on empirical comparisons on Standard online products~\cite{ohb16} dataset for image retrieval, our hand-crafted recall@k surrogate~\cite{patel2022recall} performs substantially better. 

Concurrent to our research is the work of Pogan{\v{c}}i{\'c} {\em et al.}~\cite{poganvcic2020differentiation} that implements an efficient backward pass through black box implementations of combinatorial solvers with linear objective functions. This work was extended in~\cite{rolinek2020optimizing} for optimizing rank-based evaluation metrics such as recall and average precision, where the efficacy of their approach was shown on metric learning and object detection benchmarks. When empirically compared on the standard benchmarks on the task of metric learning, our proposed approach of recall@k surrogate~\cite{patel2022recall} (Chapter \ref{chapter:rs@k}) attains substantially better results.

\chapter{Learning Surrogates via Deep Embedding}
\label{chapter:ls}

For many practical problems in computer vision, models are trained with simple proxy losses, which may not align with the evaluation metric. The evaluation metric may not always be differentiable, prohibiting its use as a loss function. An example of a non-differentiable metric is the visible surface discrepancy (VSD)~\cite{hodan2018bop} used to evaluate 6D object pose estimation methods. Another example is the edit distance (ED) defined by counting unit operations (addition, deletion, and substitution) necessary to transform one text string into another and is a common choice for evaluating scene text recognition methods \cite{karatzas2015icdar,karatzas2013icdar,nayef2019icdar2019}. Since ED is non-differentiable, the methods use either CTC \cite{graves2006connectionist} or per-character cross-entropy~\cite{baek2019wrong} as the proxy loss. Yet another popular non-differentiable metric is the intersection over union (IoU) used to compare the predicted and the ground truth bounding boxes when evaluating object detection methods. Although these methods typically resort to using proxy losses such as {\em smooth-L$_{1}$} \cite{ren2015faster} or {\em L$_{2}$} \cite{redmon2016you}, Rezatofighi~\etal~\cite{rezatofighi2019generalized} demonstrate that there is no strong correlation between {\em L$_{n}$} objectives and IoU. Further, Yu~\etal~\cite{yu2016unitbox} show that IoU accounts for a bounding box as a whole whereas regressing using an {\em L$_{n}$} proxy loss treats each point independently.

\begin{figure}[t!]
    \centering
    \includegraphics[width=0.9\textwidth, trim={2.7cm 1.7cm 1cm 0cm},clip]{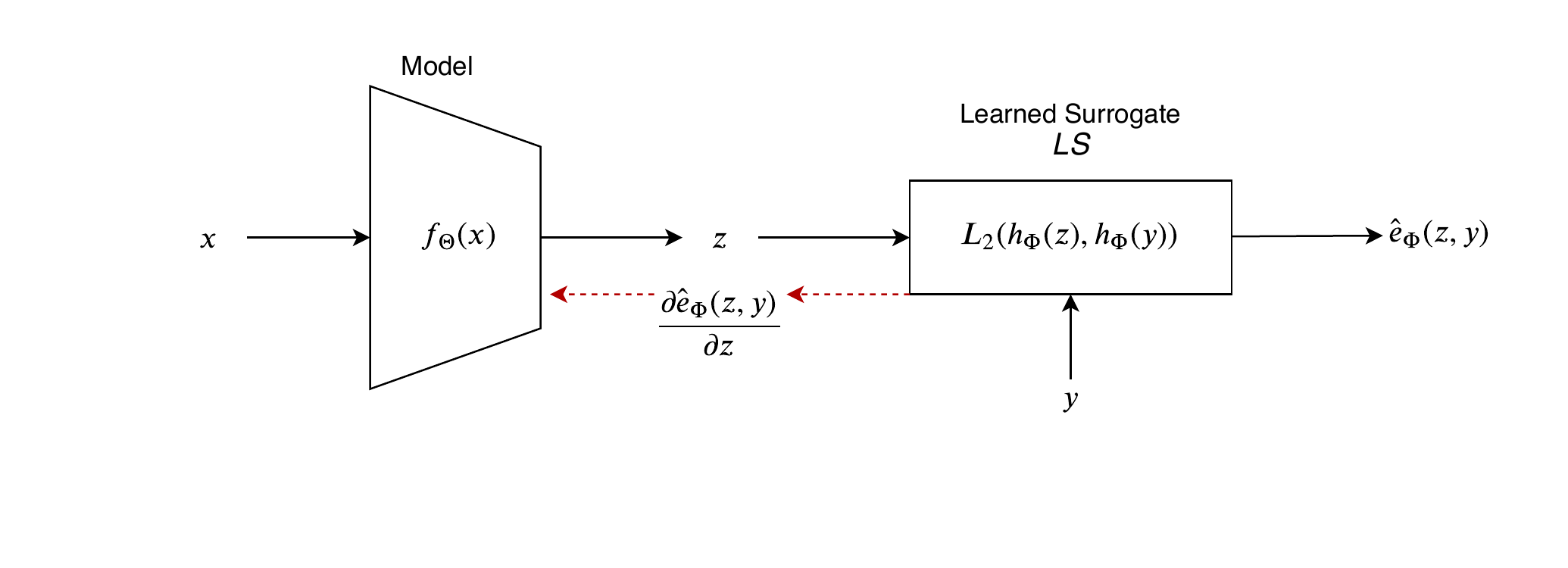}
    \caption{For the input $x$ with the corresponding ground-truth $y$, the model being trained outputs $z=f_{\Theta}(x)$. The learned surrogate provides a differentiable approximation of the evaluation metric: $\hat{e}_{\Phi}(z,y)=L_{2}(h_{\Phi}(z), h_{\Phi}(y))$,
    %where $\hat{e}_{\Phi}$ is the approximation,
    where $h_{\Phi}$ is a learned deep embedding model, and $h_{\Phi}(z)$ and $h_{\Phi}(y)$ are embedding representations for the prediction and the ground truth, respectively. Model $f_{\Theta}(x)$ for the target task (\emph{e.g.} scene text recognition or detection) is trained with the gradients from the surrogate: $\frac{\partial(\hat{e}_{\Phi}(z,y))}{\partial z}$.}
    \label{fig:overall_lsl}
\end{figure}

For popular metrics such as IoU, hand-crafted differentiable approximations have been designed \cite{yu2016unitbox,rezatofighi2019generalized}. However, hand-crafting a surrogate is not scalable as it requires domain expertise and may involve task-specific assumptions and simplifications. The IoU-loss introduced in \cite{yu2016unitbox,rezatofighi2019generalized} allows for optimization on the evaluation metric directly but makes a strong assumption about the bounding boxes to be axis-aligned. In numerous practical applications such as aerial image object detection~\cite{xia2018dota}, scene text detection~\cite{karatzas2015icdar} and visual object tracking~\cite{kristan2019seventh}, the bounding boxes may be rotated and the methods for such tasks revert to using simple but non-optimal proxy loss functions such as {\em smooth-L$_{1}$}~\cite{ma2018arbitrary,buvsta2018e2e,azimi2018towards}.

To address the aforementioned issues, this chapter proposes to learn a differentiable surrogate that approximates the evaluation metric and use the learned surrogate to optimize the model for the target task. The metric is approximated via a deep embedding, where the Euclidean distance between the prediction and the ground truth corresponds to the value of the metric. The mapping to the embedding space is realized by a neural network, which is learned using only the value of the metric. Gradients of this value with respect to the inputs are not required for learning the surrogate. In fact, the gradients may not even exist, as is the case of the edit distance metric. Throughout this chapter, we refer to the proposed method for training with learned surrogates as ``LS". Figure \ref{fig:overall_lsl} provides an overview of the proposed method.

In this chapter, the focus on a post-tuning setup, where a model that has converged on a proxy loss is tuned with LS. We consider two different optimization tasks: post-tuning with a learned surrogate for the edit distance (LS-ED) and the IoU of rotated bounding boxes (LS-IoU). To the best of our knowledge, we are the first to optimize directly on these evaluation metrics.

The rest of the chapter is structured as follows. Related work is reviewed in Section~\ref{sec:ls_related_work}, the technique for learning the surrogate and training with it is presented in Section~\ref{sec:ls_lsl}, experiments are shown in Section~\ref{sec:ls_experiments} and the chapter is concluded in Section~\ref{sec:ls_conclusion}.

\section{Related Work}
\label{sec:ls_related_work}

Training machine learning models by directly minimizing the evaluation metric has been shown effective on various tasks. For example, the state-of-the-art learned image compression \cite{lee2018context,balle2018variational} and super-resolution \cite{ledig2017photo,dong2015image} methods directly optimize the perceptual similarity metrics such as MS-SSIM \cite{wang2003multiscale} and the peak signal-to-noise ratio (PSNR). Certain compression methods optimize on an approximate of human perceptual similarity, which is learned in a supervised manner using annotated data \cite{patel2019deep,patel2020hierarchical}. Image classification methods \cite{krizhevsky2012imagenet,he2016deep,simonyan2014very} are typically trained with the cross-entropy loss that has been shown to align well with the misclassification rate, \emph{i.e.} the evaluation metric, under the assumption of large scale and clean data~\cite{berrada2018smooth,lapin2016loss}.

When designing evaluation metrics for practical computer vision tasks, the primary goal is to fulfil the requirements of potential applications and not to ensure the metrics being amenable to an optimization approach. As a consequence, many evaluation metrics are non-differentiable and cannot be directly minimized by the currently popular gradient-descent optimization approaches. For example, the visible surface discrepancy~\cite{hodan2018bop}, which is used to evaluate 6D object pose estimation methods, was designed to be invariant under pose ambiguity. This is achieved by calculating the error only over the visible part of the object surface, which requires a visibility test that makes the metric non-differentiable. Another example is the edit distance metric~\cite{karatzas2015icdar,DBLP:conf/icdar/GomezSGNVMBK17}, which is used to evaluate scene text recognition methods and is calculated via dynamic programming, which makes it infeasible to obtain the gradients.

There have been efforts towards approximating non-differentiable operations in a differentiable manner to enable end-to-end training. Kato~\etal~\cite{kato2018neural} proposed a neural network to approximate rasterization, allowing for a direct optimization on IoU for 3D reconstruction. Agustsson~\etal~\cite{agustsson2017soft} proposed a soft-to-hard vector quantization mechanism. It is based on soft cluster assignments during backpropagation, which allows neural networks to learn tasks involving quantization, \emph{e.g.} the image compression. Our work differs as we propose a general approach to approximate the evaluation metric, instead of approximating task-specific building blocks of neural networks.

Another line of research has focused on hand-crafting differentiable approximates of the evaluation metrics, which either align better with the metrics or enable training on them directly. Prabhavalkar~\etal~\cite{prabhavalkar2018minimum} proposed a way of optimizing attention based speech recognition models directly on word error rate. As mentioned earlier, \cite{yu2016unitbox,rezatofighi2019generalized} proposed ways for directly optimizing on intersection-over-union (IoU) as the loss for the case of axis-aligned bounding boxes.  Rahman~\etal~\cite{rahman2016optimizing} proposed a hand-crafted approximation of IoU for semantic segmentation.

Learning task-specific surrogates has been attempted. Nagendra~\etal~\cite{nagendar2018neuro} demonstrated that learning the approximate of IoU leads to better performance in the case of semantic segmentation. However, the method requires custom operations to estimate true and false positives, and false negatives, which makes the learning approach task-specific. Engilberge~\etal~\cite{DBLP:conf/cvpr/EngilbergeCPC19} proposed a learned surrogate for sorting-based tasks such as cross-modal retrieval, multi-label image classification and visual memorability ranking. Their results on sorting-based tasks suggest that learning the loss function could outperform hand-crafted losses.

More closely related to our work is the direct loss method by Hazan~\etal~\cite{hazan2010direct} where a surrogate loss is minimized by embedding the true loss as a correction term. Song~\etal~\cite{song2016training} extended this approach to the training of neural networks. However, it assumes that the loss can be disentangled into per-instance sub-losses, which is not always feasible, \emph{e.g.} the $F_{1}$ score \cite{grabocka2019learning} involves two non-decomposable functions (recall and precision). An alternative is to directly learn the amount of update values that are applied to the parameters of the prediction model. The framework proposed in \cite{li2017learning} includes a controller that uses per-parameter learning curves comprised of the loss values and derivatives of the loss with respect to each parameter. The method suffers from two drawbacks that prohibit its direct application to training on evaluation metrics: a) for large networks, it is computationally infeasible to store the learning curve of every parameter, and b) no gradient information is available for non-differentiable losses.

Our work is similar to the approach by Grabocka~\etal~\cite{grabocka2019learning}, where the evaluation metric is approximated by a neural network. Their approach differs as the network learning the surrogate takes both the prediction and the ground truth as the input and directly regresses the value of the metric. Since we formulate the task as embedding learning and train the surrogate such that the $L_{2}$ in the embedded space corresponds to the metric, our method ensures that the gradients are smaller when the prediction is closer to the ground truth. Furthermore, as illustrated in Section \ref{sec:ls_lsl}, we learn the surrogate with an additional gradient penalty term to ensure that the gradients obtained from our learned surrogate are bounded for stable training.

\section{Learning Surrogates via Deep Embedding}
\label{sec:ls_lsl}

Say that the supervised task is being learned from samples drawn uniformly from a distribution $(x,y)\sim P_{D}$. For a given input $x$ and an expected output $y$, a neural network model outputs $z=f_{\Theta}(x)$ where $\Theta$ are the model parameters learned via backpropagation as:

\begin{equation}
    \Theta_{t+1} \leftarrow \Theta_{t} - \eta \frac{\partial l(z,y)}{\partial \Theta_{t}}
\end{equation}
where $l(z,y)$ is a differentiable loss function, $t$ is the training iteration, and $\eta$ is the learning rate.

The model trained with loss $l(z,y)$ is evaluated using metric $e(z,y)$. When metric $e(z,y)$ is differentiable, it can be directly used as the loss. The technique proposed in this chapter addresses the cases when metric $e(z,y)$ is non-differentiable by learning a differentiable surrogate loss denoted as $\hat{e}_{\Phi}(z,y)$. The learned surrogate is realized by a neural network, which is differentiable and is used to optimize the model. The weight updates are:

\begin{equation}
\Theta_{t+1} \leftarrow \Theta_{t} - \eta \frac{\partial \hat{e}_{\Phi}(z,y)}{\partial \Theta_{t}}
\end{equation}

\subsection{Definition of the Surrogate}

The surrogate is defined via a learned deep embedding $h_{\Phi}$ where the Euclidean distance between the prediction $z$ and the ground truth $y$ corresponds to the value of the evaluation metric:

\begin{equation}
    \hat{e}_{\Phi}(z,y) = \left\Vert h_{\Phi}(z) - h_{\Phi}(y)\right\Vert_{2}
    \label{eq:ls_defination_ehat}
\end{equation}

\subsection{Learning the Surrogate}
\label{sec:learning_the_surrogate}

Learning the surrogate, \emph{i.e.} approximating the evaluation metric, with a deep neural network is formulated as a supervised learning task requiring three major components: a model architecture, a loss function, and a source of training data.

\subsubsection{Architecture.}
In this chapter, the architecture is designed manually, such that it is suitable for the nature of the inputs $z$ and $y$ (details are in Section~\ref{sec:ls_experiments}). Modern approaches for architecture search, \emph{e.g.}~\cite{elsken2018neural,ryoo2019assemblenet,zoph2016neural}, could yield better results but are computationally expensive.

\subsubsection{Training Loss.}

The surrogate is learned with the following objectives:
\begin{enumerate}
    \item The learned surrogate corresponds to the value of the evaluation metric: \begin{equation}
        \hat{e}_{\Phi}(z,y) \approx e(z,y)
    \end{equation}
    \item The first order derivative of the learned surrogate with respect to the prediction $z$ is close to $1$: \begin{equation}
       \left\Vert \frac{\partial \hat{e}_{\Phi}(z,y)}{\partial z} \right\Vert_{2} \approx 1
       \label{eq:ls_gp_term}
    \end{equation}
\end{enumerate}

Both objectives are realized and linearly combined in the training loss:
\begin{equation}
    \text{loss}(z,y) = \left\Vert \big(\hat{e}_{\Phi}(z,y) - e(z,y)\right\Vert_2^2 + \lambda \left(\left\Vert \frac{\partial \hat{e}_{\Phi}(z,y)}{\partial z} \right\Vert_{2} - 1\right)^{2}
    \label{eq:ls_lsl_loss}
\end{equation}

Bounding the gradients (Equation~\ref{eq:ls_gp_term}) has shown to enhance the training stability for Generative Adversarial Networks~\cite{gulrajani2017improved} and has shown to be useful for learning the surrogate. Parameters $\Phi$ of the embedding model $h_\Phi$ are learned by minimizing the loss (Equation~\ref{eq:ls_lsl_loss}).

\subsubsection{Source of Training Data.}
Source of the training data for learning the surrogate determines the quality of the approximation over the domain. The model $f_{\Theta}(x) = z$ for the supervised task is trained on samples obtained from a dataset $D$. Let us assume that $R$ is a random data generator providing examples for the learning of the surrogate, sampled uniformly in the range of the evaluation metric (see Section~\ref{sec:ls_experiments} for details). Note that $R$ is independent of $f_{\Theta}(x)$.

Three possibilities for the data source are considered:
\begin{enumerate}
    \item \textit{Global approximation}: $(z,y) \sim P_{R}$.
    \item \textit{Local approximation}: $(z,y) \sim P_{f_{\Theta}(x)}$, where $(x,y) \sim P_{D}$.
    \item \textit{Local-global approximation}: $(z,y) \sim P_{f_{\Theta}(x) \cup R}$.
\end{enumerate}

The local-global approximation yields a high quality of both the approximation and gradients (Section \ref{sec:exp_analysis}) and is therefore used in the main experiments.

\subsection{Training with the Learned Surrogate}
\label{sec:training_with_the_learned_surrogate}

The learned surrogate is used in a post-tuning setup, where model $f_{\Theta}(x)$ has been pre-trained using a proxy loss. This setup ensures that $f_{\Theta}(x)$ is not generating random outputs and thus simplifies post-tuning with the surrogate. The parameters of the surrogate $\Phi$ are initialized randomly.

Learning of the surrogate $\hat{e}_{\Phi}$ and post-tuning of the model $f_{\Theta}(x)$ are conducted alternatively. The surrogate parameters $\Phi$ are updated first while the model parameters $\Theta$ are fixed. The surrogate is learned by sampling $(z,y)$ jointly from the model and the random generator. Subsequently, the model parameters are trained while the surrogate parameters are fixed. Algorithm \ref{alg:train_with_lsl} demonstrates the overall training procedure.

\begin{algorithm}
\caption{Training with LS \textit{(local-global approximation)}}
\label{alg:train_with_lsl}
\textbf{Inputs}: Supervised data $D$, random data generator $R$, evaluation metric $e$.\\
\textbf{Hyper-parameters}: Number of update steps $I_{a}$ and $I_{b}$, learning rates $\eta_{a}$ and $\eta_{b}$, number of epochs $E$.\\
\textbf{Objective}: Train the model for a given task that is $f_{\Theta}(x)$ and the surrogate ,\emph{i.e.}, $e_{\Phi}$.
\begin{algorithmic}[1]
\State \textit{Initialize} $\Theta \leftarrow$ pre-trained weights, $\Phi \leftarrow$ random weights.
\For{epoch = 1,...,E}
    \For{i = 1,...,$I_{a}$}
        \State sample $(x,y) \sim P_{D}$, sample $(z_{r},y_{r}) \sim P_{R}$
        \State inference $z=f_{\Theta^{epoch-1}}(x)$
        \State compute loss $l_{\hat{e}} =$ {\em loss}$(z,y) + ${\em loss}$(z_{r},y_{r})$ (Equation \ref{eq:ls_lsl_loss})
        \State $\Phi^{i} \leftarrow \Phi^{i-1} - \eta_{a} \frac{\partial l_{\hat{e}}}{\partial \Phi^{i-1}}$
    \EndFor \\
   \quad \quad $\Phi \leftarrow \Phi^{I_{a}}$
    \For{i = 1,...,$I_{b}$}
        \State sample $(x,y) \sim P_{D}$
        \State inference $z = f_{\Theta^{i-1}}(x)$
        \State compute loss $l_{f} = \hat{e}_{\Phi^{epoch}}(z,y)$ (Equation \ref{eq:ls_defination_ehat})
        \State $\Theta^{i} \leftarrow \Theta^{i-1} - \eta_{b} \frac{\partial(l_{f})}{\partial \Theta^{i-1}}$
    \EndFor \\
    \quad \quad $\Theta \leftarrow \Theta^{I_{b}}$
\EndFor
\end{algorithmic}
\end{algorithm}

\section{Experiments}
\label{sec:ls_experiments}

The efficacy of LS is demonstrated on two different tasks: post-tuning with a learned surrogate for the edit distance (Section \ref{sec:exp_ed}) and for the IoU of rotated bounding boxes (Section \ref{sec:exp_iou}). This section provides details of the models for these tasks, design choices for learning the surrogates and empirical evidence showing the efficacy of LS. Unless stated otherwise, the results were obtained using the local-global approximation setup as elaborated in Algorithm \ref{alg:train_with_lsl}.

\subsection{Analysing the Learned Surrogates} \label{sec:exp_analysis}

The aspects considered for evaluating the surrogates are:
\begin{enumerate}
    \item The quality of approximation $\hat{e}_{\Phi}(z,y)$.
    \item The quality of gradients $\frac{\partial(\hat{e}_{\Phi}(z,y))}{\partial z}$.
\end{enumerate}

Both the quality of the approximation and the gradients depend on three components: an architecture, a loss function, and a source of training data (Section \ref{sec:learning_the_surrogate}). Given an architecture, the choices for the loss function to learn the surrogate and the training data are justified subsequently.

\subsubsection{Quality of approximation.} 
The quality of the approximation is judged by comparing the value of the surrogate with the value of the evaluation metric, calculated on samples obtained from model $f_{\Theta}(x)$. When learning the surrogate, higher quality of approximation is enforced by the mean squared loss between $e(z,y)$ and $\hat{e}_{\Phi}(z,y)$ (the first term on the right-hand side of Equation \ref{eq:ls_lsl_loss}). Figure \ref{fig:ls_ed_plots} (left) shows the quality of the approximation measured by the $L_{1}$ distance between the learned surrogate and the edit distance. It can be seen that the surrogate approximates the edit distance accurately (the $L_{1}$ distance drops swiftly below $0.2$, which is negligible for the edit distance).

\subsubsection{Quality of gradients.} 
Judging the quality of gradients is more complicated. When learning the surrogate, the gradient-penalty term attempts to make the gradients bounded, \emph{i.e.} to make the training stable (second term on the right-hand side of the equation \ref{eq:ls_lsl_loss}). However, this is not sufficient if the gradients do not optimize $f_{\Theta}(x)$ on the evaluation metric. We rely on the improvement or the decline in the performance of the model $f_{\Theta}(x)$ to judge the quality of the gradients. Table \ref{table:detection_results} shows that the local-global approximation leads to the largest improvements when optimizing on IoU for rotated bounding boxes.

\subsubsection{Choice of training data.}
\begin{figure}[t!]
    \centering
    \includegraphics[width=\textwidth,trim={0cm 0.5cm 0cm 0cm},clip]{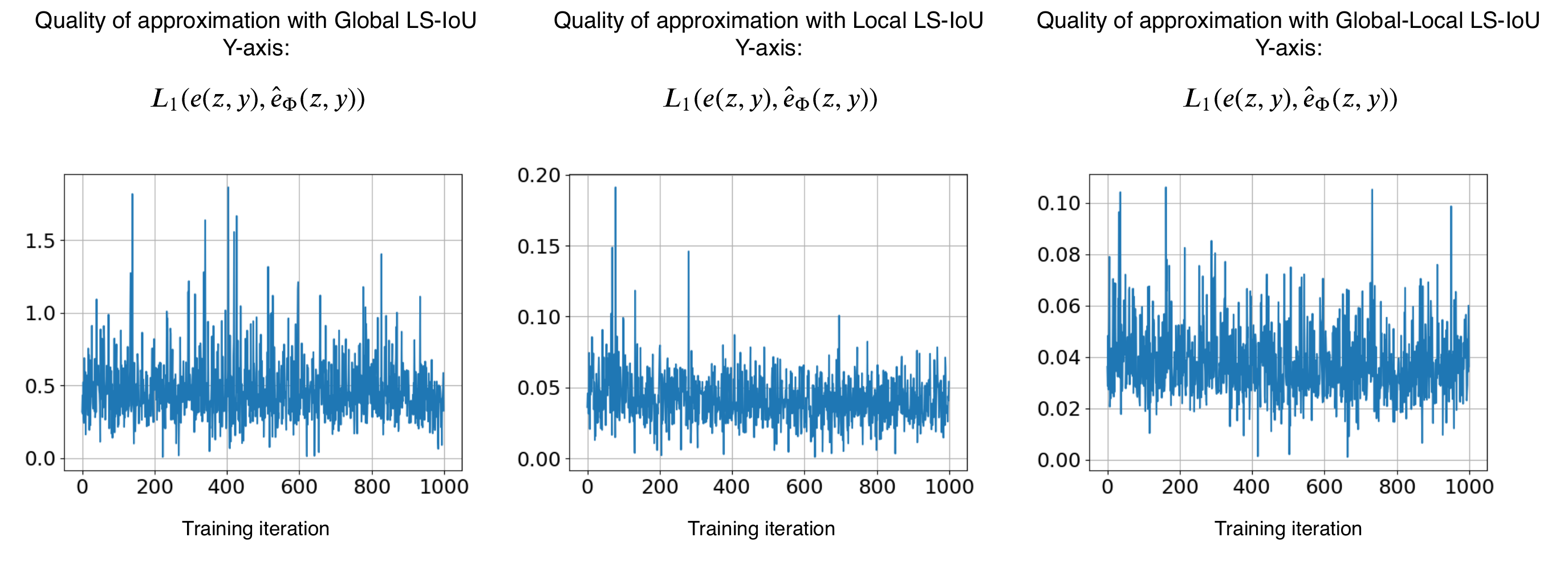}
    \caption{The error in the approximation of the IoU for rotated bounding boxes is shown for the first $1K$ iterations of the training with LS-IoU. Error is measured by the $L_{1}$ distance between IoU and the surrogate. It can be seen that the error is high for the global and low for the local and global-local approximation variants.}
    \label{fig:ls_iou_plots}
\end{figure}

Figure \ref{fig:ls_iou_plots} shows the quality of approximation with different choices of training data for learning the surrogate. These empirical observations suggest that using global approximation leads to a low quality of the approximation. This can be accounted to the domain gap between the data obtained from the random generator and the model. Using the local approximation leads to a higher quality of the approximation, however, the gradients obtained from the surrogate are not useful to train $f_{\Theta}(x)$ (Table. \ref{table:detection_results}), \emph{i.e.} although the quality of the approximation is high, the quality of gradients is not. This can be attributed to surrogate over-fitting on samples obtained from the model and losing generalization capability on samples outside this distribution. Finally, it was observed that using the local-global approximation leads to both properties -- high quality of approximation and high quality of gradients.

\begin{figure}[t!]
    \centering
    \includegraphics[width=\textwidth,trim={1cm 0cm 1cm 0cm},clip]{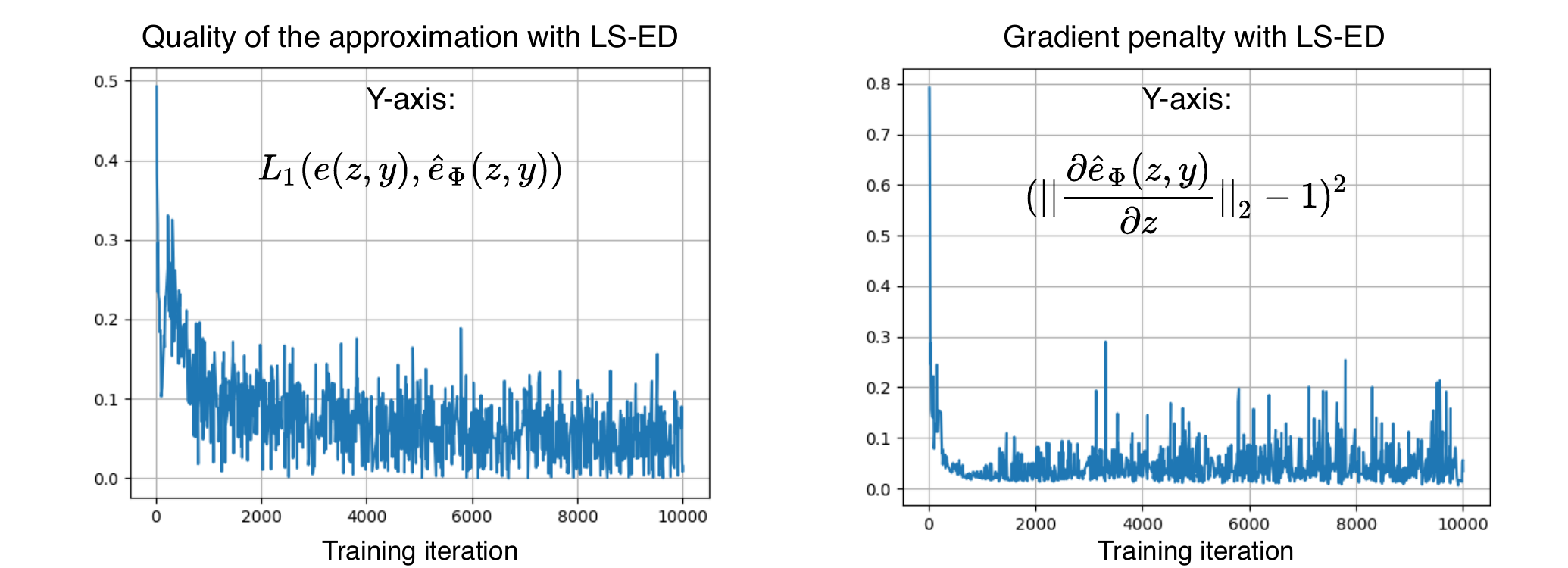}
    \caption{\textbf{Left}: The error in approximation for the first $10K$ training iterations. The error is obtained by computing the $L_{1}$ distance between the true edit distance values and the LS-ED predictions and dividing by the batch size. Note that the edit distance can only take non-negative integer values, thus the error in the range of $0-0.2$ is fairly low. \textbf{Right}: The gradient penalty term from the optimization of the LS-ED model (Equation \ref{eq:ls_lsl_loss}).}
    \label{fig:ls_ed_plots}
\end{figure}

\subsection{Post-Tuning with a Learned Surrogate for ED (LS-ED)}
\label{sec:exp_ed}

It is experimentally shown that LS can improve scene text recognition models (STR) on edit distance (ED), which is a popularly used metric to evaluate STR methods~\cite{karatzas2015icdar,karatzas2013icdar,nayef2019icdar2019}. The empirical evidence shows that post-tuning STR models with LS-ED lead to improved performance on various metrics such as accuracy, normalized edit distance, and total edit distance~\cite{DBLP:conf/icdar/GomezSGNVMBK17}.

\subsubsection{Scene Text Recognition (STR).}
Given an input image of a cropped word, the task of STR is to generate the transcription of the word. The state-of-the-art architectures for scene text recognition can be factorized into four modules \cite{baek2019wrong} (in this order): (a) transformation, (b) feature extraction, (c) sequence modelling, and (d) prediction. The feature extraction and prediction are the core modules of any STR model and are always employed. On the other hand, transformation and sequence modelling are not essential but have shown to improve the performance on benchmark datasets. Post-tuning with LS-ED is investigated for two different configurations of STR models.

The transformation module attempts to rectify the curved or tilted text, making the task easier for the subsequent modules of the model. It is learned jointly with the rest of the modules, and a popular choice is thin-plate spline (TPS) \cite{DBLP:conf/cvpr/ShiWLYB16,DBLP:conf/nips/JaderbergSZK15,DBLP:conf/bmvc/LiuCWSH16}. TPS can be either present or absent in the overall STR model. 

The feature extraction module maps the image or its transformed version to a representation that focuses on the attributes relevant for character recognition, while the irrelevant features are suppressed. Popular choices include VGG-16 \cite{simonyan2014very} and ResNet \cite{he2016deep}. It is a core module of the STR model and is always present. 

The features are the input of the sequence modelling module, which captures the contextual information within a sequence of characters for the next module to predict each character more robustly. BiLSTM \cite{hochreiter1997long} is a popular choice.

The output character sequence is predicted from the identified features of the image. The choice of the prediction module depends on the loss function used for training the STR model. Two popular choices of loss functions are CTC \cite{graves2006connectionist} (sigmoid output) or attention \cite{DBLP:conf/cvpr/ShiWLYB16} (per-character softmax output). 

Baek~\etal~\cite{baek2019wrong} provides a detailed analysis of STR models and the impact of different modules on the performance. Following \cite{baek2019wrong}, LS-ED is investigated with the state-of-the-art performing configuration, which is \textit{TPS-ResNet-BiLSTM-Attn}. To demonstrate the efficacy of LS-ED, results are also shown with \textit{ResNet-BiLSTM-Attn}, \emph{i.e.}, the transformation module is removed. Note that the CTC based prediction has been shown to consistently perform worse compared to the attention counter-part \cite{baek2019wrong}, and thus the analysis in this chapter has been narrowed down to only the attention-based prediction.

Similar to \cite{baek2019wrong}, the STR models are trained on the union of the synthetic data obtained from MJSynth \cite{DBLP:journals/corr/JaderbergSVZ14} and SynthText \cite{DBLP:conf/cvpr/GuptaVZ16} resulting in a total of $14.4$ million training examples. Furthermore, following the standard setup of \cite{baek2019wrong}, there is no fine-tuning performed in a dataset-specific manner before the final testing. Let us say that the STR model is $f_{\Theta}(x)$, such that $f_{\Theta}:\mathbb{R}^{100\times 32 \times 1}\xrightarrow{}\mathbb{R}^{|A|\times L}$. The dimensions of the input cropped word image $x$ is fixed to $100\times 32 \times 1$ (gray-scale). The output for attention based prediction module is a per-character softmax over the set of characters. Here $L$ is the maximum length of characters in the word and $|A|$ is the number of characters. During inference, argmax is performed at each character location to output the predicted text string. The ground truth $y$ is represented as a per-character one-hot vector.

The STR models are first trained with the proxy loss, \emph{i.e.}, cross-entropy for $300K$ iterations with a mini-batch size of $192$. The models are optimized using ADADELTA \cite{DBLP:journals/corr/abs-1212-5701} (same setup as \cite{baek2019wrong}). Once the training is completed these models are tuned with LS-ED on the same set of $14.4$ million training examples for another $20K$ iterations. The models trained purely on the synthetic datasets are tested on a collection of real datasets - IIIT-5K \cite{mishra2012scene}, SVT \cite{wang2011end}, ICDAR'03 \cite{lucas2003icdar}, ICDAR'13 \cite{karatzas2013icdar}, ICDAR'15 \cite{karatzas2015icdar}, SVTP \cite{quy2013recognizing}  and CUTE \cite{risnumawan2014robust} datasets. 

\begin{figure}[t!]
    \centering
    \includegraphics[width=\textwidth]{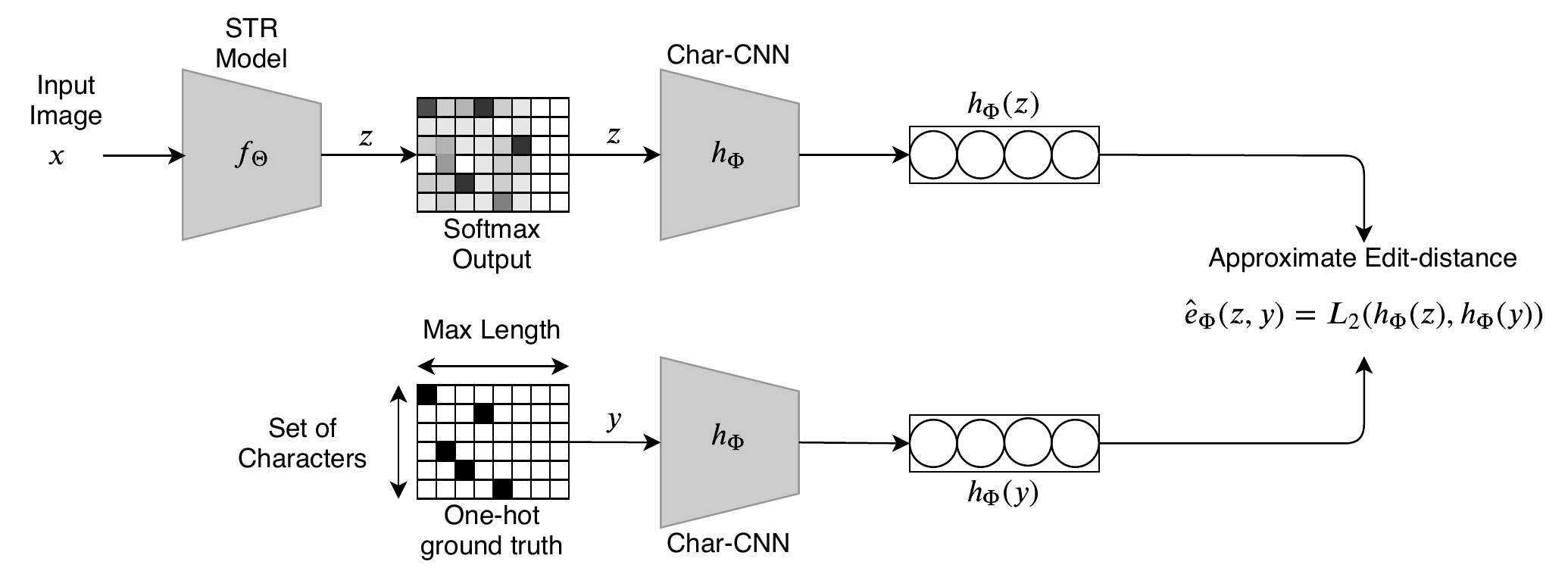}
    \caption{Training scene text recognition (STR) models with LS-ED. The output of the STR model $z_{|A|\times L}$ and the ground-truth $y_{|A|\times L}$ ($L$ is the maximum length of the word and $A$ is the set of characters) are fed to the Char-CNN embedding model to obtain embedding vectors, $h_{\Phi}(z)$ and $h_{\Phi}(y)$ respectively. The approximate edit distance value is obtained by computing $\hat{e}_{\Phi}(z,y) = L_{2}(h_{\Phi}(z),h_{\Phi}(y))$.}
    \label{fig:lsl_ed}
\end{figure}

\subsubsection{LS-ED architecture.}
Char-CNN architecture~\cite{DBLP:conf/nips/ZhangZL15} is used for learning the deep embedding $h_{\Phi}$. It consists of five $1D$ convolution layers equipped with LeakyReLU activation \cite{DBLP:journals/corr/XuWCL15} followed by two fully connected layers. The embedding $h_{\Phi}$ maps the input such that $h_{\Phi}: \mathbb{R}^{|A|\times L} \xrightarrow{} \mathbb{R}^{1024}$. Note that since $h_{\Phi}$ constitutes of convolution and fully-connected layers, it is differentiable and allows for backpropagation to the STR model. In feed-forward, the two embeddings for the ground-truth $y$ (one-hot) and the model prediction $z$ (softmax) are obtained by performing feed-forward through $h_{\Phi}$ and an approximate of edit distance is computed by measuring the $L_{2}$ between the two vectors (Figure \ref{fig:lsl_ed}).

\subsubsection{Post-tuning with LS-ED.}
A random generator is designed for this task, which generates a pair of words $(z_{r}, y_{r})$ and ensures uniform sampling in the range of the true error. It was observed that the uniform sampling is essential to avert over-fitting of the learned surrogate on a certain range of the true metric. For the edit distance metric $e(z,y) \in \{0,...,b\}$ ($b$ being the maximum possible value), the generator samples a word randomly from a text corpus and distorts the words by performing random addition, deletion, and substitution operations.

The post-tuning of the STR model $f_{\Theta}(x)$ with LS-ED follows Algorithm~\ref{alg:train_with_lsl}. For the case of the edit distance, there is a significant domain gap between the samples obtained from the STR model ($z$) and the random generator ($z_{r}$). This is because the random generator operates directly on the text string, \textit{i.e.}, $z_{r}$ is one-hot representation. Thus, using the global approximation setting yields a low quality of the approximation. Further, it was observed that training the surrogate purely with the data generated from the STR model, \emph{i.e.}, local approximation, leads to a good approximation but does not lead to an improvement in the performance of the STR model, which indicates a low quality of gradients.

Finally as described in Algorithm \ref{alg:train_with_lsl}, the local-global approximation is used. The quality of approximation and the gradient penalty from post-tuning with LS-ED are shown in Figure \ref{fig:ls_ed_plots}. Note that the edit distance value is a whole number and the surrogate attempts to approximate it, thus the error in approximation as shown in Figure \ref{fig:ls_ed_plots} is low. The quality of the gradients can be seen by improvement in the performance of the STR models. Thus the local-global approximation guides to a high quality of both the approximation and gradients.

The results for the two configurations of STR models, \emph{i.e.}, \textit{ResNet-BiLSTM-Attn} and \textit{TPS-ResNet-BiLSTM-Attn}, are shown in Table \ref{table:resnet_bilstm_attn} and Table \ref{table:ls_tps_resnet_bilstm_attn}, respectively. It can be observed that LS-ED improves the performance of the STR models on all metrics. The most significant gains are observed on total-edit distance (TED) as the surrogate attempts to minimize its approximation.

\begin{table}[t!]
\begin{center}
\resizebox{\textwidth}{!}{
\begin{tabular}{ c | c | l | l | l}
\toprule
\textbf{\specialcell{Test \\ Data}} & \textbf{\specialcell{Loss \\ Function}} & \textbf{$\uparrow$ \specialcell{Acc.}} & 
\textbf{$\uparrow$ \specialcell{NED}} &
\textbf{$\downarrow$ \specialcell{TED}}\\
\midrule
IIIT-5K  & Cross-Entropy & $84.300$ & $0.954$ & $945$\\
IIIT-5K  & LS-ED  & $86.300$ \textcolor{green}{$+2.37\%$} & $0.953$ \textcolor{red}{$-0.10\%$} &  $837$ \textcolor{green}{$+11.42\%$}\\
\midrule
SVT  & Cross-Entropy & $84.699$ & $0.940$ & $229$\\
SVT  & LS-ED  & $86.399$ \textcolor{green}{$+2.00\%$} & $0.947$ \textcolor{green}{$+0.74\%$} & $196$ \textcolor{green}{$+14.41\%$}\\
\midrule
ICDAR'03 & Cross-Entropy & $92.558$ & $0.972$ & $151$\\
ICDAR'03 & LS-ED  & $94.070$ \textcolor{green}{$+1.63\%$} & $0.977$ \textcolor{green}{$+0.51\%$} & $119$ \textcolor{green}{$+26.89\%$}\\
\midrule
ICDAR'13 & Cross-Entropy & $89.754$ & $0.949$ & $260$ \\
ICDAR'13 & LS-ED  & $91.133$ \textcolor{green}{$+1.53\%$}& $0.960$ \textcolor{green}{$+1.15\%$}& $157$ \textcolor{green}{$+39.61\%$}\\
\midrule
ICDAR'15 & Cross-Entropy & $71.452$ & $0.889$ & $1135$ \\
ICDAR'15 & LS-ED  & $74.655$ \textcolor{green}{$+4.48\%$} & $0.899$ \textcolor{green}{$+1.12\%$} & $1013$ \textcolor{green}{$+10.74\%$}\\
\midrule
SVTP & Cross-Entropy & $74.109$ & $0.891$ & $424$ \\
SVTP & LS-ED  & $77.519$ \textcolor{green}{$+4.60\%$} & $0.901$ \textcolor{green}{$+1.22\%$} & $381$ \textcolor{green}{$+10.14\%$} \\
\midrule
CUTE & Cross-Entropy & $68.293$ & $0.838$ & $285$ \\
CUTE & LS-ED  & $71.777$ \textcolor{green}{$+5.10\%$} & $0.868$ \textcolor{green}{$+3.57\%$} & $234$ \textcolor{green}{$+17.89\%$}\\
\midrule
\end{tabular}}
\end{center}
\caption{
ResNet-BiLSTM-Attn: The models are evaluated on IIIT-5K \cite{mishra2012scene}, SVT \cite{wang2011end}, ICDAR'03 \cite{lucas2003icdar}, ICDAR'13 \cite{karatzas2013icdar}, ICDAR'15 \cite{karatzas2015icdar}, SVTP \cite{quy2013recognizing}  and CUTE \cite{risnumawan2014robust} datasets. The results are reported using accuracy \textbf{Acc.} (higher is better), normalized edit distance \textbf{NED} (higher is better) and total edit distance \textbf{TED} (lower is better).
Relative gains are shown in \textcolor{green}{green} and relative declines in \textcolor{red}{red}.
}
\label{table:resnet_bilstm_attn}
\end{table}

\begin{table}[t!]
\begin{center}
\resizebox{\textwidth}{!}{
\begin{tabular}{ c | c | l | l | l}
\toprule
\textbf{\specialcell{Test \\ Data}} & \textbf{\specialcell{Loss \\ Function}} & \textbf{$\uparrow$ \specialcell{Acc.}} & 
\textbf{$\uparrow$ \specialcell{NED}} &
\textbf{$\downarrow$ \specialcell{TED}}\\
\midrule
IIIT-5K  & Cross-Entropy & $87.500$ & $0.961$ & $722$\\
IIIT-5K  & LS-ED  & $87.933$ \textcolor{green}{$+0.49\%$} & $0.963$ \textcolor{green}{$+0.20\%$} & $645$  \textcolor{green}{$+10.66\%$}\\
\midrule
SVT  & Cross-Entropy & $87.172$ & $0.952$ & $180$\\
SVT  & LS-ED  & $86.708$ \textcolor{red}{$-0.53$} & $0.954$ \textcolor{green}{$+0.21\%$} & $163$ \textcolor{green}{$+9.44\%$}\\
\midrule
ICDAR'03 & Cross-Entropy & $94.302$ & $0.979$ & $110$\\
ICDAR'03 & LS-ED  & $94.535$ \textcolor{green}{$+0.24\%$} & $0.981$ \textcolor{green}{$+0.20\%$} & $99$       \textcolor{green}{$+10.00\%$}\\
\midrule
ICDAR'13 & Cross-Entropy & $92.020$ & $0.966$ & $137$ \\
ICDAR'13 & LS-ED  & $92.299$ \textcolor{green}{$+0.30\%$} & $0.979$ \textcolor{green}{$+1.34\%$} & $108$ \textcolor{green}{$+21.16\%$} \\
\midrule
ICDAR'15 & Cross-Entropy & $78.520$ & $0.915$ & $868$ \\
ICDAR'15 & LS-ED  & $78.410$ \textcolor{red}{$-0.14\%$} & $0.915$ \textcolor{black}{$\pm0.00\%$} & $837$ \textcolor{green}{$+3.57\%$} \\
\midrule
SVTP & Cross-Entropy & $78.605$ & $0.912$ & $346$ \\
SVTP & LS-ED  & $79.225$ \textcolor{green}{$+0.78\%$} & $0.913$ \textcolor{green}{$+0.10\%$} & $333$ \textcolor{green}{$+3.75\%$}\\
\midrule
CUTE & Cross-Entropy & $73.171$ & $0.871$ & $224$ \\
CUTE & LS-ED  & $74.216$ \textcolor{green}{$+1.42\%$} & $0.875$ \textcolor{green}{$+0.45\%$} & $219$ \textcolor{green}{$+2.23\%$}\\
\bottomrule
\end{tabular}}
\end{center}
\caption{
TPS-ResNet-BiLSTM-Attn: The models are evaluated on IIIT-5K \cite{mishra2012scene}, SVT \cite{wang2011end}, ICDAR'03 \cite{lucas2003icdar}, ICDAR'13 \cite{karatzas2013icdar}, ICDAR'15 \cite{karatzas2015icdar}, SVTP \cite{quy2013recognizing}  and CUTE \cite{risnumawan2014robust} datasets.
The results are reported using accuracy \textbf{Acc.} (higher is better), normalized edit distance \textbf{NED} (higher is better) and total edit distance \textbf{TED} (lower is better).
Relative gains are shown in \textcolor{green}{green} and relative declines in \textcolor{red}{red}.
}
\label{table:ls_tps_resnet_bilstm_attn}
\end{table}

\begin{table}
\begin{center}
\resizebox{\textwidth}{!}{
\begin{tabular}{ l | l | l | l}
\toprule
\textbf{\specialcell{Loss \\ Function}} & \textbf{$\uparrow$ \specialcell{Recall}} & \textbf{$\uparrow$ \specialcell{Precision}} & \textbf{$\uparrow$ \specialcell{$F_{1}$ score}}\\
\midrule
 {\em Smooth-$L_{1}$}  & $71.21\%$ & $84.71\%$ & $77.37\%$ \\
 LS-IoU (global) & $66.97\%$ \textcolor{red}{$-5.95\%$} & $84.71\%$ \textcolor{black}{$\pm 0.00\%$} & $74.81\%$ \textcolor{red}{$-3.30\%$}\\
 LS-IoU (local) & $70.92\%$ \textcolor{red}{$-0.40\%$} & $86.60\%$ \textcolor{green}{$+2.23\%$} & $77.98\%$ \textcolor{green}{$+0.78\%$}\\
 LS-IoU (local-global) & $76.79\%$ \textcolor{green}{$+7.83\%$} & $84.93\%$ \textcolor{green}{$+0.25\%$} & $80.66\%$ \textcolor{green}{$+4.25\%$}\\
\bottomrule
\end{tabular}}
\end{center}
\caption{RRPN-ResNet-50 \cite{ma2018arbitrary,ma2019rrpn}: Evaluations on Incidental Scene Text ICDAR'15 \cite{karatzas2015icdar}. Relative gains are shown in \textcolor{green}{green} and relative declines in \textcolor{red}{red}.}
\label{table:detection_results}
\end{table}

\subsection{Post-Tuning with a Learned Surrogate for IoU (LS-IoU)}
\label{sec:exp_iou}

It is experimentally demonstrated that LS can optimize scene text detection models on intersection-over-union (IoU) for rotated bounding boxes. IoU is a popular metric used to evaluate the object detection \cite{redmon2016you,ren2015faster} and scene text detection models \cite{ma2018arbitrary,buvsta2018e2e,DBLP:conf/cvpr/LiuLYCQY18,karatzas2015icdar,DBLP:conf/icdar/GomezSGNVMBK17}. Gradients for IoU can be hand-crafted for the case of axis-aligned bounding boxes \cite{yu2016unitbox,rezatofighi2019generalized}, however, it is complex to design the gradients for rotated bounding boxes. The learned surrogate of IoU allows backpropagation for rotated bounding boxes. For the task of rotated scene text detection on ICDAR'15 \cite{karatzas2015icdar}, it is shown that post-tuning the text detection model with LS-IoU leads to improvement on recall, precision, and $F_{1}$ score.

\subsubsection{Scene Text Detection.} 
Given a natural scene image, the objective is to obtain precise word-level rotated bounding boxes. The method proposed by Ma~\etal~\cite{ma2018arbitrary} is used for the task. It extends Faster-RCNN \cite{ren2015faster} based object detector to incorporate rotations. This is achieved by adding angle priors in anchor boxes to enable rotated region proposals. A sampling strategy using IoU compares these proposals with the ground truth and filter the positive and the negative proposals. Only the filtered proposals are used for the loss computation.

The positive proposals are regressed to fit precisely with the ground truth. Through rotated region-of-interest (RROI) pooling, the features corresponding to the proposals are obtained and used for text/no-text binary classification. The overall loss function for training in \cite{ma2018arbitrary} is defined as a linear combination of classification loss (negative log-likelihood) and regression loss ({\em smooth-L$_{1}$}).

The publicly available implementation of \cite{ma2018arbitrary,ma2019rrpn} is used with the original hyper-parameter settings -- the model is trained for $140K$ iterations using the SGD optimizer and batch-size of $1$. The model is trained on a union of ICDAR'15 \cite{karatzas2015icdar} and ICDAR-MLT \cite{nayef2019icdar2019} datasets, providing $6295$ training images.

\subsubsection{LS-IoU architecture.} 
The embedding model for LS-IoU consists of five fully-connected layers with ReLU activation \cite{DBLP:journals/jmlr/GlorotBB11}. A rotated bounding box is represented with six parameters, two for the coordinates of the centre of the box, two for the height and the width and two for {\em cosine} and {\em sine} of the rotation angle. The centre coordinates and the dimensions of the box are normalized with image dimensions to make the representation invariant to the image resolution.

The embedding model maps the representation of a positive box proposal and the matching ground-truth into a vector as $h_{\Phi}: \mathbb{R}^{6}\xrightarrow{} \mathbb{R}^{16}$. The approximation of the IoU between two bounding boxes is computed by the $L_{2}$ distance between the two vector representations.

\subsubsection{Post-tuning with LS-IoU.} 
The random generator for LS-IoU samples rotated bounding boxes from the set of training labels and modifies the boxes by changing the centre locations, dimensions, and rotation angle within certain bounds to create a distorted variant. Since uniform sampling over the range of IoU is difficult, we store roughly $3$ million such examples along with the IoU values and sample from this collection.

Note that since the overall loss for training \cite{ma2018arbitrary} is a combination of a regression loss and a classification loss, LS-IoU only replaces the regression component ({\em smooth-$L_{1}$}) with the learned surrogate for IoU. For post-tuning with LS-IoU, the results are shown for all three setups, that is, global approximation, local approximation and global-local approximation (Algorithm \ref{alg:train_with_lsl}). For each of these, the model trained with proxy losses is post-tuned with LS-IoU for $20K$ iterations. The quality of the approximations for the first $1K$ iterations of the training is shown in Figure \ref{fig:ls_iou_plots}. Since the range of IoU is in $[0,1]$, it can be seen that the error is high for the global approximation. For both local and global-local, the quality of the approximation is significantly better (roughly $10$ times lower error).

As mentioned earlier, the quality of gradients is judged by the improvement or deterioration of the model ($f_{\Theta}(x)$) post-tuned with LS-IoU. The results for scene text detection on the ICDAR'15 \cite{karatzas2015icdar} dataset are shown in Table \ref{table:detection_results}. It is observed that post-tuning the detection model with LS-IoU (global) leads to deterioration. Post-tuning with LS-IoU (local) improves the precision but makes recall worse. Finally, LS-IoU (local-global) from Algorithm \ref{alg:train_with_lsl} improves both the precision and recall, boosting the $F_{1}$ score by relative $4.25\%$.

\section{Conclusions}
\label{sec:ls_conclusion}

A technique is proposed for training neural networks by minimizing learned surrogates that approximate the target evaluation metric. The effectiveness of the proposed technique has been demonstrated in a post-tuning setup, where a trained model is tuned on the learned surrogate. Improvements have been achieved on the challenging tasks of scene-text recognition and detection. By post-tuning, the model with LS-ED, relative improvements of up to $39\%$ on the total edit distance has been achieved. On detection, post-tuning with LS-IoU has shown to provide a relative gain of $4.25\%$ on the $F_{1}$ score.
\chapter{FEDS - Filtered Edit Distance Surrogate}
\label{chapter:feds}

\begin{figure}[H]
    \centering
    \includegraphics[width=\textwidth]{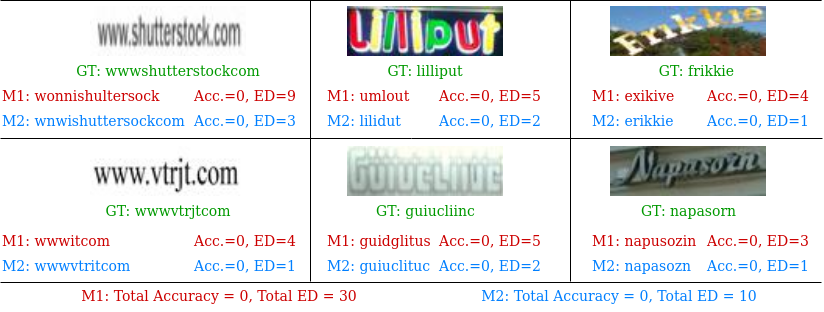}
    \caption{Accuracy and edit distance comparison for different predictions of scene text recognition (STR). For the scene text images, \textcolor{green}{green} shows the ground truth, \textcolor{red}{red} shows the prediction from a STR model \textcolor{red}{M1} and \textcolor{blue}{blue} shows the predictions from another STR model \textcolor{blue}{M2}. For these examples accuracy ranks both the models equally, however, it can be clearly seen that for the predictions in blue vocabulary search or Google search will succeed.
    }
    \label{fig:acc_vs_ed}
\end{figure}

Deep neural networks are trained by back-propagating gradients \cite{rumelhart1986learning}, which requires the loss function to be differentiable. However, the task-specific objective is often defined via an evaluation metric, which may not be differentiable. The evaluation metric's design is to fulfill the application requirements, and for the cases where the evaluation metric is differentiable, it is directly used as a loss function. For scene text recognition (STR), accuracy and edit distance are popular evaluation metric choices. Accuracy rewards the method if the prediction exactly matches the ground truth. Whereas edit distance (ED) is defined by counting addition, subtraction, and substitution operations, required to transform one string into another. As shown in Figure \ref{fig:acc_vs_ed}, accuracy does not account for partial correctness. Note that the low {\em ED} errors from M2 can be easily corrected by a dictionary search in a word-spotting setup \cite{patel2016dynamic}. Therefore, edit distance is a better metric, especially when the state-of-the-art is saturated on the benchmark datasets \cite{karatzas2013icdar,karatzas2015icdar,nayef2019icdar2019}.

\begin{figure}
    \centering
    \includegraphics[width=\textwidth]{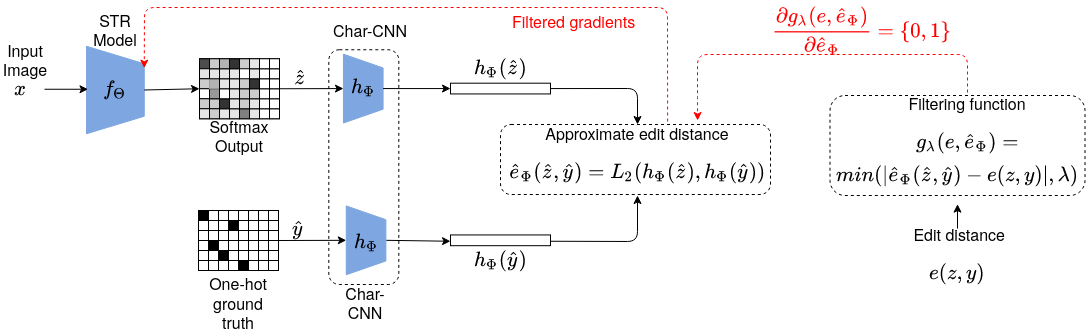}
    \caption{Overview of the proposed post-tuning procedure. $x$ is the input to the STR model $f_{\Theta}(x)$ with output $\hat{z}$. $y$ is the ground truth, $\hat{y}$ is the ground truth expressed as one-hot, $e(z,y)$ is the evaluation metric, $\hat{e}_{\Phi}(\hat{z}, \hat{y})$ is the learned surrogate and $g_{\lambda}(e, \hat{e}_{\Phi})$ is the filtering function. The approximations from the learned surrogate are checked against the edit distance by the filtering function. The STR model is not trained on the samples where the surrogate is incorrect.}
    \label{fig:feds_overall}
\end{figure}

When the evaluation metric is non-differentiable, a proxy loss is employed, which may not align well with the evaluation metric. Edit distance is computed via dynamic programming and is non-differentiable. Therefore, it can not be used as a loss function for training deep neural networks. The proxy loss used for training STR models is per-character cross-entropy or Connectionist Temporal Classification (CTC) \cite{graves2006connectionist}. The models trained with cross-entropy or CTC may have a sub-optimal performance on edit distance as they optimize a different objective.

The aforementioned issue can be addressed by learning a surrogate, e.g. \cite{patel2020learning}, where a model trained with the proxy loss is post-tuned on a learned surrogate of the evaluation metric. In \cite{patel2020learning}, post-tuning has shown significant improvement in performance on the evaluation metric. While Patel ~\etal \cite{patel2020learning} have paid attention to learning the surrogate, none was given to robustly train the neural network with the surrogate. In the training procedure, \cite{patel2020learning} assumes that the learned surrogate robustly estimates the edit distance for all samples. Since the surrogate is learned via supervised training, it is prone to overfitting on the training distribution and may fail on out-of-the-distribution samples. In hope for better generalizability of the surrogate, \cite{patel2020learning} makes use of a data generator to train the surrogate, which requires extra engineering effort. This chapter shows that the learned edit distance surrogate often fails, leading to noisy training.

As an improvement, this chapter proposes \textbf{F}iltered \textbf{E}dit \textbf{D}istance \textbf{S}urrogate. In FEDS, the STR model is trained only on the samples where the surrogate approximates the edit distance within a small error bound. This is achieved by computing the edit distance for a training sample and comparing it with the approximation from the surrogate. The comparison is realized by a ramp-function, which is piece-wise differentiable, allowing for end-to-end training. Figure \ref{fig:feds_overall} provides an overview of the proposed method. The proposed training method simplifies the training and eliminates the need for a data generator to learn a surrogate.

The rest of the chapter is structured as follows. Related work is reviewed in Section~\ref{sec:feds_related_work}, the technique for robustly training with the learned surrogate of ED is presented in Section~\ref{sec:feds}, experiments are shown in Section~\ref{sec:feds_experiments} and the chapter is concluded in Section~\ref{sec:feds_conclusions}.

\section{Related Work}
\label{sec:feds_related_work}

Scene text recognition (STR) is the task of recognizing text from images against complex backgrounds and layouts. STR is an active research area; comprehensive surveys can be found in \cite{ye2014text,baek2019wrong,long2020scene}. Before deep learning, STR methods focused on recognizing characters via sliding window, and hand-crafted features \cite{wang2011end,wang2010word,yao2014strokelets}. Deep learning based STR methods have made a significant stride in improving model architectures that can handle both regular (axis-aligned text) and irregular text (complex layout, such as perspective and curved text). Selected relevant methods are discussed subsequently.

\paragraph{Convolutional models for STR.} Among the first deep learning STR methods was the work of Jaderberg ~\etal \cite{jaderberg2014deep}, where a character-centric CNN \cite{lecun1998gradient} predicts a text/no-text score, a character, and a bi-gram class. Later this work was extended to word-level recognition \cite{jaderberg2016reading} where the CNN takes a fixed dimension input of the cropped word and outputs a word from a fixed dictionary. Bušta ~\etal \cite{buvsta2018e2e,busta2017deep} proposed a fully-convolutional STR model, which operates on variable-sized inputs using bi-linear sampling \cite{jaderberg2015spatial}. The model is trained jointly with a detector in a multi-task learning setup using CTC \cite{graves2006connectionist} loss. Gomez ~\etal \cite{gomez2017lsde} trains an embedding model for word-spotting, such that, the euclidean distance between the representations of two images corresponds to the edit-distance between their text strings. This embedding model differs from FEDS as it operates on images instead of STR model's predictions and is not used to train a STR model.

\paragraph{Recurrent models for STR.} Shi ~\etal \cite{shi2016end} and He ~\etal \cite{he2016reading} were among the first to propose end-to-end trainable, sequence-to-sequence models \cite{sutskeverSqeuence} for STR. An image of a cropped word is seen as a sequence of varying length, where convolutional layers are used to extract features and recurrent layers to predict a label distribution. Shi ~\etal \cite{shi2016robust} later combined the CNN-RNN hybrid with spatial transformer network \cite{jaderberg2015spatial} for better generalizability on irregular text. In \cite{shi2018aster}, Shi ~\etal adapted Thin-Plate-Spline \cite{bookstein1989principal} for STR, leading to an improved performance on both regular and irregular text (compared to \cite{shi2016robust}). While \cite{shi2018aster,shi2016robust} rectify the entire text image, Liu ~\etal \cite{liu2018char} detects and rectifies each character. This is achieved via a recurrent RoIWarp layer, which sequentially attends to a region of the feature map that corresponds to a character. Li ~\etal \cite{li2019show} passed the visual features through an attention module before decoding via an LSTM. MaskTextSpotter \cite{liao2019mask} solves detection and recognition jointly; the STR module consists of two branches while the first uses local visual features, the second utilizes contextual information in the form of attention. Litman ~\etal \cite{litman2020scatter} utilizes a stacked block architecture with intermediate supervision during training, which improves the encoding of contextual dependencies, thus improving the performance on the irregular text.

\paragraph{Training data.} Annotating scene text data in real images is complex and expensive. As an alternative, STR methods often use synthetically generated data for training. Jaderberg ~\etal \cite{jaderberg2014deep} generated $8.9$ million images by rendering fonts, coloring the image layers, applying random perspective distortion, and blending it to a background. Gupta ~\etal \cite{gupta2016synthetic} placed rendered text on natural scene images; this is achieved by identifying plausible text regions using depth and segmentation information. Patel ~\etal \cite{patel12018e2e} further extended this to multi-lingual text. The dataset of \cite{gupta2016synthetic} was proposed for training scene text detection; however, it is also useful for improving STR models \cite{baek2019wrong}. Long ~\etal \cite{long2020unrealtext} used a 3D graphics engine to generate scene text data. The 3D synthetic engine allows for better text region proposals as scene information such as normal and objects meshes are available. Their analysis shows that compared to \cite{gupta2016synthetic}, more realistic looking diverse images (contains shadow, illumination variations, etc.) are more useful for STR models. As an alternative to synthetically generate data, Janouskova ~\etal \cite{klara} leverages weakly annotated images to generate pseudo scene text labels. The approach uses an end-to-end scene text model to generate initial labels, followed by a heuristic neighborhood search to match imprecise transcriptions with weak annotations.

As discussed, significant work has been done towards improving the model architectures \cite{litman2020scatter,baek2019wrong,shi2016end,shi2016robust,shi2018aster,jaderberg2015spatial,jaderberg2016reading,busta2017deep,buvsta2018e2e,liao2019scene,zhan2019esir,yang2019symmetry,wang2020decoupled,yu2020towards,qiao2020seed,yue2020robustscanner} and obtaining data for training \cite{klara,jaderberg2016reading,gupta2016synthetic,long2020unrealtext,gomez2019selective}.

Limited attention has been paid to the loss function. Most deep learning based STR methods rely on per-character cross-entropy or CTC loss functions \cite{graves2006connectionist,baek2019wrong}. While in theory and under an assumption of infinite training data, these loss functions align with accuracy \cite{lapin2016loss}, there is no concrete evidence of their alignment with edit-distance. In comparison to the related work, this chapter makes an orthogonal contribution, building upon learning surrogates \cite{patel2020learning}, this chapter proposes a robust training procedure for better optimization of STR models on edit distance.

\section{FEDS: Filtered Edit Distance Surrogate }
\label{sec:feds}

\subsection{Background}
\label{sec:backgound}

The samples for training the scene text recognition (STR) model are drawn from a distribution $(x,y) \sim U_{D}$. Here, $x$ is the image of a cropped word, and $y$ is the corresponding transcription. An end-to-end trainable deep model for STR, denoted by $f_{\Theta}(x)$ predicts a soft-max output $\hat{z}=f_{\Theta}(x)$, $f_{\Theta}:\mathbb{R}^{W\times H \times 1}\xrightarrow{}\mathbb{R}^{|A|\times L}$. Here $W$ and $H$ are the dimensions of the input image, $A$ is the set of characters, and $L$ is the maximum possible length of the word.

For training, the ground truth $y$ is converted to one-hot representation $\hat{y}^{|A| \times L}$. Cross entropy (CE) is a popular choice of the loss function \cite{baek2019wrong}, which provides the loss for each character:
\begin{equation}
    \text{{\em CE}}(\hat{z}, \hat{y}) = - \frac{1}{L|A|} \sum_{i=1}^{L} \sum_{j=1}^{|A|}\hat{y}_{i,j} log(\hat{z}_{i,j})
\end{equation}

Patel ~\etal \cite{patel2020learning} learns the surrogate of edit distance via a learned deep embedding $h_{\Phi}$, where the Euclidean distance between the prediction and the ground truth corresponds to the value of the edit distance, which provides the edit distance surrogate, denoted by $\hat{e}_{\Phi}$:
\begin{equation}
    \hat{e}_{\Phi}(\hat{z}, \hat{y}) = \left\Vert h_{\Phi}(\hat{z}) - h_{\Phi}(\hat{y}) \right\Vert_{2}
    \label{eq:defination_ehat}
\end{equation}

where $h_{\Phi}$ is the Char-CNN \cite{zhang2015character,patel2020learning} with parameters $\Phi$. Note that the edit distance surrogate is defined on the one-hot representation of the ground truth and the soft-max prediction from the STR model.

\subsection{Learning edit distance surrogate}
\label{sec:lsed}

\subsubsection{Objective.}
To fairly demonstrate the improvements using the proposed FEDS, the loss for learning the surrogate is the same as LS-ED \cite{patel2020learning}:
\begin{enumerate}
    \item The learned edit distance surrogate should correspond to the value of the edit distance: \begin{equation}
        \hat{e}_{\Phi}(\hat{z}, \hat{y}) \approx e(z,y)
    \end{equation}
    where $e(z,y)$ is the edit distance defined on the string representation of the prediction and the ground truth.
    \item The first order derivative of the learned edit distance surrogate with respect to the STR model prediction $\hat{z}$ is close to $1$: \begin{equation}
       \left\Vert \frac{\partial \hat{e}_{\Phi}(\hat{z},\hat{y})}{\partial \hat{z}} \right\Vert_{2} \approx 1
       \label{eq:feds_gp_term}
    \end{equation}
\end{enumerate}

Bounding the gradients (Equation~\ref{eq:feds_gp_term}) has shown to enhance the training stability for Generative Adversarial Networks~\cite{gulrajani2017improved} and has shown to be useful for learning the surrogate \cite{patel2020learning}.

Both objectives are realized and linearly combined in the training loss:
\begin{equation}
    \text{loss}(\hat{z},\hat{y}) = w_{1}\left\Vert \big(\hat{e}_{\Phi}(\hat{z},\hat{y}) - e(z,y)\right\Vert_2^2 + w_{2} \left(\left\Vert \frac{\partial \hat{e}_{\Phi}(\hat{z},\hat{y})}{\partial \hat{z}} \right\Vert_{2} - 1\right)^{2}
    \label{eq:feds_lsl_loss}
\end{equation}

\subsubsection{Training data.}
Patel ~\etal \cite{patel2020learning} uses two sources of data for learning the surrogate - the pre-trained STR model and a random generator. The random generator provides a pair of words and their edit distance and ensures uniform sampling in the range of the edit distance. The random generator helps the surrogate to generalize better, leading to an improvement in the final performance of the STR model.

The proposed FEDS does not make use of a random generator, reducing the effort and the computational cost. FEDS learns the edit distance surrogate only on the samples obtained from the STR model:
\begin{equation}
    (\hat{z},\hat{y}) \sim f_{\Theta}(x) \; | \; (x, y) \sim U_{D}
\end{equation}

\subsection{Robust Training}
\label{sec:robust_training}

The filtering function $g_{\lambda}$ is defined on the surrogate and the edit distance, parameterized by a scalar $\lambda$ that acts as a threshold to determine the quality of the approximation from the surrogate. The filtering function is defined as:

\begin{equation}
    g_{\lambda}(e(z,y),\hat{e}_{\Phi}(\hat{z}, \hat{y})) = \min(|\hat{e}_{\Phi}(\hat{z}, \hat{y}) - e(z,y)|, \lambda) \; | \; \lambda > 0
    \label{eq:filtering_fuc}
\end{equation}

\begin{figure}
    \centering
    \includegraphics[width=0.6\textwidth]{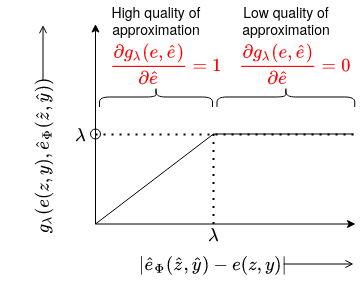}
    \caption{The filtering function enforces zero gradients for the samples that are hard for the surrogate (low quality of approximation). STR model is trained only on the samples where the quality of the approximation from the edit distance surrogate is high.}
    \label{fig:g_lambda}
\end{figure}

The filtering function is piece-wise differentiable, as can be seen in Figure \ref{fig:g_lambda}. For the samples where the quality of approximation from the surrogate is low, the gradients are zero, and the STR model is not trained on those samples. Whereas for samples where the quality of the approximation is within the bound of $\lambda$, the STR model is trained to minimize the edit distance surrogate.

Learning of the ED surrogate $\hat{e}_{\Phi}$ and post-tuning of the STR model $f_{\Theta}(x)$ are conducted alternatively. The surrogate is learned first for $I_{a}$ number of iterations while the STR model is fixed. Subsequently, the STR model is trained using the surrogate and the filtering function, while the ED surrogate parameters are kept fixed. Algorithm \ref{alg:feds} and Figure \ref{fig:feds_overall} demonstrate the overall training procedure with FEDS.

\begin{algorithm}[t!]
\caption{Post-tuning with FEDS}
\label{alg:feds}
\textbf{Inputs}: Supervised data $D$, evaluation metric $e$.\\
\textbf{Hyper-parameters}: Number of update steps $I_{a}$ and $I_{b}$, learning rates $\eta_{a}$ and $\eta_{b}$, number of epochs $E$.\\
\textbf{Objective}: Robustly post-tune the STR model, {\em i.e.}, $f_{\Theta}(x)$ and learn the edit distance surrogate, \emph{i.e.}, $\hat{e}_{\Phi}$.
\begin{algorithmic}[1]
\State \textit{Initialize} $\Theta \leftarrow$ pre-trained weights, $\Phi \leftarrow$ random weights.
\For{epoch = 1,...,E}
    \For{i = 1,...,$I_{a}$}
        \State sample, $(x,y) \sim U_{D}$
        \State inference, $\hat{z}=f_{\Theta^{epoch-1}}(x)$
        \State compute loss, $l_{\hat{e}} =$ {\em loss}$(\hat{z}, \hat{y})$ (Equation \ref{eq:feds_lsl_loss})
        \State update ED surrogate, $\Phi^{i} \leftarrow \Phi^{i-1} - \eta_{a} \frac{\partial l_{\hat{e}}}{\partial \Phi^{i-1}}$
    \EndFor \\
   \quad \quad $\Phi \leftarrow \Phi^{I_{a}}$
    \For{i = 1,...,$I_{b}$}
        \State sample, $(x,y) \sim U_{D}$
        \State inference, $\hat{z} = f_{\Theta^{i-1}}(x)$
        \State compute ED from the surrogate, $\hat{e} = \hat{e}_{\Phi^{epoch}}(\hat{z}, \hat{y})$ (Equation \ref{eq:defination_ehat})
        \State compute ED, $e = e(z, y)$
        \State computer loss, $l_{f} = g_{\lambda}(e, \hat{e})$ (Equation \ref{eq:filtering_fuc})
        \State update STR model, $\Theta^{i} \leftarrow \Theta^{i-1} - \eta_{b} \frac{\partial(l_{f})}{\partial \Theta^{i-1}}$
    \EndFor \\
    \quad \quad $\Theta \leftarrow \Theta^{I_{b}}$
\EndFor
\end{algorithmic}
\end{algorithm}

\section{Experiments}
\label{sec:feds_experiments}

\subsection{FEDS model}
\label{sec:feds_architecture}

The model for learning the deep embedding, {\em i.e.}, $h_{\Phi}$ is kept same as \cite{patel2020learning}. A Char-CNN architecture~\cite{zhang2015character} is used with five $1D$ convolution layers, LeakyReLU \cite{DBLP:booktitles/corr/XuWCL15} and two {\em FC} layers. The embedding model, $h_{\Phi}$, maps the input to a $1024$ dimensions, $h_{\Phi}: \mathbb{R}^{|A|\times L} \xrightarrow{} \mathbb{R}^{1024}$. Feed forward (Equation \ref{eq:defination_ehat}), generates embeddings for the ground-truth $\hat{y}$ (one-hot) and model prediction $\hat{z}$ (soft-max) and an approximation of edit distance is computed by $L_{2}$ distance between the two embedding.

\subsection{Scene Text Recognition model}
Following the survey on STR, \cite{baek2019wrong}, the state-of-the-art model ASTER is used \cite{shi2018aster}, which contains four modules: (a) transformation, (b) feature extraction, (c) sequence modeling, and (d) prediction. Baek~\etal~\cite{baek2019wrong} provides a detailed analysis of STR models and the impact of different modules on the performance.

\subsubsection{Transformation.} Operates on the input image and rectifies the curved or tilted text, easing the recognition for the subsequent modules. The two popular variants include Spatial Transformer \cite{jaderberg2015spatial} and Thin Plain Spline (TPS) \cite{shi2018aster}. TPS employs a smooth spline interpolation between a set of fiducial points, which are fixed in number. Following the analysis of Shi ~\etal \cite{shi2018aster} \cite{baek2019wrong}, the STR model used employs TPS.

\subsubsection{Feature extraction.} Involves a Convolutional Neural Network \cite{lecun1998gradient}, that extracts the features from the image transformed by TPS. Popular choices include VGG-16 \cite{simonyan2014very} and ResNet \cite{he2016deep}. Follwoing \cite{baek2019wrong}, the STR model used employs ResNet for the ease of optimization and good performance. 

\subsubsection{Sequence modeling.} Captures the contextual information within a sequence of characters; this module operates on the features extracted from a ResNet. The STR model used employs BiLSTM \cite{hochreiter1997long}.

\subsubsection{Prediction.} The predictions are made based on the identified features of the image. The prediction module depends on the loss function used for training. CTC loss requires the prediction to by sigmoid, whereas cross-entropy requires the prediction to be a soft-max distribution over the set of characters. The design of FEDS architecture (Section \ref{sec:feds_architecture}) requires a soft-max distribution.

FEDS and LS-ED \cite{patel2020learning} are investigated with the state-of-the-art performing configuration of the STR model, which is \textit{TPS-ResNet-BiLSTM-Attn}.

\subsection{Training and Testing data}

The STR models are trained on synthetic and pseudo labeled data and are evaluated on real-world benchmarks. Note that the STR models are not fine-tuned on evaluation datasets (same as \cite{baek2019wrong}).
\subsubsection{Training data.} The experiments make use of the following synthetic and pseudo labeled data for training:
\begin{itemize}
    \item \textbf{MJSynth} \cite{jaderberg2014deep} (synthetic). $8.9$ million synthetically generated images, obtained by rendering fonts, coloring the image layers, applying random perspective distortion, and blending it to a background.
    \item \textbf{SynthText} \cite{gupta2016synthetic} (synthetic). $5.5$ million text instance by placing rendered text on natural scene images. This is achieved by identifying plausible text regions using depth and segmentation information.
    \item \textbf{Uber-Text} \cite{klara} (pseudo labels). $138$K real images from Uber-Text \cite{zhang2017uber} with pseudo labels obtained using \cite{klara}.
    \item \textbf{Amazon book covers} \cite{klara} (pseudo labels). $1.5$ million real images from amazon book covers with pseudo labels obtained using \cite{klara}.
\end{itemize}

\subsubsection{Testing data.} The models trained purely on the synthetic and pseudo labelled datasets are tested on a collection of real datasets. This includes regular scene text - IIIT-5K \cite{mishra2012scene}, SVT \cite{wang2011end}, ICDAR'03 \cite{lucas2003icdar} and ICDAR'13 \cite{karatzas2013icdar}, and irregular scene text ICDAR'15 \cite{karatzas2015icdar}, SVTP \cite{quy2013recognizing}  and CUTE \cite{risnumawan2014robust}. 

\subsection{Implementation details}
\label{sec:implementation_details}

The analysis of the proposed FEDS and LS-ED \cite{patel2020learning} is conducted for two setups of training data. First, similar to \cite{baek2019wrong}, the STR models are trained on the union of the synthetic data obtained from MJSynth \cite{jaderberg2014deep}, and SynthText \cite{gupta2016synthetic} resulting in a total of $14.4$ million training examples. Second, additional pseudo labeled data \cite{klara} is used to obtain a stronger baseline.

The STR models are first trained with the proxy loss, \emph{i.e.}, cross-entropy for $300K$ iterations with a mini-batch size of $192$. The models are optimized using ADADELTA \cite{DBLP:booktitles/corr/abs-1212-5701}. Once the training is complete, these models are tuned with FEDS (Algorithm \ref{alg:feds}) on the same training set for another $20K$ iterations. For learning the edit distance surrogate the weights in the loss (Equation \ref{eq:feds_lsl_loss}) are set as $w_{1}=1, w_{2}=0.1$. Note that the edit distance value is a non-negative integer, therefore, optimal range for $\lambda$ is $(0,0.5)$. Small value of $\lambda$ filters out substantial number of samples, slowing down the training, whereas, large values of $\lambda$ allows a noisy training. Therefore, the threshold for the filtering function (Equation \ref{eq:filtering_fuc}) is set as $\lambda=0.25$, {\em i.e.}, in the middle of the optimal range.

\subsection{Quality of the edit distance surrogate}
\label{sec:quality_of_ed}

\begin{figure}[H]
    \centering
    \includegraphics[width=\textwidth]{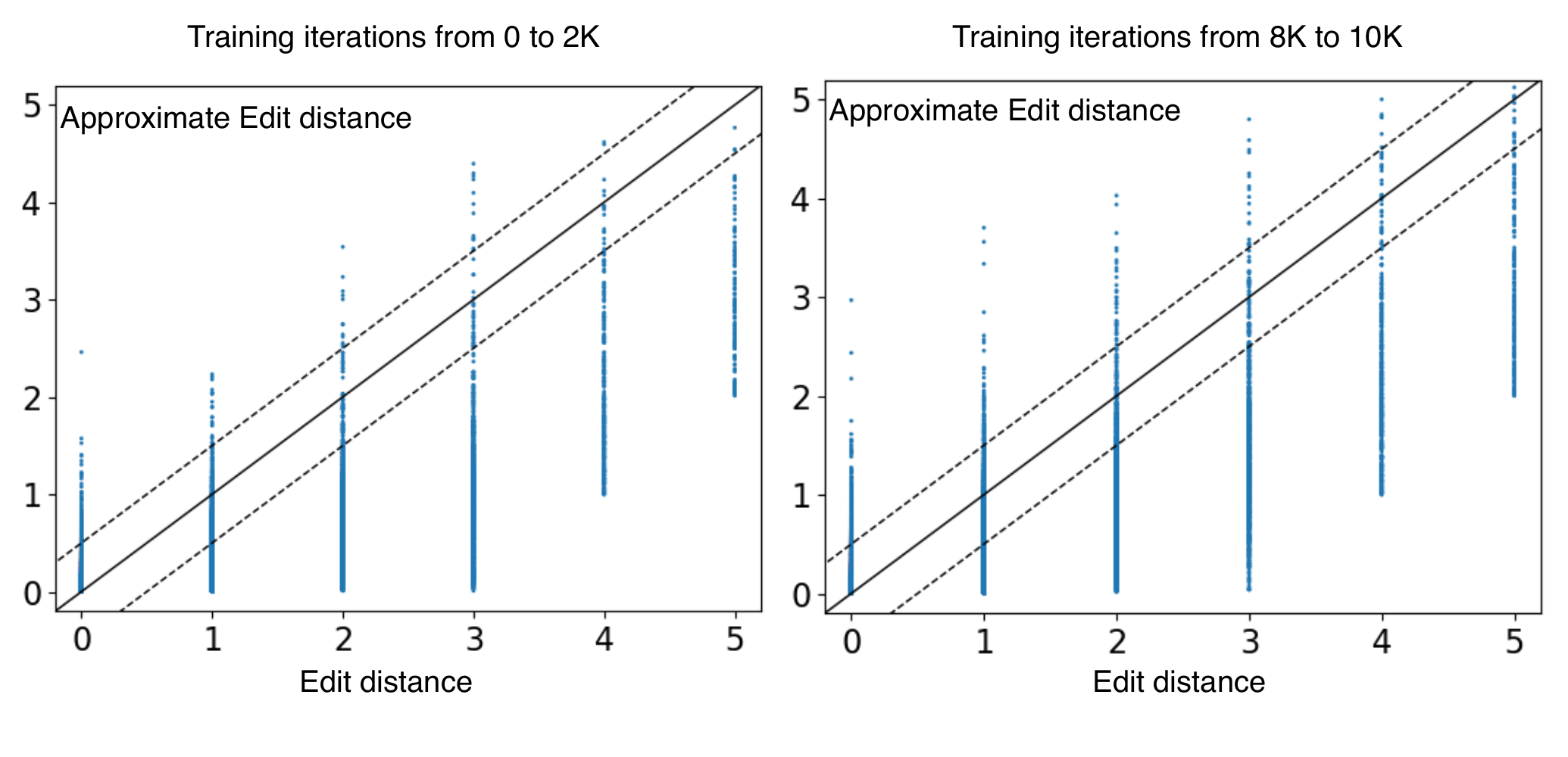}
    \caption{A comparison between the true edit distance and the approximated edit distance is shown. Each point represents a training sample for the STR model. The solid line represents an accurate approximation of the edit distance. The dotted lines represent the filtering in FEDS. \textbf{Left}: Plot for the first 2K iterations of the STR model training. \textbf{Right}: Plot for iterations from 8K to 10K of the STR model training.}
    \label{fig:e_vs_ehat}
\end{figure}

Figure \ref{fig:e_vs_ehat} shows a comparison between the edit distance and the approximation from the surrogate. As the training progresses, the approximation improves, {\em i.e.}, more samples are closer to the solid line. The dotted lines represent the filtering in FEDS, {\em i.e.}, only the samples between the dotted lines contribute to the training of the STR model. Note that the surrogate fails for a large fraction of samples; therefore, the training without the filtering (as done in LS-ED \cite{patel2020learning}) is noisy.

\subsection{Quantitative results}
\label{sec:quant_results}

Table \ref{table:feds_tps_resnet_bilstm_attn} shows the results with LS-ED \cite{patel2020learning} and the proposed FEDS in compression with the standard baseline \cite{baek2019wrong,shi2018aster}. For the training, only the synthetic datasets \cite{jaderberg2014deep,gupta2016synthetic} are used. Both LS-ED \cite{patel2020learning} and FEDS improve the performance on all evaluation metrics. Most significant gains are observed on total edit distance as the surrogate approximates it. In comparison with LS-ED, significant gains are observed with the proposed FEDS. On average, FEDS provides an improvement of $11.2\%$ on the total edit distance and $0.98\%$ on accuracy (an equivalent of $9.5\%$ error reduction).

\begin{table}[t!]
\caption{STR model trained with MJSynth \cite{jaderberg2014deep} and SynthText \cite{gupta2016synthetic}. Evaluation on IIIT-5K \cite{mishra2012scene}, SVT \cite{wang2011end}, IC'03 \cite{lucas2003icdar}, IC'13 \cite{karatzas2013icdar}, IC'15 \cite{karatzas2015icdar}, SVTP \cite{quy2013recognizing}  and CUTE \cite{risnumawan2014robust}. The results are reported using accuracy \textbf{Acc.} (higher is better), normalized edit distance \textbf{NED} (higher is better) and total edit distance \textbf{TED} (lower is better). Relative gains are shown in \textcolor{blue}{blue} and relative declines in \textcolor{red}{red}.}
\begin{center}
\resizebox{0.95\textwidth}{!}{
\begin{tabular}{ c | l | l | l | l}
\toprule
\multicolumn{1}{c|}{\textbf{\specialcell{Test \\ Data}}} 
& \multicolumn{1}{c|}{\textbf{\specialcell{Loss \\ Function}}} 
& \multicolumn{1}{c|}{\textbf{$\uparrow$ \specialcell{Acc.}}} 
& \multicolumn{1}{c|}{\textbf{$\uparrow$ \specialcell{NED}}} 
& \multicolumn{1}{c}{\textbf{$\downarrow$ \specialcell{TED}}}\\
\midrule
\multirow{3}{*}{IIIT-5K (3000)}  
& Cross-Entropy \cite{baek2019wrong} 
& $87.1$ 
& $0.959$ 
& $772$ \\

  & LS-ED \cite{patel2020learning} 
  & $88.0$ \textcolor{blue}{$+1.03\%$} 
  & $0.962$ \textcolor{blue}{$+0.31\%$} 
  & $680$ \textcolor{blue}{$+11.9\%$} \\
  
  & FEDS  
  & $88.8$ \textcolor{blue}{$+1.95\%$} 
  & $0.966$ \textcolor{blue}{$+0.72\%$}
  & $591$ \textcolor{blue}{$+23.44\%$} \\
\midrule

\multirow{3}{*}{SVT (647)}  
& Cross-Entropy \cite{baek2019wrong} 
& $87.2$ 
& $0.953$ 
& $175$ \\

 & LS-ED \cite{patel2020learning} 
 & $87.3$ \textcolor{blue}{$+0.11\%$} 
 & $0.954$ \textcolor{blue}{$+0.10\%$} 
 & $161$ \textcolor{blue}{$+8.00\%$} \\

 & FEDS  
 & $88.7$ \textcolor{blue}{$+1.72\%$} 
 & $0.957$ \textcolor{blue}{$+0.41\%$} 
 & $147$ \textcolor{blue}{$+16.0\%$} \\
 
\midrule
\multirow{3}{*}{IC'03 (860)} 
& Cross-Entropy \cite{baek2019wrong} 
& $95.1$ 
& $0.981$ 
& $105$  \\

 & LS-ED \cite{patel2020learning} 
 & $95.3$ \textcolor{blue}{$+0.21\%$} 
 & $0.982$ \textcolor{blue}{$+0.10\%$} 
 & $89$ \textcolor{blue}{$+15.2\%$} \\

 & FEDS  
 & $95.4$ \textcolor{blue}{$+0.31\%$} 
 & $0.983$ \textcolor{blue}{$+0.20\%$} 
 & $87$ \textcolor{blue}{$+17.1\%$} \\
 
\midrule
\multirow{3}{*}{IC'03 (867)} 
& Cross-Entropy \cite{baek2019wrong} 
& $95.1$ 
& $0.982$ 
& $102$ \\

 & LS-ED  \cite{patel2020learning} 
 & $95.2$ \textcolor{blue}{$+0.10\%$} 
 & $0.983$ \textcolor{blue}{$+0.10\%$} 
 & $90$ \textcolor{blue}{$+11.7\%$} \\

 & FEDS  
 & $95.0$ \textcolor{red}{$-0.10\%$} 
 & $0.981$ \textcolor{red}{$-0.10\%$} 
 & $81$ \textcolor{blue}{$+20.5\%$} \\
\midrule

\multirow{3}{*}{IC'13 (857)} 
& Cross-Entropy \cite{baek2019wrong} 
& $92.9$ 
& $0.979$ 
& $110$ \\

 & LS-ED \cite{patel2020learning} 
 & $93.9$ \textcolor{blue}{$+1.07\%$} 
 & $0.981$ \textcolor{blue}{$+0.20\%$} 
 & $97$ \textcolor{blue}{$+11.8\%$} \\

 & FEDS 
 & $93.8$ \textcolor{blue}{$+0.96\%$} 
 & $0.985$ \textcolor{blue}{$+0.61\%$} 
 & $99$ \textcolor{blue}{$+10.0\%$} \\

\midrule
\multirow{3}{*}{IC'13 (1015)} 
& Cross-Entropy \cite{baek2019wrong} 
& $92.2$ 
& $0.966$ 
& $140$ \\

 & LS-ED \cite{patel2020learning} 
 & $93.1$ \textcolor{blue}{$+0.97\%$} 
 & $0.969$ \textcolor{blue}{$+0.31\%$} 
 & $123$ \textcolor{blue}{$+12.1\%$} \\
 
 & FEDS 
 & $92.6$ \textcolor{blue}{$+0.43\%$} 
 & $0.969$ \textcolor{blue}{$+0.31\%$} 
 & $118$ \textcolor{blue}{$+15.7\%$} \\

\midrule
\multirow{3}{*}{IC'15 (1811)} 
& Cross-Entropy \cite{baek2019wrong} 
& $77.9$ 
& $0.915$ 
& $880$ \\

 & LS-ED \cite{patel2020learning} 
 & $78.2$ \textcolor{blue}{$+0.38\%$} 
 & $0.915$ - 
 & $851$ \textcolor{blue}{$+3.29\%$}\\

 & FEDS  
 & $78.5$ \textcolor{blue}{$+0.77\%$} 
 & $0.919$ \textcolor{blue}{$+0.43\%$} 
 & $820$ \textcolor{blue}{$+6.81\%$}\\

\midrule
\multirow{3}{*}{IC'15 (2077)} 
& Cross-Entropy \cite{baek2019wrong} 
& $75.0$ 
& $0.884$ 
& $1234$ \\

 & LS-ED \cite{patel2020learning} 
 & $75.3$ \textcolor{blue}{$+0.39\%$} 
 & $0.883$ \textcolor{red}{$-0.11\%$} 
 & $1210$ \textcolor{blue}{$+1.94\%$} \\
 
 & FEDS  
 & $75.7$ \textcolor{blue}{$+0.93\%$} 
 & $0.888$ \textcolor{blue}{$+0.45\%$} 
 & $1176$ \textcolor{blue}{$+4.70\%$}\\

\midrule
\multirow{3}{*}{SVTP (645)} 
& Cross-Entropy \cite{baek2019wrong} 
& $79.2$
& $0.912$ 
& $340$\\

 & LS-ED  \cite{patel2020learning} 
 & $80.0$ \textcolor{blue}{$+1.01\%$} 
 & $0.915$ \textcolor{blue}{$+0.32\%$} 
 & $327$ \textcolor{blue}{$+3.82\%$}\\

 & FEDS  
 & $80.9$ \textcolor{blue}{$+2.14\%$} 
 & $0.919$ \textcolor{blue}{$+0.76\%$} 
 & $307$ \textcolor{blue}{+$9.70\%$}\\

\midrule
\multirow{3}{*}{CUTE (288)} 
& Cross-Entropy \cite{baek2019wrong} 
& $74.9$
& $0.881$ 
& $221$\\

 & LS-ED \cite{patel2020learning} 
 & $75.6$ \textcolor{blue}{$+0.93\%$} 
 & $0.885$ \textcolor{blue}{$+0.45\%$} 
 & $204$ \textcolor{blue}{$+7.69\%$}\\ 
 
 & FEDS 
 & $75.3$ \textcolor{blue}{$+0.53\%$} 
 & $0.891$ \textcolor{blue}{$+1.13\%$} 
 & $197$ \textcolor{blue}{$+10.8\%$}\\

\midrule
\multirow{3}{*}{TOTAL} 
& Cross-Entropy \cite{baek2019wrong} 
& $85.6$
& $0.941$
& $4079$\\  

 & LS-ED \cite{patel2020learning}
 & $86.1$ \textcolor{blue}{$+0.61\%$} 
 & $0.942$ \textcolor{blue}{$+0.18\%$}
 & $3832$ \textcolor{blue}{$+6.05\%$}\\
 
 & FEDS 
 & $86.5$ \textcolor{blue}{$+0.98\%$}
 & $0.946$ \textcolor{blue}{$+0.48\%$} 
 & $3623$ \textcolor{blue}{$+11.2\%$}\\

\bottomrule
\end{tabular}}
\end{center}
\label{table:feds_tps_resnet_bilstm_attn}
\end{table}

Table \ref{table:feds_tps_resnet_bilstm_attn_pgt} presents the results with LS-ED \cite{patel2020learning} and FEDS in compression with a stronger baseline \cite{klara}. For the training, a combination of synthetic \cite{jaderberg2014deep,gupta2016synthetic} and pseudo labelled \cite{klara} data is used. LS-ED \cite{patel2020learning} provides a limited improvement of $2.91\%$ on total edit distance whereas FEDS provides a significant improvement of $7.90\%$ and an improvement of $1.01\%$ on accuracy (equivalently $7.9\%$ error reduction). Furthermore, LS-ED \cite{patel2020learning} declines the performance on ICDAR'03 \cite{lucas2003icdar} dataset.

\begin{table}[t!]
\caption{STR model trained with MJSynth \cite{jaderberg2014deep}, SynthText \cite{gupta2016synthetic} and pseudo labelled \cite{klara} data. Evaluation on IIIT-5K \cite{mishra2012scene}, SVT \cite{wang2011end}, IC'03 \cite{lucas2003icdar}, IC'13 \cite{karatzas2013icdar}, IC'15 \cite{karatzas2015icdar}, SVTP \cite{quy2013recognizing}  and CUTE \cite{risnumawan2014robust}. The results are reported using accuracy \textbf{Acc.} (higher is better), normalized edit distance \textbf{NED} (higher is better) and total edit distance \textbf{TED} (lower is better). Relative gains are shown in \textcolor{blue}{blue} and relative declines in \textcolor{red}{red}.}
\small
\begin{center}
\resizebox{0.95\textwidth}{!}{
\begin{tabular}{ c | l | l | l | l}
\toprule
\multicolumn{1}{c|}{\textbf{\specialcell{Test \\ Data}}} 
& \multicolumn{1}{c|}{\textbf{\specialcell{Loss \\ Function}}} 
& \multicolumn{1}{c|}{\textbf{$\uparrow$ \specialcell{Acc.}}} 
& \multicolumn{1}{c|}{\textbf{$\uparrow$ \specialcell{NED}}} 
& \multicolumn{1}{c}{\textbf{$\downarrow$ \specialcell{TED}}}\\
\midrule
\multirow{3}{*}{IIIT-5K (3000)}  
& Cross-Entropy \cite{baek2019wrong} 
& $91.7$ 
& $0.973$ 
& $550$\\
  
  & LS-ED \cite{patel2020learning} 
  & $91.8$ \textcolor{blue}{$+0.14\%$} 
  & $0.973$ - 
  & $539$ \textcolor{blue}{$+2.00\%$}\\
  
  & FEDS  
  & $92.2$ \textcolor{blue}{$+0.54\%$} 
  & $0.975$ \textcolor{blue}{$+0.20\%$} 
  & $479$ \textcolor{blue}{$+12.9\%$}\\
\midrule

\multirow{3}{*}{SVT (647)}  
& Cross-Entropy \cite{baek2019wrong} 
& $91.8$ 
& $0.970$ 
& $107$\\

 & LS-ED \cite{patel2020learning} 
 & $91.8$ - 
 & $0.971$ \textcolor{blue}{$+0.10\%$} 
 & $100$ \textcolor{blue}{$+6.54\%$}\\
 
 & FEDS  
 & $92.1$ \textcolor{blue}{$+0.32\%$}
 & $0.971$ \textcolor{blue}{$+0.10\%$} 
 & $102$ \textcolor{blue}{$+4.67\%$}\\
 
\midrule
\multirow{3}{*}{IC'03 (860)} 
& Cross-Entropy \cite{baek2019wrong} 
& $95.6$ 
& $0.984$ 
& $85$\\

 & LS-ED \cite{patel2020learning} 
 & $95.5$ \textcolor{red}{$-0.01\%$} 
 & $0.983$ \textcolor{red}{$-0.10\%$} 
 & $91$ \textcolor{red}{$-7.05\%$}\\
 
 & FEDS  
 & $96.2$ \textcolor{blue}{$+0.62\%$} 
 & $0.986$ \textcolor{blue}{$+0.20\%$} 
 & $73$ \textcolor{blue}{$+14.1\%$}\\
 
\midrule
\multirow{3}{*}{IC'03 (867)} 
& Cross-Entropy \cite{baek2019wrong} 
& $95.7$ 
& $0.984$ 
& $89$\\

 & LS-ED  \cite{patel2020learning} 
 & $95.7$ \textcolor{blue}{$+0.03\%$} 
 & $0.984$ - 
 &  $91$ \textcolor{red}{$-2.24\%$}\\
 
 & FEDS  
 & $96.4$ \textcolor{blue}{$+0.73\%$} 
 & $0.987$ \textcolor{blue}{$+0.30\%$} 
 & $77$ \textcolor{blue}{$+13.4\%$}\\
\midrule

\multirow{3}{*}{IC'13 (857)} 
& Cross-Entropy \cite{baek2019wrong} 
& $95.4$ 
& $0.988$ 
& $65$\\

 & LS-ED \cite{patel2020learning} 
 & $96.3$ \textcolor{blue}{$+1.03\%$} 
 & $0.989$ \textcolor{blue}{$+0.10\%$} 
 &  $55$ \textcolor{blue}{$+15.3\%$}\\
 
 & FEDS 
 & $96.5$ \textcolor{blue}{$+1.15\%$} 
 & $0.989$ \textcolor{blue}{$+0.10\%$} 
 &  $57$ \textcolor{blue}{$+12.3\%$}\\
 
\midrule
\multirow{3}{*}{IC'13 (1015)} 
& Cross-Entropy \cite{baek2019wrong} 
& $94.1$ 
& $0.975$ 
& $97$ \\

 & LS-ED \cite{patel2020learning} 
 & $94.8$ \textcolor{blue}{$+0.82\%$} 
 & $0.975$ - 
 &  $87$ \textcolor{blue}{$+10.3\%$}\\
 
 & FEDS 
 & $95.3$ \textcolor{blue}{$+1.27\%$} 
 & $0.975$ -
 &  $90$ \textcolor{blue}{$+7.21\%$}\\
 
\midrule
\multirow{3}{*}{IC'15 (1811)} 
& Cross-Entropy \cite{baek2019wrong} 
& $82.8$ 
& $0.939$ 
& $614$\\

 & LS-ED \cite{patel2020learning} 
 & $83.2$ \textcolor{blue}{$+0.56\%$} 
 & $0.939$ - 
 & $599$ \textcolor{blue}{$+2.44\%$}\\
 
 & FEDS  
 & $83.8$ \textcolor{blue}{$+1.20\%$} 
 & $0.942$ \textcolor{blue}{$+0.31\%$} 
 &  $578$ \textcolor{blue}{$+5.86\%$}\\
 
\midrule
\multirow{3}{*}{IC'15 (2077)} 
& Cross-Entropy \cite{baek2019wrong} 
& $80.0$ 
& $0.908$ 
& $961$\\

 & LS-ED \cite{patel2020learning} 
 & $80.4$ \textcolor{blue}{$+0.54\%$} 
 & $0.908$ - 
 &  $944$ \textcolor{blue}{$+1.76\%$} \\
 
 & FEDS  
 & $80.9$ \textcolor{blue}{$+1.12\%$} 
 & $0.91$ \textcolor{blue}{$+0.22\%$} 
 & $929$ \textcolor{blue}{$+3.32\%$}\\
 
\midrule
\multirow{3}{*}{SVTP (645)} 
& Cross-Entropy \cite{baek2019wrong} 
& $82.4$ 
& $0.930$ 
&  $271$\\

 & LS-ED  \cite{patel2020learning} 
 & $83.4$ \textcolor{blue}{$+1.22\%$} 
 & $0.933$ \textcolor{blue}{$+0.32\%$} 
 & $258$ \textcolor{blue}{$+4.79\%$}\\
 
 & FEDS  
 & $84.0$ \textcolor{blue}{$+1.94\%$} 
 & $0.935$ \textcolor{blue}{$+0.53\%$} 
 & $248$ \textcolor{blue}{$+8.48\%$}\\
 
\midrule
\multirow{3}{*}{CUTE (288)} 
& Cross-Entropy \cite{baek2019wrong} 
& $77.3$  
& $0.883$  
&  $211$\\

 & LS-ED \cite{patel2020learning} 
 & $77.3$ \textcolor{blue}{$+0.06\%$} 
 & $0.885$ \textcolor{blue}{$+0.22\%$}
 & $197$ \textcolor{blue}{$+6.63\%$}\\
 
 & FEDS 
 & $79.0$ \textcolor{blue}{$+2.19\%$} 
 & $0.898$ \textcolor{blue}{$+1.69\%$} 
 & $176$ \textcolor{blue}{$+16.5\%$}\\
 
\midrule
\multirow{3}{*}{TOTAL} 
& Cross-Entropy \cite{baek2019wrong} 
& $88.7$ 
& $0.953$
& $3050$\\

 & LS-ED \cite{patel2020learning}
 & $89.0$ \textcolor{blue}{$+0.41\%$} 
 & $0.954$ \textcolor{blue}{$+0.62\%$}
 & $2961$ \textcolor{blue}{$+2.91\%$}\\
 
 & FEDS 
 & $89.6$ \textcolor{blue}{$+1.01\%$} 
 & $0.956$ \textcolor{blue}{$+0.35\%$}
 & $2809$ \textcolor{blue}{$+7.90\%$}\\
 
\bottomrule
\end{tabular}}
\end{center}
\label{table:feds_tps_resnet_bilstm_attn_pgt}
\end{table}

\subsection{Qualitative results}
\label{sec:qual_results}

\begin{figure}[H]
    \centering
    \includegraphics[width=\textwidth]{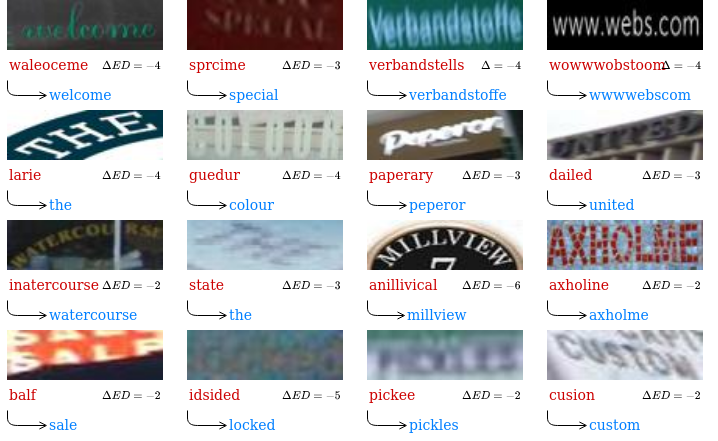}
    \caption{Randomly chosen examples from the test set where FEDS improves the STR model trained with cross-entropy. \textcolor{red}{Red} shows the incorrect predictions from the baseline model, \textcolor{blue}{blue} shows the correct prediction after post-tuning with FEDS and the arrow indicates post-tuning with FEDS.}
    \label{fig:feds_qual_good}
\end{figure}

Figure \ref{fig:feds_qual_good} shows randomly picked qualitative examples where FEDS leads to an improvement in the edit distance. Notice that the predictions from the baseline model are incorrect in all the examples. After post-tuning with FEDS, the predictions are correct, {\em i.e.}, perfectly match with the ground truth.

\begin{figure}[H]
    \centering
    \includegraphics[width=\textwidth]{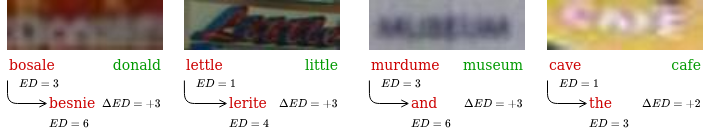}
    \caption{Four worst examples out of $12$K samples in the test set where FEDS leads to an increase in the edit distance. \textcolor{red}{Red} shows the incorrect predictions, \textcolor{green}{green} shows the ground truth and the arrow indicates post-tuning with FEDS.}
    \label{fig:feds_qual_bad}
\end{figure}

Figure \ref{fig:feds_qual_bad} shows hand-picked examples where FEDS leads to a maximum increase in the edit distance (ED increases). Notice that in these examples, the predictions from the baseline model are also incorrect. Furthermore, the input images are nearly illegible for a human.

\section{Conclusions}
\label{sec:feds_conclusions}
This chapter makes an orthogonal contribution to the trend of scene text recognition progress. It proposes a method to robustly post-tune a STR model using a learned surrogate of edit distance. The empirical results demonstrate an average improvement of $11.2 \%$ on total edit distance and an error reduction of $9.5\%$ on accuracy on a standard baseline \cite{baek2019wrong}. Improvements of $7.9\%$ on total edit distance and an error reduction of $10.3\%$ on accuracy are shown on a stronger baseline \cite{klara} that uses additional weakly supervised data for training.

\chapter{Recall@k Surrogate Loss with Large Batches and Similarity Mixup}
\label{chapter:rs@k}
\chaptermark{Recall@k Surrogate Loss}

\newcommand{\nn}[1]{\ensuremath{\text{NN}_{#1}}\xspace}
\def\l1{\ensuremath{\ell_1}\xspace}
\def\l2{\ensuremath{\ell_2}\xspace}

\def\roxf{$\mathcal{R}$Oxford\xspace}
\def\rox{$\mathcal{R}$Oxf\xspace}
\def\ro{$\mathcal{R}$O\xspace}
\def\rpar{$\mathcal{R}$Paris\xspace}
\def\rpa{$\mathcal{R}$Par\xspace}
\def\rp{$\mathcal{R}$P\xspace}
\def\rdis{$\mathcal{R}$1M\xspace}

\newcommand\resnet[3]{\ensuremath{\prescript{#2}{}{\mathtt{R}}{#1}_{\scriptscriptstyle #3}}\xspace}

\newcommand*\OK{\ding{51}}

\newenvironment{narrow}[1][1pt]
	{\setlength{\tabcolsep}{#1}}
	{\setlength{\tabcolsep}{6pt}}

\newcommand{\alert}[1]{{\color{red}{#1}}}
\newcommand{\gio}[1]{{\color{blue}{#1}}}
%\newcommand{\replace}[2]{{\color{gray}{#1}}{\color{red}{#2}}}

%--------------------------------------------------------------------------------
%algorithm
%\newcommand{\comment} [1]{{\color{orange} \Comment     #1}} % colored comment

%--------------------------------------------------------------------

\newcommand{\head}[1]{{\smallskip\noindent\bf #1}}
\newcommand{\equ}[1]{(\ref{equ:#1})\xspace}

\newcommand{\red}[1]{{\color{red}{#1}}}
\newcommand{\blue}[1]{{\color{blue}{#1}}}
\newcommand{\green}[1]{{\color{green}{#1}}}
\newcommand{\gray}[1]{{\color{gray}{#1}}}

%--------------------------------------------------------------------

\newcommand{\tran}{^\top}
\newcommand{\mtran}{^{-\top}}
\newcommand{\zcol}{\mathbf{0}}
\newcommand{\zrow}{\zcol\tran}

\newcommand{\ind}{\mathds{1}}
\newcommand{\expect}{\mathbb{E}}
\newcommand{\nat}{\mathbb{N}}
\newcommand{\zahl}{\mathbb{Z}}
\newcommand{\real}{\mathbb{R}}
\newcommand{\proj}{\mathbb{P}}
\newcommand{\prob}{\mathbf{Pr}}

\newcommand{\mif}{\textrm{if }}
\newcommand{\other}{\textrm{otherwise}}
\newcommand{\minimize}{\textrm{minimize }}
\newcommand{\maximize}{\textrm{maximize }}

\newcommand{\id}{\operatorname{id}}
\newcommand{\const}{\operatorname{const}}
\newcommand{\sgn}{\operatorname{sgn}}
\newcommand{\var}{\operatorname{Var}}
\newcommand{\mean}{\operatorname{mean}}
\newcommand{\trace}{\operatorname{tr}}
\newcommand{\diag}{\operatorname{diag}}
\newcommand{\vect}{\operatorname{vec}}
\newcommand{\cov}{\operatorname{cov}}

\newcommand{\softmax}{\operatorname{softmax}}
\newcommand{\clip}{\operatorname{clip}}

\newcommand{\defn}{\mathrel{:=}}
\newcommand{\peq}{\mathrel{+\!=}}
\newcommand{\meq}{\mathrel{-\!=}}

\newcommand{\floor}[1]{\left\lfloor{#1}\right\rfloor}
\newcommand{\ceil}[1]{\left\lceil{#1}\right\rceil}
\newcommand{\inner}[1]{\left\langle{#1}\right\rangle}
\newcommand{\norm}[1]{\left\|{#1}\right\|}
\newcommand{\frob}[1]{\norm{#1}_F}
\newcommand{\card}[1]{\left|{#1}\right|\xspace}
\newcommand{\diff}{\mathrm{d}}
\newcommand{\der}[3][]{\frac{d^{#1}#2}{d#3^{#1}}}
\newcommand{\pder}[3][]{\frac{\partial^{#1}{#2}}{\partial{#3^{#1}}}}
\newcommand{\ipder}[3][]{\partial^{#1}{#2}/\partial{#3^{#1}}}
\newcommand{\dder}[3]{\frac{\partial^2{#1}}{\partial{#2}\partial{#3}}}

\newcommand{\wb}[1]{\overline{#1}}
\newcommand{\wt}[1]{\widetilde{#1}}

\def\nsp{\hspace{-3pt}}
\def\zsp{\hspace{0pt}}
\def\xssp{\hspace{1pt}}
\def\ssp{\hspace{3pt}}
\def\msp{\hspace{6pt}}
\def\lsp{\hspace{12pt}}
\def\xlsp{\hspace{20pt}}

\newcommand{\cA}{\mathcal{A}}
\newcommand{\cB}{\mathcal{B}}
\newcommand{\cC}{\mathcal{C}}
\newcommand{\cD}{\mathcal{D}}
\newcommand{\cE}{\mathcal{E}}
\newcommand{\cF}{\mathcal{F}}
\newcommand{\cG}{\mathcal{G}}
\newcommand{\cH}{\mathcal{H}}
\newcommand{\cI}{\mathcal{I}}
\newcommand{\cJ}{\mathcal{J}}
\newcommand{\cK}{\mathcal{K}}
\newcommand{\cL}{\mathcal{L}}
\newcommand{\cM}{\mathcal{M}}
\newcommand{\cN}{\mathcal{N}}
\newcommand{\cO}{\mathcal{O}}
\newcommand{\cP}{\mathcal{P}}
\newcommand{\cQ}{\mathcal{Q}}
\newcommand{\cR}{\mathcal{R}}
\newcommand{\cS}{\mathcal{S}}
\newcommand{\cT}{\mathcal{T}}
\newcommand{\cU}{\mathcal{U}}
\newcommand{\cV}{\mathcal{V}}
\newcommand{\cW}{\mathcal{W}}
\newcommand{\cX}{\mathcal{X}}
\newcommand{\cY}{\mathcal{Y}}
\newcommand{\cZ}{\mathcal{Z}}

\newcommand{\vA}{\mathbf{A}}
\newcommand{\vB}{\mathbf{B}}
\newcommand{\vC}{\mathbf{C}}
\newcommand{\vD}{\mathbf{D}}
\newcommand{\vE}{\mathbf{E}}
\newcommand{\vF}{\mathbf{F}}
\newcommand{\vG}{\mathbf{G}}
\newcommand{\vH}{\mathbf{H}}
\newcommand{\vI}{\mathbf{I}}
\newcommand{\vJ}{\mathbf{J}}
\newcommand{\vK}{\mathbf{K}}
\newcommand{\vL}{\mathbf{L}}
\newcommand{\vM}{\mathbf{M}}
\newcommand{\vN}{\mathbf{N}}
\newcommand{\vO}{\mathbf{O}}
\newcommand{\vP}{\mathbf{P}}
\newcommand{\vQ}{\mathbf{Q}}
\newcommand{\vR}{\mathbf{R}}
\newcommand{\vS}{\mathbf{S}}
\newcommand{\vT}{\mathbf{T}}
\newcommand{\vU}{\mathbf{U}}
\newcommand{\vV}{\mathbf{V}}
\newcommand{\vW}{\mathbf{W}}
\newcommand{\vX}{\mathbf{X}}
\newcommand{\vY}{\mathbf{Y}}
\newcommand{\vZ}{\mathbf{Z}}

\newcommand{\va}{\mathbf{a}}
\newcommand{\vb}{\mathbf{b}}
\newcommand{\vc}{\mathbf{c}}
\newcommand{\vd}{\mathbf{d}}
\newcommand{\ve}{\mathbf{e}}
\newcommand{\vf}{\mathbf{f}}
\newcommand{\vg}{\mathbf{g}}
\newcommand{\vh}{\mathbf{h}}
\newcommand{\vi}{\mathbf{i}}
\newcommand{\vj}{\mathbf{j}}
\newcommand{\vk}{\mathbf{k}}
\newcommand{\vl}{\mathbf{l}}
\newcommand{\vm}{\mathbf{m}}
\newcommand{\vn}{\mathbf{n}}
\newcommand{\vo}{\mathbf{o}}
\newcommand{\vp}{\mathbf{p}}
\newcommand{\vq}{\mathbf{q}}
\newcommand{\vr}{\mathbf{r}}
\newcommand{\Vs}{\mathbf{s}}
\newcommand{\vt}{\mathbf{t}}
\newcommand{\vu}{\mathbf{u}}
\newcommand{\vv}{\mathbf{v}}
\newcommand{\vw}{\mathbf{w}}
\newcommand{\vx}{\mathbf{x}}
\newcommand{\vy}{\mathbf{y}}
\newcommand{\vz}{\mathbf{z}}

\newcommand{\vone}{\mathbf{1}}
\newcommand{\vzero}{\mathbf{0}}

\newcommand{\valpha}{{\boldsymbol{\alpha}}}
\newcommand{\vbeta}{{\boldsymbol{\beta}}}
\newcommand{\vgamma}{{\boldsymbol{\gamma}}}
\newcommand{\vdelta}{{\boldsymbol{\delta}}}
\newcommand{\vepsilon}{{\boldsymbol{\epsilon}}}
\newcommand{\vzeta}{{\boldsymbol{\zeta}}}
\newcommand{\veta}{{\boldsymbol{\eta}}}
\newcommand{\vtheta}{{\boldsymbol{\theta}}}
\newcommand{\viota}{{\boldsymbol{\iota}}}
\newcommand{\vkappa}{{\boldsymbol{\kappa}}}
\newcommand{\vlambda}{{\boldsymbol{\lambda}}}
\newcommand{\vmu}{{\boldsymbol{\mu}}}
\newcommand{\vnu}{{\boldsymbol{\nu}}}
\newcommand{\vxi}{{\boldsymbol{\xi}}}
\newcommand{\vomikron}{{\boldsymbol{\omikron}}}
\newcommand{\vpi}{{\boldsymbol{\pi}}}
\newcommand{\vrho}{{\boldsymbol{\rho}}}
\newcommand{\vsigma}{{\boldsymbol{\sigma}}}
\newcommand{\vtau}{{\boldsymbol{\tau}}}
\newcommand{\vupsilon}{{\boldsymbol{\upsilon}}}
\newcommand{\vphi}{{\boldsymbol{\phi}}}
\newcommand{\vchi}{{\boldsymbol{\chi}}}
\newcommand{\vpsi}{{\boldsymbol{\psi}}}
\newcommand{\vomega}{{\boldsymbol{\omega}}}

\newcommand{\rLambda}{\mathrm{\Lambda}}
\newcommand{\rSigma}{\mathrm{\Sigma}}

%--------------------------------------------------------------------
% Add a period to the end of an abbreviation unless there's one
% already, then \xspace.
\makeatletter
\DeclareRobustCommand\onedot{\futurelet\@let@token\@onedot}
\def\@onedot{\ifx\@let@token.\else.\null\fi\xspace}
\def\eg{\emph{e.g}\onedot} \def\Eg{\emph{E.g}\onedot}
\def\ie{\emph{i.e}\onedot} \def\Ie{\emph{I.e}\onedot}
\def\cf{\emph{cf}\onedot} \def\Cf{\emph{C.f}\onedot}
\def\etc{\emph{etc}\onedot} \def\vs{\emph{vs}\onedot}
\def\wrt{w.r.t\onedot} \def\dof{d.o.f\onedot}
\def\etal{\emph{et al}\onedot}
\makeatother

Minimization of a loss that is a function of the test-time evaluation metric has shown to be beneficial in deep learning for numerous computer vision and natural language processing tasks. Examples include intersection-over-union as a loss that boosts performance for object detection~\cite{yjw+16,rig+19} and semantic segmentation~\cite{nsb+18}, and structural similarity~\cite{mae+18}, peak signal-to-noise ratio~\cite{bms+18} and perceptual~\cite{pam+21} as reconstruction losses for image compression that give better results according to the respective evaluation metrics. 

\begin{figure}
\begin{center}
\includegraphics[width=0.85\textwidth]{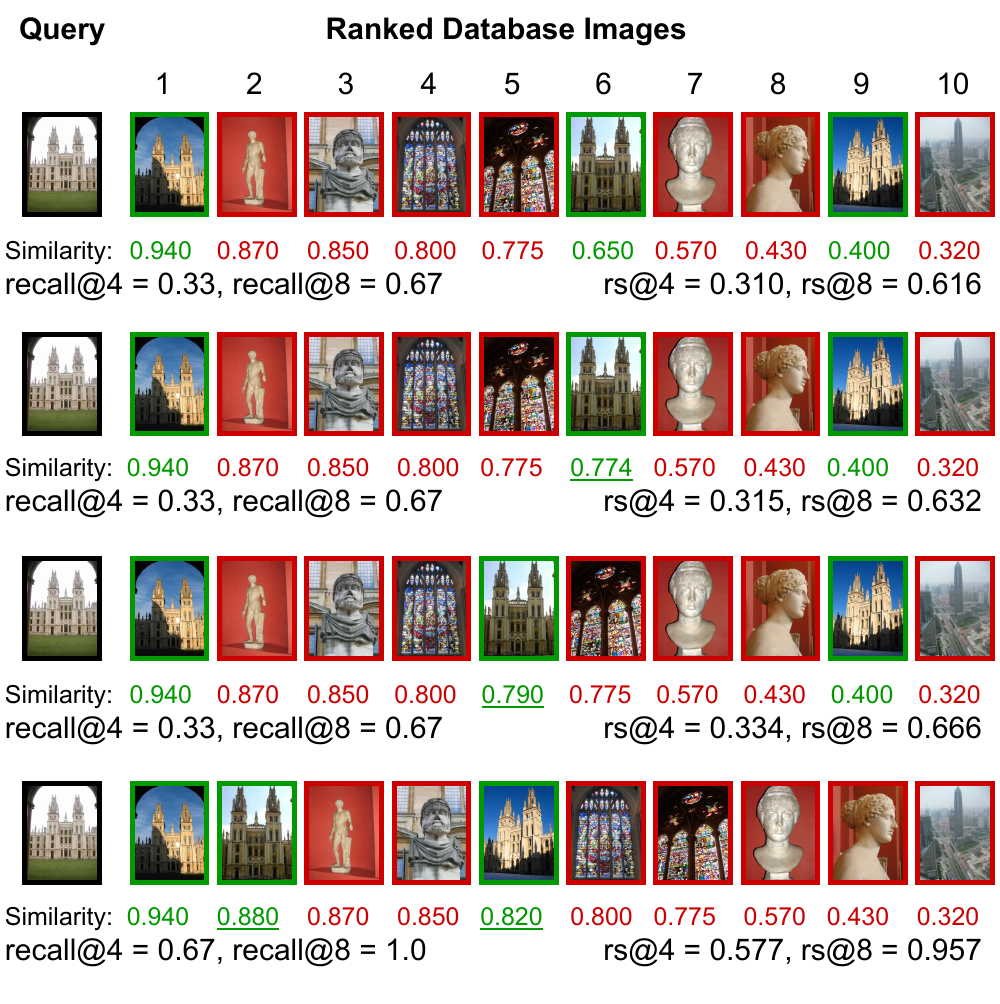}
\caption{A comparison between recall@k and rs@k, the proposed differentiable recall@k surrogate. Examples show a query, the ranked database images sorted according to the similarity and the corresponding values for recall@k and rs@k and their dependence on similarity score change. Note that the values of recall@k and rs@k are close. Changes to similarity and ranking in some cases may not affect the original recall@k but can affect the surrogate, with the latter having a more significant impact than the former. Similarity values of all negatives are fixed for ease of understanding. The similarity values of the positives that were changed in rows 2, 3 and 4 are \underline{underlined}.
\label{fig:teaser}}
\end{center}
\end{figure}

Training deep networks via gradient descent on the evaluation metric is not possible when the metric is non-differentiable. Deep learning methods resort to a proxy loss, a differentiable function, as a workaround, which empirically leads to a reasonable performance but may not align well with the evaluation metric. Examples exist in object detection~\cite{yjw+16}, scene text recognition~\cite{phm+20,pm+21}, machine translation~\cite{bbx+17} and image retrieval~\cite{bxk+20,pgr+19}.

This chapter deals with the training of image retrieval posed  as deep metric learning and Euclidean search in the learned image embedding space. It is the task of ranking all database examples according to the relevance to a query, which is of vital importance for many applications. The standard evaluation metrics are precision and recall in the top retrieved results and the mean Average Precision (mAP).
These metrics are standard in information retrieval, they reflect the quality of the retrieved results and allow for flexibility to focus either on the few top results or the whole ranked list of examples, respectively.
Recall at top-$k$ retrieved results, denoted by \emph{recall@k} in the following, is the primary focus of this work.

The problem related to the optimization of non-differentiable evaluation metrics applies to recall@k as well. 
% Re-visit Kristina's rephrased sentence.
Estimating the position of positive images in the list of retrieved results and counting how many positives appear inside a short-list of a fixed size involves non-differentiable operations.
% using step functions, which are non-differentiable. 
Note that methods for training on non-differentiable losses, such as actor-critic~\cite{bbx+17} and learning surrogates~\cite{phm+20} are not directly applicable to recall@k. This is due to the fact that these methods are limited to decomposable functions, where a per-example performance measure is available. Such an attempt is made by Engilberge~\etal~\cite{ecp+19}, where an LSTM learns sorting-based metrics, but is not adapted in consequent work due to slow training. As an alternative, deep metric learning approaches for image retrieval often use ranking proxy losses, termed pairwise losses. In the embedding space, loss functions such as contrastive~\cite{hcl06}, triplet~\cite{skp+15}, and margin~\cite{wms+17} pull the examples from the same class closer to one another and push the examples from a different class away. These losses are hand-crafted to reflect the objectives of the retrieval task and, consequently, the evaluation metric. The loss value depends on the image-to-image similarity for image pairs or triplets and does not take into account the whole ranked list of examples. Changes in the similarity value without any change in the overall ranking alter the loss value indicate that they are not well correlated with ranking~\cite{bxk+20}. Recent methods focus on optimizing Average Precision (AP) and use a surrogate function as a loss~\cite{hls18,rar+19,rmp+20,chx+19,bxk+20}. A surrogate of an evaluation metric is a function that approximates it in a differentiable manner. 

The proposed method attains state-of-the-art results for 4 fine-grained retrieval datasets, namely iNaturalist~\cite{vms+18}, VehicleID~\cite{vms+18}, SOP~\cite{ohb16} and Cars196~\cite{ksd+13}, and 2 instance-level retrieval datasets, namely Revisited Oxford and Paris~\cite{rit+18}. 
This is accomplished by the demonstrated synergy between the three following elements.
First, a new loss that is proposed as a surrogate of an established retrieval evaluation metric, namely recall at top $k$, and is experimentally shown to consistently outperform existing competitors. 
A comparison between the evaluation metric and the proposed loss is shown in Figure \ref{fig:teaser}.
Second, the use of a very large batch size, in the order of several thousand large resolution images on a single GPU. This is inspired by the instance-level retrieval literature \cite{rar+19} and is introduced for the first time in the context of fine-grained categorization. 
In a recent work of verifying prior results in deep metric learning for fine-grained categorization~\cite{mbl20} the batch-size is considered fixed to a single and small value among a large set of comparisons for different losses; in this work we reach batch-sizes that are two orders of magnitude larger than in the work of Musgrave~\etal~\cite{mbl20}.
The third elements is the proposed mixup regularization technique that is computationally efficient and that virtually enlarges the batch. Its efficiency is obtained by operating on the very last stage of similarity estimation, \ie scalar similarities are mixed, while its applicability goes beyond the combination with the proposed loss in this work.
The proposed loss is used for training widely used ResNet architectures~\cite{hzr+16} but also recent vision-transformers (ViT)~\cite{dbk+21}. The superiority of this loss compared to existing losses is demonstrated with both architectures, while with ViT-B/16 top results are achieved at lower throughput than with ResNet.

The rest of the chapter is structured as follows: related work specific to metric learning is provided in Section \ref{sec:rs@k_related_work}, the proposed recall@k surrogate along with the similarity mixup is presented in Section \ref{sec:rs@k_method}, experimental restuls for metric learning, instance-level search, fine-grained recognition and ablation studies are presented in Section \ref{sec:rs@k_experiments}, finally, the conclusions for the proposed contributions of the chapter are made in Section \ref{sec:rs@k_conclusions}. 

\section{Related work}
\label{sec:rs@k_related_work}

In this section, the related work is reviewed for two different families of deep metric learning approaches regarding the type of loss that is optimized, namely classification losses and pairwise losses. 
Given an embedding network that maps input images to a high dimensional space, in the former, the loss is a function of the embedding and the corresponding category label of a single image, while in the latter, the loss is a function of the distance, or similarity, between two embeddings and the corresponding pairwise label. 
Prior work for mixup~\cite{zcd+17} techniques related to embedding learning is reviewed too. 

\paragraph{Classification losses.} The work of Zhai and Wu~\cite{zw18} supports that the standard classification loss, \ie cross-entropy (CE) loss is a strong approach for deep metric learning. Their finding is supported by the use of layer normalization and class-balanced sampling. In the domain of metric learning for faces, several different classification losses are proposed, such as SphereFace~\cite{lwy+17}, CosFace~\cite{wwz+18} and ArcFace~\cite{dgx+19}, where contributions are in the spirit of large margin classification. Despite the specificity of the domain, such losses are applicable beyond faces. Another variant is the Neighborhood Component Analysis (NCA) loss that is used in the work of Movshovitz-Attias~\etal~\cite{mtl+17}, which is later improved~\cite{tdt20} by temperature-based scaling and faster update of the class prototype vectors, also called proxies in their work. The restriction of a single prototype vector per class is dropped by Qian~\etal~\cite{qss+19} who stores multiple representatives per category. 

Classification losses, in contrast to pairwise losses, perform the optimization independently per image. An exception is the work of Elezi~\etal~\cite{evt+20} where a similarity propagation module captures group interactions within the batch. Then, cross-entropy loss is used, which now comes with significant improvements by taking into account such interactions. This is recently improved~\cite{sel21} by replacing the propagation module with an attention model.
The relation between CE loss and some of the widely used pairwise losses is studied from a mutual information point of view~\cite{brz+20}. CE loss is viewed as approximate bound-optimization for minimizing pairwise losses; CE maximizes mutual information, and so do these pairwise losses, which are reviewed in the following. 

\paragraph{Pairwise losses.} The first pairwise loss introduced for this task is the so-called contrastive loss~\cite{hcl06}, where embeddings of relevant pairs are pushed as close as possible, while those of non-relevant ones are pushed far enough. Since the target task is typically a ranking one, the triplet loss~\cite{skp+15}, a popular and widely used loss, improves that by forming training triplets in the form of anchor, positive and negative examples. The loss is a function of the difference between anchor-to-positive and anchor-to-negative distances and is zero if such a difference is large enough, therefore satisfying the objectives of a ranking task for this triplet. Optimization over all pairs or triplets is not tractable and is observed to be sub-optimal~\cite{wms+17}. As a result, a lot of attention is paid to finding informative pairs and triplets~\cite{mbl20,rms+20,sxj+15,sohn16,lxz+19}, which typically includes heuristics. Several other losses are suggested in the literature~\cite{wms+17,wzw+17,sxj+15} and are added to the long list of hand-designed proxy losses which target to learn embeddings that transfer well to a ranking or a similar task. 

A few cases follow a principled approach for obtaining a loss that is appropriate for ranking tasks. This is the case with the work of Ustinova~\etal~\cite{ul16} where the goal is to minimize the probability that the similarity between embeddings of a non-relevant pair is larger than that of a relevant one. This probability is approximated by the quantization of the range of possible similarities and the histogram loss, which is estimated within a single batch. Their work dispenses with the need for any kind of sampling for mini-batch construction. 
An information-theoretic loss function, called RankMI~\cite{kpd+20}, maximizes the mutual information between the samples within the same semantic class using a neural network. 
Another principled approach focuses on optimizing AP, which is a standard retrieval evaluation metric. A smooth approximation of it is often used in the literature~\cite{rmp+20,hls18,rar+19}, while the work of Brown~\etal~\cite{bxk+20} is the closest to ours. In combination with such AP-based losses, a large batch size is crucial, which meets the limitations set by the hardware. Such limitations are overcome in the work of Revaud~\etal~\cite{rar+19} who uses a batch of $4,000$ high-resolution images.

\paragraph{Embedding mixup.} Manifold mixup~\cite{vlb+19}, which involves mixing~\cite{zcd+17} intermediate representations and labels of two examples, has demonstrated to improve generalizability for supervised learning by encouraging smoother decision boundaries. Such techniques are investigated for embedding learning and image retrieval by mixing the embedding of two examples. Duan~\etal~\cite{dzl+18} uses adversarial training to synthesize additional negative samples from the observed negatives. Kalantidis ~\etal~\cite{ksp+20} synthesize hard-negatives for contrastive self-supervised learning by mixing the embedding of the two hardest negatives and also mixing them with the query itself. Zheng~\etal~\cite{zcl+19} uses a linear interpolation between the embeddings to manipulate the hardness levels. In the work of Gu ~\etal~\cite{gk+20}, two embedding vectors from the same class are used to generate symmetrical synthetic examples and hard-negative mining is performed within the set of original and the synthetic examples. This is further extended to proxy-based losses, where the embedding of examples from different classes and labels is mixed to generate synthetic proxies~\cite{gkk+21}. Linearly interpolating labels entails the risk of generating false negatives if the interpolation factor is close to $0$ or $1$. Such limitations are overcome in the work of Venkataramanan ~\etal~\cite{vpa+21}, which generalizes mixing examples from different classes for pairwise loss functions. The proposed {\em SiMix} approach differs from the aforementioned techniques as it operates on the similarity scores instead of the embedding vectors, does not require training an additional model, making it computationally efficient. Furthermore, unlike the existing mixup techniques, it uses a synthetic sample in the roles of a query, positive and negative example. 

\section{Method}
\label{sec:rs@k_method}

This section presents the task of image retrieval and the proposed approach for learning image embeddings.

\paragraph{Task.} We are given a query example $q\in \cX$ and a collection of examples $\Omega \subset \cX$, also called database, where $\cX$ is the space of all images. The set of database examples that are positive or negative to the query are denoted by $P_q$ and $N_q$, respectively, with $\Omega=P_q \cup N_q$. Ground-truth information for the positive and negative sets per query is obtained according to discrete class labels per example, \ie if two examples come from the same class, then they are considered positive to each other, otherwise negative. This is the case for all (training or testing) databases used in this work. Terms example and image are used interchangeably in the following text. In image retrieval, all database images are ranked according to similarity to the query $q$, and the goal is to rank positive examples before negative ones.

\paragraph{Deep image embeddings.} 
Image embeddings, otherwise called descriptors, are generated by function $f_{\theta}: \cX \rightarrow \real^d$. In this work, function $f_\theta$ is a deep fully  convolutional neural network or a vision transformer mapping input images of any size or aspect ratio to an $L_{2}$-normalized $d$-dimensional embedding. Embedding for image $x$ is denoted by $\vx = f_\theta(x)$. Parameter set $\theta$ of the network is learned during the training. Similarity between a query $q$ and a database image $x$ is computed by the dot product of the corresponding embeddings and is denoted by $s(q,x) = \vq^\top \vx$, also denoted as $s_{qx}$ for brevity.

\paragraph{Evaluation metric.} Recall@k is one of the standard metrics to evaluate image retrieval methods. For query $q$, it is defined as a ratio of the number of relevant (positive) examples within the top-k ranked examples to the total number of relevant  examples for $q$ given by $|P_q|$.
It is denoted by $R_\Omega^k(q)$ when computed for query $q$ and database $\Omega$ and can be expressed as

\begin{equation}
    R_\Omega^k(q) = \frac{\sum\limits_{x \in P_q} H (k - r_\Omega(q,x))}{|P_q|},
    \label{equ:recall}
\end{equation}
where $r_\Omega(q,x)$ is the rank of example $x$ when all database examples in $\Omega$ are ranked according to similarity to query $q$. Function $H(.)$ is the Heaviside step function, which is equal to 0 for negative values, otherwise equal to 1. The rank of example $x$ is computed by

\begin{equation}
    r_\Omega(q,x) = 1 + \sum\limits_{z \in \Omega, z \neq x} H(s_{qz} - s_{qx}),
    \label{equ:rank}
\end{equation}

Therefore, \equ{recall} can now be expressed as
\begin{equation}
R_\Omega^k(q) = \frac{\sum\limits_{x \in P_q} H (k  - 1 - \sum\limits_{z \in \Omega, z \neq x} H(s_{qz} - s_{qx}) )}{|P_q|}.
\label{equ:recall_full}
\end{equation}

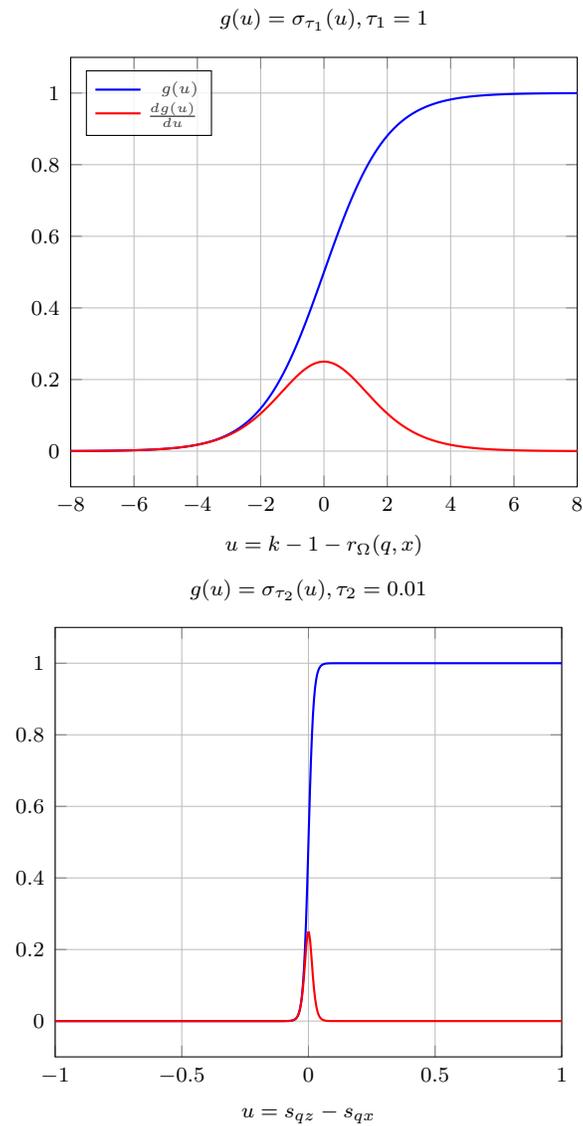
\begin{figure}
\begin{center}
\begin{tikzpicture}[declare function={sigma(\x)=1/(1+exp(-\x));sigmap(\x)=sigma(\x)*(1-sigma(\x));}]
\begin{axis}[%
	width=0.57\linewidth,
	height=0.5\linewidth,
	xlabel={$u = k -1 - r_\Omega(q,x)$},
	title={$g(u)= \sigma_{\tau_1}(u), \tau_1=1$},
	legend cell align={left},
	legend pos=north west,
    legend style={cells={anchor=east}, font =\tiny, fill opacity=0.8, row sep=-2.5pt},
    xmin = -8,
    xmax = 8,
    domain=-8:8
]
\addplot[blue,mark=none,samples=500]   (x,{sigma(x/1.0)});\addlegendentry{$g(u)$}
\addplot[red,mark=none,samples=500]   (x,{sigmap(x/1.0)});\addlegendentry{$\frac{dg(u)}{du}$}
\end{axis}
\end{tikzpicture}
\hspace{-5pt}
% \raisebox{12pt}
{
\begin{tikzpicture}[declare function={sigma(\x)=1/(1+exp(-\x));sigmap(\x)=sigma(\x)*(1-sigma(\x));}]
\begin{axis}[%
	width=0.57\linewidth,
	height=0.5\linewidth,
	xlabel={$u= s_{qz}-s_{qx}$ },
	title={$g(u)= \sigma_{\tau_2}(u), \tau_2 =0.01$},
	legend cell align={left},
	legend pos=north west,
    legend style={cells={anchor=east}, font =\small, fill opacity=0.8, row sep=-2.5pt},
    xmin = -1,
    xmax = 1,
    domain=-1:1
]
\addplot[blue,mark=none,samples=500]   (x,{sigma(x/0.01)});
\addplot[red,mark=none,samples=500]   (x,{sigmap(x/0.01)});
\end{axis}
\end{tikzpicture}
}
\caption{The two sigmoid functions which replace the Heaviside step function for counting the positive examples in the short-list of size $k$ (top) and for estimating the rank of examples (bottom).
\label{fig:sigmoids}
}
\end{center}
\end{figure}

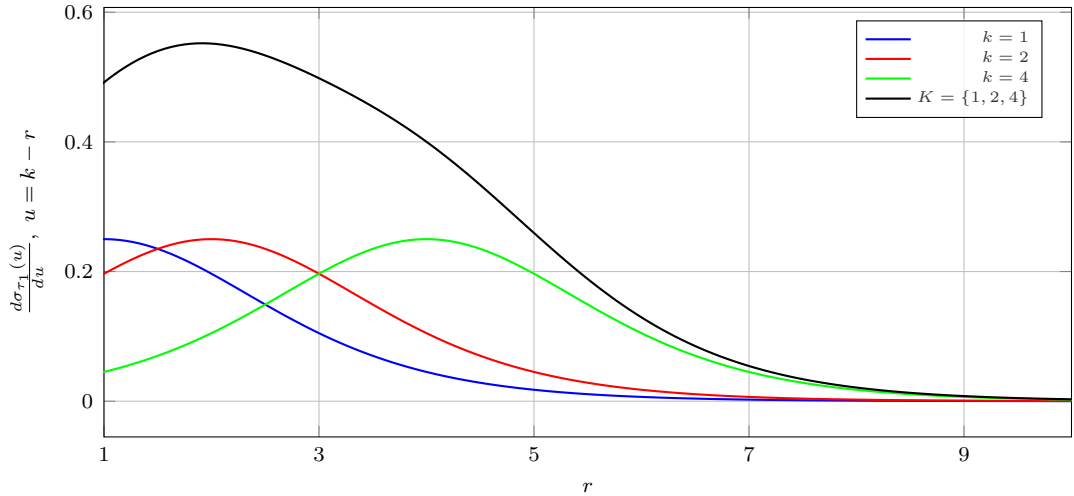
\begin{figure}
\begin{center}
\begin{tikzpicture}[declare function={sigma(\x)=1/(1+exp(-\x));sigmap(\x)=sigma(\x)*(1-sigma(\x));}]
\begin{axis}[%
	width=0.99\linewidth,
	height=0.5\linewidth,
	xlabel={$r$},
	ylabel={$\frac{d\sigma_{\tau_1}(u)}{du},~u=k-r$},
	legend cell align={left},
	legend pos=north east,
    legend style={cells={anchor=east}, font =\tiny, fill opacity=0.8, row sep=-2.5pt},
    xtick = {1,3,5,7,9},
    xmin = 1,
    xmax = 10,
    domain=0:30
]
\addplot[blue,mark=none,samples=500]   (x,{sigmap(x-1)});\addlegendentry{$k=1$}
\addplot[red,mark=none,samples=500]   (x,{sigmap(x-2)});\addlegendentry{$k=2$}
\addplot[green,mark=none,samples=500]   (x,{sigmap(x-4)});\addlegendentry{$k=4$}
\addplot[black,mark=none,samples=500]   (x,{sigmap(x-1)+sigmap(x-2)+sigmap(x-4)});\addlegendentry{$K=\{1,2,4\}$};
\end{axis}
\end{tikzpicture}
\caption{Gradient magnitude of the sigmoid used to count the positive examples in the short-list of size $k$ versus the rank $r$ (equal to $r_\Omega(q,x)$, see \equ{rank}) of a positive example $x$. It shows how much a positive example is pushed towards lower ranks depending on its current rank. In the case of multiple values for $k$, the total gradient is equivalent to the sum of the separate ones.
\label{fig:deriv}}
\end{center}
\end{figure}

\paragraph{Recall@k surrogate loss.} The computation of recall in \equ{recall_full} involves the use of the Heaviside step function. The gradient of the Heaviside step function is a Dirac delta function. Hence, direct optimization of recall with back-propagation is not feasible. A common smooth approximation of the Heaviside step function is provided by the logistic function~\cite{km+15,ikm+15,ikm+17}, a common sigmoid function $\sigma_{\tau}: \real \rightarrow \real$ controlled by temperature $\tau$, which is given by
\begin{equation}
\sigma_\tau(u) = \frac{1}{1+e^{-\frac{u}{\tau}}},
\end{equation}
where large (small) temperature value leads to worse (better) approximation and denser (sparser) gradient. This approximation is common in the machine learning literature for several tasks~\cite{sh+09,gls+16,mmt+17} and also appears in the approximation of the Average Precision evaluation metric~\cite{bxk+20}, which is used for the same task as ours. By replacing the step function with the sigmoid function, a smooth approximation of recall is obtained as
\begin{equation}
     \tilde{R}_\Omega^k(q) = \frac{\sum\limits_{x \in P_q} \sigma_{\tau_1} (k  - 1 - \sum\limits_{\substack{z \in \Omega\\ z \neq x}} \sigma_{\tau_2}(s_{qz} - s_{qx}) )}{|P_q|},
\label{equ:smooth_recall}
\end{equation}
which is differentiable and can be used for training with back-propagation. The two sigmoids have different function domains and, therefore, different temperatures (see Figure~\ref{fig:sigmoids}). The minimized single-query loss in a mini-batch $B$, with size $M=|B|$, and query $q\in B$ is given by 
\begin{equation}
L^k(q) = 1- \tilde{R}_{B\setminus q}^k(q).
\end{equation}
while incorporation of multiple values of $k$ is performed in the loss given by 
\begin{equation}
L^K(q) = \frac{1}{|K|}\sum_{k \in K} L^k(q).
\label{equ:loss}
\end{equation}
Figure~\ref{fig:deriv} shows the impact of using single or multiple values for $k$. 

All examples in the mini-batch are used as queries and the average loss over all queries is minimized during the training.
The proposed loss is referred to as \emph{Recall@k Surrogate loss}, or RS@k loss for brevity.

To allow for 0 loss when $k$ is smaller than the number of positives (note that exact recall@k is less than 1 by definition), we slightly modify \equ{smooth_recall} during the training. Instead of dividing by $|P_q|$, we divide by $\min(k, |P_q|)$, and, consequently, we clip values larger than $k$ in the numerator to avoid negative loss values.

\paragraph{Similarity mixup (SiMix).} Given original batch $B$, virtual batch $\hat{B}$ is created by mixing all pairs of positive examples in the original batch.
Embeddings of examples $x\in B$ and $z \in B$ are used to generate mixed embedding
\begin{equation}
    \vv_{xz\alpha} = \alpha \vx + (1-\alpha) \vz \quad | \quad \alpha \sim U(0,1),
\end{equation}
for a virtual example that is denoted by $xz\alpha \in \hat{B}$.
The similarity of an original example $w\in B$ to the virtual example $xz\alpha \in \hat{B}$ is given by
\begin{equation}
s(w,xz\alpha) = \vw^\top \vv_{xz\alpha} = \alpha s_{wx} + (1-\alpha) s_{wz}, 
\label{equ:simix1} 
\end{equation}
where the original and virtual examples can be the query and database examples, respectively, or vice versa. 
In case both examples are virtual, \eg $xz\alpha_1 \in \hat{B}$ used as a query and $yw\alpha_2 \in \hat{B}$ as a part of the database, then their similarity is given by
\begin{align}
s(xz\alpha_1,yw\alpha_2) &= \vv_{xz\alpha_1}^\top \vv_{yw\alpha2} \nonumber\\ 
        &= \alpha_1 \alpha_2 s_{xy} + (1-\alpha_1) (1- \alpha_2) s_{zw} \nonumber\\
        &+ \alpha_1 (1-\alpha_2) s_{xw}+(1-\alpha_1) \alpha_2 s_{zy}.
\label{equ:simix2} 
\end{align}
The pairwise similarities that appear on the right-hand side of the previous formulas, \eg $s_{wx}$ and $s_{wz}$ in \equ{simix1}, are computed from the embeddings of the original, non-virtual examples and are also required for the computation of the RS@k without any virtual examples. Therefore, the mini-batch is expanded to $B \cup \hat{B}$ by adding virtual examples without the need for explicit construction of the corresponding embeddings or computation of the similarity via dot product; simple mixing of the corresponding pairwise scalar similarities is enough. SiMix reduces to mixing pairwise similarities due to the lack of re-normalization of the mixed embeddings, which is different to existing practice in prior work~\cite{vpa+21,gk+20,gkk+21,ksp+20} and brings training efficiency benefits.

Virtual examples are created only between examples of the same classes and are labeled according to the class of the original examples that are mixed. Virtual examples are used both as queries and as database examples, while mixing is applied to all pairs of positive examples inside a mini-batch.

\paragraph{Overview.} An overview of the training process with the proposed loss and SiMix is given in Algorithm \ref{alg:main}. In case SiMix is not used, then lines \ref{lin:vbatch}, \ref{lin:vsim1}, \ref{lin:vsim2} and \ref{lin:expand} are skipped. It is assumed that each image in training is labeled to a class. Mini-batches of size $M$ are generated by randomly sampling $m$ images per class out of $\nicefrac{M}{m}$ sampled classes.

\begin{algorithm}
\caption{Training with RS@k and SiMix.}
% \footnotesize
% \algrenewcommand\algorithmicindent{0.5em}%
% \begin{algorithmic}[1]
% \Procedure{Train-SR@k}{$X$, $Y$, $M$, $m$}
% \color{gray}
% \State $X: $ training images
% \State $Y: $ class labels
% \State $M: $ mini-batch size
% \State $m: $ number of images per class in mini-batch
% \color{black}
% \State
% \State $\theta \gets$ initialize according to pre-training \comment{use ImageNet}
% \For{$\text{iteration} \in [1,\ldots, \text{number-of-iterations}]$}
% \State $loss \gets 0$ \comment{set batch loss to zero}
% \State $B \gets$ \Call{Batch-sampler}{$X$, $Y$, $M$, $m$} 
% \State \algorithmicfor { $(x,z) \in B \times B$} \algorithmicdo { compute $s_{xz}$}  \comment{all pairs within batch}%
% \For{$q \in B$} \comment {use each image in the batch as query}
% \State $B_q \gets B \setminus q$ \comment {exclude query from the database}
% \State $loss \gets loss + L^K(q)$ \comment {Recall@k loss}
% \EndFor
% \State $\theta \gets $ \Call{Minimize}{$\frac{L}{|B|}$} \comment {SGD update}
% \EndFor
% \EndProcedure
% \end{algorithmic}
\footnotesize
\algrenewcommand\algorithmicindent{0.5em}%
\begin{algorithmic}[1]
\Procedure{Train-RS@k}{$X$, $Y$, $M$, $m$}
\color{gray}
\State $X: $ training images
\State $Y: $ class labels
\State $M: $ mini-batch size
\State $m: $ number of images per class in mini-batch
\color{black}
\State
\State $\theta \gets$ initialize according to pre-training 
\Comment{use ImageNet}
\For{$\text{iteration} \in [1,\ldots, \text{number-of-iterations}]$}
\State $loss \gets 0$ 
\Comment{set batch loss to zero}
\State $B \gets$ \Call{Batch-sampler}{$X$, $Y$, $M$, $m$} 
\State $\hat{B} \gets$ \Call{Virtual-batch}{$B$} 
\Comment{enumerate virtual examples} \label{lin:vbatch}
\State \algorithmicfor { $(x,z) \in B \times B$} \algorithmicdo { compute $s(x,z)$ }  
\Comment{use $\vx^\top \vz$} 
\State \algorithmicfor { $(x,z) \in B \times \hat{B}$} \algorithmicdo { compute $s(x,z)$ }  
\Comment{use \equ{simix1}} \label{lin:vsim1}
\State \algorithmicfor { $(x,z) \in \hat{B} \times \hat{B}$} \algorithmicdo { compute $s(x,z)$ }  
\Comment{use \equ{simix2}} \label{lin:vsim2}
\State $B \gets B \cup \hat{B}$ 
\Comment{expand batch with virtual examples} \label{lin:expand}
\For{$q \in B$} 
\Comment {use each image in the batch as query}
\State $B_q \gets B \setminus q$ 
\Comment{exclude query from the database}
\State $loss \gets loss + L^K(q)$ 
\Comment{Recall@k loss \equ{loss}}
\EndFor
\State $\theta \gets $ \Call{Minimize}{$\frac{loss}{|B|}$} 
\Comment{SGD update}
\EndFor
\EndProcedure
\end{algorithmic}
\label{alg:main}
\end{algorithm}

\section{Experiments on Retrieval Benchmarks}
\label{sec:rs@k_experiments}

\subsection{Datasets}
\label{sec:rs@k_datasets}

\begin{table}
\setlength\extrarowheight{1pt}
\begin{center}
\small
\begin{tabular}{l|r|r|r}
    \hline
    \textbf{Dataset} & \textbf{\#Images} &  \textbf{\#Classes} & \textbf{\#Avg}\\
    \hline\hline
    iNaturalist Train~\cite{vms+18} & $325,846$ & $5,690$ & $57.3$ \\
    iNaturalist Test~\cite{vms+18} & $136,093$ & $2,452 $ & $55.5$ \\
    VehicleID Train~\cite{ltw+16} & $110,178$ & $13,134$ & $8.4$ \\
    VehicleID Test~\cite{ltw+16} & $40,365 $ & $4,800$ & $8.4$ \\
    SOP Train~\cite{ohb16} & $59,551 $ &  $11,318$ & $5.3$ \\
    SOP Test~\cite{ohb16} & $60,502$ &  $11,316$ & $5.3$ \\
    Cars196 Train~\cite{ksd+13} & $8,054$ & $98$ & $82.1$ \\
    Cars196 Test~\cite{ksd+13} & $8,131$ & $98$ & $82.9$ \\
    \hline
    $\mathcal{R}$Oxford~\cite{rit+18} & $4,993$ & $11$ & n/a \\
    $\mathcal{R}$Paris~\cite{rit+18} & $6,322$ & $11$ & n/a \\
    GLDv1~\cite{nas+17} & $1,060,709$ & $12,894$ & $82.3$ \\
    \hline
\end{tabular}
\end{center}
\caption{Dataset composition for training and evaluation.}
\label{tab:datasets}
\end{table}

The training and evaluation is performed on four widely used image retrieval benchmarks, namely iNaturalist~\cite{vms+18}, PKU VehicleID~\cite{ltw+16}, Stanford Online Products~\cite{ohb16} (SOP) and Stanford Cars~\cite{ksd+13} (Cars196). Recall at top $k$ retrieved images,  denoted by r@k, is one of the standard evaluation metrics in these benchmarks.  Metric r@k is 1 if at least one positive image appears in the top $k$ list, otherwise 0. The metric is averaged across all queries. Note that this is different from the standard definition of recall in \equ{recall}. 

iNaturalist~\cite{vms+18} is firstly used by Brown~\etal~\cite{bxk+20}, whose setup we follow: $5,690$ classes for training and $2,452$ classes for testing. For VehicleID, according to the standard setup~\cite{ltw+16}, $13,134$ classes are used for training, and the evaluation is conducted on the predefined small ($800$ classes), medium ($1600$ classes) and large ($2400$ classes) test sets. For SOP~\cite{ohb16} and Cars196~\cite{ksd+13}, the standard experimental setup of Song~\etal~\cite{sxj+15} is followed. The first half of the classes are used for training and the rest for testing, resulting in $11,318$ classes for SOP and $98$ for Cars196.

The method is evaluated for instance-level search on Revisited Oxford ($\mathcal{R}$Oxford) and Paris ($\mathcal{R}$Paris) benchmark~\cite{rit+18}, where the evaluation metric is mean Average Precision (mAP). The training uses the Google Landmarks dataset (GLDv1)~\cite{nas+17} to perform a comparison with the work of Revaud~\etal~\cite{rar+19} and their AP loss. The validation is performed according to the work of Tolias~\etal~\cite{tjc20}. 

The number of examples, classes, and average number of examples per class can be found in Table. \ref{tab:datasets}. Note that these datasets are diverse in the number of training examples, the number of classes, and the number of examples per class, ranging from class balanced~\cite{ksd+13} to long-tailed~\cite{vms+18}.

\subsection{Implementation details}
\label{sec:rs@k_implementation_details}

Implementation details are identical for the four image retrieval benchmarks but differ for  $\mathcal{R}$Oxford/$\mathcal{R}$Paris to follow and compare to prior work~\cite{rar+19}. Differences are clarified when needed.

\paragraph{Architecture.} An ImageNet~\cite{dsl+09} pre-trained ResNet-50~\cite{hzr+16} is used as the backbone for deep image embeddings. Building on the standard implementation of~\cite{rms+20}, the BatchNorm parameters are kept frozen during the training. After the convolutional layers, Generalized mean pooling~\cite{rtc19} and layer normalization~\cite{bkh+16} are used,  similar to~\cite{tdt20}. For vision transformers~\cite{dbk+21} ViT-B/32 and ViT-B/16 with an ImageNet-21k initialization from the timm library~\cite{rw2019timm} are used. The last layer of the model is a $d$ dimensional fully connected (FC) layer with $L_{2}$ normalization. In the case of $\mathcal{R}$Oxford/$\mathcal{R}$Paris, ResNet-101~\cite{hzr+16} is used, layer normalization is not added, while the FC layer is initialized with the result of whitening~\cite{rtc19}.

\paragraph{Training hyper-parameters.} For ResNet architectures, Adam optimizer~\cite{kb15} is used and for vision transformers, AdamW~\cite{lh+19} is used. We follow the standard class-balanced-sampling~\cite{mbl20,bxk+20,tdt20} with $4$ samples per class for all the datasets, while classes with less than $4$ samples are not used for training. Unless stated otherwise, the batch size for training is set $4,000$ for all datasets but Cars196 where it is equal to $4\times\#\text{classes}=392$. Following the setup of ProxyNCA++~\cite{tdt20}, the training set is split into training and validation by using the first half of the classes for training and the other half for validation. With this split, a grid search determines the learning rate, decay steps, decay size and the total number of epochs. Once the hyper-parameters are fixed, training is conducted once on the entire training set and evaluated on the test set. When training on GLDv1 and testing on $\mathcal{R}$Oxford/$\mathcal{R}$Paris, the batch size is set to $4096$~\cite{rar+19}, and training is performed for 500 batches, while other training hyper-parameters are set as in the work and GitHub implementation of Radenovic \etal~\cite{rtc19}. Note that the hyper-parameters for each dataset will be released with the implementation.

\paragraph{RS@k hyper-parameters.} The proposed Recall@k Surrogate (RS@k) loss \equ{smooth_recall} contains three hyper-parameters: sigmoid temperature $\tau_{2}$ - applied on similarity differences, sigmoid temperature $\tau_{1}$ - applied on ranks and the set of values for $k$ for which the loss is computed. Both sigmoid temperatures are kept fixed across all the experiments as $\tau_{2}=0.01$ (same as~\cite{bxk+20}) and $\tau_{1}=1$. The values of $k$ are kept fixed as $k=\{1,2,4,8,16\}$ without SiMix and $k=\{1,2,4,8,12,16,20,24,28,32\}$ with SiMix. For GLDv1~\cite{nas+17}, this is $k=\{1,2,4\}$, and $k=\{1,2,4,8\}$, respectively. The values of $k$ are studied in the supplementary materials and the sigmoid temperature $\tau_{1}$ are investigated in Section \ref{sec:effect_hyperparams}, where it is observed that the method is not very sensitive to these hyper-parameters. 

\paragraph{Large batch size.} To dispense with the GPU hardware constraints and manage to train with the large batch size, we follow the multistage back-propagation of Revaud~\etal~\cite{rar+19}. A forward pass is performed to obtain all embeddings while intermediate tensors are discarded from memory. Then, the loss is computed, and so are the gradients \wrt the embeddings. Finally, each of the embeddings is recomputed, this time allowing the propagation of the gradients. Note that there is no implementation online of this approach and that the code of this work will become publicly available. Algorithm \ref{alg:main} does not include such implementation details, but it is compatible with such an extension. The batch-size impact for the proposed RS@k loss function is validated in Section \ref{sec:effect_hyperparams}.

\paragraph{Discussion.} The methods in the literature use different embedding sizes, $d$, therefore, the models for the RS@k loss are trained with two embedding sizes of $d=128$ and $d=512$ for image retrieval benchmarks \cite{vms+18,ltw+16,ohb16,ksd+13}, and $d=2048$ for $\mathcal{R}$Oxford/$\mathcal{R}$Paris \cite{rit+18}, to allow a fair comparison. In the standard split, the image retrieval benchmarks~\cite{ksd+13,ohb16,ltw+16,vms+18} do not contain an explicit validation set; as a result, image retrieval methods often tune the hyper-parameters on the test set, leading to the issue of training with test set feedback. This issue has been studied in~\cite{mbl20}, which proposes to train different methods with identical hyper-parameters. The setup of~\cite{mbl20} is not directly usable for experiments with the RS@k loss, as large batch sizes are crucial to estimate recall@k accurately. Furthermore, their setup does not allow mixup. Therefore, instead of following~\cite{mbl20}, the issue is eliminated by using a part of the training set for validation as described above.

\subsection{Evaluation}
\label{sec:evaluation}

Unless otherwise stated, the results of the competing methods are taken from the original papers. Methods marked with a $\dag$ were trained by us, using the same implementation as used for the RS@k loss. The results on image retrieval benchmarks~\cite{ksd+13,ohb16,ltw+16,vms+18} are compared with the methods that use either ResNet-50~\cite{hzr+16} or Inception network~\cite{slj+15}. ResNet-50~\cite{hzr+16} is represented as $R_{50}^{d}$ in the tables and the standard Inception network~\cite{slj+15} as $I_{1}^{d}$, the Inception network with BatchNorm as  $I_{3}^{d}$ (same as ~\cite{tdt20}). Here $d$ is the embedding size. On all the datasets, the performance of the baseline, Smooth-AP (SAP)~\cite{bxk+20}, is also reported with Generalized mean pooling~\cite{rtc19} and layer normalization~\cite{bkh+16}, shown as SAP\textsuperscript{\dag} (+Gem +LN). This is to eliminate any performance boost in the comparisons that were caused by the architecture. Note that unless otherwise stated in our experiments, the batch size for SAP is set as $384$, the same as the original implementation~\cite{bxk+20}. Further, we demonstrate the performance of SAP and RS@k on ViT-B architectures. The variant of ViT-B that uses a patch size of $32\times32$ is denoted by ViT-B/32 and the one that uses a patch size of $16\times16$ by ViT-B/16.

\begin{table*}
\begin{center}
\footnotesize
\setlength{\tabcolsep}{5pt}
    \setlength\extrarowheight{+0pt}
    \begin{tabular}{@{\zsp}l@{\xssp}l@{\xssp}|cccc}
    \hline
    \multirow{3}[1]{*}{Method}& \multirow{3}[1]{*}{Arch.$^\text{dim}$} & \multicolumn{4}{c}{iNaturalist \cite{vms+18}} \\
    \cline{3-6}
    & & \multicolumn{4}{c}{r@k}\\ \cline{3-6}
    & & 1 & 4 & 16 & 32 \\
    \hline \hline
    
	ProxyNCA \cite{mtl+17} & {\scriptsize$I_{1}^{128}$} &
	$\underline{61.6}$ & 
	$\underline{77.4}$ & 
	$\underline{87.0}$ & 
	$90.6$ 
	\\
	
	Margin \cite{wms+17} & {\scriptsize$R_{50}^{128}$} &
	$58.1$ & 
	$75.5$ & 
	$86.8$ & 
	$\underline{90.7}$ 

	\\

	\hline
	RS@k\textsuperscript{\dag} & {\scriptsize$R_{50}^{128}$} &
	$69.3$  &
	$82.9$  &
	$90.6$  &
	$93.1$ 
	
	\\
	
	RS@k\textsuperscript{\dag} +SiMix & {\scriptsize$R_{50}^{128}$} &
	$\bm{69.6}$ & 
	$\bm{83.3}$ & 
	$\bm{91.2}$ & 
	$\bm{93.8}$ 
		\\

	& & 
	\tiny{\textcolor{blue}{$+21\%$}} &
	\tiny{\textcolor{blue}{$+26\%$}} &
	\tiny{\textcolor{blue}{$+32\%$}} &
	\tiny{\textcolor{blue}{$+33\%$}} 
	
	\\

    \hline \hline
	FastAP \cite{chx+19} & {\scriptsize$R_{50}^{512}$} &
	$60.6$ & 
	$77.0$ & 
	$87.2$ & 
	$90.6$ 
	\\
			
	Blackbox AP \cite{rmp+20} & {\scriptsize$R_{50}^{512}$} &
	$62.9$ &
	$79.0$ &
	$88.9$ &
	$92.1$ 
	\\

	SAP \cite{bxk+20} & {\scriptsize$R_{50}^{512}$} &
	$\underline{67.2}$ &
	$\underline{81.8}$ &
	$\underline{90.3}$ &
	$\underline{93.1}$ 
	
	\\
	
	SAP\textsuperscript{\dag} \cite{bxk+20} {\tiny +GeM +LN} & {\scriptsize$R_{50}^{512}$} &
	$68.7$ &
	$82.7$ &
	$90.9$ &
	$93.5$ 
	
	\\

	\hline	

	RS@k\textsuperscript{\dag} & {\scriptsize$R_{50}^{512}$} &
	$71.2$ &
	$84.0$ &
	$91.3$ &
	$93.6$ 
	
	\\

	RS@k\textsuperscript{\dag} +SiMix & {\scriptsize$R_{50}^{512}$} &
	$\bm{71.8}$ &
	$\bm{84.7}$ &
	$\bm{91.9}$ &
	$\bm{94.3}$ 
	
	\\
	
	& & 
	\tiny{\textcolor{blue}{$+14\%$}} &
	\tiny{\textcolor{blue}{$+16\%$}} &
	\tiny{\textcolor{blue}{$+16\%$}} &
	\tiny{\textcolor{blue}{$+17\%$}} 
	
	\\

    \hline \hline
    
    SAP\textsuperscript{\dag} \cite{bxk+20} & {\scriptsize ViT-B/32$^{512}$} &
    $72.2$ &
	$84.6$ &
	$91.6$ &
	$93.9$ 
	
	\\

    RS@k\textsuperscript{\dag} & {\scriptsize ViT-B/32$^{512}$} & 
    $75.9$ & 
    $87.1$ &
    $93.1$ &
    $95.1$ 

    \\
    
	\hline
	
    SAP\textsuperscript{\dag} \cite{bxk+20} & {\scriptsize ViT-B/16$^{512}$} &
	$79.1$ &
	$89.0$ &
	$94.2$ &
	$95.8$ 
	
	\\

    RS@k\textsuperscript{\dag} & {\scriptsize ViT-B/16$^{512}$} & 
    $83.9$ & 
    $92.1$ &
    $95.9$ &
    $97.2$ 
    
    \\
    \hline
    \end{tabular}
    \caption{Recall@$k(\%)$ on iNaturalist~\cite{vms+18}. Best results are shown with \textbf{bold}, previous state-of-the-art with \underline{underline} and relative gains over the state-of-the-art in \% of error reduction with \textcolor{blue}{blue} and relative declines in \textcolor{red}{red}. Methods marked with $\dag$ were trained using the same pipeline by us.}
    \label{tab:metriclearning_inat}
\end{center}
\end{table*}

\paragraph{iNaturalist.} The results on iNaturalist~\cite{vms+18} species recognition are presented in Table \ref{tab:metriclearning_inat}. The performances of the competing methods are taken from~\cite{bxk+20}, which uses the official implementations of these methods. It can be clearly seen that the RS@k outperforms classification and pairwise losses, including the three AP approximation losses, reaching the recall@1 score of $71.8\%$ with SiMix, an error reduction of $14\%$.

\begin{table*}
\begin{center}
\footnotesize
\setlength{\tabcolsep}{5pt}
    \setlength\extrarowheight{+0pt}
    \begin{tabular}{@{\zsp}l@{\xssp}l@{\xssp}|cccc}
    \hline
    \multirow{3}[1]{*}{Method}& \multirow{3}[1]{*}{Arch.$^\text{dim}$} & \multicolumn{4}{c}{SOP \cite{ohb16}} \\
    \cline{3-6}
    & & \multicolumn{4}{c}{r@k}\\ \cline{3-6}
   & & $10^0$ & $10^1$ & $10^2$ & $10^3$ \\
    \hline \hline
    
	ProxyNCA \cite{mtl+17} & {\scriptsize$I_{1}^{128}$} &
	$73.7$ & 
	- & 
	- & 
	- 	
	\\
	
	Margin \cite{wms+17} & {\scriptsize$R_{50}^{128}$} &
	$72.7$ &
	$86.2$ & 
	$93.8$ & 
	$98.0$ 
	\\
	
	Divide \cite{skp+15} & {\scriptsize$R_{50}^{128}$} &
	$75.9$ & 
	$88.4$ & 
	$94.9$ &
	$98.1$ 
	\\
	
	MIC \cite{rbo+19} & {\scriptsize$R_{50}^{128}$} &
	$77.2$ &
	$89.4$ &
	$95.6$ &
	-  
	\\
	
	Cont. w/M \cite{wzh+20} & {\scriptsize$R_{50}^{128}$} &
	$\underline{80.6}$ &
	$\underline{91.6}$ &
	$\underline{96.2}$ &
	$\underline{98.7}$ 
	\\
	
	\hline
	RS@k\textsuperscript{\dag} & {\scriptsize$R_{50}^{128}$} &
	$80.6$ &
	$91.6$ &
	$96.4$ &
	$\bm{98.8}$ 
	\\
	
	RS@k\textsuperscript{\dag} +SiMix & {\scriptsize$R_{50}^{128}$} &
	$\bm{80.9}$ &
	$\bm{91.7}$ &
	$\bm{96.5}$ &
	$\bm{98.8}$ 
	\\

	& & 
	\tiny{\textcolor{blue}{$+1.5\%$}} &
	\tiny{\textcolor{blue}{$+1.2\%$}} &
	\tiny{\textcolor{blue}{$+7.9\%$}} &
	\tiny{\textcolor{blue}{$+7.7\%$}} 
	\\
	
    \hline \hline
	FastAP \cite{chx+19} & {\scriptsize$R_{50}^{512}$} &
	$76.4$ &
	$89.0$ & 
	$95.1$ & 
	$98.2$ 
	\\
	
	MS \cite{whh+19} & {\scriptsize$I_{3}^{512}$} &
	$78.2$ &
	$90.5$ &
	$96.0$ &
	$98.7$ 
	\\
	
	NormSoftMax \cite{zw18} & {\scriptsize$R_{50}^{512}$} &
	$78.2$ &
	$90.6$ &
	$96.2$ &
	- 
	\\
	
	Blackbox AP \cite{rmp+20} & {\scriptsize$R_{50}^{512}$} &	
	$78.6$ &
	$90.5$ &
	$96.0$ &
	$98.7$ 	
	\\
	
	Cont. w/M \cite{wzh+20} & {\scriptsize$I_{3}^{512}$} &
	$79.5$ &
	$90.8$ &
	$96.1$ &
	$98.7$ 
	\\
	
	HORDE \cite{jph+19} & {\scriptsize$R_{50}^{512}$} &
	$80.1$ &
	$91.3$ &
	$96.2$ &
	- 
	\\
	
	ProxyNCA++ \cite{tdt20} & {\scriptsize$R_{50}^{512}$} &
	$\underline{80.7}$ &
	$\underline{92.5}$ &
	$96.7$ &
	$98.9$ 
	\\

	SAP \cite{bxk+20} & {\scriptsize$R_{50}^{512}$} &
	$80.1$ &
	$91.5$ &
	$\underline{96.6}$ & 
	$\underline{99.0}$ 
	\\
	
	SAP\textsuperscript{\dag} \cite{bxk+20} {\tiny +GeM +LN} & {\scriptsize$R_{50}^{512}$} &
	$80.3$ &
	$92.0$ &
	$96.9$ &
	$99.0$ 
	\\
    
	\hline

	RS@k\textsuperscript{\dag} & {\scriptsize$R_{50}^{512}$} &	
	$\bm{82.8}$ & 
	$\bm{92.9}$ & 
	$\bm{97.0}$ & 
	$99.0$ 
	\\

	RS@k\textsuperscript{\dag} +SiMix & {\scriptsize$R_{50}^{512}$} &	
	$82.1$ & 
	$92.8$ &
	$\bm{97.0}$ &
	$\bm{99.1}$ 
	\\
	
	& & 	
	\tiny{\textcolor{blue}{$+11\%$}} &
	\tiny{\textcolor{blue}{$+5.3\%$}} &
	\tiny{\textcolor{blue}{$+12\%$}} &
	\tiny{\textcolor{blue}{$+10\%$}} 
	\\

    \hline \hline
    
    SAP\textsuperscript{\dag} \cite{bxk+20} & {\scriptsize ViT-B/32$^{512}$} &
        $83.7$ &
	$94.0$ &
	$97.8$ &
	$99.3$ 
	\\

    RS@k\textsuperscript{\dag} & {\scriptsize ViT-B/32$^{512}$} & 

    $85.1$ & 
    $94.6$ &
    $98.0$ &
    $99.3$ 

    \\

	\hline
	
    SAP\textsuperscript{\dag} \cite{bxk+20} & {\scriptsize ViT-B/16$^{512}$} &
	
	$86.6$ &
	$95.4$ &
	$98.4$ &
	$99.5$ 
	\\

    RS@k\textsuperscript{\dag} & {\scriptsize ViT-B/16$^{512}$} & 
    $88.0$ & 
    $96.1$ &
    $98.6$ &
    $99.6$
    
    \\

    \hline
    \end{tabular}
    \caption{Recall@$k(\%)$ on Stanford Online Products (SOP)~\cite{ohb16}. Best results are shown with \textbf{bold}, previous state-of-the-art with \underline{underline} and relative gains over the state-of-the-art in \% of error reduction with \textcolor{blue}{blue} and relative declines in \textcolor{red}{red}. Methods marked with $\dag$ were trained using the same pipeline by us.}
    \label{tab:metriclearning_sop}
\end{center}
\end{table*}

\paragraph{SOP.} The performance on SOP~\cite{ohb16} is presented in Table \ref{tab:metriclearning_sop}, along with the comparisons with the competing methods. The proposed RS@k loss demonstrates clear state-of-the-art results, surpassing ProxyNCA++~\cite{tdt20} by $2.0\%$ on recall@1, an error reduction of $10.4\%$. If a smaller batch size, equal to $384$, is used for RS@k, it reaches a performance of $81.2\%$, $92.2\%$, $96.9\%$ and $99.0\%$ on r@$10^{0}$, r@$10^{1}$, r@$10^{2}$ and r@$10^{3}$ respectively. This result shows that large batch size helps in improving the performance, but RS@k outperforms the competing methods even with smaller batch size.

\begin{table*}
\begin{center}
\footnotesize
\setlength{\tabcolsep}{5pt}
    \setlength\extrarowheight{+0pt}
    \begin{tabular}{@{\zsp}l@{\xssp}l@{\xssp}|cc|cc|cc}
    \hline
    \multirow{3}[1]{*}{Method}& \multirow{3}[1]{*}{Arch.$^\text{dim}$} & \multicolumn{6}{c}{VehicleID \cite{ltw+16}} \\
    \cline{3-8}
    & & \multicolumn{6}{c}{r@k}\\ \cline{3-8}
    & & \multicolumn{2}{c|}{Small} & \multicolumn{2}{c|}{Medium} & \multicolumn{2}{c}{Large} \\
    & & 1 & 5 & 1 & 5 & 1 & 5 \\
    \hline \hline    
	Divide \cite{skp+15} & {\scriptsize$R_{50}^{128}$} &
    $87.7$ &
    $92.9$ &
    $85.7$ &
    $90.4$ &
    $82.9$ &
    $90.2$ 
	\\
	
	MIC \cite{rbo+19} & {\scriptsize$R_{50}^{128}$} &
    $86.9$ &
    $93.4$ &
    - &
    - &
    $82.0$ &
    $91.0$ 
	\\
	
	Cont. w/M \cite{wzh+20} & {\scriptsize$R_{50}^{128}$} &
    $\underline{94.7}$ &
    $\underline{96.8}$ &
    $\underline{93.7}$ &
    $\underline{95.8}$ &
    $\underline{93.0}$ &
    $\underline{95.8}$ 
	\\
	
	\hline
	RS@k\textsuperscript{\dag} & {\scriptsize$R_{50}^{128}$} &
    $\bm{95.6}$ &
    $\bm{97.8}$ &
    $\bm{94.4}$ &
    $\bm{96.8}$ &
    $\bm{93.5}$ &
    $\bm{96.6}$     
	\\

	RS@k\textsuperscript{\dag} +SiMix & {\scriptsize$R_{50}^{128}$} &
    $95.4$ &
    $97.5$ &
    $93.8$ &
    $96.6$ &
    $93.0$ &
    $96.2$ 
	\\

	& & 
	\tiny{\textcolor{blue}{$+17\%$}} &
	\tiny{\textcolor{blue}{$+31\%$}} &
	\tiny{\textcolor{blue}{$+11\%$}} &
	\tiny{\textcolor{blue}{$+24\%$}} &
	\tiny{\textcolor{blue}{$+7.1\%$}} &
	\tiny{\textcolor{blue}{$+19\%$}} 
	\\
	
    \hline \hline
	FastAP \cite{chx+19} & {\scriptsize$R_{50}^{512}$} &
    $91.9$ &
    $96.8$ &
    $90.6$ &
    $95.9$ &
    $87.5$ &
    $95.1$     
	\\
	
	Cont. w/M \cite{wzh+20} & {\scriptsize$I_{3}^{512}$} &
    $94.6$ &
    $96.9$ &
    $\underline{93.4}$ &
    $96.0$ &
    $\underline{93.0}$ &
    $96.1$ 
	\\
		
	SAP \cite{bxk+20} & {\scriptsize$R_{50}^{512}$} &
    $\underline{94.9}$ &
    $\underline{97.6}$ &
    $93.3$ &
    $\underline{96.4}$ &
    $91.9$ &
    $\underline{96.2}$ 
	\\
	
	SAP\textsuperscript{\dag} \cite{bxk+20} {\tiny +GeM +LN} & {\scriptsize$R_{50}^{512}$} &
    $94.2$ &
    $97.2$ &
    $92.7$ &
    $96.2$ &
    $91.0$ &
    $95.8$ 
	\\
 
	\hline

	RS@k\textsuperscript{\dag} & {\scriptsize$R_{50}^{512}$} &
    $\bm{95.7}$ &
    $\bm{97.9}$ &
    $\bm{94.6}$ &
    $\bm{96.9}$ &
    $\bm{93.8}$ &
    $\bm{96.6}$ 
	\\

	RS@k\textsuperscript{\dag} +SiMix & {\scriptsize$R_{50}^{512}$} &
    $95.3$ &
    $97.7$ &
    $94.2$ &
    $96.5$ &
    $93.3$ &
    $96.4$ 
	\\
	
	& & 
	\tiny{\textcolor{blue}{$+16\%$}} &
	\tiny{\textcolor{blue}{$+13\%$}} &
	\tiny{\textcolor{blue}{$+18\%$}} &
	\tiny{\textcolor{blue}{$+14\%$}} &
	\tiny{\textcolor{blue}{$+11\%$}} &
	\tiny{\textcolor{blue}{$+10\%$}} 
	\\

    \hline \hline
    
    SAP\textsuperscript{\dag} \cite{bxk+20} & {\scriptsize ViT-B/32$^{512}$} &
	$94.8$ &
	$97.7$ &
	$93.5$ &
	$96.8$ &
	$92.1$ &
	$96.3$ 	
	\\

    RS@k\textsuperscript{\dag} & {\scriptsize ViT-B/32$^{512}$} & 
    $95.1$ & 
    $97.7$ &
    $94.1$ &
    $96.7$ &
    $93.2$ &
    $96.5$     
    \\

	\hline
	
    SAP\textsuperscript{\dag} \cite{bxk+20} & {\scriptsize ViT-B/16$^{512}$} &
	$95.5$ &
	$97.7$ &
	$94.2$ &
	$96.9$ &
	$93.1$ &
	$96.6$ 
	\\

    RS@k\textsuperscript{\dag} & {\scriptsize ViT-B/16$^{512}$} & 
    $96.2$ & 
    $98.0$ &
    $95.2$ &
    $97.2$ &
    $94.7$ &
    $97.1$ 
    \\
    \hline
    \end{tabular}
    \caption{Recall@$k(\%)$ on PKU VehicleID~\cite{ltw+16}. Best results are shown with \textbf{bold}, previous state-of-the-art with \underline{underline} and relative gains over the state-of-the-art in \% of error reduction with \textcolor{blue}{blue} and relative declines in \textcolor{red}{red}. Methods marked with $\dag$ were trained using the same pipeline by us.}
    \label{tab:metriclearning_vid}
\end{center}
\end{table*}

\paragraph{VehicleID.} The results on VehicleID~\cite{ltw+16} are presented in Table \ref{tab:metriclearning_vid}. RS@k outperforms the competing methods both with and without SiMix. Better results were observed without SiMix where RS@k reaches recall@1 performance of $95.7\%$, $94.6\%$ and $93.8\%$ on the small, medium, and large test sets, respectively.

\begin{table*}
\begin{center}
\footnotesize
\setlength{\tabcolsep}{5pt}
    \setlength\extrarowheight{+0pt}
    \begin{tabular}{@{\zsp}l@{\xssp}l@{\xssp}|cccc}
    \hline
    \multirow{3}[1]{*}{Method}& \multirow{3}[1]{*}{Arch.$^\text{dim}$} & \multicolumn{4}{c}{Cars196 \cite{ksd+13}}\\
    \cline{3-6}
    & & \multicolumn{4}{c}{r@k}\\ \cline{3-6}
    & & 1 & 2 & 4 & 8\\
    \hline \hline
    
	ProxyNCA \cite{mtl+17} & {\scriptsize$I_{1}^{128}$} &
	$73.2$ &
	$82.4$ &
	$86.4$ &
	$88.7$
	\\
	
	Margin \cite{wms+17} & {\scriptsize$R_{50}^{128}$} &
	$\underline{79.6}$ &
	$\underline{86.5}$ &
	$\underline{91.9}$ &
	$\underline{95.1}$
	\\
				
	\hline
	RS@k\textsuperscript{\dag} & {\scriptsize$R_{50}^{128}$} &
    $78.1$ &
    $85.8$ &
    $91.1$ &
    $94.5$
	\\
 
	RS@k\textsuperscript{\dag} +SiMix & {\scriptsize$R_{50}^{128}$} &
    $\bm{84.7}$ &
    $\bm{90.9}$ &
    $\bm{94.7}$ &
    $\bm{96.9}$ 
	\\

	& & 
	\tiny{\textcolor{blue}{$+25\%$}} &
	\tiny{\textcolor{blue}{$+33\%$}} &
	\tiny{\textcolor{blue}{$+35\%$}} &
	\tiny{\textcolor{blue}{$+37\%$}}
	\\

    \hline \hline	
	MS \cite{whh+19} & {\scriptsize$I_{3}^{512}$} &
	$84.1$ &
	$90.4$ &
	$94.0$ &
	$96.1$
	\\
	
	NormSoftMax \cite{zw18} & {\scriptsize$R_{50}^{512}$} &
	$84.2$ &
	$90.4$ &
	$94.4$ &
	$96.9$
	\\
			
	HORDE \cite{jph+19} & {\scriptsize$R_{50}^{512}$} &
	$86.2$ &
	$91.9$ &
	$95.1$ &
	$97.2$
	\\
	
	ProxyNCA++ \cite{tdt20} & {\scriptsize$R_{50}^{512}$} &
	$\underline{86.5}$ &
	$\underline{92.5}$ &
	$\underline{95.7}$ &
	$\underline{\bm{97.7}}$
	\\

	SAP \cite{bxk+20} & {\scriptsize$R_{50}^{512}$} &
    $76.1$ &
    $84.3$ &
    $89.8$ &
    $93.8$
	\\
	
	SAP\textsuperscript{\dag} \cite{bxk+20} {\tiny +GeM +LN} & {\scriptsize$R_{50}^{512}$} &    
    $78.2$ &
    $85.6$ &
    $90.8$ &
    $94.3$
	\\
    
	\hline
	RS@k\textsuperscript{\dag} & {\scriptsize$R_{50}^{512}$} &
    $80.7$ &
    $88.3$ &
    $92.8$ &
    $95.7$
	\\

	RS@k\textsuperscript{\dag} +SiMix & {\scriptsize$R_{50}^{512}$} &
    $\bm{88.2}$ &
    $\bm{93.0}$ &
    $\bm{95.9}$ &
    $97.4$
	\\
	
	& & 
	\tiny{\textcolor{blue}{$+13\%$}} &
	\tiny{\textcolor{blue}{$+6.7\%$}} &
	\tiny{\textcolor{blue}{$+4.7\%$}} &
	\tiny{\textcolor{red}{$-13\%$}}
	\\

    \hline \hline
    
    SAP\textsuperscript{\dag} \cite{bxk+20} & {\scriptsize ViT-B/32$^{512}$} &
	$78.1$ &
	$85.7$ &
	$91.0$ &
	$94.8$
	\\

    RS@k\textsuperscript{\dag} & {\scriptsize ViT-B/32$^{512}$} & 
    $78.1$ &
    $86.4$ &
    $92.3$ &
    $95.6$
    \\

	\hline
	
    SAP\textsuperscript{\dag} \cite{bxk+20} & {\scriptsize ViT-B/16$^{512}$} &
	$86.2$ &
	$92.1$ &
	$95.1$ &
	$97.2$
	\\

    RS@k\textsuperscript{\dag} & {\scriptsize ViT-B/16$^{512}$} & 
    $89.5$ &
    $94.2$ &
    $96.6$ &
    $98.3$
    \\
    \hline
    \end{tabular}
    \caption{Recall@$k(\%)$ on Stanford Cars (Cars196)~\cite{ksd+13}. Best results are shown with \textbf{bold}, previous state-of-the-art with \underline{underline} and relative gains over the state-of-the-art in \% of error reduction with \textcolor{blue}{blue} and relative declines in \textcolor{red}{red}. Methods marked with $\dag$ were trained using the same pipeline by us.}
    \label{tab:metriclearning_cars196}
\end{center}
\end{table*}

\paragraph{Cars196.} Evaluation on a small scale dataset, Cars196~\cite{ksd+13} is presented in the Table \ref{tab:metriclearning_cars196}. We train SAP with a batch size of $392$; it provides a performance of $79.5\%$, $86.6\%$, $91.2\%$, and $94.4\%$ and when combined with SiMix a performace of $85.4\%$, $91.0\%$, $94.3\%$ and $96.7\%$ on r@$1$, r@$2$, r@$4$ and r@$8$ respectively. SiMix makes a large difference in performance for both RS@k and SAP~\cite{bxk+20}, primarily because of a smaller batch size ($392$), as constrained by the low number of classes. With SiMix, RS@k reaches the state-of-the-art results on three out of four recall@k values. If the batch size is further increased to $588$ by changing the number of samples per class from $4$ to $6$, then RS@k provides a larger gain with performance $88.3\%$, $93.3\%$, $95.9\%$ and $97.6\%$.

\paragraph{Results with ViT-B.} The results by replacing the ResNet-50~\cite{hzr+16} backbone with a ViT-B~\cite{dbk+21} for SAP~\cite{bxk+20} and the proposed RS@k are also shown in Tables \ref{tab:metriclearning_inat}, \ref{tab:metriclearning_sop}, \ref{tab:metriclearning_vid} and \ref{tab:metriclearning_cars196}. With an exception of ViT-B/32 on VehicleID and Cars196 datasets, the use of ViT-B backbone leads to better performance for both methods, compared to the ResNet counterpart. It can be clearly seen that RS@k outperforms SAP~\cite{bxk+20} on all datasets. ViT-B/16 when trained with RS@k shows unprecedented performance on all datasets reaching recall@$1$ score of $83.9\%$ on iNaturalist~\cite{vms+18}, $88.0\%$ on SOP~\cite{ohb16}, $96.2\%$ on VehicleID~\cite{ltw+16} (small) and $89.5\%$ on Cars196~\cite{ksd+13}. Note that while ResNet-50 has $24.5$ M parameters and operates with $8.12$ GMac/image, ViT-B has $87.8$ M parameters and operates with $4.36$ and $16.8$ GMac/image for ViT-B/32 and ViT-B/16 respectively.

\paragraph{Concurrent work.} The method of learning intra-batch connections for deep metric learning~\cite{sel21} achieves r@1 of $81.4\%$ on the SOP and $88.1\%$ on Cars196 dataset. The approach for Grouplet embedding learning~\cite{zlx+21} obtains r@1 of $82.0\%$ on SOP and $91.5\%$ on Cars196. The metric mixup approach~\cite{vpa+21} reports the best results of $81.3\%$ r@1 on SOP in combination with ProxyNCA++~\cite{tdt20} and $89.6\%$ on Cars196 which is in combination with MS~\cite{whh+19}.

\begin{table*}
\tablestyle{3pt}{1.3}
\begin{center}
\resizebox{\textwidth}{!}{
\setlength\extrarowheight{+0pt}
\newcommand{\xdagger}{^{\dagger}}
\def\arraystretch{1.0}%  1 is the default, row height
\small
\begin{tabular}	{l@{\msp}r@{\msp}r@{\msp}r@{\msp}c@{\msp}c@{\msp}c@{\msp}c@{\msp}}
\hline
\multirow{2}{*}{Arch.} & 
\multirow{2}{*}{Loss} & 
\multirow{2}{*}{Train-set} & 
\multirow{2}{*}{} & 
\multicolumn{2}{c}{\ro} &
\multicolumn{2}{c}{\ro\hspace{-3pt}+\rdis}
\\ 
\cline{5-8}

& & & & med  & hard  & med  & hard \\ 

\hline \hline

GeM$\ast$ & 
AP \cite{hls18} & 
Landmarks-clean~\cite{bsc+14}\cite{gar+17} & 
~\cite{rar+19}/~\cite{tjc20} & 
$67.1$ & 
$42.3$ & 
$47.8$ & 
$22.5$  
\\  

GeM$\ast$ & 
AP \cite{hls18} & 
GLDv1~\cite{nas+17} & 
~\cite{rar+19}/github & 
$66.3$ & 
$42.5$ & 
- & 
-    
\\

GeM\dag &
SAP \cite{bxk+20} & 
GLDv1~\cite{nas+17} & 
~\cite{bxk+20} & 
$67.9$ & 
$46.3$ & 
$49.5$ & 
$25.8$ 
\\     

GeM\dag & 
RS@k & 
GLDv1~\cite{nas+17} & 
ours & 
$68.3$ & 
$46.1$ & 
$50.1$ & 
$25.8$ 
\\

GeM+SiMix\dag & 
RS@k & 
GLDv1~\cite{nas+17} & 
ours & 
$68.4$ & 
$45.3$ & 
$51.0$ & 
$26.4$
\\            
 \hline
\end{tabular}
}
\caption{Performance comparison (mAP\%) on $\mathcal{R}$Oxford with 1m distractor images ($\mathcal{R}$1m). $\ast$ denotes that the FC layer is not part of the training but is added afterward to implement whitening. Batch size is 4096 for all methods; SiMix virtually increases it to 10240. ResNet101 is used as a backbone for all methods.
\label{tab:rop_oxford}}
\end{center}
\end{table*}

\begin{table*}
\tablestyle{3pt}{1.3}
\begin{center}
\resizebox{\textwidth}{!}{
\setlength\extrarowheight{+0pt}
\newcommand{\xdagger}{^{\dagger}}
\def\arraystretch{1.0}%  1 is the default, row height
\small
\begin{tabular}	{l@{\msp}r@{\msp}r@{\msp}r@{\msp}c@{\msp}c@{\msp}c@{\msp}c@{\msp}}
\hline
\multirow{2}{*}{Arch.} & 
\multirow{2}{*}{Loss} & 
\multirow{2}{*}{Train-set} & 
\multirow{2}{*}{} & 
\multicolumn{2}{c}{\rpa} & 
\multicolumn{2}{c}{\rp\hspace{-3pt}+\rdis}     
\\ 
\cline{5-8}

& & & & med  & hard  & med  & hard \\ 

\hline \hline

GeM$\ast$ & 
AP \cite{hls18} & 
Landmarks-clean~\cite{bsc+14}\cite{gar+17} & 
~\cite{rar+19}/~\cite{tjc20} & 
$80.3$ & 
$60.9$ & 
$51.9$ & 
$24.6$ 
\\  

GeM$\ast$ & 
AP \cite{hls18} & 
GLDv1~\cite{nas+17} & 
~\cite{rar+19}/github & 
$80.2$ & 
$60.8$ &
-  
& - 
\\

GeM\dag &
SAP \cite{bxk+20} & 
GLDv1~\cite{nas+17} & 
~\cite{bxk+20} & 
$81.7$ &
$63.3$ &
$57.4$ & 
$29.8$ 
\\     

GeM\dag & 
RS@k & 
GLDv1~\cite{nas+17} & 
ours & 
$82.1$ &
$63.9$ & 
$57.9$ & 
$30.2$ 
\\

GeM+SiMix\dag & 
RS@k & 
GLDv1~\cite{nas+17} & 
ours & 
$81.2$ & 
$62.4$ & 
$58.7$ & 
$31.1$
\\            
 \hline
\end{tabular}
}
\caption{Performance comparison (mAP\%) on $\mathcal{R}$Paris with 1m distractor images ($\mathcal{R}$1m). $\ast$ denotes that the FC layer is not part of the training but is added afterward to implement whitening. Batch size is 4096 for all methods; SiMix virtually increases it to 10240. ResNet101 is used as a backbone for all methods.
\label{tab:rop_paris}}
\end{center}
\end{table*}

\begin{table*}
\tablestyle{3pt}{1.3}
\begin{center}
\resizebox{\textwidth}{!}{
\setlength\extrarowheight{+0pt}
\newcommand{\xdagger}{^{\dagger}}
\def\arraystretch{1.0}%  1 is the default, row height
\small
\begin{tabular}	{l@{\msp}r@{\msp}r@{\msp}r@{\msp}c@{\msp}c@{\msp}}
\hline
\multirow{2}{*}{Arch.} & 
\multirow{2}{*}{Loss} & 
\multirow{2}{*}{Train-set} & 
\multirow{2}{*}{} & 
\multicolumn{2}{c}{Mean} 
\\ 
\cline{5-6}

& & & & all & \rdis \\ 

\hline \hline

GeM$\ast$ & 
AP \cite{hls18} & 
Landmarks-clean~\cite{bsc+14}\cite{gar+17} & 
~\cite{rar+19}/~\cite{tjc20} & 
$49.7$ & 
$36.7$ 
\\  

GeM$\ast$ & 
AP \cite{hls18} & 
GLDv1~\cite{nas+17} & 
~\cite{rar+19}/github & 
- & 
- 
\\

GeM\dag &
SAP \cite{bxk+20} & 
GLDv1~\cite{nas+17} & 
~\cite{bxk+20} & 
$52.7$ & 
$40.6$ 
\\     

GeM\dag & 
RS@k & 
GLDv1~\cite{nas+17} & 
ours & 
$53.1$ & 
$41.0$ 
\\

GeM+SiMix\dag & 
RS@k & 
GLDv1~\cite{nas+17} & 
ours & 
$53.1$ & 
$41.8$ 
\\            
 \hline
\end{tabular}
}
\caption{Performance comparison (mAP\%) on $\mathcal{R}$Oxford and $\mathcal{R}$Paris with 1m distractor images ($\mathcal{R}$1m). Mean performance is reported across all setups or the large-scale setups only. $\ast$ denotes that the FC layer is not part of the training but is added afterward to implement whitening. Batch size is 4096 for all methods; SiMix virtually increases it to 10240. ResNet101 is used as a backbone for all methods.
\label{tab:rop_mean}}
\end{center}
\end{table*}

\paragraph{$\mathcal{R}$Oxford/$\mathcal{R}$Paris.} Tables~\ref{tab:rop_oxford}, \ref{tab:rop_paris}, and \ref{tab:rop_mean} summarizes a comparison with AP-based losses in the literature on $\mathcal{R}$Oxford/$\mathcal{R}$Paris with and without distractor images. The comparison is performed with GLDv1 as a training set whose performance is reported for the work of Revaud \etal~\cite{rar+19} in their GitHub page, while the \emph{landmarks-clean dataset} is avoided as all initial images are not publicly available at the moment. During the training performed by us, training images are down-sampled to have a maximum resolution of $1024 \times1024$. 
The inference is performed with multi-resolution descriptors at three scales with up-sampling and down-sampling by a factor of $\sqrt{2}$.
Note that SAP is not evaluated on these datasets in the original work and this experiment is performed by us, which outperforms the previously used AP loss~\cite{hls18}. RS@k, with or without the SiMix, increases the performance by a small margin.

\subsection{Effect of hyper-parameters}
\label{sec:effect_hyperparams}

We study the impact of hyper-parameter on the Cars196 dataset~\cite{ksd+13} since it is the smallest compared to the others and has the lowest training time. 

\begin{figure}
\centering
\raisebox{3pt}{
\input{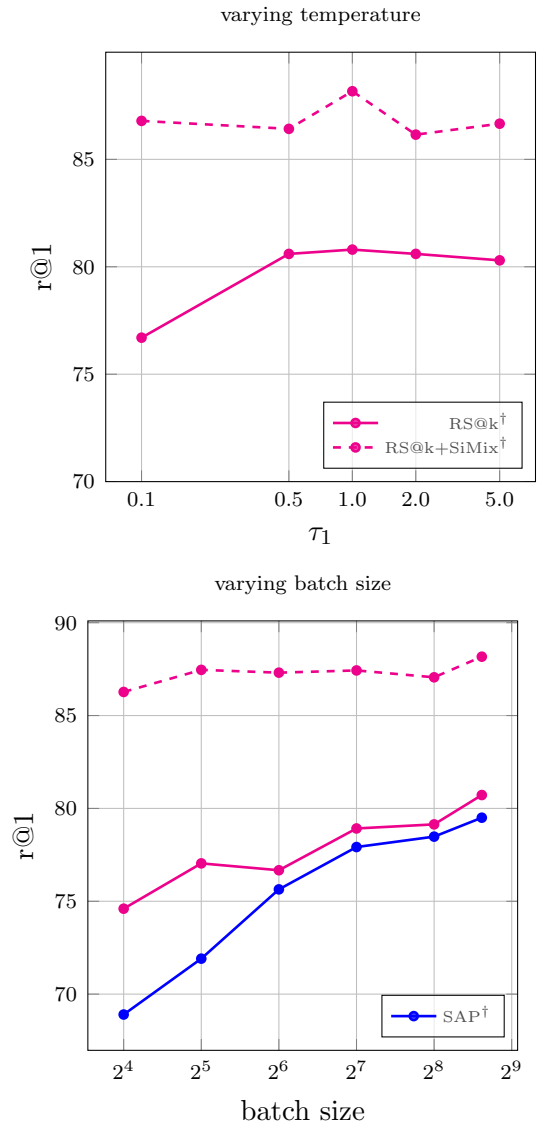}
\begin{tikzpicture}
\begin{axis}[%
	width=0.5\linewidth,
	height=0.5\linewidth,
	xlabel={\small $\tau_{1}$},
	ylabel={\small r@1},
	title={varying temperature},
	legend cell align={left},
	legend pos=south east,
    legend style={cells={anchor=east}, font =\tiny, fill opacity=0.7, row sep=-2.5pt},
   	xtick={0.1,0.5,1.0,2.0,5.0},
   	xticklabels={0.1,0.5,1.0,2.0,5.0},
   	xmode=log,
   	ymin=70,
    grid=both,
]
	\addplot[color=magenta,     solid, mark=*,  mark size=1.5, line width=1.0] table[x=temp, y expr={\thisrow{cars_r1}}] 
	\yfccLambda;
	\addlegendentry{RS@k$^{\dag}$};
	\addplot[color=magenta,     dashed, mark=*,  mark size=1.5, line width=1.0] table[x=temp, y expr={\thisrow{cars_simix_r1}}] \yfccLambda;
	\addlegendentry{RS@k+SiMix$^{\dag}$};
\end{axis}
\end{tikzpicture}
}
\pgfplotstableread{
 		bs_sop	sop_r1     bs_cars     cars_r1      cars_simix_r1   cars_r1_sap
 		16		72.9	    16          74.60       86.27           68.9
 		32		77.3	    32          77.04       87.46           71.91
 		64		79.0	    64          76.67       87.31           75.64
 		128		79.7	    128         78.92       87.43           77.92
 		256		80.5        256         79.14       87.06           78.48
 		512     81.2        392         80.72       88.17           79.50
 		1024    82.0        nan         nan         nan             nan
 		2048    82.5        nan         nan         nan             nan
 		4096    82.8        nan         nan         nan             nan
 		8192    82.4        nan         nan         nan             nan
 		% 16384   80.97       nan         nan         nan             nan
 	}{\yfccLambda}
\begin{tikzpicture}
\begin{axis}[%
	width=0.5\linewidth,
	height=0.5\linewidth,
	xlabel={\small batch size},
	ylabel={\small r@1},
	title={varying batch size},
	legend cell align={left},
	legend pos=south east,
    legend style={cells={anchor=east}, font =\tiny, fill opacity=0.7, row sep=-2.5pt},
   	xtick={16,32,64,128,256,512,1024,2048,4096},
   	xmode=log,
   	log basis x={2},
    grid=both,
]
	\addplot[color=blue,     solid, mark=*,  mark size=1.5, line width=1.0] table[x=bs_cars, y expr={\thisrow{cars_r1_sap}}]
	\yfccLambda;
	\addlegendentry{SAP$^{\dag}$};
	\addplot[color=magenta,     solid, mark=*,  mark size=1.5, line width=1.0] table[x=bs_cars, y expr={\thisrow{cars_r1}}] 
	\yfccLambda;
	\addplot[color=magenta,     dashed, mark=*,  mark size=1.5, line width=1.0] table[x=bs_cars, y expr={\thisrow{cars_simix_r1}}]
	\yfccLambda;
\end{axis}
\end{tikzpicture}
\caption{The effect of sigmoid temperature $\tau_{1}$ applied on ranks (top) and of batch size (bottom). Results are shown on Cars196~\cite{ksd+13}.}
\label{fig:ab_tau1_bs}
\end{figure}

\paragraph{Sigmoid temperature $\tau_{1}$ - applied on ranks.}
The effect of the sigmoid temperature $\tau_{1}$ is summarized in Figure \ref{fig:ab_tau1_bs} (top). For both setups of with and without SiMix, $\tau_{1}=1.0$ gives best results while higher and lower values lead to a decline.

\paragraph{Batch size.}
The effect of the varying batch size is shown in Figure \ref{fig:ab_tau1_bs} (bottom). It demonstrates that large batch size leads to better results. A significant performance boost is observed with the use of SiMix, especially in the small batch size regime, which comes at a small extra computation. A comparison with SAP~\cite{bxk+20} is also shown in this figure. Note that on smaller batch sizes, the proposed RS@k outperforms SAP with a larger margins.

\paragraph{Values for {\em k}.} The study for the set of values of {\em k} used for RS@k loss can be found in Table \ref{tab:Cars196_ablation}. The results RS@$\{1\}$, RS@$\{1,2\}$, RS@$\{1,2,4\}$ and RS@$\{1,2,4,8\}$ suggest that adding larger values of {\em k} leads to decline in the performance. However, RS@$\{1,2,4,8,16\}$ gives on an average the same results as RS@$\{1\}$, with higher performance on larger {\em k} values. Comparing the entries RS@$\{4,8,16\}$ with RS@$\{1,2,4,8,16\}$ suggests that the use of small values, such as $k=1$ or $k=2$, is crucial as the performance drops significantly when these values are removed. Further removing $k=4$ (RS@$\{8,16\}$) does not change the performance. However, removing $k=8$ (RS@$\{16\}$) leads to a significant decline in the performance.

\begin{table}[H]
  \tablestyle{3pt}{1}
    \setlength\extrarowheight{-2pt}
   \begin{tabular}{l|c|c|c|c|c|c}
    \hline
    \multicolumn{1}{l}{Method} &  r@1   & r@2 &  r@4  & r@8 & r@16 & Avg\\
    	\hline\hline
    	
        RS@$\{1\}$\textsuperscript{\dag} &
        $81.1$ &
        $87.7$ &
        $92.0$ &
        $95.0$ &
        $96.9$ &
        $90.5$
        \\
        
        RS@$\{1,2\}$\textsuperscript{\dag} &
        $80.2$ &
        $87.2$ &
        $91.9$ &
        $95.0$ &
        $97.2$ &
        $90.3$
        \\
	
	    RS@$\{1,2,4\}$\textsuperscript{\dag} &
        $79.6$ &
        $86.5$ &
        $91.2$ &
        $94.5$ &
        $96.8$ &
        $89.7$
        \\
        
        RS@$\{1,2,4,8\}$\textsuperscript{\dag} &
        $79.3$ &
        $86.3$ &
        $91.0$ &
        $94.5$ &
        $96.9$ &
        $89.6$
        \\
        
        RS@$\{1,2,4,8,16\}$\textsuperscript{\dag} &
        $80.8$ &
        $87.6$ &
        $92.2$ &
        $95.0$ &
        $97.1$ &
        $90.5$
        \\
        
        RS@$\{2,4,8,16\}$\textsuperscript{\dag} &
        $80.3$ &
        $87.5$ &
        $92.3$ &
        $95.4$ &
        $97.5$ &
        $90.6$
        \\
        
        RS@$\{4,8,16\}$\textsuperscript{\dag} &
        $79.6$ &
        $87.1$ &
        $91.7$ &
        $95.0$ &
        $97.3$ &
        $90.1$
        \\
        RS@$\{8,16\}$\textsuperscript{\dag} &
        $79.6$ &
        $87.1$ &
        $91.7$ &
        $95.0$ &
        $97.3$ &
        $90.1$
        \\
        RS@$\{16\}$\textsuperscript{\dag} &
        $75.8$ & 
        $83.9$ &
        $89.8$ &
        $93.6$ &
        $96.4$ &
        $87.9$
        \\
	\hline
    \end{tabular}
    \caption{Varying the set of values of {\em k}. Results on Cars196~\cite{ksd+13}. In all experiments, $\tau_{1}=1$ and $\tau_{2}=0.01$.}
    \label{tab:Cars196_ablation}
\end{table}

\paragraph{Impact of SiMix}
Our results suggest that SiMix leads to a larger performance gain on smaller datasets, where batch size is restricted by the total number of classes. Results are summarized in Table \ref{tab:sap_vs_rsk}, where we additionally report results on CUB which has small (100) number of training classes. On Cars196 dataset, RS@k attains a r@1 of $80.7\%$ without and $88.2\%$ with SiMix (an absolute improvement for $7.5\%$). Similarly on CUB200, RS@k attains a r@1 of $63.8\%$ without and $69.5\%$ with SiMix (an absolute improvement of $5.7\%$).

\begin{table}[H]
\setlength\extrarowheight{-0pt}
\begin{center}
\resizebox{\textwidth}{!}{
\begin{tabular}{l|l|l|l|l}
    \hline
    \textbf{Dataset} & \textbf{\# Training Samples} & \textbf{SAP\textsuperscript{\dag}} & \textbf{RS@k\textsuperscript{\dag}} & \textbf{RS@k\textsuperscript{\dag}} + SiMix \\
    \hline
    iNaturalist & $325,846$ & $70.7$ & $71.2$ & $71.8$\\
    VehicleID & $110,178$ & $95.5$ & $95.7$ & $95.3$\\
    SOP & $59,551$ & $81.3$ & $82.8$ & $82.1$\\
    Cars196 & $8,054$ & $79.5$ & $80.7$ & $88.2$\\
    CUB200 & $5,864$ & $63.6$ & $63.8$ & $69.5$\\
    \hline
    $\mathcal{R}$Oxf \& $\mathcal{R}$Par ($1$m) & $1,060,709$ & $40.6$ & $41.0$ & $41.8$\\
    \hline
\end{tabular}}
\end{center}
\caption{Recall@1 (in $\%$) with batch size of $\min(4000, 4\times \#\text{classes})$ for iNaturalist, VehicleID, SOP, Cars196 and CUB200. mAP (in $\%$) with batch size of $4096$ for $\mathcal{R}$Oxford and  $\mathcal{R}$Paris with $1$ million distractor samples.}
\label{tab:sap_vs_rsk}
\end{table}

\section{Experiments on Fine-Grained Classification}

This section compares training a softmax image classifier explicitly and training an image retrieval system, which is subsequently used for nearest neighbour classification. 
The resolution of images, pre-trained weights and number of training epochs are kept the same across the two setups for a fair comparison.

\begin{figure}
\includegraphics[width=.99\textwidth]{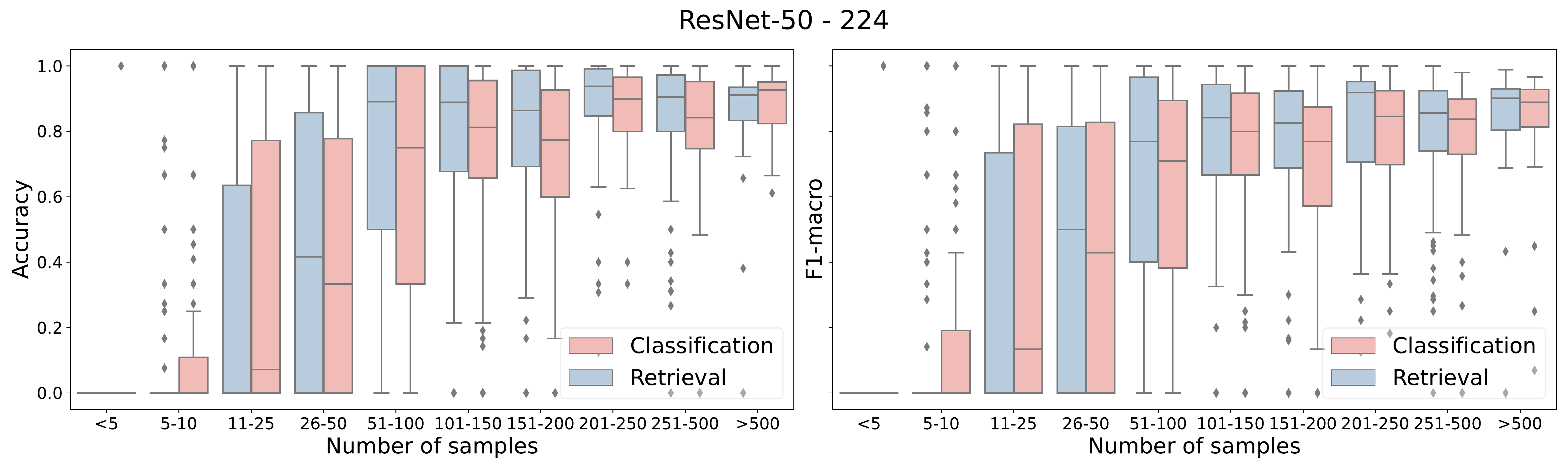}
\includegraphics[width=.99\textwidth]{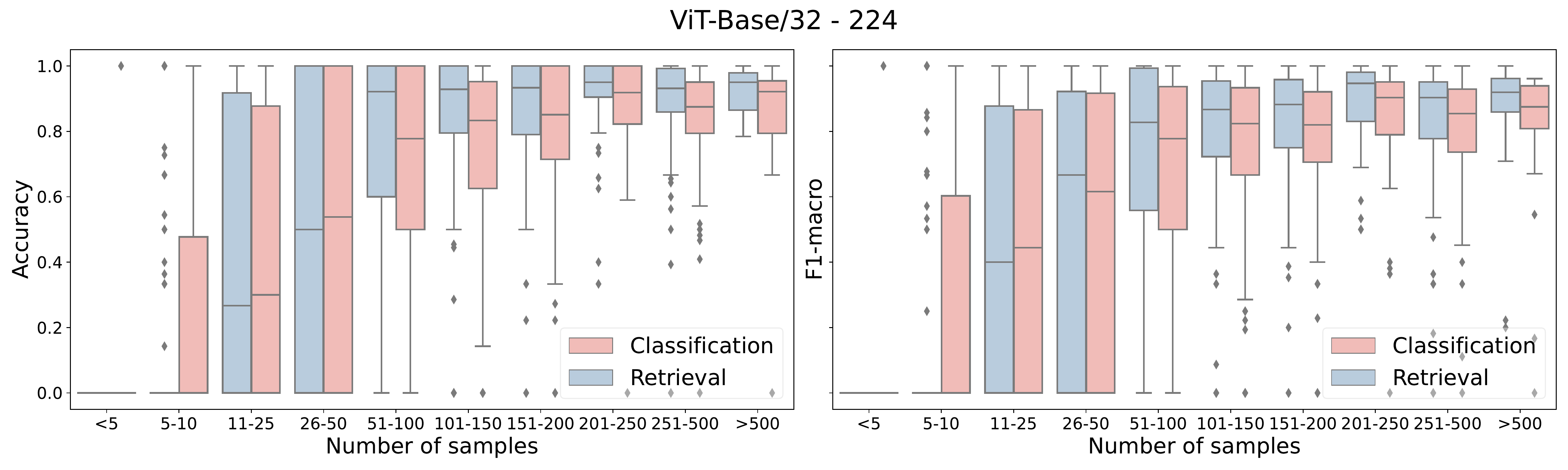}
\includegraphics[width=.99\textwidth]{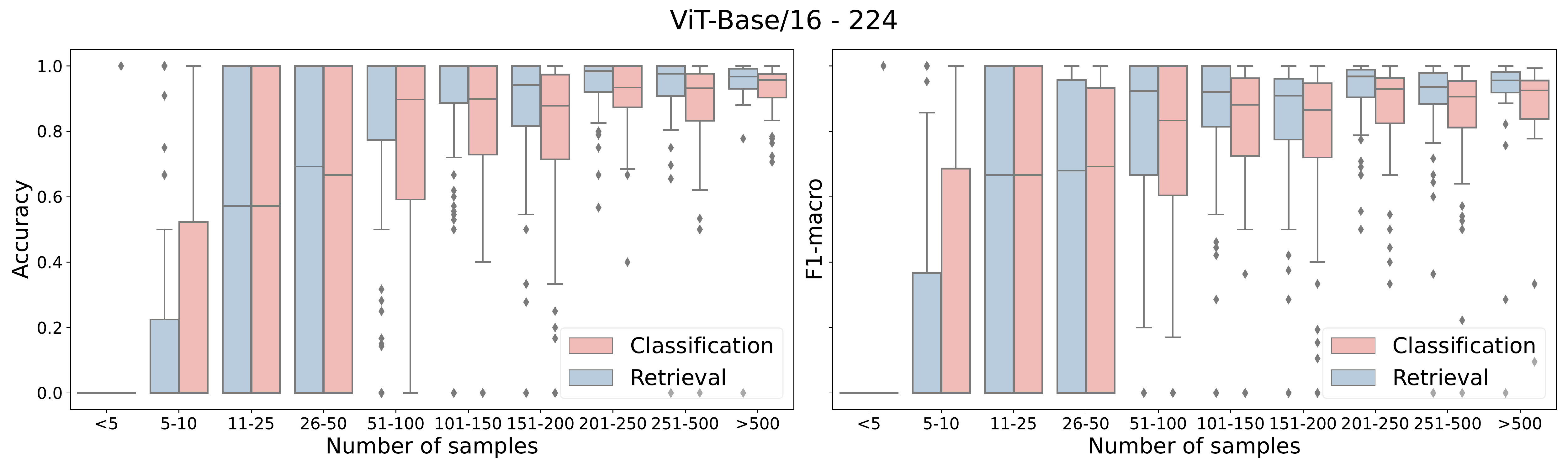}
\caption{Classification performance (F1 and Accuracy) as box-plot for three backbone architectures and Classification Retrieval approaches. Tested on PlantCLEF2017 test set with input resolution of $224\times224$.}
\label{fig:perf_boxplot}
\end{figure}

Overall, the proposed retrieval approach achieved superior performance in all measured scenarios. Notably, the ViT-Base/16 feature extractor architecture achieved a higher classification accuracy with a margins of $0.28\%$, $4.13\%$, and $10.25\%$ on ExpertLifeCLEF 2018~\cite{goeau2018overview}, PlantCLEF 2017~\cite{goeau2017plant} and iNat2018--Plantae~\cite{vms+18}, respectively.
Besides, the macro-F1 performance differences margin is noticeably higher---$1.85\%$ for ExpertLifeCLEF 2018 and $12.23\%$ for iNat2018--Plantae datasets.
Even though the standard classification approach performs better on classes with fewer samples (See Table\,\ref{fig:perf_boxplot}), common species with high a-prior probability are frequently wrongly predicted. 
This is primarily due to the high-class imbalance preserved in the dataset mimicked by the deep neural network optimized via SoftMax Cross-Entropy Loss.
Thus, the results of the standard image classification approach performs way worst in case of the macro-F1 score.
Full comparison of the classification and retrieval-based methods and their appropriate recognition scores are listed in Table\,\ref{table:results_classification_retrieval}. Three architectures---ResNet-50, ViT-Base/32, and ViT-Base/16 are evaluated. It can be seen from the results that for all selected architectures, retrieval leads to a better performance.

\begin{table}[tbh]
\centering
\renewcommand{\arraystretch}{1.3}
\caption{Classification (C) vs Retrieval (R). All models were trained for 100 epochs with fixed image size ($224\times224$). No test-time augmentations were used. The most confident image prediction is used for all images belonging to the same observation.}
\resizebox{\textwidth}{!}{
\begin{tabular}{ l c  c c  c c  c c }
\multicolumn{2}{c}{} & \multicolumn{2}{c}{ExpertLifeCLEF 2018} & \multicolumn{2}{c}{PlantCLEF 2017} & \multicolumn{2}{c}{iNat2018--Plantae}  \\
 \cline{3-5}
 \cline{6-8}
    \multicolumn{1}{l}{Architecture} & \multicolumn{1}{c}{Method} & Acc. &  macro F1 & Acc &  macro F1 & Acc &  macro F1 \\
  \hline
    ResNet-50       & C &   59.87 & 55.11 & 77.89 & 54.48 & 57.73 & 52.69\\ % CHECKED
    ViT-Base/32     & C &   65.21 & 60.29 & 80.68 & 59.18 & 57.24 & 53.17 \\ % CHECKED 
    ViT-Base/16     & C &   71.71 & 67.35 & 84.48 & 65.40 & 67.42 & 64.51 \\ 
    \hline
    ResNet-50       & R &   60.15 & 56.30 & 80.27 & 55.57 & 57.95 & 56.32 \\ % CHECKED
    ViT-Base/32     & R &   66.48 & 61.49 & 84.89 & 60.79 & 63.12 & 61.24 \\ % CHECKED
    ViT-Base/16     & R &   \textbf{71.99} & \textbf{69.20} &  \textbf{88.61} & \textbf{66.39} & \textbf{77.67} & \textbf{76.74} \\
    \hline
\end{tabular}}
\label{table:results_classification_retrieval}
\end{table}

\begin{figure}
\includegraphics[width=.99\textwidth]{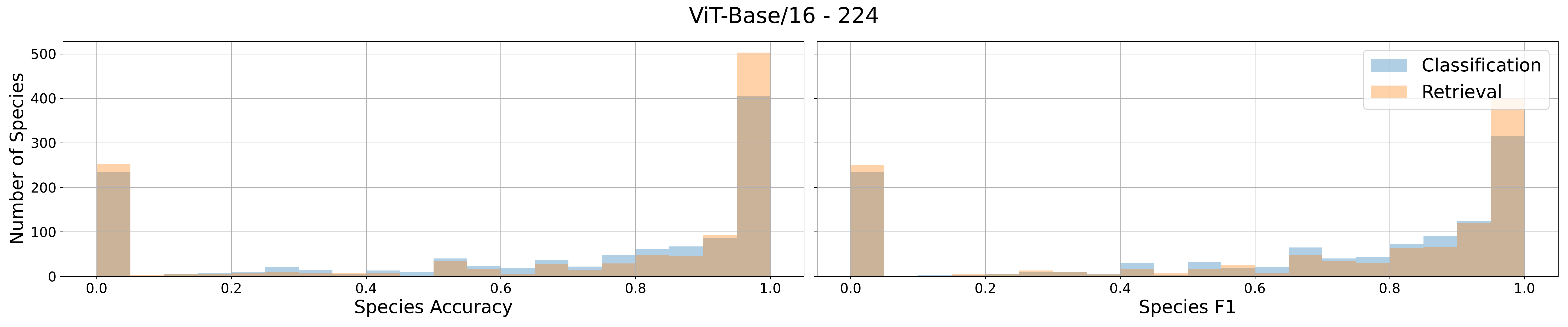}
\caption{Species-wise classification performance histogram within a given 5\% interval, evaluated on the PlantCLEF2017 test set with ViT-Base/16 backbone and Classification and Retrieval approaches.}
\label{fig:perf_histogram}
\end{figure}

\begin{figure}[b]
  \centering
  \setlength{\tabcolsep}{2pt}
  \def\arraystretch{1.05}
\resizebox{\textwidth}{!}{
  \begin{tabular}{c|ccccc}
    Target & Top 1 & Top 2 & Top 3 & Top 4 & Top 5 \\
    \setlength{\fboxsep}{0pt}%
    \setlength{\fboxrule}{0.75pt}%
    \fcolorbox{black}{white}{\includegraphics[width=2.65cm, height=2.25cm]{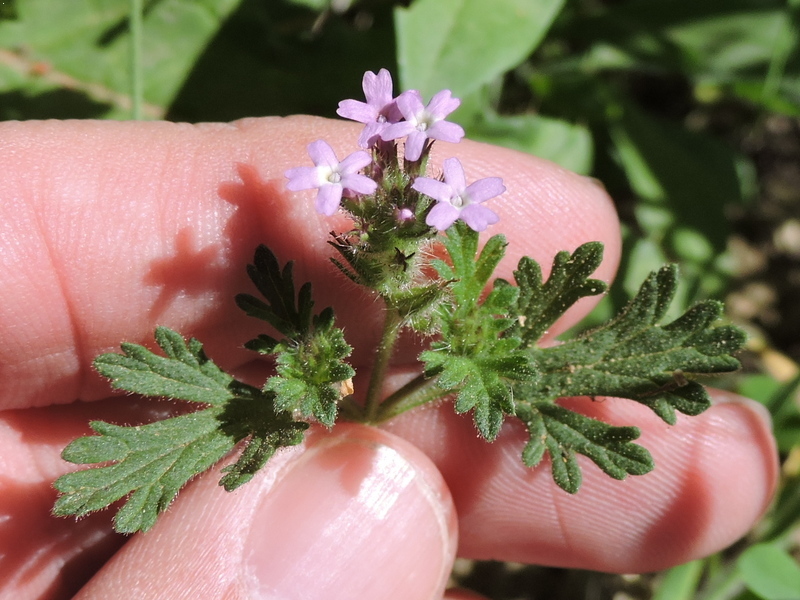}} \hspace{0.75pt} & \hspace{0.75pt}
    \setlength{\fboxsep}{0pt}%
    \setlength{\fboxrule}{0.75pt}%
    \fcolorbox{green}{white}{\includegraphics[width=2.65cm, height=2.25cm]{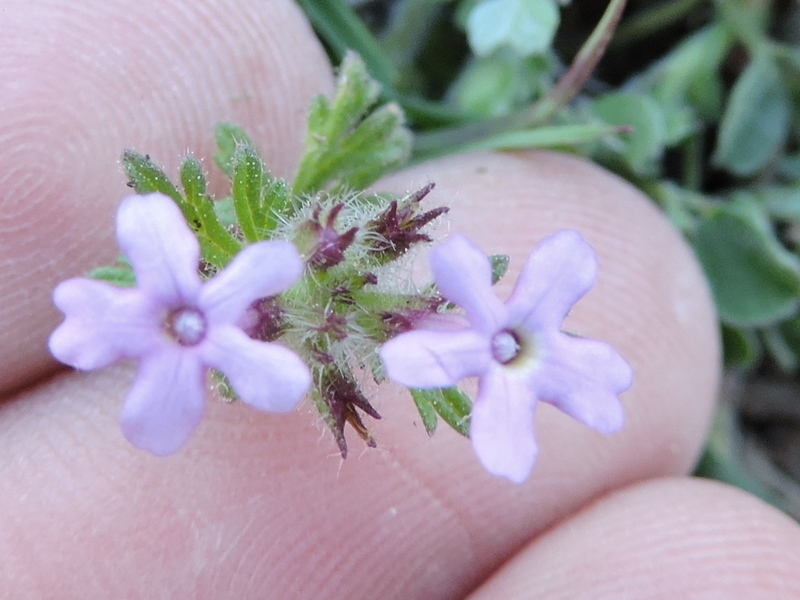}} &
    \setlength{\fboxsep}{0pt}%
    \setlength{\fboxrule}{0.75pt}%
    \fcolorbox{green}{white}{\includegraphics[width=2.65cm, height=2.25cm]{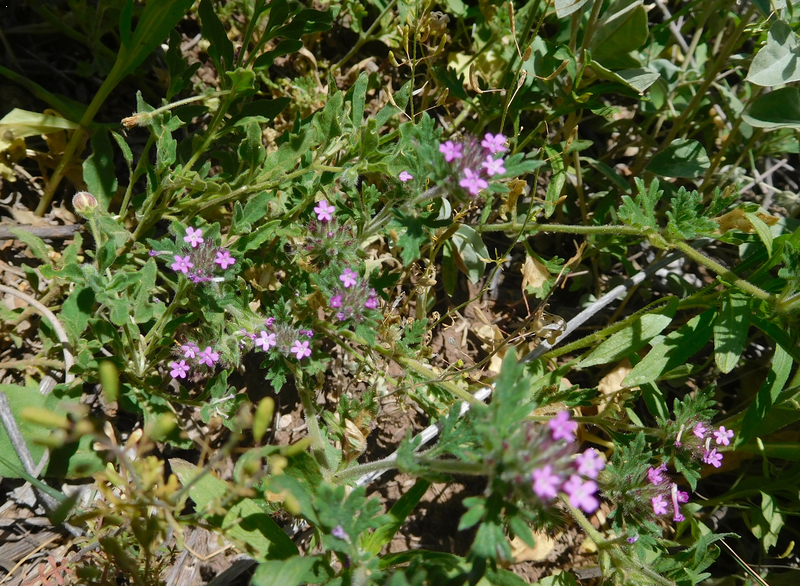}} &
    \setlength{\fboxsep}{0pt}%
    \setlength{\fboxrule}{0.75pt}%
    \fcolorbox{green}{white}{\includegraphics[width=2.65cm, height=2.25cm]{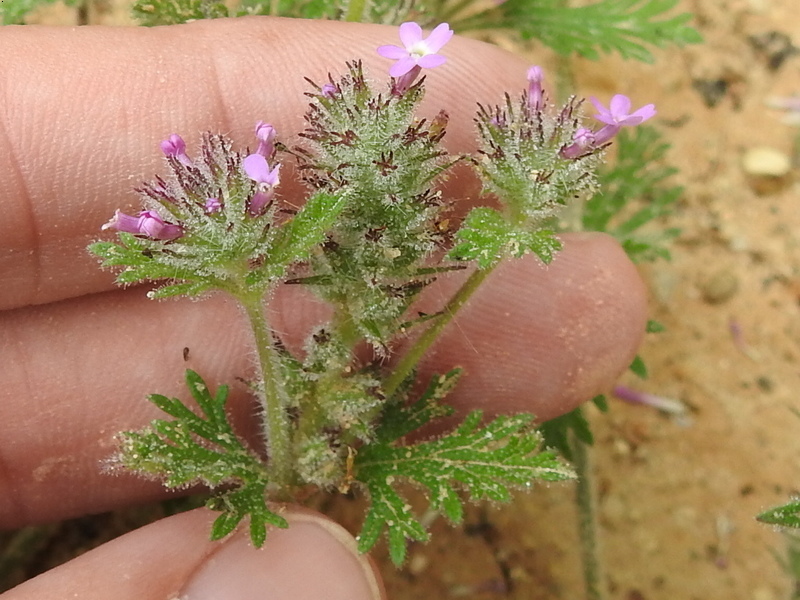}} &
    \setlength{\fboxsep}{0pt}%
    \setlength{\fboxrule}{0.75pt}%
    \fcolorbox{red}{white}{\includegraphics[width=2.65cm, height=2.25cm]{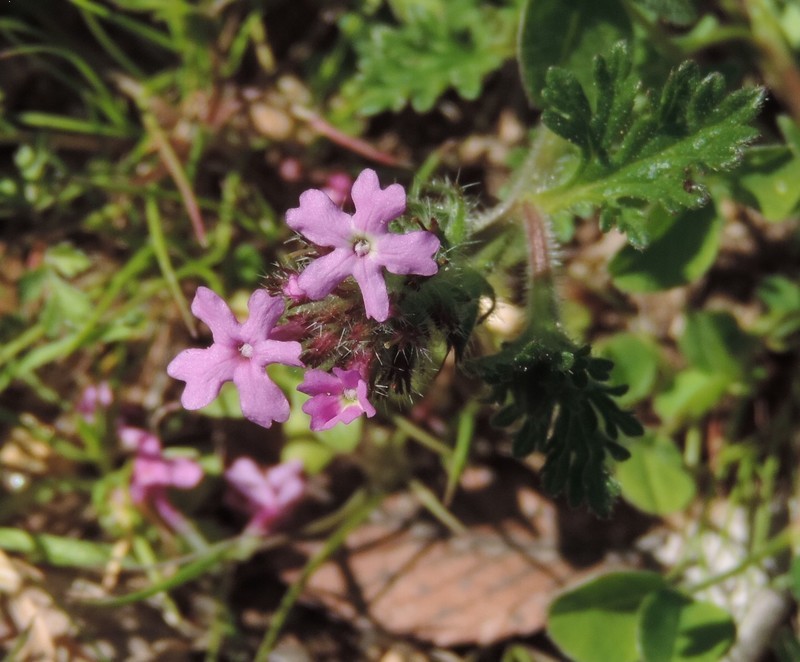}} &
    \setlength{\fboxsep}{0pt}%
    \setlength{\fboxrule}{0.75pt}%
    \fcolorbox{green}{white}{\includegraphics[width=2.65cm, height=2.25cm]{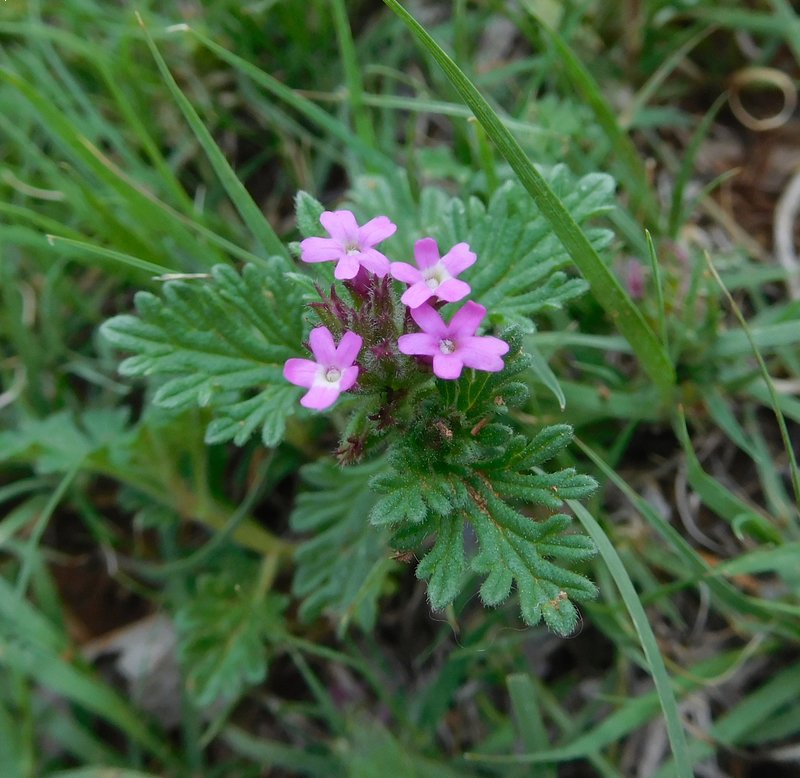}} \\
    
    \setlength{\fboxsep}{0pt}%
    \setlength{\fboxrule}{0.75pt}%
    \fcolorbox{black}{white}{\includegraphics[width=2.65cm, height=2.25cm]{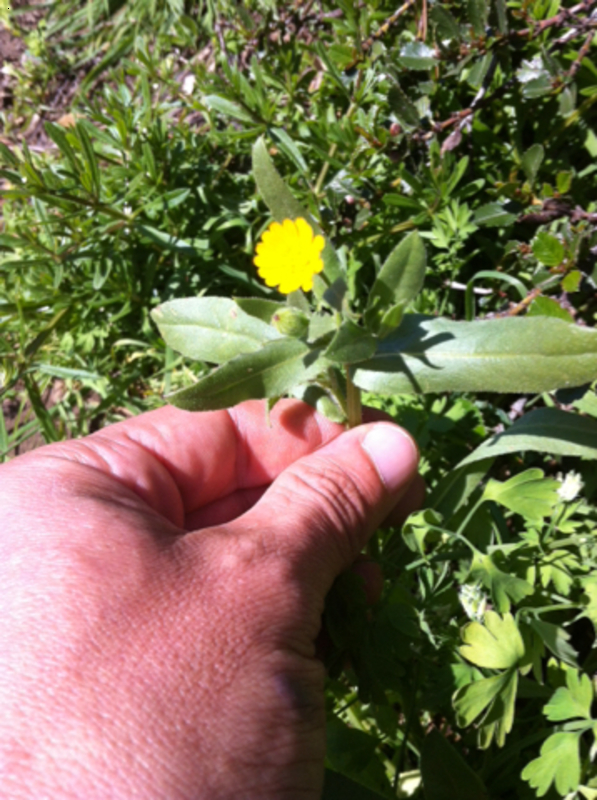}} \hspace{0.75pt} & \hspace{0.75pt}
    \setlength{\fboxsep}{0pt}%
    \setlength{\fboxrule}{0.75pt}%
    \fcolorbox{green}{white}{\includegraphics[width=2.65cm, height=2.25cm]{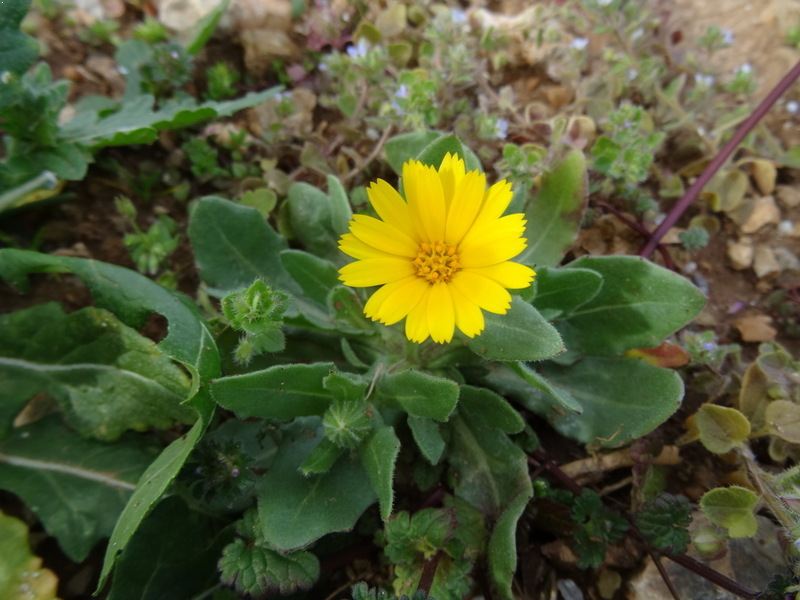}} &
    \setlength{\fboxsep}{0pt}%
    \setlength{\fboxrule}{0.75pt}%
    \fcolorbox{green}{white}{\includegraphics[width=2.65cm, height=2.25cm]{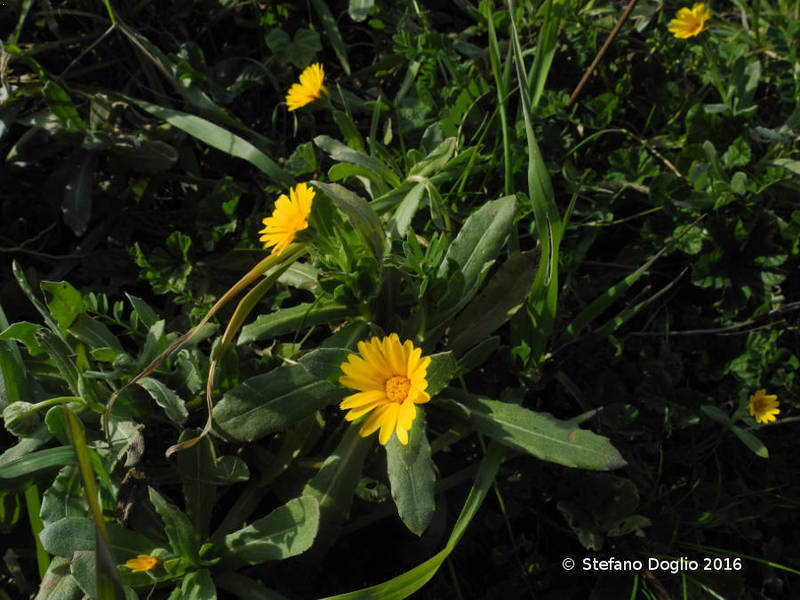}} &
    \setlength{\fboxsep}{0pt}%
    \setlength{\fboxrule}{0.75pt}%
    \fcolorbox{green}{white}{\includegraphics[width=2.65cm, height=2.25cm]{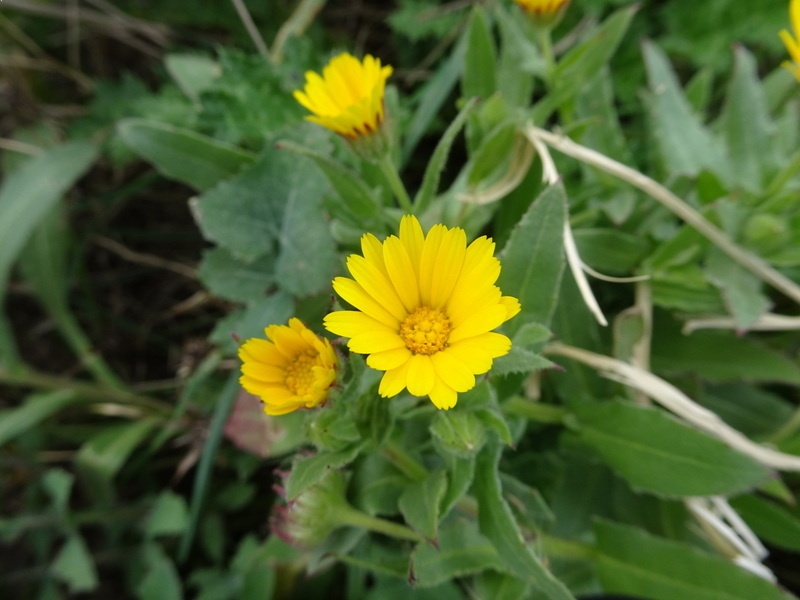}} &
    \setlength{\fboxsep}{0pt}%
    \setlength{\fboxrule}{0.75pt}%
    \fcolorbox{red}{white}{\includegraphics[width=2.65cm, height=2.25cm]{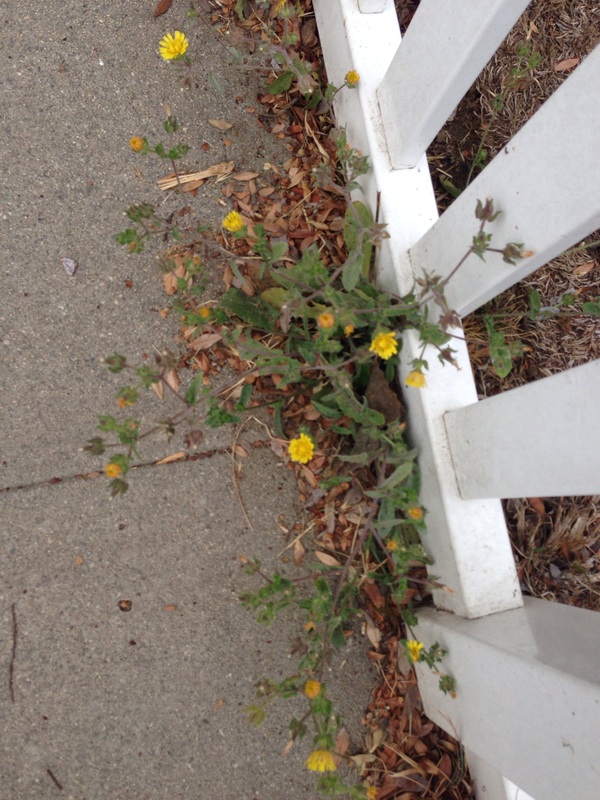}} &
    \setlength{\fboxsep}{0pt}%
    \setlength{\fboxrule}{0.75pt}%
    \fcolorbox{green}{white}{\includegraphics[width=2.65cm, height=2.25cm]{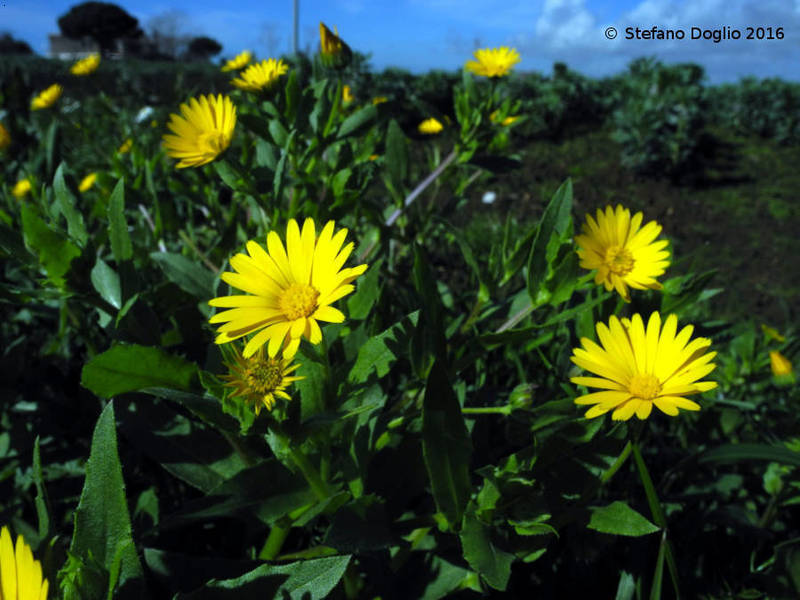}} \\
    
    \setlength{\fboxsep}{0pt}%
    \setlength{\fboxrule}{0.75pt}%
    \fcolorbox{black}{white}{\includegraphics[width=2.65cm, height=2.25cm]{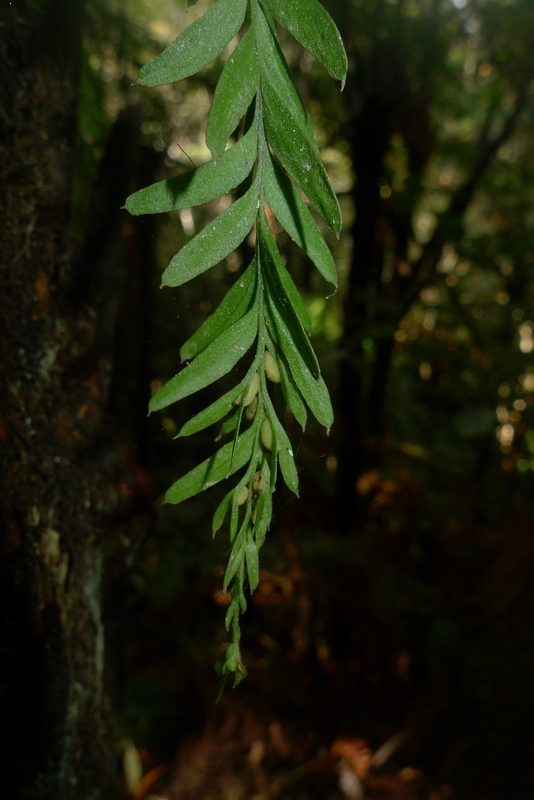}} \hspace{0.75pt} & \hspace{0.75pt}
    \setlength{\fboxsep}{0pt}%
    \setlength{\fboxrule}{0.75pt}%
    \fcolorbox{green}{white}{\includegraphics[width=2.65cm, height=2.25cm]{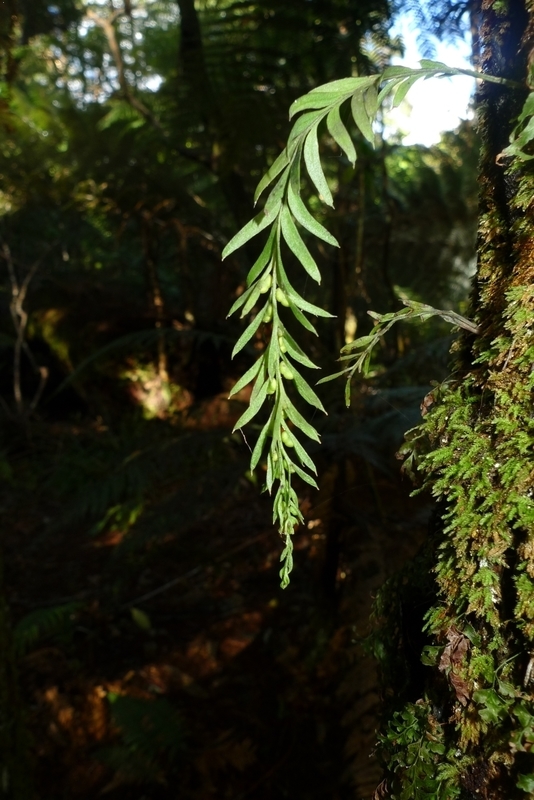}} &
    \setlength{\fboxsep}{0pt}%
    \setlength{\fboxrule}{0.75pt}%
    \fcolorbox{green}{white}{\includegraphics[width=2.65cm, height=2.25cm]{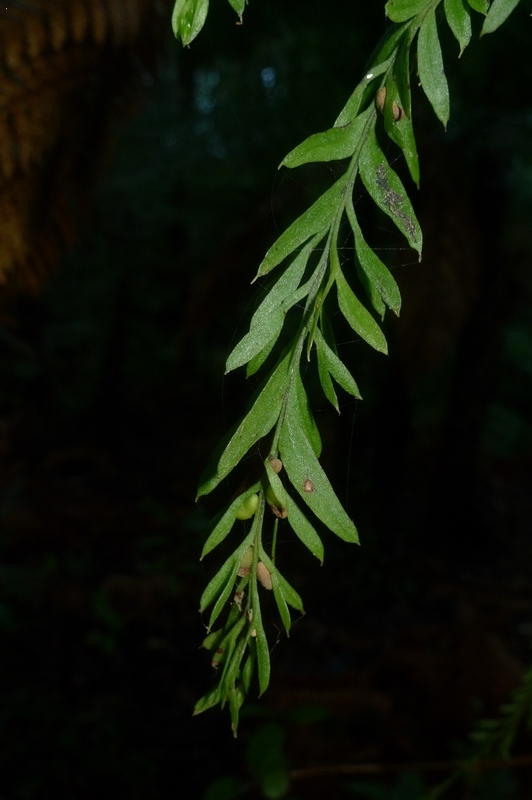}} &
    \setlength{\fboxsep}{0pt}%
    \setlength{\fboxrule}{0.75pt}%
    \fcolorbox{green}{white}{\includegraphics[width=2.65cm, height=2.25cm]{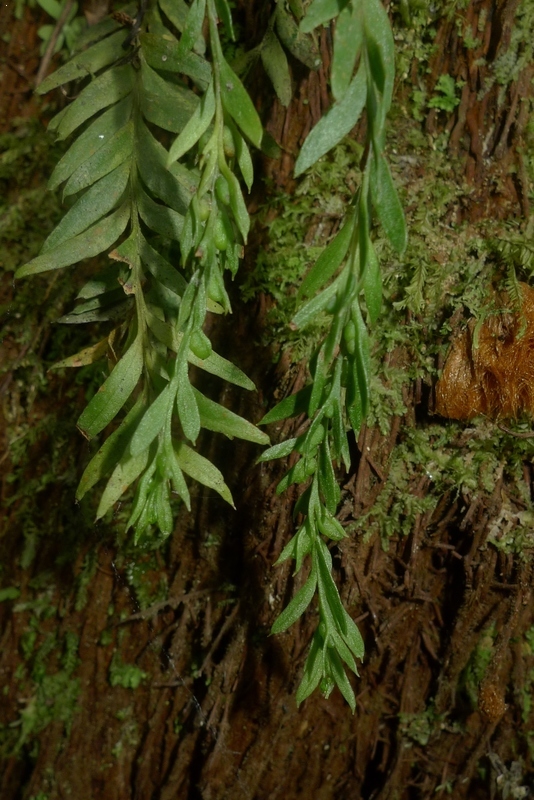}} &
    \setlength{\fboxsep}{0pt}%
    \setlength{\fboxrule}{0.75pt}%
    \fcolorbox{red}{white}{\includegraphics[width=2.65cm, height=2.25cm]{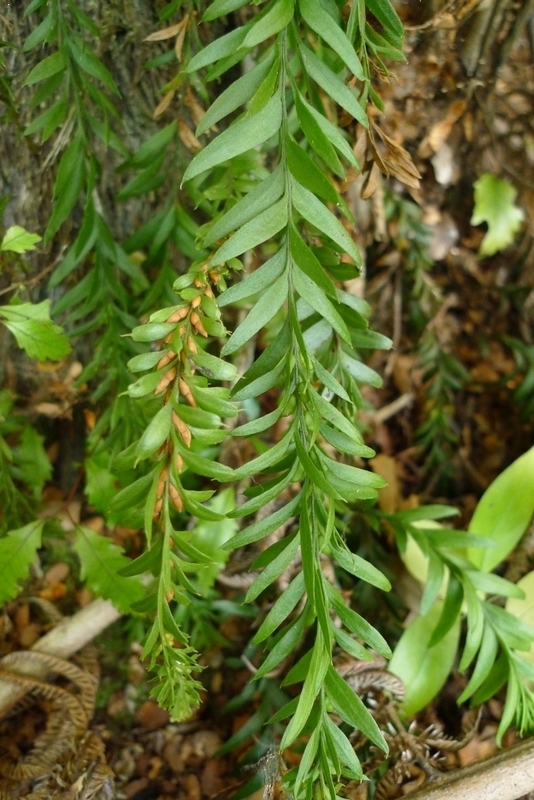}} &
    \setlength{\fboxsep}{0pt}%
    \setlength{\fboxrule}{0.75pt}%
    \fcolorbox{green}{white}{\includegraphics[width=2.65cm, height=2.25cm]{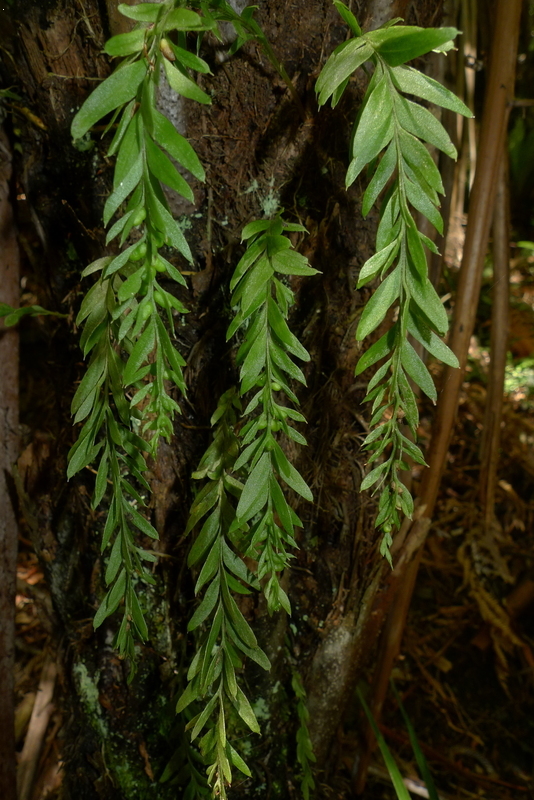}} \\

    \setlength{\fboxsep}{0pt}%
    \setlength{\fboxrule}{0.75pt}%
    \fcolorbox{black}{white}{\includegraphics[width=2.65cm, height=2.25cm]{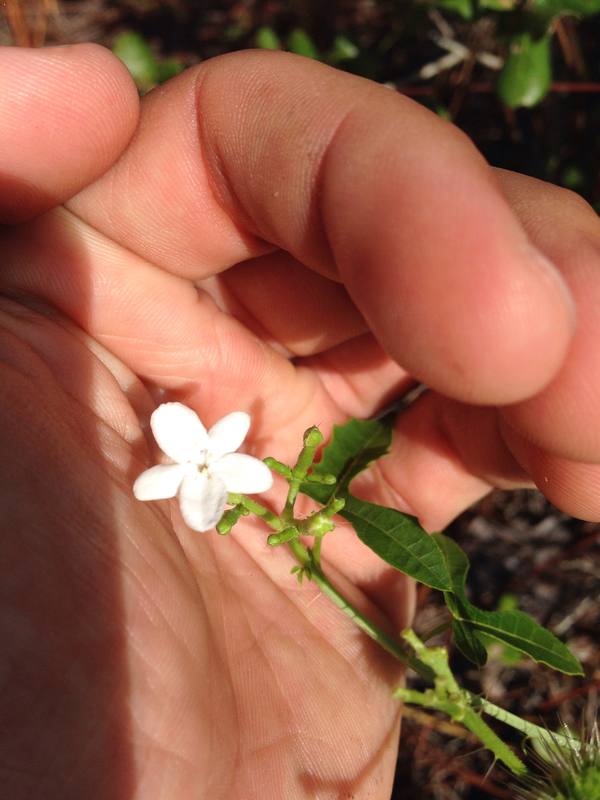}} \hspace{0.75pt} & \hspace{0.75pt}
    \setlength{\fboxsep}{0pt}%
    \setlength{\fboxrule}{0.75pt}%
    \fcolorbox{green}{white}{\includegraphics[width=2.65cm, height=2.25cm]{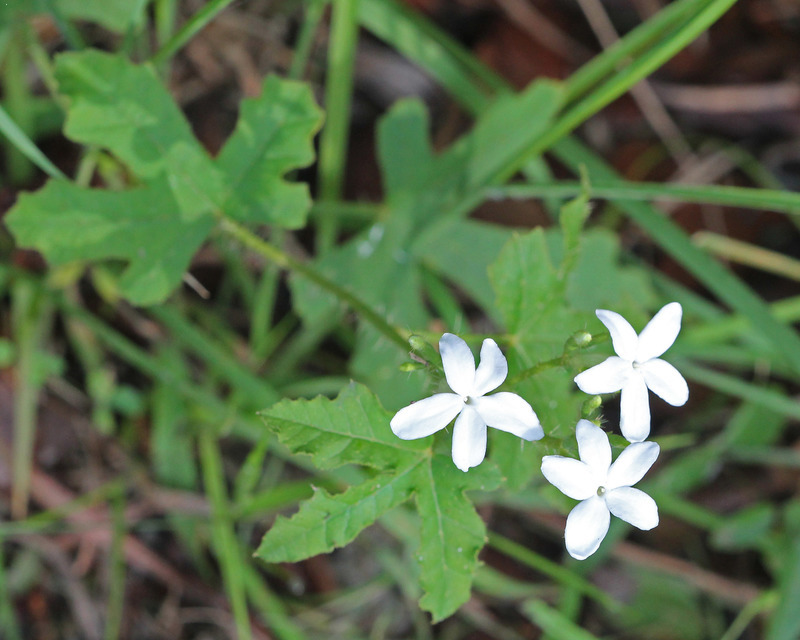}} &
    \setlength{\fboxsep}{0pt}%
    \setlength{\fboxrule}{0.75pt}%
    \fcolorbox{green}{white}{\includegraphics[width=2.65cm, height=2.25cm]{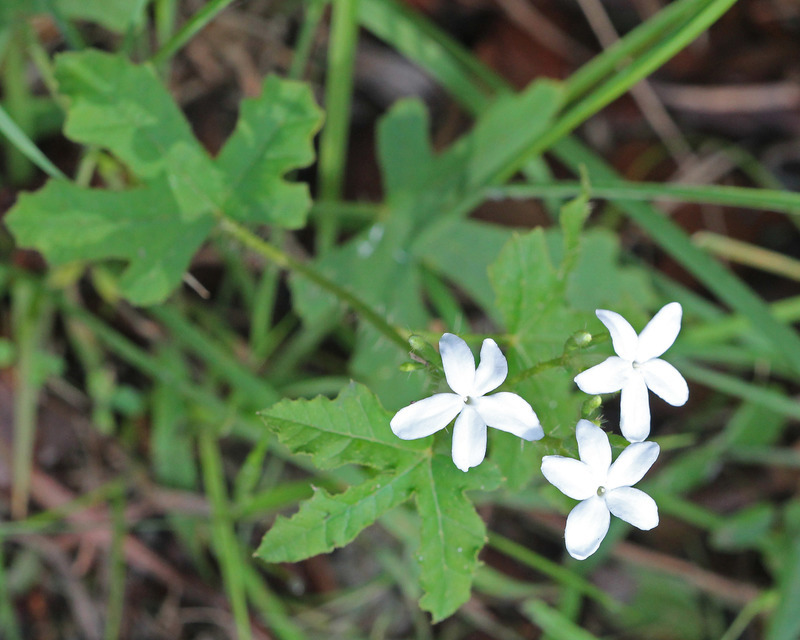}} &
    \setlength{\fboxsep}{0pt}%
    \setlength{\fboxrule}{0.75pt}%
    \fcolorbox{green}{white}{\includegraphics[width=2.65cm, height=2.25cm]{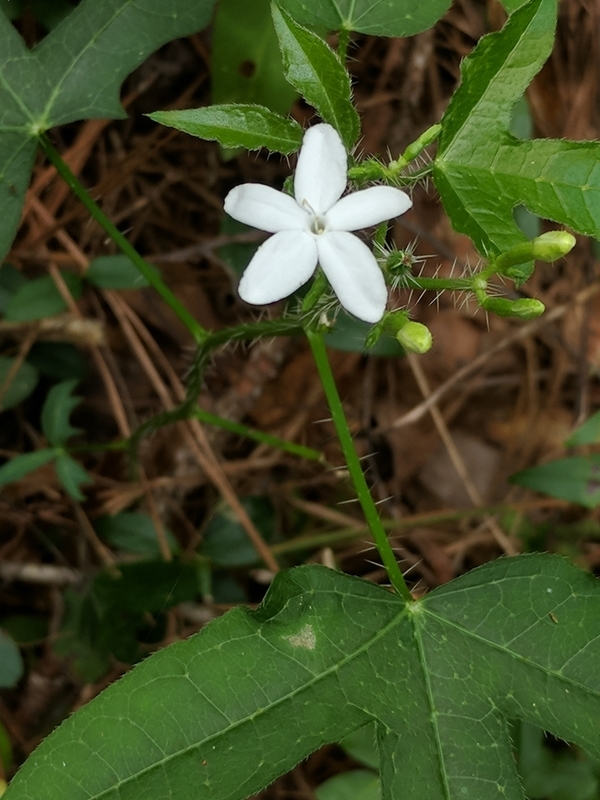}} &
    \setlength{\fboxsep}{0pt}%
    \setlength{\fboxrule}{0.75pt}%
    \fcolorbox{green}{white}{\includegraphics[width=2.65cm, height=2.25cm]{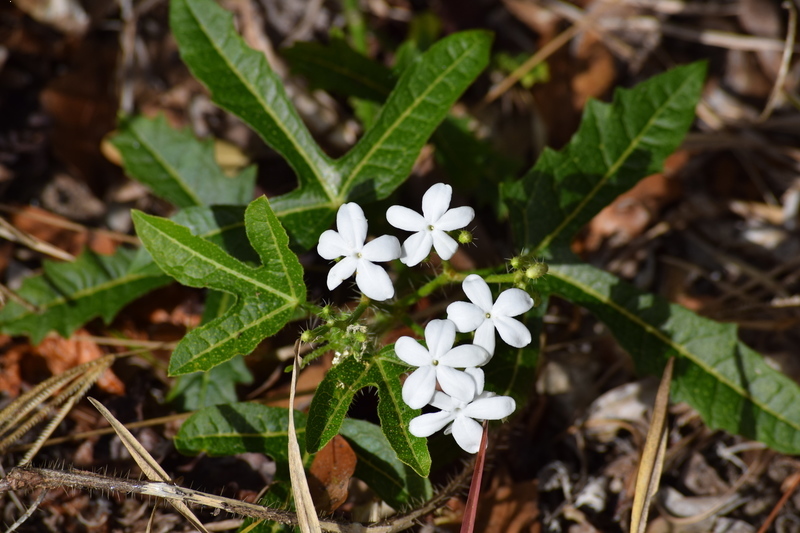}} &
    \setlength{\fboxsep}{0pt}%
    \setlength{\fboxrule}{0.75pt}%
    \fcolorbox{green}{white}{\includegraphics[width=2.65cm, height=2.25cm]{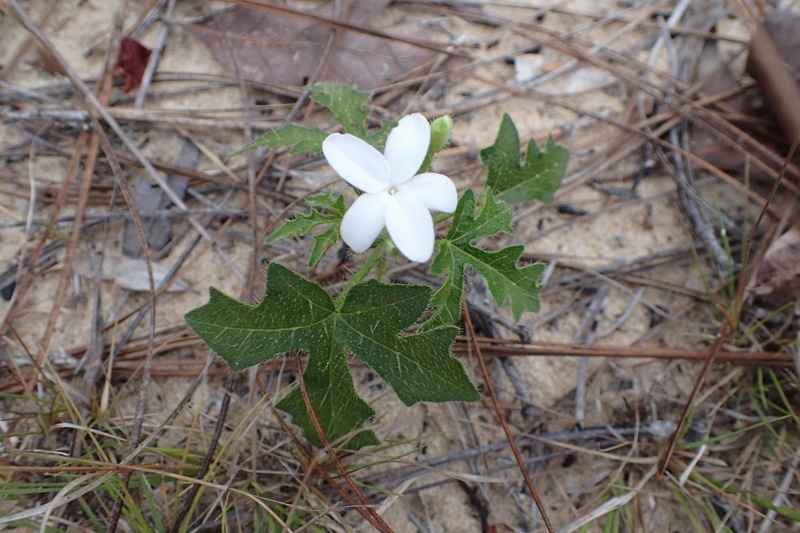}} \\

    \setlength{\fboxsep}{0pt}%
    \setlength{\fboxrule}{0.75pt}%
    \fcolorbox{red}{white}{\includegraphics[width=2.65cm, height=2.25cm]{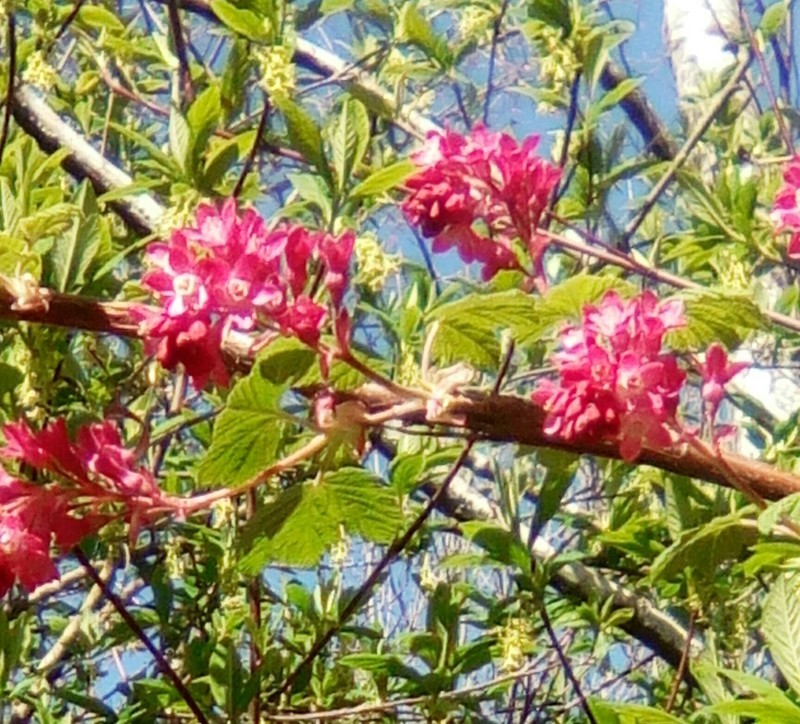}} \hspace{0.75pt} & \hspace{0.75pt}
    \setlength{\fboxsep}{0pt}%
    \setlength{\fboxrule}{0.75pt}%
    \fcolorbox{green}{white}{\includegraphics[width=2.65cm, height=2.25cm]{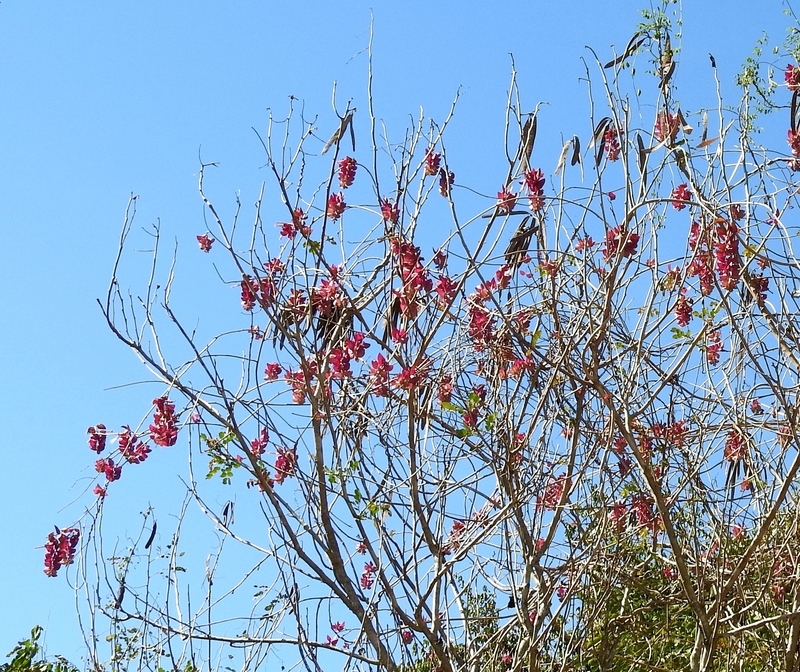}} &
    \setlength{\fboxsep}{0pt}%
    \setlength{\fboxrule}{0.75pt}%
    \fcolorbox{green}{white}{\includegraphics[width=2.65cm, height=2.25cm]{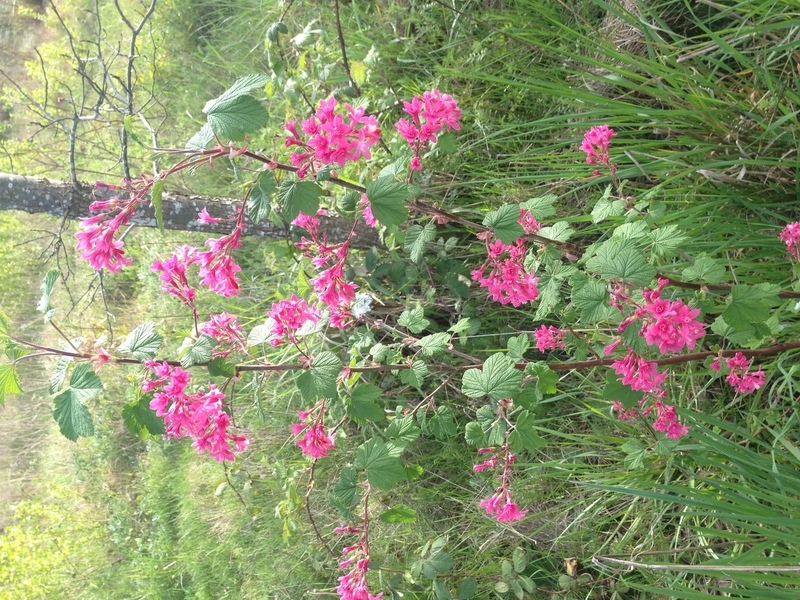}} &
    \setlength{\fboxsep}{0pt}%
    \setlength{\fboxrule}{0.75pt}%
    \fcolorbox{green}{white}{\includegraphics[width=2.65cm, height=2.25cm]{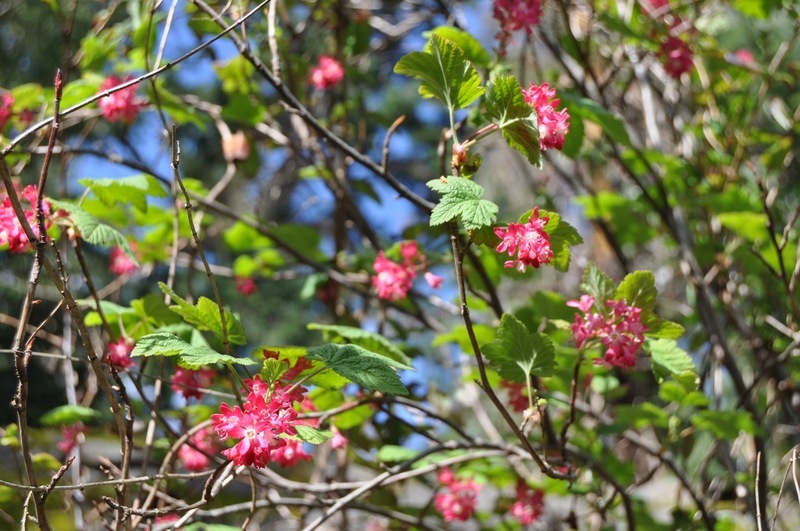}} &
    \setlength{\fboxsep}{0pt}%
    \setlength{\fboxrule}{0.75pt}%
    \fcolorbox{red}{white}{\includegraphics[width=2.65cm, height=2.25cm]{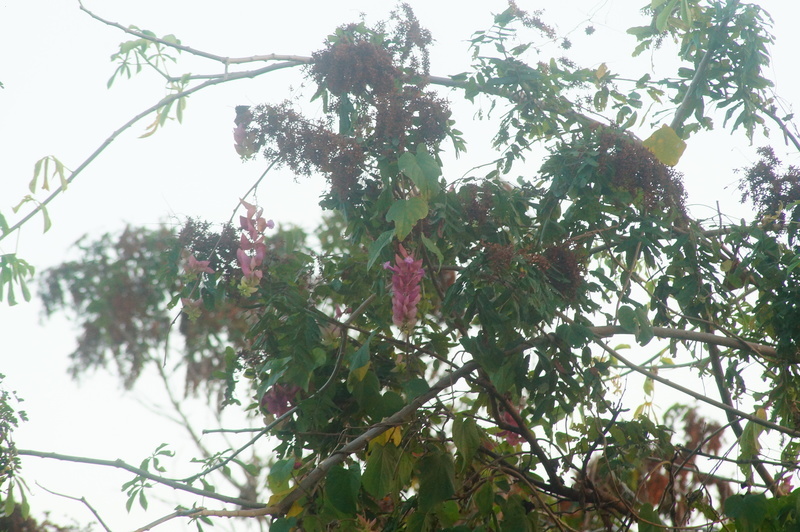}} &
    \setlength{\fboxsep}{0pt}%
    \setlength{\fboxrule}{0.75pt}%
    \fcolorbox{green}{white}{\includegraphics[width=2.65cm, height=2.25cm]{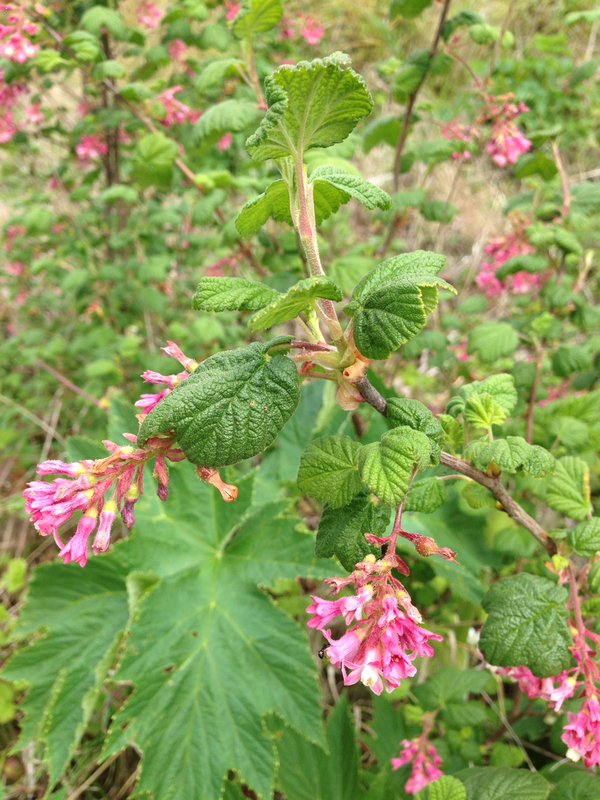}} \\

    \setlength{\fboxsep}{0pt}%
    \setlength{\fboxrule}{0.75pt}%
    \fcolorbox{black}{white}{\includegraphics[width=2.65cm, height=2.25cm]{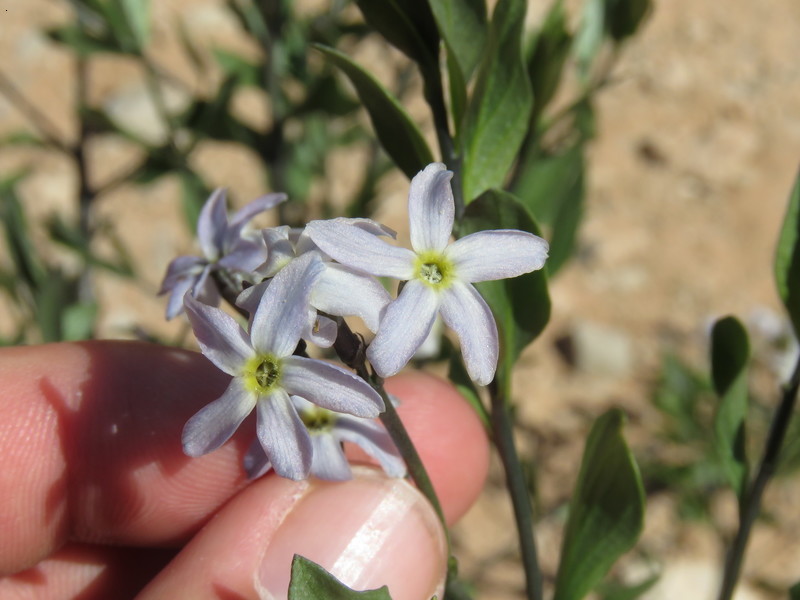}} \hspace{0.75pt} & \hspace{0.75pt}
    \setlength{\fboxsep}{0pt}%
    \setlength{\fboxrule}{0.75pt}%
    \fcolorbox{green}{white}{\includegraphics[width=2.65cm, height=2.25cm]{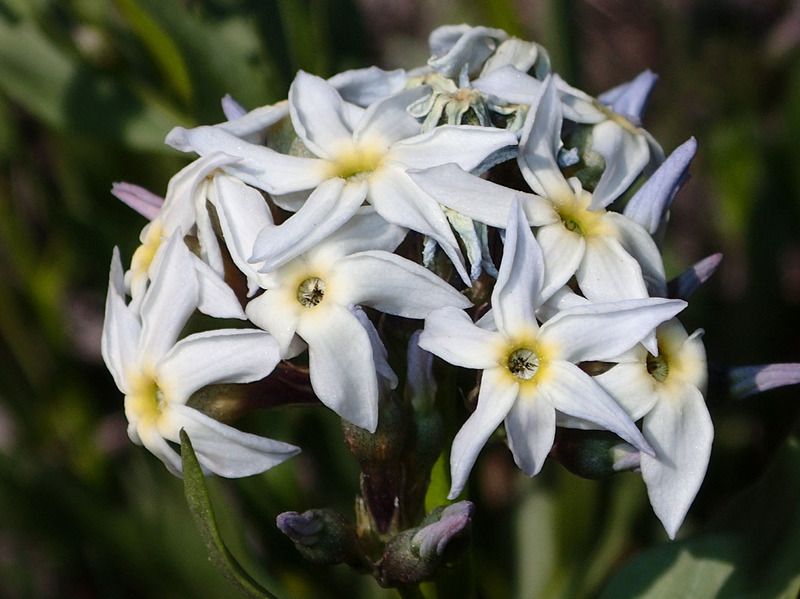}} &
    \setlength{\fboxsep}{0pt}%
    \setlength{\fboxrule}{0.75pt}%
    \fcolorbox{green}{white}{\includegraphics[width=2.65cm, height=2.25cm]{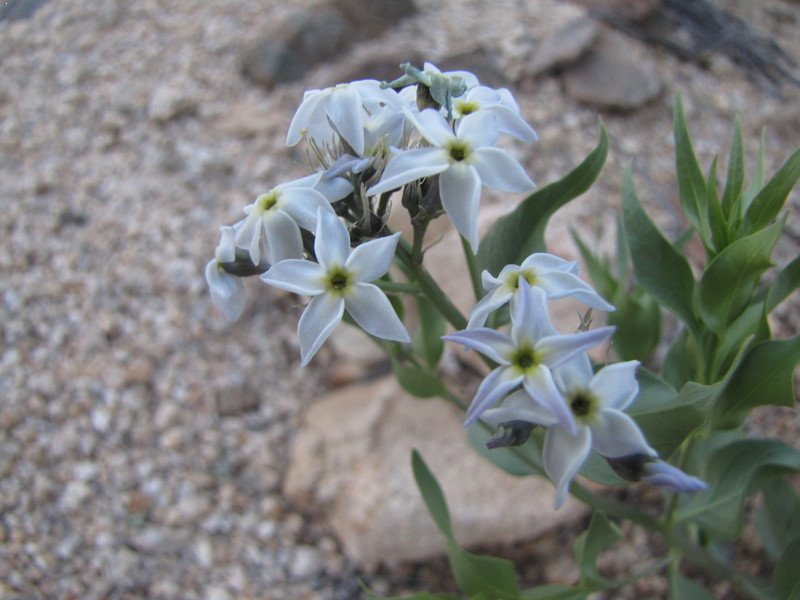}} &
    \setlength{\fboxsep}{0pt}%
    \setlength{\fboxrule}{0.75pt}%
    \fcolorbox{green}{white}{\includegraphics[width=2.65cm, height=2.25cm]{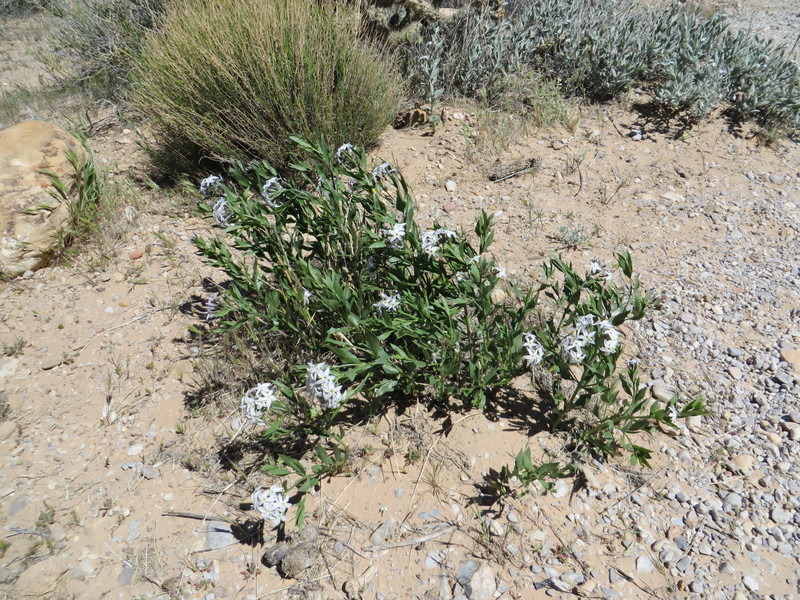}} &
    \setlength{\fboxsep}{0pt}%
    \setlength{\fboxrule}{0.75pt}%
    \fcolorbox{green}{white}{\includegraphics[width=2.65cm, height=2.25cm]{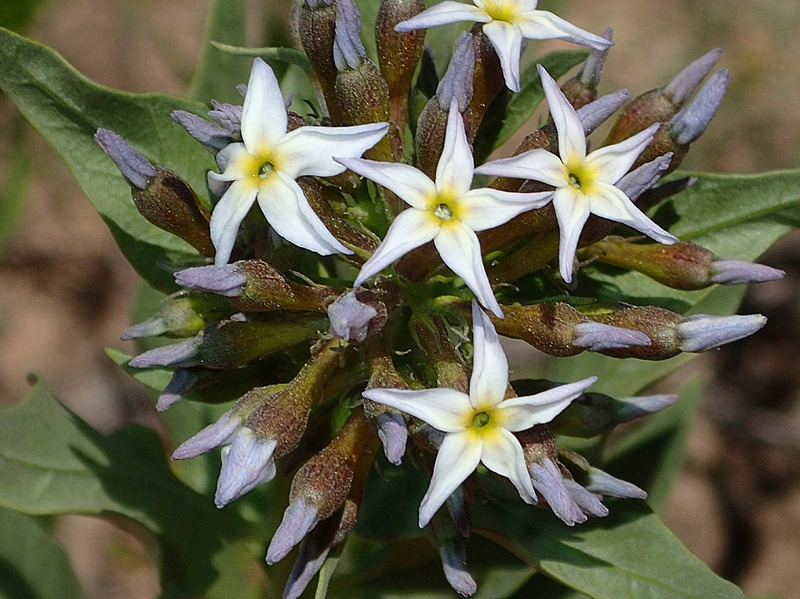}} &
    \setlength{\fboxsep}{0pt}%
    \setlength{\fboxrule}{0.75pt}%
    \fcolorbox{green}{white}{\includegraphics[width=2.65cm, height=2.25cm]{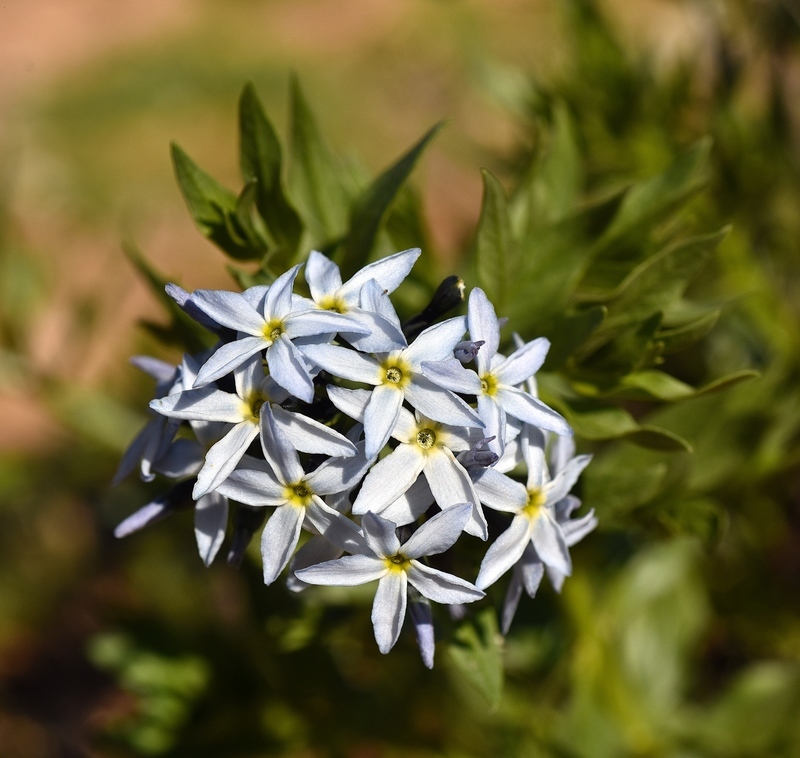}} \\

    \setlength{\fboxsep}{0pt}%
    \setlength{\fboxrule}{0.75pt}%
    \fcolorbox{black}{white}{\includegraphics[width=2.65cm, height=2.25cm]{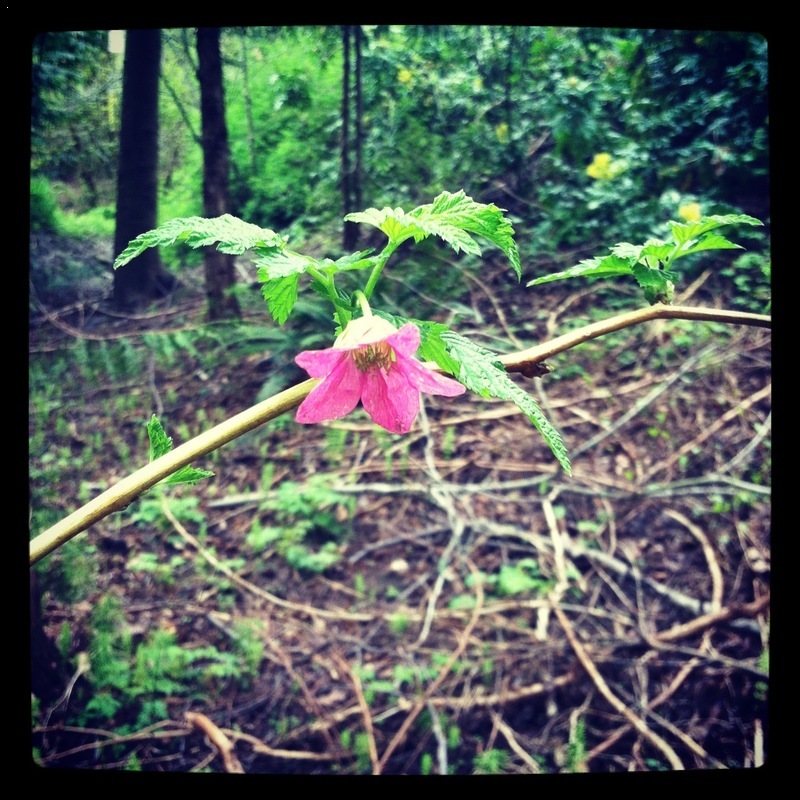}} \hspace{0.75pt} & \hspace{0.75pt}
    \setlength{\fboxsep}{0pt}%
    \setlength{\fboxrule}{0.75pt}%
    \fcolorbox{green}{white}{\includegraphics[width=2.65cm, height=2.25cm]{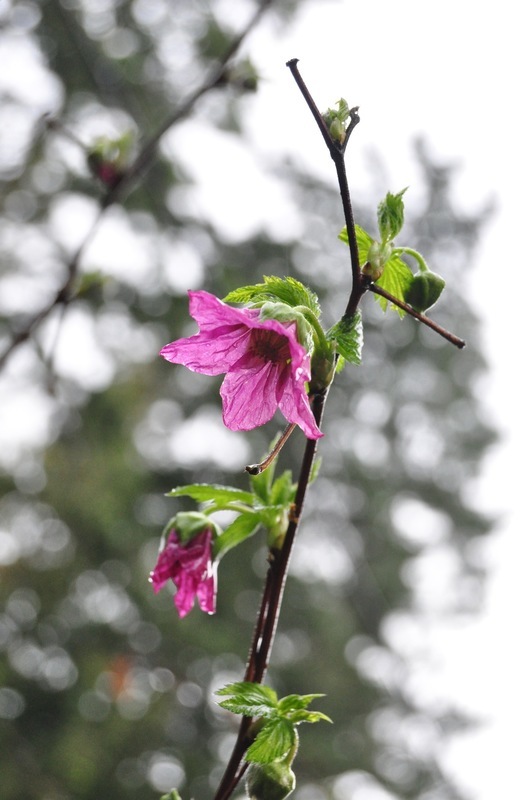}} &
    \setlength{\fboxsep}{0pt}%
    \setlength{\fboxrule}{0.75pt}%
    \fcolorbox{green}{white}{\includegraphics[width=2.65cm, height=2.25cm]{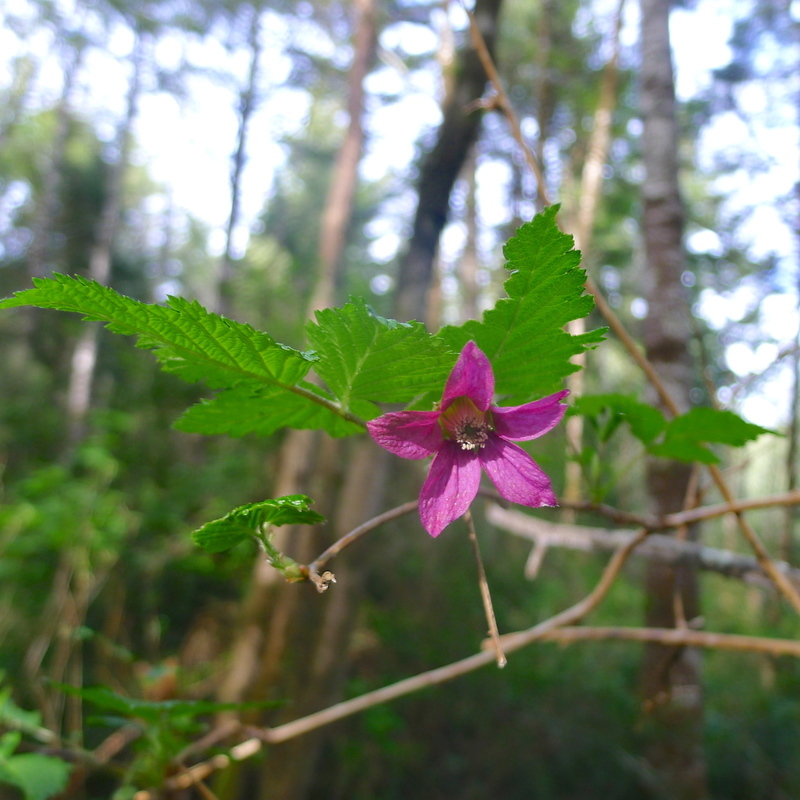}} &
    \setlength{\fboxsep}{0pt}%
    \setlength{\fboxrule}{0.75pt}%
    \fcolorbox{green}{white}{\includegraphics[width=2.65cm, height=2.25cm]{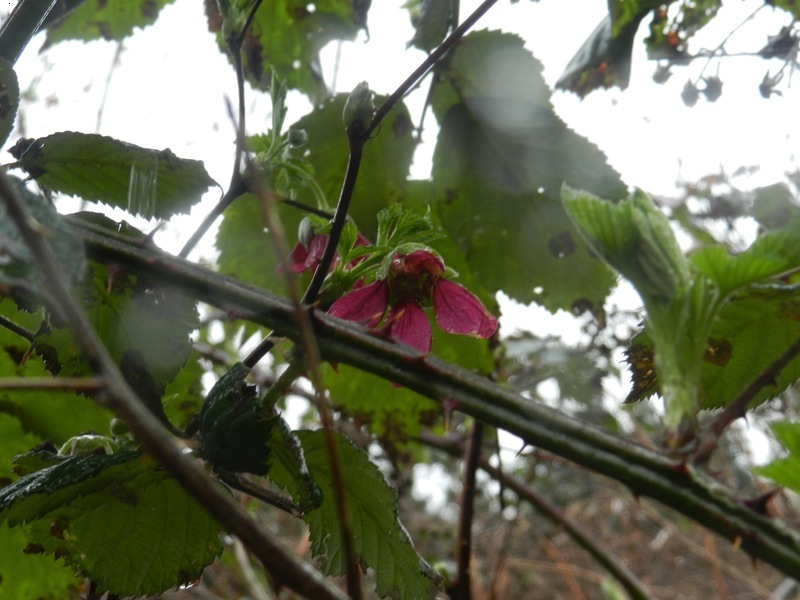}} &
    \setlength{\fboxsep}{0pt}%
    \setlength{\fboxrule}{0.75pt}%
    \fcolorbox{green}{white}{\includegraphics[width=2.65cm, height=2.25cm]{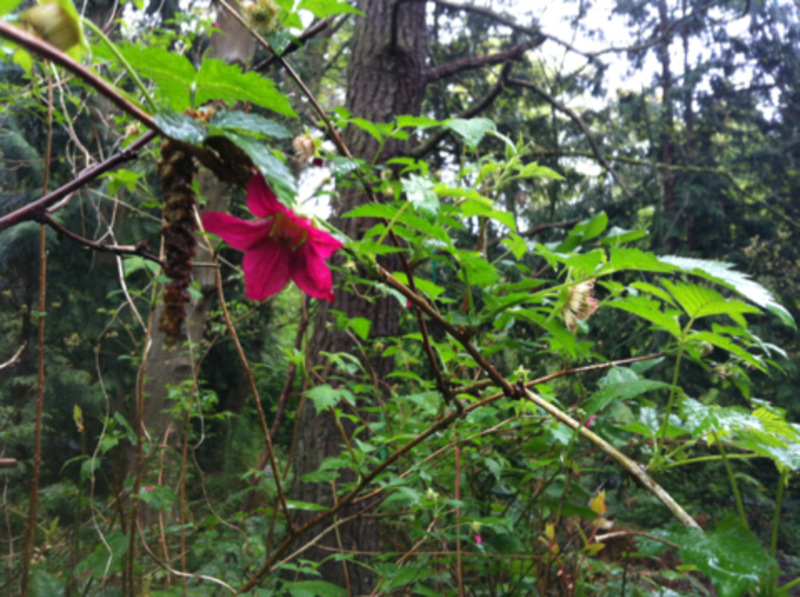}} &
    \setlength{\fboxsep}{0pt}%
    \setlength{\fboxrule}{0.75pt}%
    \fcolorbox{green}{white}{\includegraphics[width=2.65cm, height=2.25cm]{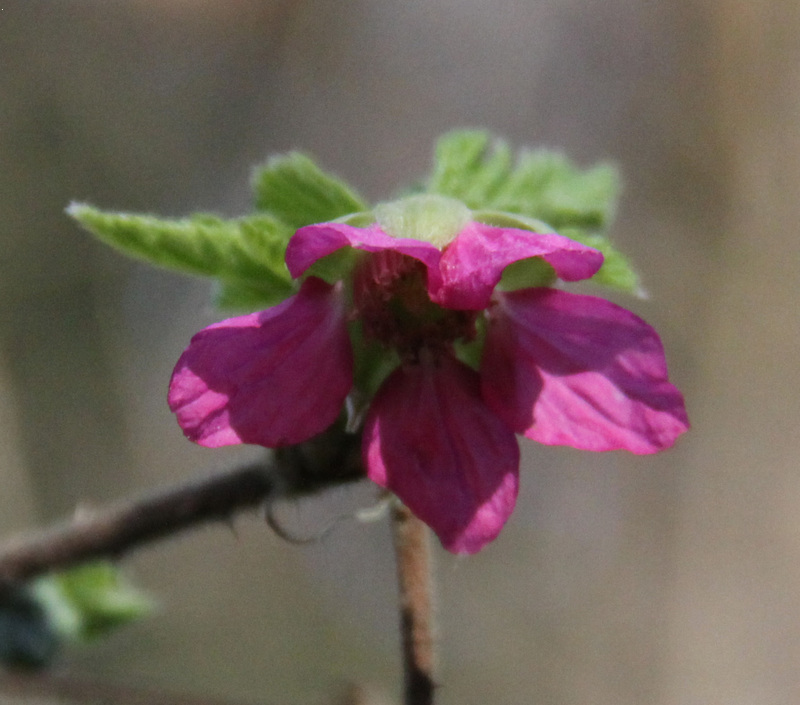}} \\

    \setlength{\fboxsep}{0pt}%
    \setlength{\fboxrule}{0.75pt}%
    \fcolorbox{black}{white}{\includegraphics[width=2.65cm, height=2.25cm]{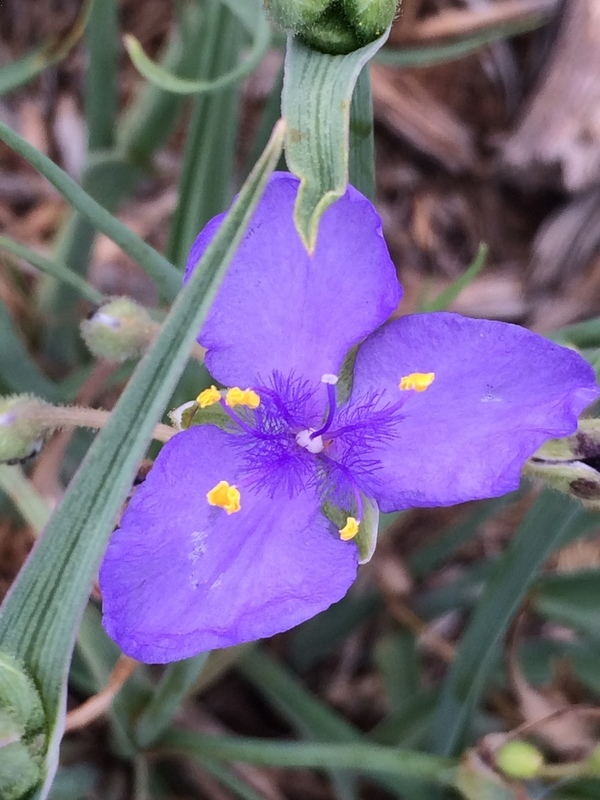}} \hspace{0.75pt} & \hspace{0.75pt}
    \setlength{\fboxsep}{0pt}%
    \setlength{\fboxrule}{0.75pt}%
    \fcolorbox{green}{white}{\includegraphics[width=2.65cm, height=2.25cm]{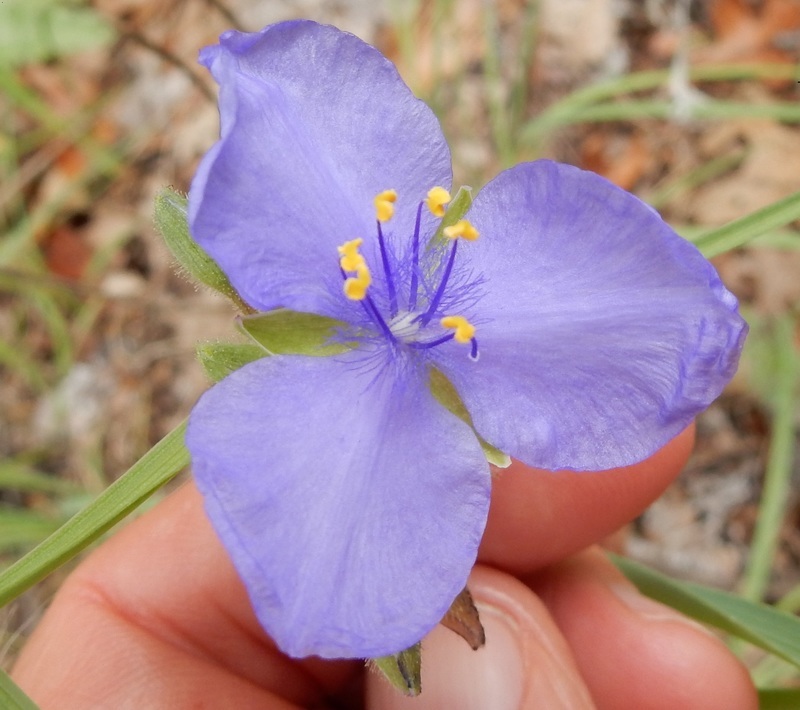}} &
    \setlength{\fboxsep}{0pt}%
    \setlength{\fboxrule}{0.75pt}%
    \fcolorbox{green}{white}{\includegraphics[width=2.65cm, height=2.25cm]{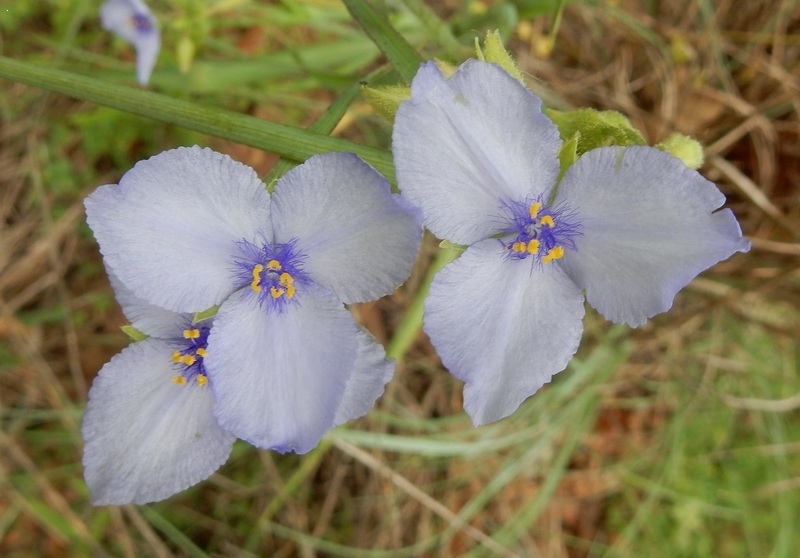}} &
    \setlength{\fboxsep}{0pt}%
    \setlength{\fboxrule}{0.75pt}%
    \fcolorbox{green}{white}{\includegraphics[width=2.65cm, height=2.25cm]{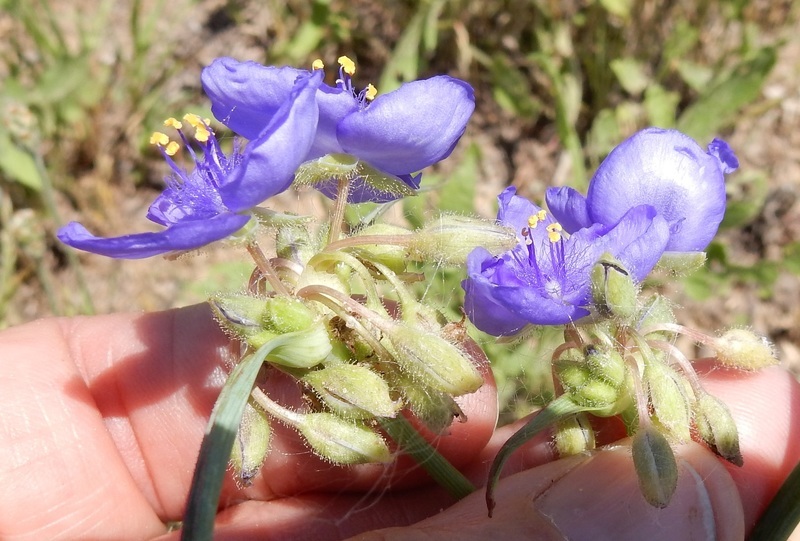}} &
    \setlength{\fboxsep}{0pt}%
    \setlength{\fboxrule}{0.75pt}%
    \fcolorbox{green}{white}{\includegraphics[width=2.65cm, height=2.25cm]{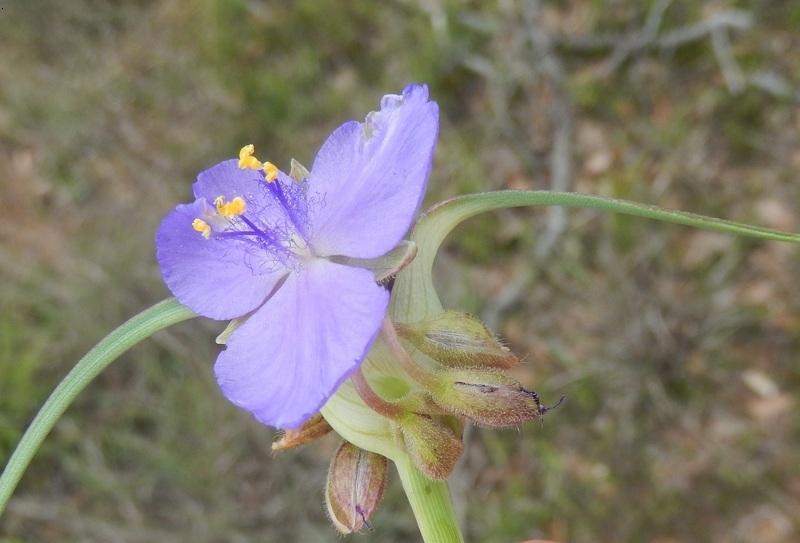}} &
    \setlength{\fboxsep}{0pt}%
    \setlength{\fboxrule}{0.75pt}%
    \fcolorbox{green}{white}{\includegraphics[width=2.65cm, height=2.25cm]{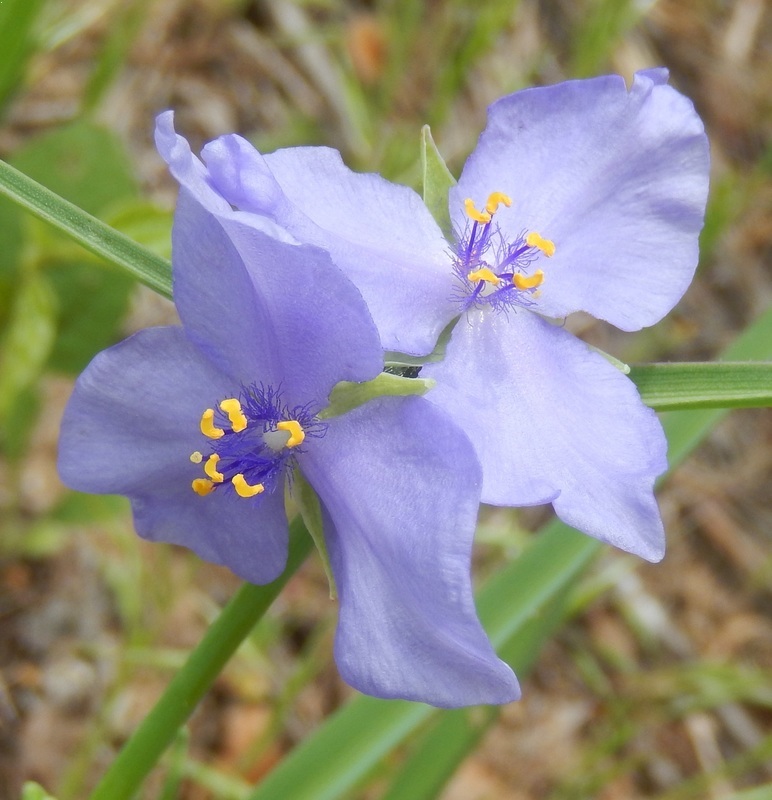}}
  \end{tabular}}
  \caption{Qualitative examples from the retrieval approach on iNaturalist dataset. The left most column shows samples from the test set followed by five nearest neighbours in the learned embedding space from the training set.}
  \label{fig:background_samples}
\end{figure}

Training an image retrieval system and subsequently performing a nearest neighbour classification is a competitive alternative, with better results than direct classification. The prediction obtained via a nearest neighbour search is more interpretable as the samples contributing to the prediction can be visualized. Therefore, a retrieval-based approach is more suitable if utilized within the humans in the loop. On the other hand, the softmax predictions of a standard neural network classifier allow for simple post-processing procedures such as averaging, prior shift adaptation, etc., which are yet to be explored for the retrieval approach, and which noticeably improve the final recognition accuracy of the standard classifiers.

Overall, using image-retrieval has clear advantages, e.g., recovering relevant nearest-neighbour labelled samples, providing ranked class predictions, and allows user or experts to visually verify the species based on the k-nearest neighbours 
Besides, the retrieval approach naturally supports open-set recognition problems, i.e., the ability to extend or modify the set of recognised classes after the training stage. The set of classes may change e.g. as a results of modifications to biological taxonomy. New classes are introduced simply by adding training images with the new label, whereas in the standard approach, the classification head needs re-training.
On the negative side, the retrieval approach requires, on top of running the deep net to extract the embedding, to execute the nearest neighbour search efficiently, increasing the overall complexity of the fine-grained recognition system.

Contrary to our expectations, the error analysis in Figure \ref{fig:perf_boxplot} shows that the retrieval approach does not bring an improvement in classifying images from classes with few training samples. Figure \ref{fig:perf_histogram} shows that retrieval has a very high accuracy for a higher number of species, but it also fails for a higher number of species.

\section{Conclusions}
\label{sec:rs@k_conclusions}

This work has presented image embedding learning for retrieval using a novel surrogate loss function for the recall@k metric. State-of-the-art results were achieved on a number of standard benchmarks. Training with very large batch size, up to 4k images, has shown to be highly beneficial. The batch size is further increased, in a virtual way, with a newly proposed mixup approach that acts directly on the scalar similarities. This approach offers a boost in performance at a small increase of the computational cost, while its applicability goes beyond the proposed loss. The implementation of the proposed Recall@k Surrogate loss, proposed similarity mixup, along with the training procedure that allows the use of large batch sizes on a single GPU by sidestepping memory constraints, is available at \href{https://github.com/yash0307/RecallatK_surrogate}{https://github.com/yash0307/RecallatK\_surrogate}.
\newcommand{\z}{\mathbf{z}}
\newcommand{\x}{\mathbf{x}}
\newcommand{\SimCLR}{\text{SimCLR}\xspace}
\newcommand{\SupCon}{\text{SupCon}\xspace}
\newcommand{\CE}{\text{CE}\xspace}
\newcommand{\SPCE}{\text{SPCE}\xspace}
\newcommand{\clSup}{\text{clSup}\xspace}
\newcommand{\pt}{\text{pt}\xspace}
\newcommand{\SupC}{\text{SupC}\xspace}
\renewcommand{\th}{\boldsymbol{\theta}}
\renewcommand{\c}{\mathbf{c}}
\newcommand{\ExtSupCon}{\tt{ESupCon}\xspace}
\newcommand{\ESupCon}{\text{ESupCon}\xspace}
\newcommand{\tSPCE}{\tt{SPCE}\xspace}
\newcommand{\tCE}{\tt{CE}}
\newcommand{\tsupcontt}{\tt{SupCon+Tt}\xspace}
\newcommand{\tsupconce}{\tt{SupCon+CE}\xspace}
\newcommand{\tsupconceN}{\tt{SupCon+CE(n)}\xspace}
\newcommand{\replace}[2]{\sout{#1}{\textcolor{red}{#2}}}

\chapter{Contrastive Classification and Representation Learning with Probabilistic Interpretation}
\label{chapter:esupcon}
\chaptermark{Extended Supervised Contrastive Learn...}

\begin{figure*}[t]
    \centering
\includegraphics[width=\textwidth]{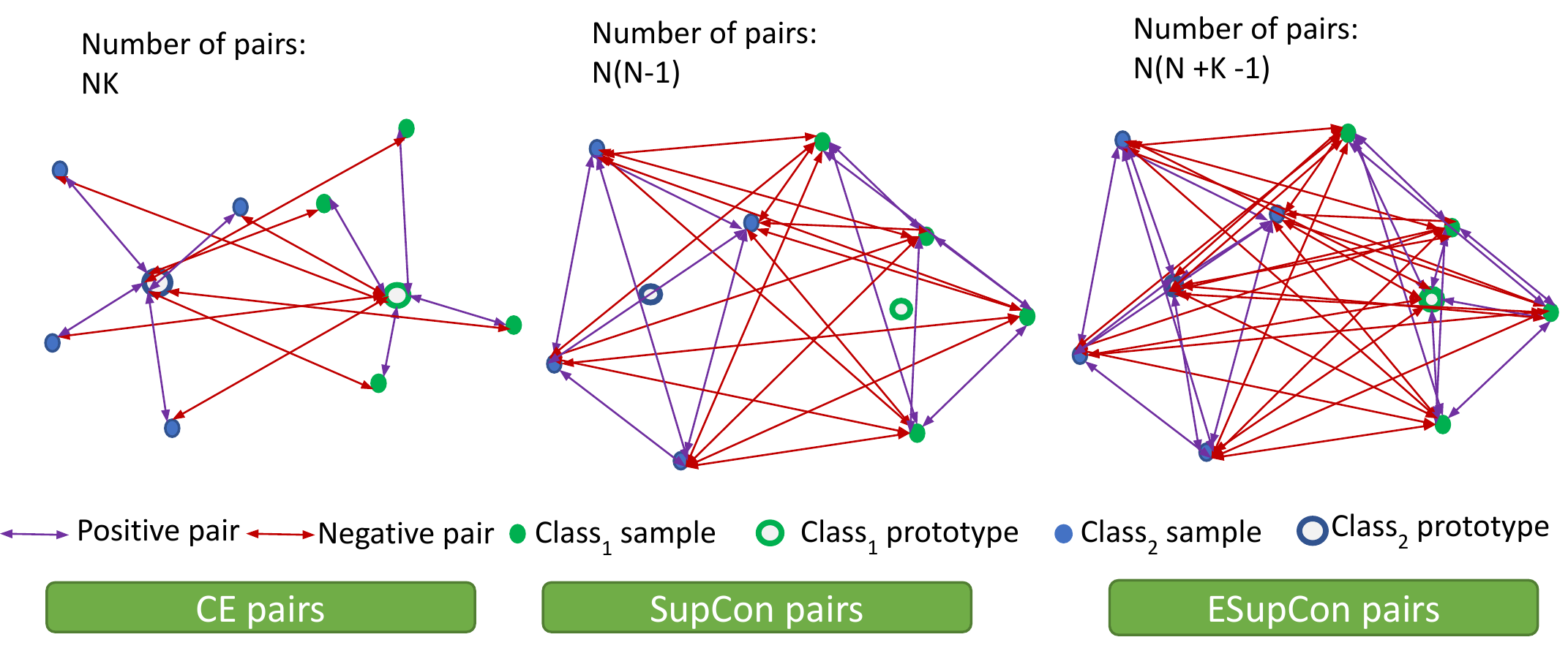}
\caption{Illustration of the possible number of pairs that each loss accesses during training in the learned embedding space, N is the batch size and K is the number of classes.
CE pairs are only defined through classes weights while in SupCon each sample forms positive pairs with its class samples and negative pairs with samples from other classes.
For our ESupCon in addition to the positive and negative samples pairs, class weights (prototypes) form positive pairs with corresponding class samples and negative pairs with other classes samples. Note that here we don't consider augmentations. }%
    \label{fig:illustration}%
\end{figure*}
Representation learning is a powerful tool to create an embedding space that is beneficial for performing downstream tasks e.g., classification or retrieval.  Contrastive representation learning first proposed by ~\cite{chopra2005learning} is a dominant successful line for representation learning. It divides the data into pairs of positive (similar)  and negative (unrelated) samples with the objective of maximizing the similarity of positive pairs samples and minimize it for negative pairs.   

More recently, contrastive learning has become a key component of methods for self-supervised learning~\cite{chen2020simple,kalantidis2020hard,chen2020big,caron2021emerging} and has shown impressive performance~\cite{caron2020unsupervised,caron2021emerging} that is very close to the supervised learning counterpart with cross entropy loss.
Moreover, it was  shown that supervised contrastive  learning marginally outperforms  the \textit{cross entropy loss} in fully supervised image classification~\cite{khosla2020supervised}. Not only for standard supervised classification but it has been applied in continual learning~\cite{davari2022probing}, Out of Distribution Detection~\cite{winkens2020contrastive}, Domain Adaptation~\cite{chen2022contrastive} and many more  showing superior performance to  cross entropy based counterpart. 

Minimizing \textit{cross entropy} (CE) loss is widely used in training deep neural network classifiers,
derived as the maximum likelihood estimate (MLE) of classifier's parameters $\th$ to approximate posterior probabilities $\hat p(\text{class}|\text{observation})$.
~\cite{boudiaf2020unifying} draw the connection among popular pairwise-distance losses and the cross entropy loss, showing that all of them are related to maximizing the mutual information (MI) between the learned embeddings and the corresponding samples' labels.

We emphasize the advantages of the probabilistic interpretation of the CE loss in classification problems.
Such explicit probabilistic interpretation is missing within the embedding spaces trained by popular contrastive learning methods. The posterior estimates $\hat p(\text{class}|\text{observation})$ can be utilized when combining classifiers~\cite{kittler1998combining,breiman1996bagging,ju2018relative}, in adaptation to prior shift~\cite{saerens2002adjusting,sulc2019improving,alexandari2020maximum,sipka2021hitchhiker}, in knowledge distillation~\cite{hinton2015distilling};  out-of-distribution detection~\cite{hendrycks2016baseline} and in many other problems.

In this chapter, we suggest that one possible reason for the improved performance of supervised contrastive learning  is the inherent access to a large number of samples pairs, 
while the ``pairs'' within the softmax \CE loss are centered around the linear classifier weights. Here we draw an analogy with proxy based loss and consider the linear classifier weights optimized  in the softmax \CE loss as proxies for learning the samples representations.  Proxy base training that utilizes proxies instead of the direct sample to sample relationship is simple and faster to converge, however,  it doesn't leverage the rich data to data similarities as the supervised contrastive loss. 
We refer to Figure~\ref{fig:illustration} for an illustration on this assumption.
We hypothesize that the access to more pairs during training might lead to a better convergence and less overfitting resulting in the advantages hinted in recent work~\cite{khosla2020supervised,graf2021dissecting}. 
 
Hence, to combine the advantages of contrastive representation learning via pairwise losses and the clear probabilistic interpretation of classifiers trained by cross entropy minimization, we present the following  contributions:
First, we consider the weights of the last linear classification layer as prototypes of each class. 
We show that adding a simple  term corresponding to maximizing the similarity between the prototypes and their class samples, leads to an assignment of the prototypes to the mean of each class samples with momentum updates of representation.
This is optimized during the representation training with a supervised contrastive loss~\cite{khosla2020supervised}, resulting in a nearest prototype classifier~\cite{wohlhart2013optimizing}.
Second, we propose an extension to the \textit{supervised contrastive loss}~(\SupCon)~\cite{khosla2020supervised},
where samples of a given class form positive pairs with their class prototype and other classes samples correspond to negative pairs.

We show that the resulting objective   combines in its formulation  the \SupCon loss~\cite{khosla2020supervised} and the standard \CE loss on prototypes related pairs,  preserving the probabilistic interpretation of the predictions. We refer to this loss as \ESupCon (short for \textit{Extended Supervised Contrastive loss}).

Third, we revisit the Simplified Pairwise Cross Entropy (\SPCE) loss, proposed in the theoretical analysis of~\cite{boudiaf2020unifying}, and compare it with standard \CE loss and  the supervised contrastive learning loss in an extensive experimental evaluation. 

In our experimental evaluation, we not only consider the fully supervised setting but also for the first time a number of challenging settings (low sample regime, imbalanced data and noisy labels). To the best of our knowledge, this is the first comprehensive evaluation of \SupCon loss and the standard  \CE loss in addition to our proposed extensions.
 
We show \ESupCon is  more powerful as a training objective than the standard \CE loss while maintaining a probabilistic interpretation.
and is more robust in challenging and low sample settings. 
Surprisingly, our simple  prototypes similarity term  is more robust than \CE loss for learning a linear classifier after \SupCon in most of the imbalanced and noisy data experiments. 

In the following, we  describe the closely Related Work, then provide a short Background on pairwise losses and the link to Cross Entropy loss, followed by our extension to Supervised Contrastive Loss.
We validate and compare different studied losses in the Experiments, and summarize  our contributions and limitations in Conclusion.

\section{Related Work}
\label{sec:esupcon_relatedwork}
CE loss is a standard and powerful training objective to optimize deep neural networks for classification-related problems. 
For long, the CE loss was believed to be more effective than representation learning losses \eg, metric learning based losses.
For example,~\cite{boudiaf2020unifying} studied the relation of \CE loss to contrastive metric learning losses and showed that the \CE loss also has a contrastive and a tightness part.
The authors suggested that \CE ``does it all'' and that it is easier to optimize compared to its contrastive-learning counterparts.
Recently, self-supervised learning losses have shown great success~\cite{chen2021exploring,grill2020bootstrap,chen2020simple, he2020momentum,chen2020big,grill2020bootstrap,caron2020unsupervised,caron2021emerging} as pretraining methods with only a small performance gap to that of fully supervised learning. The core of the self-supervised methods is the use of rich data augmentation methods to construct positive pairs corresponding to augmented version of a given sample. 
Closer to our work, \cite{li2020prototypical,caron2020unsupervised}  construct clusters and establish cluster assignments through prototypes while learning the embedding space. 
It remains unclear how these losses can be extended to the supervised setting as in our case. 

Inspired by the self-supervised \SimCLR loss~\cite{chen2020simple}, \cite{khosla2020supervised} introduced a new supervised contrastive learning method called \SupCon, which achieved superior results compared to the standard \CE minimization, and which has been shown to be more generalizable and robust to noise. However, the method is only used to train the image representation  and still relies on the \CE loss to train the linear classifier afterwards.
\CE-based training suffers from  known issues of noise sensitive, overfitting~\cite{berrada2018smooth}, and being less transferable than the representation learning counterparts~\cite{chen2020simple}.
Recently, \cite{graf2021dissecting} investigated the difference between the \SupCon~\cite{khosla2020supervised} loss and the \CE loss in the geometry of the targeted representation. 
It was shown that both losses target the same geometric solution, however, \SupCon converges much closer to the target leading to a better generalization performance.

As such, starting from the nice suggested characteristics of the \SupCon loss based training, we propose and study alternatives that can train the whole network (representation and classifier) end-to-end, while preserving both the performance improvements of contrastive representation learning and the clear probabilistic interpretation of the \CE loss. 
We start by considering the classes weights as prototypes for each class samples. We learn these prototypes while maximizing positive pairs similarities and minimizing negative pairs similarities.  Our work hence can be seen as combination of proxy (prototype) based and pairwise based contrastive representation learning. 
Proxy based losses resort to learning a set of  proxies as representative of clusters or classes of samples and optimize the similarities to these proxies rather than the data to data similarities. Proxy NCA~\cite{movshovitz2017no} was the first proxy base metric learning method, it is an approximation of NCA (Neares Component Analysis) using proxies. We note that in the case of learning with class level labels the Proxy NCA matches learning with Softmax Cross Entropy loss when the last classification layer is without a bias term and its weights  are  normalized vectors. Proxy anchor loss~\cite{kim2020proxy}, attempts to combine the benefits of both proxy-based and pairwise losses. While in the main loss formulation only similarities to proxies are considered,     the magnitude of the loss gradient w.r.t. each sample is scaled by the  corresponding proxy similarity proportional to other samples-proxies similarities. 
In general, proxy based losses do not use the proxy at test time and it is unknown how they perform for classification or whether there can exist any probabilistic interpretations. 
Circle loss~\cite{sun2020circle} presents a unified framework for both pairwise and proxy based losses but it adopts an adaptive scaling of the loss depending on how much a given similarity is deviated from its optimum. In doing so, Circle loss abandons the probabilistic interpretation of a sample assignment to its prototype (proxy). 

\section{Background}
\label{sec:esupcon_background}
In this section, we describe recent self-supervised and supervised contrastive losses and the connection with CE loss.
\subsection{Pairwise Losses}
Contrastive losses work with pairs of  embeddings that are pulled together if a pair is positive (related embeddings) and pulled further apart otherwise~\cite{chopra2005learning}. 
 Consider the following: 1) a random data augmentation module that for each sample $\x$ generates two differently augmented samples, 2) a neural network encoder $f$ that maps an augmented input sample $\x$ to its  feature representation: $f(\x)=\z, \z\in \mathbb{R}^d$.  We  start by outlining SimCLR~\cite{chen2020simple}, a popular, effective and simple self supervised contrastive loss to lay the ground for our work:
\begin{equation}
\ell_{\SimCLR}=\frac{1}{2N}\sum_i^N \ell_{\SimCLR}(\z_i,\z_{i+N}) +\ell_{\SimCLR}(\z_{i+N},\z_i),
\end{equation}
\begin{equation}
\begin{split}
\ell_{\SimCLR}(\z_i,\z_j) &= -\log\frac{\exp (\similarity(\z_i,\z_j)/\tau)}{\sum_{k\ne i} \exp(\similarity(\z_i,\z_k)/\tau)},
\end{split}
\end{equation}
where $\tau$ is the temperature scaling term, $N$ is the mini batch size, and the pairs $(\z_i,\z_j)$ consist of features of two differently augmented views of the same data example and $\similarity(\z_i,\z_j) = \dfrac{\z_i^\top \z_j}{||\z_i||\cdot||\z_j|| }$ is the cosine similarity.
Assuming normalized embedding vectors $\z_i$, this pairwise loss is:
\begin{equation}
\ell_{\SimCLR}(\z_i,\z_j) = -\z_i^\top\z_j/\tau + \log \sum_{k\ne i}\exp\left(\z_i^\top\z_k/\tau\right).
\end{equation}
Note that the first term corresponding to the positive pair is the tightness term  and the second one is the contrastive term.
The aforementioned self-supervised batch contrastive approach was extended in~\cite{khosla2020supervised} to the fully supervised setting with the Supervised Contrastive Loss:
\begin{equation}
\ell_\SupCon=\frac{1}{2N}\sum_i^{2N}\ell_\SupCon(\z_i, P_i),
\end{equation}
\begin{equation}\label{eq:supcon}
\begin{split}
&\ell_\SupCon(\z_i, P_i)=\\
&-\frac{1}{|P_i|}\sum_{\z_p\in P_i} \log\frac{\exp(\similarity(\z_i,\z_p)/\tau)}{\sum_{j\ne i}\exp(\similarity(\z_i,\z_j)/\tau)} =\\
&\frac{1}{|P_i|}\sum_{\z_p\in P_i} \left(-{(\z_i^\top\z_p)}{/\tau} + \log \sum_{j\ne i}\exp\left(({\z_i^\top\z_j)}{/\tau}\right)\right),
\end{split}
\end{equation}
where $P_i$ is the set of representations $\z_p$ forming positive pairs for the $i$-th sample, and the index $j$ iterates over all (original and augmented) samples. SupCon loss is expressed as the average  of the loss defined on each positive pair where in this supervised setting, the positive pairs are formed of augmented views and other samples of the same class. 
The authors showed that the supervised contrastive learning achieves excellent results in image classification, improving ImageNet classification accuracy with ResNet-50 by 0.5\% compared to the best results achieved by training with the \CE loss.

\subsection{Cross Entropy and Pairwise Cross Entropy}

The cross entropy~(\CE) loss is a common choice for training classifiers, as its minimization leads to the maximum likelihood estimate of the classifier parameters for estimating the posterior probabilities $\hat p(\text{class} | \text{observation})$.

For $N$ samples of $K$ classes, and a single-label softmax classifier, the \CE loss can be defined as follows:
\begin{equation}
\begin{split}
\ell_\CE &= \frac{1}{N}\sum\limits_{i=1}^N \ell_\CE(\z_i) = -\frac{1}{N}\sum\limits_{i=1}^N\log  \dfrac{\exp{\th_{y_i}^\top\z_i}}{\sum\limits_{k=1}^K \exp{\th_k^\top \z_i} }\\
&= -\frac{1}{N}\sum\limits_{i=1}^N \th_{y_i}^\top \z_i + \frac{1}{N}\sum\limits_{i=1}^N \log \sum\limits_{k=1}^K \exp{\th_k^\top \z_i},
\end{split}
\label{eq:l_CE}
\end{equation}
where $\z_i$ is sample feature for the $i$-th observation having label $y_i\in\{1,\dots,K\}$, and $\th=(\th_1, \dots,\th_K)$ are the parameters of the last fully connected  layer, assuming that no bias term is used.

The \textit{Simplified Pairwise Cross Entropy} (SPCE) loss was introduced  in~\cite{boudiaf2020unifying} as a variant of the \CE loss~\eqref{eq:l_CE}:
\begin{equation}
\begin{split}
\ell_\SPCE &=
-\frac1N\sum_{i=1}^N \log\dfrac{\exp\left(\frac1N\sum_{j:y_j=y_i}\z_j^\top\z_i\right)}{\sum\limits_{k=1}^K \exp\left(\frac1N\sum_{j:y_j=k}\z_j^\top\z_i\right)}.
\end{split}
\label{eq:spce}
\end{equation}
When training the feature encoder with the $\ell_\SPCE$ loss, the classifier weights $\th$ can be estimated directly from the class feature means $\c_k$.
Moreover, the class posterior probabilities $p(k|\z_i)$ also can be estimated explicitly:
\begin{equation}
p(k|\z_i) = 
    \dfrac{\exp\left(\frac{1}{N} \sum_{j:y_j=k}\z_j^\top\z_i\right) }{\sum\limits_{c=1}^K \exp\left( \frac{1}{N}\sum_{j:y_j=c}\z_j^\top\z_i \right)}.
\end{equation}
In the experimental section, we will evaluate \SPCE loss and compare it with \SupCon.
Differently from \SPCE, with \SupCon, one needs to train a classifier on top of the learned representation as a posthoc process. In the following we will discuss and propose alternatives to  jointly learn the classifier  and the feature extraction parameters.

\section{Learning a Classifier Jointly with Representation Learning}
\label{sec:esupcon_jointlearning}
Representation learning under  \SupCon or \SPCE losses targets  grouping one class samples together while pushing  away samples of other classes. In fact, both losses contain  tightness and contrastive terms and  fulfill similar objectives to  that of the cross entropy loss.

Assuming that forcing samples of different classes to lie far apart  is achieved by the contrastive part of \SupCon or \SPCE, in order to learn the parameters of the classifier, one can consider the weight vectors of the linear classifier as prototypes and optimize these prototypes to be closest to the samples of the class they represent (with solely a tightness term).
We assume that both the samples representations  and the classifier weights are normalized vectors and that the classifier is linear with no bias term. We define the following loss to learn the desired prototypes:
\begin{equation}
    \ell_\text{tt}=\frac{1}{N}\sum_i^{N}\ell_\text{tt}(\z_i,\th_{y_i})=\frac{1}{N} \sum_i^{N} -\z_i^\top\th_{y_i}.
    \label{eq:tt}
\end{equation}
Note that the number of samples in~\eqref{eq:tt} might differ from $N$ (\eg, due to augmentation), in which case  $N$  should be replaced by the corresponding number of samples.
With that assumption, the classifier we use is a nearest prototype classifier i.e., assigning a test sample to the class of the nearest prototype.
Note that $\ell_\text{tt}$  resembles only the tightness part of the \CE loss~\eqref{eq:l_CE}. 
% If we inspect the gradient of this loss w.r.t. the classifier weights, we get the following:
The gradient of the $\ell_\text{tt}$ loss w.r.t.\@ the classifier weights can be directly derived from~\eqref{eq:tt}:
\begin{equation}
    \frac{\partial\ell_\text{tt}}{\partial\th_k}=
    -\frac{1}{N} \sum_{i: y_i=k}\z_i.
\end{equation}
Through minimizing this loss jointly with the representation learning loss, we update the classifier weights using the following iterative formula:
\begin{equation}
    \th^{0}_k=\eta\frac{1}{N}\sum_{i: y_i=k}\z_i^0,\;\;\; 
    \th^{t+1}_k=\th^t_k +\eta\frac{1}{N}\sum_{i: y_i=k}\z_i^{t+1},
\end{equation}
where $t$ is the iteration index and $\eta$ is the learning rate.
Note that this is equivalent to setting (up to a constant) the class weights $\th_k$ to the  class features mean $\c_k$ with momentum updates, where the new prototype combines the new iteration representation mean with the previous iteration mean. % (the first term of the previous assignment add eqref). 
We will compare the minimization of the $\ell_\text{tt}$ loss jointly with the the representation learning loss vs.\@ simply setting the classifier weights $\th_k$ to the hard mean $\c_k$ for each class $k$.

\section{Extended Supervised Contrastive Learning}\label{sec:ESupcon}
Here we aim at extending the \SupCon loss to include the classes prototypes being learned.
For this, we propose to consider an explicit linear classification layer with parameters $\th=(\th_1, \dots,\th_K)$ in the optimization of the supervised contrastive loss~\eqref{eq:supcon}.
Note that here we consider the embeddings $\z_i$ and the class prototypes $\th_k$ in the same feature space.
A class prototype $\th_k$ should  represent as closely as possible its class features. Hence  a  prototype similarity  with its class features should be maximized  and minimized with other classes features. To achieve this we propose to construct the following prototype-feature pair  $(\z_i,\th_{y_i})$ with sample representation $\z_i$ $(y_i=k)$ as a positive pair. 
Now we define the following loss on a positive prototype-feature pair: 
\begin{equation}
  \begin{aligned}[b]
&\ell_{\pt}(\z_i,\th_{y_i}) = -\z_i^\top\th_{y_i}\\ 
&+ \log\left(\sum_{k=1}^{K}\exp(\z_i^\top\th_k)+\sum_{j=1:j\ne i}^{2N} \exp(\z_i^\top\z_j)\right).
\end{aligned}
\label{eq:l_clSup_pairwise}
\end{equation}
Note that SupCon loss on a positive pair of samples is defined as follows:
\begin{equation}
\ell_\SupCon(\z_i,\z_p)=-\z_i^\top\z_p + \log \sum_{j\ne i}\exp({\z_i^\top\z_j)}.
\label{eq:l_SupCon_pairwise}
\end{equation}

Here we omit the temperature $\tau$ for clarity and for a better connection to the \CE loss.
In \eqref{eq:l_clSup_pairwise} we have extended the  set of existing data representations $\z_i$ with the class prototypes $\th_l$. 
Following the same analogy and constructing all positive prototype-feature pairs, the prototype loss  for a class weight $\th_k$ will be defined as follows.
\begin{equation}
\ell_{\pt}(\th_k)=\dfrac{1}{2N_k}\sum_{i:y_i=k} \ell_{\pt}(\z_i,\th_{k}).
    \label{eq:l_clSup}
\end{equation}
Note that the number of summation terms in~\eqref{eq:l_clSup} is $2N_k$ (where $N_k$ is the number of the non-augmented samples in $k$-th class), since the samples in \SupCon are considered with their augmentations.
Having the loss defined per prototype $\th_k$, we can define the full objective function that optimizes the encoder (representation backbone) parameters jointly with the classifier parameters $\th$ as:
\begin{equation}\label{eq:Esupcon}
\ell_{\ESupCon}=\frac{1}{2N+K}\left(\sum_{k=1}^{K}  \ell_\pt(\th_k)  +\sum_{i}^{2N}\ell_\SupCon(\z_i,P_i)\right).
\end{equation}
Next we show that our proposed prototype loss $\ell_{\pt}(\z_i,\th_{k})$ for a given positive pair can be expressed in terms of \SupCon loss on that positive pair and \CE loss  on the concerned sample . 
Let us define the following: 
\begin{equation}
  \begin{aligned}[b]
  T&=\z_i^\top\th_{y_i},\\
  C_1&=\sum_{k=1}^{K} \exp(\z_i^\top\th_k),\\
  C_2&=\sum_{j=1:{j\ne i}}^{2N} \exp(\z_i^\top\z_j),\\
  \exp(\ell_\CE(\z_i)) &= \exp(-T+\log(C_1))\\
        &=\exp(-T)C_1,\\
  \exp(\ell_\SupCon(\z_i, \th_{y_i})) &= \exp\left(-T+\log(C_2 +\exp(T)\right)\\
        &=\exp(-T)(C_2+\exp(T)),
  \end{aligned}
\end{equation}
where $T$ is the tightness term, $C_1$ is the first contrastive term and $C_2$ is the second contrastive term, $\ell_\CE(\z_i)$ is the CE loss for a sample $\z_i$, and 
% $\ell_\text{SupC*}$ is
the SupCon loss $\ell_\SupCon(\z_i,\th_{y_i})$ is estimated after including $\th_{y_i}$ into the pool of representations.

Then our loss for the $(\z_i,\th_{y_i})$ pair can be expressed as:
\begin{equation}
\begin{split}
&\ell_{\pt}(\z_i,\th_{y_i}) = -T + \log(C_1 + C_2)\\
&=\log\left(\exp(-T +\log(C_1 + C_2))\right)\\
&=\log\left(\exp(-T)(C_1 + C_2)\right)\\
&=\log\left(\exp(-T) (C_1 + C_2 + \exp(T) -\exp(T))\right) \\
&=\log(\exp(-T)C_1 + \exp(-T)(C_2 + \exp(T))\\
&-\exp(-T)\exp(T))\\
&=\log\left(\exp(\ell_\CE(\z_i))+  \exp(\ell_\SupCon(\z_i,\th_{y_i}))-1\right).
\end{split}
\end{equation}

As such, minimizing $\ell_{\pt}(\z_i,\th_{y_i})$ is minimizing the log sum  exponential (LSE) of cross entropy loss and supervised contrastive loss for a given positive pair $(\z_i,\th_{y_i})$, a smooth approximation to the max function.
Note that $\ell_{\pt}(\z_i,\th_{y_i})=0\;\iff\; \ell_{\CE}(\z_i)=\ell_{\SupCon}(\z_i,\th_{y_i})=0$.

We refer to the loss in \eqref{eq:Esupcon} as \ESupCon, short for Extended Supervised Contrastive learning.
In the following, we will extensively compare the different studied loss functions. 

\section{Experiments}\label{sec:exp}
\begin{table}[t]
    \begin{center}
    \resizebox{\textwidth}{!}{
        \begin{tabular}{l|l|l|l|l|l}
    \toprule
    Method &
    CIFAR-10 &
    CIFAR-100 &
    Tiny ImageNet &
     Caltech256 &
    Avg.
    \\

        \midrule
    {\tCE}  &
    $95.39$ & %
    $76.36$ &% 1.76
    $65.76$&%4.22
    $55.9$\ &%10.85
{$-$}
    \\
    \midrule
    *{\tsupconce}  & 
    % 0.1100, -0.4600, -0.2000,  2.0100
    $95.50$ \textcolor{blue}{$+0.11$} & 
    $75.90$ \textcolor{red}{$-0.46$} &
    $65.56$ \textcolor{red}{$-0.20$} &
    $57.91$ \textcolor{blue}{$+2.01$} &
    \textcolor{blue}{$+0.36$}
    \\
    
    *{\tsupconceN} &
    %([-0.1200, -1.7900, -4.0700, -2.98])
    $95.27$ \textcolor{red}{$-0.12$} & 
    $74.57$ \textcolor{red}{$-1.79$} &
    $61.69$ \textcolor{red}{$-4.07$} &
     $52.92$ \textcolor{red}{$-2.98$} &
    \textcolor{red}{$-1.52$}
    \\
    
    *{\tsupcontt} &
    %[-0.1900, -1.5600, -6.1000,  1.5200
    $95.20$ \textcolor{red}{$-0.19$} &
    $74.80$ \textcolor{red}{$-1.56$} & 
    $59.66$ \textcolor{red}{$-6.1$} &
    $57.42$ \textcolor{blue}{$1.52$} &
    \textcolor{red}{$-2.24$} 
    \\
    \midrule
    
    {\tSPCE} & 
    %0.2300,  1.7900,  0.7600, -7.44
    ${95.62}$ \textcolor{blue}{$+0.23$} &
    $\underline{78.15}$ \textcolor{blue}{$+ 1.79$} &
    $\underline{66.52}$ \textcolor{blue}{$+0.76$} &
     ${48.46}$ \textcolor{red}{$-7.44$} &
    \textcolor{red}{$-1.16$}
    \\
    
    {\tSPCE}({\tt M}) &
    % -0.0900,   1.1300,   0.5200 ,-7.52
    $95.30$ \textcolor{red}{$-0.09$} & 
    $77.49$ \textcolor{red}{$+1.13$} &
    $66.28$ \textcolor{blue}{$+0.52$} &
     $48.37$ \textcolor{red}{$-7.52$}&
    \textcolor{red}{$-1.49$}
    \\

    {\ExtSupCon} & 
    %[0.5100, 0.5600, 0.4400, 2.37
   $\underline{ 95.9}$ \textcolor{blue}{$+0.51$} & 
    $76.92$ \textcolor{blue}{$+0.56$} & 
    $66.2$ \textcolor{blue}{$+0.44$} &
     $\underline{58.27}$ \textcolor{blue}{$+2.37$} &
    \textcolor{blue}{$\underline{+0.97}$}
    \\
    \bottomrule
    \end{tabular}}
    \caption{Accuracy $(\%)$ of the different studied and proposed losses on fully labelled  datasets. * indicates the use of a projection head. Absolute gains over cross entropy are reported \textcolor{blue}{blue} and absolute declines in \textcolor{red}{red}. The last column shows an average improvement or decline over {\CE}, across the datasets.}
    \label{tab:full_dataset}
    \end{center}
\end{table}

This section serves to compare the performance of deep models trained under the different objective functions discussed earlier including tightness loss term~\eqref{eq:tt} and \ESupCon~\eqref{eq:Esupcon}. Our goal is to perform an extensive evaluation of the different losses behaviour not only under fully supervised setting but also under more challenging yet more plausible settings, namely limited data, imbalanced data and noisy labels settings. For the purpose of this experimental validation,  we focus on the object recognition problem.  

\subsection{Datasets}
We consider Cifar-100, Cifar-10~\cite{krizhevsky2009learning}, Tiny ImageNet~\cite{tinyimgnet} (a subset of $200$ classes from ImageNet~\cite{deng2009imagenet}, rescaled to the $32\times32$) datasets and Caltech256~\cite{griffin2007caltech}.
We refer to the supplementary materials for more results.

\begin{table*}[t]
    \begin{center}
    \setlength{\tabcolsep}{2pt}
    \resizebox{\textwidth}{!}{
        \begin{tabular}{l|l|l|l|l|l|l|l|l|l|l}
    \toprule
    \multirow{2}[1]{*}{Method} &
    \multicolumn{3}{c|}{CIFAR-10} &
    \multicolumn{3}{c|}{CIFAR-100} &
    \multicolumn{3}{c|}{Tiny ImageNet} &
    \multirow{2}[1]{*}{Avg.}
    \\
    \cline{2-10}
     & 
     $N=2\text{K}$ &
     $N=5\text{K}$ &
     $N=10\text{K}$ &
     $N=8\text{K}$ &
     $N=10\text{K}$ &
     $N=20\text{K}$ &
     $N=20\text{K}$ &
     $N=50\text{K}$ &
     $N=70\text{K}$ &
     \\
    \midrule 

    % {\tCE} &
    % $44.10$ &%28.02
    % $63.02$ &% 69.91
    % $78.68$ &%85.08
    % $33.51$ &%43.67
    % $40.61$ &%51.09
    % $55.30$ &%64.31
    % $35.50$ &%44.29
    % $52.78$ &%57.19
    % $57.53$ &%60.94
    % -
    % \\
    \midrule
        {\tCE} &
    $28.02$ &%-16.08
    $69.91$&% 6.88
    $85.08$ &%6.39
    $43.67$ &%10.16
    $51.09$&%10.48
    $64.31$&%9.01
    $44.29$ &%8.79
    $57.19$&%4.409
    $60.94$ &%3.40
   -
    \\
    \midrule
    *{\tsupconce}&
    %44.2500, 12.4600,  2.9500,  7.2900,  3.4000,  0.0800, -0.2900,  2.0500, 1.9400
    $72.27$ \textcolor{blue}{$+44.25$} & 
    $82.37$ \textcolor{blue}{$+12.46$} &
    $88.03$ \textcolor{blue}{$+2.95$} &
    $50.96$ \textcolor{blue}{$+7.2$} &
    $54.49$ \textcolor{blue}{$+3.4$} &
    $64.39$ \textcolor{blue}{$+ 0.08$} &
    $44.00$ \textcolor{blue}{$-0.29$} &
    $59.24$ \textcolor{blue}{$+2.05$} &
    $62.88$ \textcolor{blue}{$+1.94$} &
    \textcolor{blue}{$+8.22$}
    \\

    *{\tsupconceN} &
    %43.9700, 12.8200,  2.8300,  6.9300,  2.8300, -1.0400, -0.7900,  0.4300,
     %   -1.3200
    $71.99$ \textcolor{blue}{$+43.97$} & 
    $82.73$ \textcolor{blue}{$+12.82$} &
    $87.91$ \textcolor{blue}{$+ 2.83$} &
    $50.60$ \textcolor{blue}{$+6.93$} &
    $53.92$ \textcolor{blue}{$+ 2.83$} &
    $63.27$ \textcolor{red}{$ -1.04$} &
    $43.50$ \textcolor{red}{$-0.79$} &
    $57.62$ \textcolor{blue}{$+0.43$} &
    $59.62$ \textcolor{red}{$-1.32$} &
    \textcolor{blue}{$+7.41$}
    \\
    %tensor([ 4.4150e+01,  1.3060e+01,  2.2900e+00,  7.5600e+00,  3.4000e+00,        -2.9999e-02, -4.7000e-01, -4.9800e+00, -3.0600e+00])

    *{\tsupcontt} &
    $72.17$ \textcolor{blue}{$+44.15$} &
    $82.97$ \textcolor{blue}{$+13.06$} &
    $87.37$ \textcolor{blue}{$+2.29$} &
    $51.23$ \textcolor{blue}{$+7.56$} &
    $54.49$ \textcolor{blue}{$+3.4$} &
    $64.28$ \textcolor{red}{$-0.02$} &
    $43.82$ \textcolor{red}{$-0.47$} &
    $52.21$ \textcolor{red}{$-4.98$} &
    $57.88$ \textcolor{red}{$-3.06$} &
    \textcolor{blue}{$+6.88$}
    \\
    \midrule
    
    {\tSPCE} &
    %tensor([ 3.7900,  8.6900,  1.0700,  6.4200,  2.6900,  0.5100, -3.3500, -1.4900,  -5.4500])

    $31.81$ \textcolor{blue}{$+3.79$} &
    $78.60$ \textcolor{blue}{$+8.69$}&
    $86.15$ \textcolor{blue}{$+1.07$} &
    $50.09$ \textcolor{blue}{$+6.42$} &
    $53.78$ \textcolor{blue}{$+2.69$} &
    $64.82$ \textcolor{blue}{$+0.51$} &
    $40.94$ \textcolor{red}{$-3.35$} &
    $55.70$ \textcolor{red}{$-1.49$} &
    $55.49$ \textcolor{red}{$-5.45$} &
    \textcolor{blue}{$+1.43$}
    \\
    
    % \ExtSupCon &
    % $72.42$ \textcolor{blue}{$+28.3$} &
    % $82.03$ \textcolor{blue}{$+19.0$} &
    % $87.04$ \textcolor{blue}{$+8.3$} &
    % $50.10$ \textcolor{blue}{$+16.5$} &
    % $53.33$ \textcolor{blue}{$+12.7$} &
    % $63.21$ \textcolor{blue}{$+7.9$} &
    % $42.61$ \textcolor{blue}{$+7.1$} &
    % $55.30$ \textcolor{blue}{$+2.5$} &
    % $59.97$ \textcolor{blue}{$+2.4$} &
    % \textcolor{blue}{$+11.6$}
    % \\
   {\ExtSupCon} & 
   %46.0600, 13.9800,  3.7500,  4.5900,  1.4900, -1.1900, -0.1200,  1.4700, 1.6800])

       $74.08$ \textcolor{blue}{$+46.06$} &
    $83.89$ \textcolor{blue}{$+13.98$} &
    $88.83$ \textcolor{blue}{$+3.75$} &
    $48.26$ \textcolor{blue}{$+4.59$} &
    $52.58$ \textcolor{blue}{$+1.49$} &
    $63.12$ \textcolor{red}{$-1.19$} &
    $44.17$ \textcolor{red}{$ -0.12$} &
    $58.66$ \textcolor{blue}{$+1.47$} &
    $62.62$ \textcolor{blue}{$+ 1.68$} &
    \textcolor{blue}{$+7.97$}
   \\
    \bottomrule
    \end{tabular}}
    \caption{Accuracy $(\%)$ on CIFAR-10, CIFAR-100 and Tiny ImageNet for a low-sample training scenario, where $N$ represents the number of samples used for the training. Absolute gains over cross entropy are reported in \textcolor{blue}{blue} and absolute declines in \textcolor{red}{red}. * indicates the use of a projection head. The last column shows an average improvement or decline over cross entropy ({\CE}), across the datasets and the settings.}
    \label{tab:lowsamples}
    \end{center}
\end{table*}
\begin{table*}[t]
    \begin{center}
    \setlength{\tabcolsep}{2pt}
    \resizebox{\textwidth}{!}{
        \begin{tabular}{l|l|l|l|l|l|l|l|l|l|l}
    \toprule
    \multirow{2}[1]{*}{Method} &
    \multicolumn{3}{c|}{CIFAR-10} &
    \multicolumn{3}{c|}{CIFAR-100} &
    \multicolumn{3}{c|}{Tiny ImageNet} &
    \multirow{2}[1]{*}{Avg.}
    \\
    \cline{2-10}
     & 
     $\text{IR}=0.05$ &
     $\text{IR}=0.1$ &
     $\text{IR}=0.5$ &
     $\text{IR}=0.05$ &
     $\text{IR}=0.1$ &
     $\text{IR}=0.5$ &
     $\text{IR}=0.05$ &
     $\text{IR}=0.1$ &
     $\text{IR}=0.5$ &
     \\
    \midrule 

    % {\tCE} &
    % $81.50$ &%82.85
    % $86.48$ &% 87.83
    % $93.55$ &%93.99
    % $46.55$ &%48.57
    % $51.62$ &%54.44
    % $71.65$ &%71.19
    % $38.71$ &%40.65
    % $43.28$ &%46.16
    % $56.29$ & %60.30
    % -
    % \\
    % \midrule
    
    {\tCE} &
    $82.85$ &%
    $87.83$&% 1.34
    $93.99$&%0.43
    $48.57$&%2.02
    $54.44$&%2.82
    $71.19$ &%-0.460
    $40.65$&%1.93
    $46.16$&%2.87
    $60.30$ & %4.0
{$-$}
    \\
    \midrule
    *{\tsupconce}&
    %tensor([ -2.9100,  -0.9700,   0.3500,  -1.7800, -10.2300,  -0.0600,   4.3100,          3.4100,   2.1500])
    $79.94$ \textcolor{red}{$-2.91$} & 
    $86.86$ \textcolor{red}{$-0.97$} &
    $94.34$ \textcolor{blue}{$+0.35$} &
    $46.79$ \textcolor{red}{$-1.78$} &
    $44.21$ \textcolor{red}{$-10.23$} &
    $71.13$ \textcolor{red}{$-0.06$} &
    $44.96$ \textcolor{blue}{$+4.31$} &
    $49.57$ \textcolor{blue}{$+3.41$} &
    $62.45$ \textcolor{blue}{$+2.15$} &
    \textcolor{red}{$-0.64$}
    \\

    *{\tsupconceN} &
    %-35.0800, -40.1900,  -3.8500,  -8.5700, -14.1200, -15.2500,  -4.9600,
    %    -10.5500, -22.6000
    $47.77$ \textcolor{red}{$-35.08$} & 
    $47.64$ \textcolor{red}{$-40.19$} &
    $90.14$ \textcolor{red}{$-3.85$} &
    $40.00$ \textcolor{red}{$-8.57$} &
    $40.32$ \textcolor{red}{$ -14.12$} &
    $55.94$ \textcolor{red}{$-15.25$} &
    $35.69$ \textcolor{red}{$-4.96$} &
    $35.61$ \textcolor{red}{$-10.55$} &
    $37.70$ \textcolor{red}{$-22.60$} &
    \textcolor{red}{$-17.24$}
    \\
    
    *{\tsupcontt} &
    %2.7700,  0.9300,  0.4100,  5.8300,  2.3500, -0.8100,  3.4600,  1.2300,        -3.0000]
    $85.62$ \textcolor{blue}{$+2.7$} &
    $88.76$ \textcolor{blue}{$+0.93$} &
    $94.40$ \textcolor{blue}{$+ 0.41$} &
    $54.40$ \textcolor{blue}{$+5.83$} &
    $56.79$ \textcolor{blue}{$+ 2.35$} &
    $70.38$ \textcolor{red}{$-0.81$} &
    $44.11$ \textcolor{blue}{$+3.46$} &
    $47.39$ \textcolor{blue}{$+1.23$} &
    $57.30$ \textcolor{red}{$ -3.00$} &
    \textcolor{blue}{$+1.46$}
    \\
    \midrule
    
    {\tSPCE} &
    % 2.7700, -0.8900, -0.0400,  1.0200, -0.6600, -2.5800, -3.3800, -5.6100,         0.8400
    $85.62$ \textcolor{blue}{$+4.5$} &
    $86.94$ \textcolor{blue}{$+2.6$}&
    $93.95$ \textcolor{blue}{$+0.7$} &
    $49.59$ \textcolor{blue}{$+6.7$}  &
    $53.78$ \textcolor{blue}{$+5.4$}  &
    $68.61$ \textcolor{red}{$-1.6$} &
    $37.27$ \textcolor{red}{$+2.0$} &
    $40.55$ \textcolor{red}{$+1.5$} &
    $61.14$ \textcolor{blue}{$+3.5$} &
    \textcolor{red}{$-0.95$}
    \\
    
    % \ExtSupCon &
    % $86.00$ \textcolor{blue}{$+4.5$} &
    % $89.12$ \textcolor{blue}{$+2.6$} &
    % $94.25$ \textcolor{blue}{$+0.7$} &
    % $53.29$ \textcolor{blue}{$+6.7$} &
    % $57.05$ \textcolor{blue}{$+5.4$} &
    % $70.04$ \textcolor{red}{$-1.6$} &
    % $40.74$ \textcolor{blue}{$+2.0$} &
    % $44.86$ \textcolor{blue}{$+1.5$} &
    % $59.85$ \textcolor{blue}{$+3.5$} &
    % \textcolor{red}{$-0.95$}
  %  \\
      {\ExtSupCon} & 
      %[3.1500, 1.4300, 0.7800, 4.1700, 3.6400, 0.1800, 4.9000, 4.7400, 2.7800])
   $86.00$ \textcolor{blue}{$+3.15$} &
    $89.26$ \textcolor{blue}{$+1.43$} &
    $94.77$ \textcolor{blue}{$+0.78$} &%this number has to be updated
    $52.74$ \textcolor{blue}{$+ 4.17$} &
    $58.08$ \textcolor{blue}{$+3.64$} &
    $71.37$ \textcolor{blue}{$+0.18$} &
    $45.55$ \textcolor{blue}{$ +4.90$} &
    $50.90$ \textcolor{blue}{$+4.74$} &
    $63.08$ \textcolor{blue}{$+2.78$} &
    \textcolor{blue}{$+2.86$}
    \\
    \bottomrule
    \end{tabular}}
    \caption{Accuracy $(\%)$ on CIFAR-10, CIFAR-100 and Tiny ImageNet for an imbalanced training scenario, where IR represents the rate of imbalance. Absolute gains over cross entropy are reported \textcolor{blue}{blue} and absolute declines in \textcolor{red}{red}.* indicates the use of a projection head. The last column shows an average improvement or decline over cross entropy ({\CE}), across the datasets and the settings.}
    \label{tab:imbalancesamples}
    \end{center}
\end{table*}

\subsection{Methods and Implementation Details}
In all experiments we use ResNet50 as a main network and
 evaluate the following losses:

{\CE}: we  optimize the network parameters using the standard CE loss.
For the SupCon loss~\cite{khosla2020supervised}, we use the publicly available implementation, which uses L2-normalized outputs of a multi-layer head (FC, ReLU, FC), a projection head, on top of the embeddings used  for classification. We learn the classifier parameters using:
i) Cross entropy loss ({\tsupconce}), on the linear layer after optimizing minimizing SupCon loss.  ii) For the sake of fair comparison with other losses, we consider also cross entropy loss with no bias term,  normalized embeddings and normalized classifier weights. We denote this variant by {\tsupconceN}.
iii) Tightness loss ({\tsupcontt}), where we optimize the parameters of a linear classifier using \eqref{eq:tt} during the optimization of the rest of the network (projection head + backbone) with SupCon loss.  Note that  the gradients of the tightness loss are not propagated to the rest of the network.
 {\tSPCE}: we optimize the backbone with SPCE loss~\eqref{eq:spce} and  the classifier weights with the tightness term~\eqref{eq:tt}.
We also show the performance with directly assigning the weights to the mean of each class samples {{\tSPCE}({\tt M}) }. 

 Our {\ExtSupCon}: with~\eqref{eq:Esupcon} we optimize jointly a linear classifier and the backbone parameters.

Note that  {{\tsupconce}, {\tsupconceN} and {\tsupcontt}} use a projection head, unlike {\CE}, {\tSPCE} and {\ExtSupCon}.
All studied variants benefit from the same type of data augmentations and hyper-parameters  were estimated on Cifar-10 dataset and fixed for the rest. We refer to the supplementary materials for more details.

\subsection{Fully Supervised Classification}
We first start by comparing the different studied methods  on the standard classification setting while leveraging all the labelled training data of each dataset. Table~\ref{tab:full_dataset} shows the average test  accuracy at the end of the training on the three considered datasets. 

First,    {\ExtSupCon} outperforms {\CE} training alone, using the same number of parameters. {\tsupconce} improves over  {\CE}. {\tsupcontt} is comparable to {\tsupconceN}.

{\ExtSupCon} shows the best performance on all four datasets.
Except from Caltech dataset,  {\tSPCE} achieves superior results to {\CE}. When assigning the classifier weights directly to the mean of the features, {\tSPCE}({\tt M}), results are slightly inferior to the use of our tightness loss~\eqref{eq:tt} for training the classifier parameters. For the rest the of experiments, we show only {\tSPCE}, using the suggested tightness term to train the classifier parameters.

\subsection{Classification in Low-Sample Scenario}
\begin{table*}[h!]
    \begin{center}
    \setlength{\tabcolsep}{2pt}
    \resizebox{\textwidth}{!}{
        \begin{tabular}{l|l|l|l|l|l|l|l|l|l|l}
    \toprule
    \multirow{2}[1]{*}{Method} &
    \multicolumn{3}{c|}{CIFAR-10} &
    \multicolumn{3}{c|}{CIFAR-100} &
    \multicolumn{3}{c|}{Tiny ImageNet} &
    \multirow{2}[1]{*}{Avg.}
    \\
    \cline{2-10}
     & 
     $\text{NR}=0.5$ &
     $\text{NR}=0.3$ &
     $\text{NR}=0.2$ &
     $\text{NR}=0.5$ &
     $\text{NR}=0.3$ &
     $\text{NR}=0.2$ &
     $\text{NR}=0.5$ &
     $\text{NR}=0.3$ &
     $\text{NR}=0.2$
     \\
    \midrule 

    % {\tCE} &
    % $59.29$ &%60.88
    % $85.91$ &% 87.08
    % $87.15$ &%88.93
    % $33.99$ &%35.47
    % $55.10$ &%56.57
    % $63.23$ &%64.93
    % $28.83$ &%31.55
    % $44.65$ &%49.62
    % $51.23$ &%55.40
    % -
    % \\
    % \midrule
        {\tCE} &
    $60.88$ &%
    $87.08$ &% 
    $88.93$  &%
    $35.47$ &%1.479
    $56.57$&%1.46
    $64.93$&%1.70
    $31.55$ &%2.72
    $49.62$ &%4.96
    $55.40$ &%4.17
-
    \\
    \midrule
    *{\tsupconce}&
    %12.8000, 12.6100,  2.9900,  0.6900, -1.4900, -0.6400, -0.2600,  3.4200,         0.6600
    $48.08$  \textcolor{red}{$-12.8$} & 
    $74.47$  \textcolor{red}{$-12.61$} &
    $85.94$  \textcolor{red}{$-2.99$} &
    $34.78$  \textcolor{red}{$-0.69$} &
    $58.06$  \textcolor{blue}{$+1.49$} &
    $65.57$  \textcolor{blue}{$+0.64$} &
    $31.81$  \textcolor{blue}{$+0.26$} &
    $46.20$  \textcolor{blue}{$+3.42$} &
    $54.74$  \textcolor{blue}{$+0.66$} &
    \textcolor{red}{$-3.42$}
    \\

    *{\tsupconceN} &
    %-14.5300,  -9.6600,  -1.4800,  -1.9200,   5.8700,   2.9400,  -2.6700,
    %      5.0200,   3.2200
    $46.35$  \textcolor{red}{$-14.53$} & 
    $77.42$  \textcolor{red}{$-9.66$} &
    $87.45$  \textcolor{red}{$-1.48$} &
    $33.55$  \textcolor{red}{$-1.92$} &
    $62.44$  \textcolor{blue}{$+5.87$} &
    $67.87$  \textcolor{blue}{$+ 2.94$} &
    $28.88$  \textcolor{red}{$-2.67$}&
    $54.64$  \textcolor{blue}{$+5.02$} &
    $58.62$  \textcolor{blue}{$+3.22$} &
    \textcolor{red}{$-1.47$}
    \\
    
    *{\tsupcontt} &
    %-2.8300,  2.6200,  1.7300,  1.7600, 11.1900,  4.4800, -2.8800,  5.1900,         2.5300
    $58.05$  \textcolor{red}{$-2.83$} &
    $89.70$  \textcolor{blue}{$+2.62$} &
    $90.66$  \textcolor{blue}{$+ 1.73$} &
    $37.23$  \textcolor{blue}{$+1.76$} &
    $67.76$  \textcolor{blue}{$+11.19$} &
    $69.41$  \textcolor{blue}{$+4.48$} &
    $28.67$  \textcolor{red}{$-2.88$} &
    $54.81$  \textcolor{blue}{$+5.19$} &
    $57.93$  \textcolor{blue}{$+2.53$} &
    \textcolor{blue}{$+2.64$}
    \\
    \midrule
    
    {\tSPCE} &
    %4.7500,  1.6900,  0.0000,  1.0900,  3.7800,  0.8200, -6.2800, -7.1700,        -5.8800
    $65.63$  \textcolor{blue}{$+4.75$} &
    $88.77$  \textcolor{blue}{$+1.69$} &
    $88.93$  \textcolor{blue}{$+0.0$} &
    $36.56$  \textcolor{blue}{$+1.09$} &
    $60.35$  \textcolor{blue}{$+3.78$} &
    $65.75$  \textcolor{blue}{$+0.82$} &
    $25.27$  \textcolor{red}{$-6.28$} &
    $42.45$  \textcolor{red}{$-7.17$} &
    $49.52$  \textcolor{red}{$-5.88$} &
    \textcolor{red}{$-0.80$}
    \\
    
    % \ExtSupCon &
    % $61.17$ \textcolor{blue}{$+1.8$} &
    % $88.01$  \textcolor{blue}{$+2.1$} &
    % $89.70$  \textcolor{blue}{$+2.5$}&
    % $36.05$ \textcolor{blue}{$+2.0$} &
    % $61.66$  \textcolor{blue}{$+6.5$} &
    % $64.29$  \textcolor{blue}{$+1.0$} &
    % $30.75$  \textcolor{blue}{$+1.9$} &
    % $47.87$  \textcolor{blue}{$+3.2$} &
    % $55.91$  \textcolor{blue}{$+4.6$} &
    % \textcolor{blue}{$+2.89$}
    % \\
  {\ExtSupCon}& 
  %[-1.7000,  1.0700,  1.9900,  1.5700,  6.3700,  0.8700,  1.0050,  3.1800,         1.3900
      $59.18$ \textcolor{red}{$-1.70$} &
    $88.15$  \textcolor{blue}{$+1.07$} &
    $90.92$  \textcolor{blue}{$+1.99$}&
    $37.04$ \textcolor{blue}{$+1.57$} &%needs to be updated
    $62.94$  \textcolor{blue}{$+6.37$} &
    $65.81$  \textcolor{blue}{$+0.87$} &
    $32.555$  \textcolor{blue}{$+1.0$} &
    $52.80$  \textcolor{blue}{$+3.18$} &
    $56.79$  \textcolor{blue}{$+1.39$} &
    \textcolor{blue}{$+1.75$}
    \\
    \bottomrule
    \end{tabular}}
    \caption{Accuracy $(\%)$ on CIFAR-10, CIFAR-100 and Tiny ImageNet for a noisy training scenario,  NR represents the rate of noise. Absolute gains over cross entropy are reported in \textcolor{blue}{blue} and absolute declines in \textcolor{red}{red}. * indicates the use of a projection head. The last column shows an average improvement or decline over cross entropy ({\CE}), across the datasets and the settings.}
    \label{tab:noisysamples}
    \end{center}
\end{table*}

\begin{figure}[h!]
    \centering
    \includegraphics[width=\textwidth]{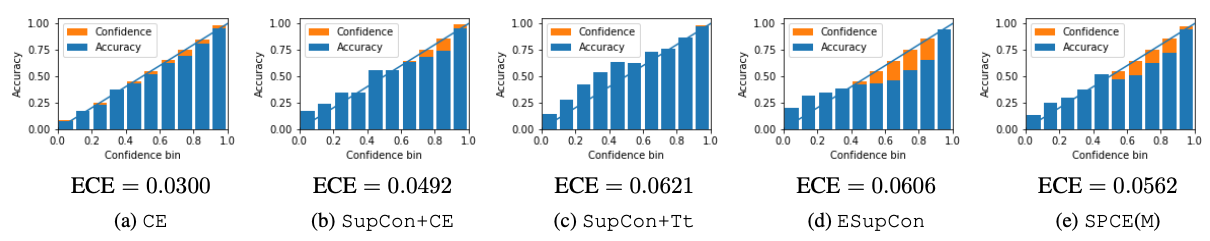}
    \caption{Reliability Diagrams and Expected Calibration Error of probabilistic classifiers learned with different studied loss functions and further calibrated by temperature scaling.}
    \label{fig:reliability_diagrams}
\end{figure}

%===========
After studying the fully labelled scenario,  here, we are interested in the performance under limited data setting. Our goal is to see how prone each method is to overfitting in low data regime and whether significant differences can be observed among the different alternatives. Table~\ref{tab:lowsamples} reports the average test accuracy on Cifar-10, Cifar-100 and Tiny ImageNet  using different numbers of training samples ($N$).

While {\CE} performance is comparable to other losses on the full data scenario, here it is significantly lower than other competitors with a gap increasing as the sample size gets smaller. Except from Tiny ImageNet, {\tsupcontt}  shows comparable performance to {\tsupconce} and is slightly inferior ($0.5\%$) to {\tsupconceN} on average. {\tSPCE} results are  better than {\CE} on Cifar-10 and Cifar-100.  {\ExtSupCon} improves significantly over {\CE} while being comparable with {\tsupconce}, however, with no projection head.
{\ExtSupCon} is much more robust than  {\tSPCE} in this setting. 
  
\subsection{Classification under Imbalanced Data}

Our goal is to compare the performance of a model trained by the different studied losses under various challenging settings beside the standard fully supervised setting. 
Here, we examine the scenario where training data are not uniformly distributed.
Some classes are undersampled while others are oversampled. Specifically, we want to test the ability of the different losses to cope with this data nature and  learn  the underrepresented classes.
We simulate this scenario by altering the training data in which half of the categories are underrepresented with a number of samples equals to the imbalance rate (IR) of other categories samples.
The test set on which we report the average accuracy remains balanced.

Table~\ref{tab:imbalancesamples} reports the average test accuracy of models trained to minimize the different losses on the three considered datasets.
For each dataset we consider imbalance rates of $0.05$, $0.1$, and $0.5$ where, for example, an imbalance rate of $0.1$ means that the size of undersampled classes samples is $0.1$ compared to the oversampled classes size.

Here it seems that  {\tsupconce} doesn't improve over {\CE} alone.    {\tSPCE}   results  are marginally lower than {\CE}. 
Our two proposed losses {\tsupcontt} and {\ExtSupCon} exhibit more robust and powerful performance compared to {\CE} with {\ExtSupCon} performing the best. 

\subsection{Classification under Noisy Data}
We continue our investigation on the different losses performance under challenging setting and test 
 another interesting scenario: classification with noisy labels.
We want to test the ability of the different training regimes to  learn generalizable decision boundaries in spite of the presence of wrongly labelled samples. 
To simulate this scenario, during training a percentage of the training data, denoted by noise rate (NR), is associated with wrong labels (shuffled labels).
As in the previous experiments, we report the results on the standard, correctly labelled, test set.
Table~\ref{tab:noisysamples} reports the average test accuracy on Cifar-10, Cifar-100 and Tiny ImageNet with noise rates of ($0.2,0.3,0.5$).
Here we obtained similar results to the imbalanced settings,  {\tsupconce} doesn't consistently improve over {\CE}, same applies for {\tSPCE}. 
Our both proposed losses improve over  {\CE} with {\tsupcontt} performing the best here. 

\subsection{General Remarks}
 We note the following on the shown results of the different losses: CE training after {\SupCon} pretraining ({\tsupconce}) improves over standard \CE in full and low data regime. However, deploying {\CE} to learn the classifier with or without  {\SupCon} pretraining is sensitive to noise and data imbalance. Interestingly, our proposed tightness term is more effective on these two scenarios, however inferior on the full and low data regime. 
In all studied settings, our proposed   {\ExtSupCon} loss improves over {\CE}  and over ({\tsupconce}) on the challenging imbalanced and noisy settings.   
In Supplementary we discuss the computational complexity of the different losses and their sensitivity to hyper-parameters.
\subsection{Classifier Outputs as Posterior Probabilities}
To access the interpretation of the classifier outputs as estimates of posterior probabilities $\hat p(\text{class}|\text{observation})$, we calibrated the outputs by temperature scaling~\cite{guo2017calibration} -- we estimated the temperature on a holdout set ($20\%$ of the test set) and computed the reliability diagram and the expected calibration error (ECE) on the remaining test samples of Cifar-100 dataset.
Results are shown in Figure~\ref{fig:reliability_diagrams}: while the standard CE loss has the lowest calibration error, all other calibrated classifiers provide reliable predictions, an interesting result given the shown performance advantage.

\section{Conclusion}
\label{sec:esupcon_conclusion}
In this work, we derive novel, robust objective functions, inspired by new evidence showing that contrastive losses improve performance over \CE. Driven by the question of whether cross entropy loss is the best option to train jointly a good representation and powerful, generalizable, decision boundaries, we start from  a recent approximation to cross entropy loss (\SPCE) with pairwise  training of representation where classifier weights can be assigned to the mean of each class features. We then suggest to learn the classifier weights under only a tightness term jointly with \SupCon representation training or \SPCE. Next, we propose an extension to \SupCon, where the classifier weights are treated as learnable prototypes in the same space as the samples embeddings, and where data points form positive pairs with their classes prototypes. We show that the proposed loss for a given pair $(\z_i, \th_k)$ is a smooth approximation to the maximum of the \CE and \SupCon losses on that pair. To this point, we test the performance of models trained with the different discussed losses under different challenging settings. We show that the proposed extensions demonstrate more robust and stable performance across different settings and datasets. As a future work, we plan to extend the experiments to object detection and image segmentation problems, as well as to test the discussed losses on Out-Of-Distribution and Continual Learning benchmarks.
\chapter{Conclusions}
\label{chapter:conclusions}

The thesis contributes to bridging the gap between the user-defined objective and the training loss when the objective is known and non-differentiable. To this end, we proposed a technique to learn the surrogate of a decomposable and non-differentiable evaluation metric (Chapter \ref{chapter:ls})~\cite{patel2020learning}. The surrogate is learned jointly with the task-specific model in an alternating training setup. The approach showed a relative improvement of up to $39$\% on the total edit distance for scene text recognition and $4.25$\% on $F_{1}$ score for scene text detection.

After observing the noisy predictions from the surrogate at the initial stage of the training, we designed an approach to train robustly with the learned surrogate (Chapter \ref{chapter:feds})~\cite{patel2021feds}. The method filters out the samples that are hard for the surrogate. With this approach, we observed the merits of training a scene text recognition model using a learned surrogate of edit distance. We attained an average relative improvement of $11.2$\% on the total edit distance and an error reduction of $9.5$\% on accuracy on several benchmarks. For comparison, our previous approach~\cite{patel2020learning} obtains an relative improvement of $6.05$\% on the total edit distance.

We introduced a surrogate loss for recall@k, a non-decomposable and non-differentiable evaluation metric (Chapter \ref{chapter:rs@k})~\cite{patel2022recall}. When combined with a novel and efficient mixup technique and training on large batch sizes, the proposed approach attained state-of-the-art results on several metric learning benchmarks and instance-level search. Further, when combined with kNN classifier, it can also serve as an effective tool for fine-grained recognition where it substantially outperforms direct classification methods equipped with performance-enhancing techniques~\cite{picek2022plant}. 

We put forward an approach for supervised contrastive classification that jointly learns the parameters of the classifier and the backbone (Chapter \ref{chapter:esupcon})~\cite{aljundi2023contrastive}. This approach leverages the robustness of contrastive training and maintains the probabilistic interpretation useful for several calibration tasks. The method outperformed standard cross-entropy and supervised contrastive losses and was shown to be robust in various challenging settings such as class imbalance, label corruption, and training with a low number of samples.

\appendix
\chapter{Abstrakt}

\noindent {(\it Automatic translation by Google Translate)}\\

Mnoho důležitých úkolů počítačového vidění je přirozeně formulováno s nediferencovatelným cílem. Proto standardní, dominantní trénovací postup neuronové sítě není použitelný, protože zpětné šíření vyžaduje gradienty cíle vzhledem k výstupu modelu. Většina metod hlubokého učení obchází problém neoptimálně použitím proxy ztráty pro trénink, který byl původně navržen pro jiný úkol a není přizpůsoben specifikům cíle. Funkce proxy ztráty se mohou, ale nemusí dobře shodovat s původním nediferencovatelným cílem. Pro nový úkol musí být navržen vhodný proxy, který nemusí být proveditelný pro laika. Tato práce přináší čtyři hlavní příspěvky k překlenutí propasti mezi nediferencovatelným cílem a funkcí ztráty tréninku. Ztrátovou funkci v celé práci označujeme jako náhradní ztrátu, pokud se jedná o diferencovatelnou aproximaci nediferencovatelného cíle.

Nejprve navrhujeme přístup k učení diferencovatelné náhrady rozložitelné a nediferencovatelné vyhodnocovací metriky. Náhradník se učí společně s modelem specifickým pro úkol střídavým způsobem. Tento přístup je ověřen na dvou praktických úlohách rozpoznávání a detekce textu scény, kde se náhradník učí aproximaci vzdálenosti úprav a průsečíku přes spojení. V nastavení po vyladění, kde je model trénovaný se ztrátou proxy dále trénován s naučeným náhradníkem, navrhovaná metoda ukazuje relativní zlepšení až o $39$ \% celkové vzdálenosti úprav pro rozpoznání textu scény a $4,25$ \% na $F_{1}$ skóre za detekci textu scény.

Za druhé, vylepšená verze tréninku s naučeným náhradníkem, kde jsou odfiltrovány tréninkové vzorky, které jsou pro náhradníka těžké. Tento přístup je ověřen pro rozpoznávání textu scény. Překonává náš předchozí přístup a dosahuje průměrného zlepšení o $11,2\%$ celkové vzdálenosti úprav a snížení chyb o $9,5\%$ v přesnosti v několika oblíbených benchmarcích. Všimněte si, že dvě navrhované metody pro učení náhradníka a školení s náhradníkem nevytvářejí žádné předpoklady o daném úkolu a mohou být potenciálně rozšířeny na nové úkoly.

Za třetí, pro reminiscenci@k, nerozložitelnou a nediferencovatelnou vyhodnocovací metriku, navrhujeme ručně vytvořenou náhradu, která zahrnuje navrhování diferencovatelných verzí operací třídění a počítání. Je také navržena účinná kombinační technika pro učení metriky, která míchá skóre podobnosti namísto vkládacích vektorů. Navrhovaná náhrada dosahuje nejmodernějších výsledků na několika metrických učeních a srovnávacích testech vyhledávání na úrovni instancí v kombinaci se školením na velkých dávkách. Dále v kombinaci s klasifikátorem kNN slouží také jako účinný nástroj pro jemnozrnné rozpoznávání, kde překonává přímé klasifikační metody.

Za čtvrté navrhujeme ztrátovou funkci nazvanou Extended SupCon, která společně trénuje parametry klasifikátoru a páteře pro řízenou kontrastní klasifikaci. Navrhovaný přístup těží z robustnosti kontrastivního učení a zachovává pravděpodobnostní interpretaci jako soft-max predikci. Empirické výsledky ukazují účinnost našeho přístupu v náročných podmínkách, jako je třídní nerovnováha, korupce štítků a školení s málo označenými údaji.

Celkově přínosy této práce činí trénování neuronových sítí škálovatelnějším – na nové úkoly téměř bezpracně, když je vyhodnocovací metrika rozložitelná, což výzkumníkům pomůže s novými úkoly. Pro nerozložitelné vyhodnocovací metriky lze pro vytváření nových náhradních prvků použít také diferencovatelné komponenty vyvinuté pro náhradního prvku pro odvolání@k, jako je třídění a počítání.

\newpage
\noindent ({\it Automatic translation by ChatGPT}) \\

Mnoho důležitých úkolů v oblasti počítačového vidění je přirozeně formulováno s nenadifferentovatelným cílem. Standardní a dominantní tréninkový postup neuronové sítě tak není použitelný, protože zpětná propagace vyžaduje gradienty objektivu vzhledem k výstupu modelu. Většina metod hlubokého učení tento problém řeší podobným způsobem pomocí proxy ztráty pro trénování, která byla původně navržena pro jiný úkol a není přizpůsobena specifikám cíle. Funkce proxy ztráty mohou nebo nemusí dobře korespondovat s původní nenadifferentovatelným cílem. Pro nový úkol musí být navržena vhodná proxy, což pro neodborníka nemusí být proveditelné. Tato práce přináší čtyři hlavní přínosy pro překlenutí rozdílu mezi nenadifferentovatelným cílem a ztrátovou funkcí pro trénování. V celé práci označujeme ztrátovou funkci jako náhradní, pokud je diferencovatelnou aproximací nenadifferentovatelného cíle.

Nejprve navrhujeme přístup pro učení diferencovatelného náhradního modelu pro rozložitelnou a nespojitou hodnotící metriku. Náhrada je společně s úkolspecifickým modelem učena střídavým způsobem. Přístup je ověřen na dvou praktických úlohách rozpoznávání a detekce textu v scéně, kde náhrada modeluje aproximaci editační vzdálenosti a překryvu-union, odpovídajících. V post-tuningovém nastavení, kde model trénovaný pomocí proxy ztráty je dále trénován s naučenou náhradou, navrhovaná metoda ukazuje relativní zlepšení až o $39$\% celkové editační vzdálenosti pro rozpoznávání textu v scéně a $4.25$\% F$_{1}$ skóre pro detekci textu v scéně.

Druhá část našeho návrhu zahrnuje vylepšenou verzi tréninku pomocí naučeného náhradního modelu, kdy jsou filtrace tvrdých tréninkových vzorků založená na náhradním modelu. Tento postup byl ověřen pro rozpoznávání textu v obrazech. Tento nový postup předčil naše předchozí řešení a dosáhl průměrného zlepšení celkové editační vzdálenosti o $11.2\%$ a snížení chybovosti o $9.5\%$ v přesnosti na několika populárních testech. Poznamenejme, že oba navržené postupy, a to učení náhradního modelu a trénink s náhradním modelem, nekladou žádné předpoklady na řešený úkol a mohou být potenciálně rozšířeny na nové úkoly.

Třetí přístup se týká metriky recall@k, což je nedekomponovatelná a nespojitá metrika. Navrhujeme ručně vytvořený náhradní funkční prvek, který zahrnuje návrh diferencovatelných verzí řazení a počítání operací. Dále navrhujeme efektivní techniku mixup pro učení metrik, která míchá podobnostní skóre místo vektorů vnoření. Navrhovaný náhradní funkční prvek dosahuje výsledků na špičkové úrovni na několika měřících a vyhledávacích úlohách na úrovni instancí, když se kombinuje s trénováním na velkých dávkách. Když se navíc použije kNN klasifikátor, slouží také jako účinný nástroj pro jemné rozlišení, kdy překonává přímé metody klasifikace.

Čtvrtým příspěvkem této práce je návrh loss funkce s názvem Extended SupCon, která společně trénuje klasifikátor a parametry základní sítě pro supervised contrastive classification. Navržený přístup využívá robustnosti contrastive learning a zachovává pravděpodobnostní interpretaci jako u soft-max predikce. Empirické výsledky ukazují účinnost našeho přístupu i v náročných podmínkách, jako je nerovnováha tříd, zkreslené štítky a trénování s málem označených dat.

Celkově přínosy této práce umožňují škálovat trénování neuronových sítí pro nové úlohy v téměř bezúdržbovém režimu, pokud je vyhodnocovací metrika rozložitelná, což pomůže výzkumníkům s novými úkoly. Pro metriky vyhodnocování, které nejsou rozložitelné, lze komponenty vyvinuté pro recall@k surrogate, jako je třídění a počítání, použít k vytváření nových surrogate.

\backmatter
\printbibliography[notcategory=fullcited]

@inproceedings{mishra2012scene,
  title={Scene text recognition using higher order language priors},
  author={Mishra, Anand and Alahari, Karteek and Jawahar, CV},
  booktitle={BMVC},
  year={2012}
}

@inproceedings{wang2011end,
  title={End-to-end scene text recognition},
  author={Wang, Kai and Babenko, Boris and Belongie, Serge},
  booktitle={ICCV},
  year={2011}
}

@inproceedings{karatzas2013icdar,
  title={ICDAR 2013 robust reading competition},
  author={Karatzas, Dimosthenis and Shafait, Faisal and Uchida, Seiichi and Iwamura, Masakazu and i Bigorda, Lluis Gomez and Mestre, Sergi Robles and Mas, Joan and Mota, David Fernandez and Almazan, Jon Almazan and De Las Heras, Lluis Pere},
  booktitle={ICDAR},
  year={2013}
}

@inproceedings{lucas2003icdar,
  title={ICDAR 2003 robust reading competitions},
  author={Lucas, Simon M and Panaretos, Alex and Sosa, Luis and Tang, Anthony and Wong, Shirley and Young, Robert},
  booktitle={ICDAR},
  year={2003}
}

@inproceedings{quy2013recognizing,
  title={Recognizing text with perspective distortion in natural scenes},
  author={Quy Phan, Trung and Shivakumara, Palaiahnakote and Tian, Shangxuan and Lim Tan, Chew},
  booktitle={ICCV},
  year={2013}
}

@article{risnumawan2014robust,
  title={A robust arbitrary text detection system for natural scene images},
  author={Risnumawan, Anhar and Shivakumara, Palaiahankote and Chan, Chee Seng and Tan, Chew Lim},
  journal={Expert Systems with Applications},
  year={2014}
}

@article{shi2016end,
  title={An end-to-end trainable neural network for image-based sequence recognition and its application to scene text recognition},
  author={Shi, Baoguang and Bai, Xiang and Yao, Cong},
  journal={PAMI},
  year={2016}
}

@article{baek2019wrong,
  title={What is wrong with scene text recognition model comparisons? dataset and model analysis},
  author={Baek, Jeonghun and Kim, Geewook and Lee, Junyeop and Park, Sungrae and Han, Dongyoon and Yun, Sangdoo and Oh, Seong Joon and Lee, Hwalsuk},
  journal={ICCV},
  year={2019}
}

@inproceedings{karatzas2015icdar,
  title={ICDAR 2015 competition on robust reading},
  author={Karatzas, Dimosthenis and Gomez-Bigorda, Lluis and Nicolaou, Anguelos and Ghosh, Suman and Bagdanov, Andrew and Iwamura, Masakazu and Matas, Jiri and Neumann, Lukas and Chandrasekhar, Vijay Ramaseshan and Lu, Shijian and others},
  booktitle={ICDAR},
  year={2015}
}

@inproceedings{shi2016robust,
  title={Robust scene text recognition with automatic rectification},
  author={Shi, Baoguang and Wang, Xinggang and Lyu, Pengyuan and Yao, Cong and Bai, Xiang},
  booktitle={CVPR},
  year={2016}
}

@inproceedings{he2016deep,
  title={Deep residual learning for image recognition},
  author={He, Kaiming and Zhang, Xiangyu and Ren, Shaoqing and Sun, Jian},
  booktitle={CVPR},
  year={2016}
}

@article{simonyan2014very,
  title={Very deep convolutional networks for large-scale image recognition},
  author={Simonyan, Karen and Zisserman, Andrew},
  journal={arXiv preprint arXiv:1409.1556},
  year={2014}
}

@inproceedings{krizhevsky2012imagenet,
  title={Imagenet classification with deep convolutional neural networks},
  author={Krizhevsky, Alex and Sutskever, Ilya and Hinton, Geoffrey E},
  booktitle={NeurIPS},
  year={2012}
}

@inproceedings{ren2015faster,
  title={Faster r-cnn: Towards real-time object detection with region proposal networks},
  author={Ren, Shaoqing and He, Kaiming and Girshick, Ross and Sun, Jian},
  booktitle={NeurIPS},
  year={2015}
}

@inproceedings{yu2016unitbox,
  title={Unitbox: An advanced object detection network},
  author={Yu, Jiahui and Jiang, Yuning and Wang, Zhangyang and Cao, Zhimin and Huang, Thomas},
  booktitle={ACM-MM},
  year={2016}
}

@inproceedings{rezatofighi2019generalized,
  title={Generalized intersection over union: A metric and a loss for bounding box regression},
  author={Rezatofighi, Hamid and Tsoi, Nathan and Gwak, JunYoung and Sadeghian, Amir and Reid, Ian and Savarese, Silvio},
  booktitle={CVPR},
  year={2019}
}

@inproceedings{lapin2016loss,
  title={Loss functions for top-k error: Analysis and insights},
  author={Lapin, Maksim and Hein, Matthias and Schiele, Bernt},
  booktitle={CVPR},
  year={2016}
}

@article{berrada2018smooth,
  title={Smooth loss functions for deep top-k classification},
  author={Berrada, Leonard and Zisserman, Andrew and Kumar, M Pawan},
  journal={ICLR},
  year={2018}
}

@article{ma2018arbitrary,
  title={Arbitrary-oriented scene text detection via rotation proposals},
  author={Ma, Jianqi and Shao, Weiyuan and Ye, Hao and Wang, Li and Wang, Hong and Zheng, Yingbin and Xue, Xiangyang},
  journal={IEEE Transactions on Multimedia},
  year={2018}
}

@misc{ma2019rrpn,
    author = {Jianqi Ma},
    title = {{RRPN in pytorch}},
    year = {2019},
    howpublished = {\url{https://github.com/mjq11302010044/RRPNpytorch}}
}

@inproceedings{hodan2018bop,
  title={Bop: Benchmark for 6d object pose estimation},
  author={Hodan, Tomas and Michel, Frank and Brachmann, Eric and Kehl, Wadim and GlentBuch, Anders and Kraft, Dirk and Drost, Bertram and Vidal, Joel and Ihrke, Stephan and Zabulis, Xenophon and others},
  booktitle={ECCV},
  year={2018}
}

@article{nayef2019icdar2019,
  title={ICDAR2019 Robust Reading Challenge on Multi-lingual Scene Text Detection and Recognition--RRC-MLT-2019},
  author={Nayef, Nibal and Patel, Yash and Busta, Michal and Chowdhury, Pinaki Nath and Karatzas, Dimosthenis and Khlif, Wafa and Matas, Jiri and Pal, Umapada and Burie, Jean-Christophe and Liu, Cheng-lin and others},
  journal={arXiv preprint arXiv:1907.00945},
  year={2019}
}

@article{elsken2018neural,
  title={Neural architecture search: A survey},
  author={Elsken, Thomas and Metzen, Jan Hendrik and Hutter, Frank},
  journal={arXiv preprint arXiv:1808.05377},
  year={2018}
}

@inproceedings{deng2009imagenet,
  title={Imagenet: A large-scale hierarchical image database},
  author={Deng, Jia and Dong, Wei and Socher, Richard and Li, Li-Jia and Li, Kai and Fei-Fei, Li},
  booktitle={CVPR},
  year={2009}
}

@inproceedings{lin2014microsoft,
  title={Microsoft coco: Common objects in context},
  author={Lin, Tsung-Yi and Maire, Michael and Belongie, Serge and Hays, James and Perona, Pietro and Ramanan, Deva and Doll{\'a}r, Piotr and Zitnick, C Lawrence},
  booktitle={ECCV},
  year={2014}
}

@inproceedings{caron2018deep,
  title={Deep clustering for unsupervised learning of visual features},
  author={Caron, Mathilde and Bojanowski, Piotr and Joulin, Armand and Douze, Matthijs},
  booktitle={ECCV},
  year={2018}
}

@article{gidaris2018unsupervised,
  title={Unsupervised representation learning by predicting image rotations},
  author={Gidaris, Spyros and Singh, Praveer and Komodakis, Nikos},
  journal={ICLR},
  year={2018}
}

@inproceedings{gomez2017self,
  title={Self-supervised learning of visual features through embedding images into text topic spaces},
  author={Gomez, Lluis and Patel, Yash and Rusi{\~n}ol, Mar{\c{c}}al and Karatzas, Dimosthenis and Jawahar, CV},
  booktitle={CVPR},
  year={2017}
}

@article{rumelhart1986learning,
  title={Learning representations by back-propagating errors},
  author={Rumelhart, David E and Hinton, Geoffrey E and Williams, Ronald J},
  journal={Nature},
  year={1986},
}

@inproceedings{xia2018dota,
  title={DOTA: A large-scale dataset for object detection in aerial images},
  author={Xia, Gui-Song and Bai, Xiang and Ding, Jian and Zhu, Zhen and Belongie, Serge and Luo, Jiebo and Datcu, Mihai and Pelillo, Marcello and Zhang, Liangpei},
  booktitle={CVPR},
  year={2018}
}

@inproceedings{kristan2019seventh,
  title={The seventh visual object tracking vot2019 challenge results},
  author={Kristan, Matej and Matas, Jiri and Leonardis, Ales and Felsberg, Michael and Pflugfelder, Roman and Kamarainen, Joni-Kristian and Cehovin Zajc, Luka and Drbohlav, Ondrej and Lukezic, Alan and Berg, Amanda and others},
  booktitle={ICCV Workshops},
  year={2019}
}

@inproceedings{graves2006connectionist,
  title={Connectionist temporal classification: labelling unsegmented sequence data with recurrent neural networks},
  author={Graves, Alex and Fern{\'a}ndez, Santiago and Gomez, Faustino and Schmidhuber, J{\"u}rgen},
  booktitle={ICML},
  year={2006}
}

@inproceedings{redmon2016you,
  title={You only look once: Unified, real-time object detection},
  author={Redmon, Joseph and Divvala, Santosh and Girshick, Ross and Farhadi, Ali},
  booktitle={CVPR},
  year={2016}
}

@inproceedings{buvsta2018e2e,
  title={E2e-mlt-an unconstrained end-to-end method for multi-language scene text},
  author={Bu{\v{s}}ta, Michal and Patel, Yash and Matas, Jiri},
  booktitle={ACCV},
  year={2018}
}

@inproceedings{azimi2018towards,
  title={Towards multi-class object detection in unconstrained remote sensing imagery},
  author={Azimi, Seyed Majid and Vig, Eleonora and Bahmanyar, Reza and K{\"o}rner, Marco and Reinartz, Peter},
  booktitle={ACCV},
  year={2018}
}

@inproceedings{wu2018learning,
  title={Learning to teach with dynamic loss functions},
  author={Wu, Lijun and Tian, Fei and Xia, Yingce and Fan, Yang and Qin, Tao and Jian-Huang, Lai and Liu, Tie-Yan},
  booktitle={NeurIPS},
  year={2018}
}

@article{li2017learning,
  title={Learning to optimize neural nets},
  author={Li, Ke and Malik, Jitendra},
  journal={arXiv preprint arXiv:1703.00441},
  year={2017}
}

@inproceedings{kato2018neural,
  title={Neural 3d mesh renderer},
  author={Kato, Hiroharu and Ushiku, Yoshitaka and Harada, Tatsuya},
  booktitle={CVPR},
  year={2018}
}

@inproceedings{agustsson2017soft,
  title={Soft-to-hard vector quantization for end-to-end learning compressible representations},
  author={Agustsson, Eirikur and Mentzer, Fabian and Tschannen, Michael and Cavigelli, Lukas and Timofte, Radu and Benini, Luca and Gool, Luc V},
  booktitle={NeurIPS},
  year={2017}
}

@inproceedings{nagendar2018neuro,
  title={Neuro-IoU: Learning a Surrogate Loss for Semantic Segmentation.},
  author={Nagendar, Gattigorla and Singh, Digvijay and Balasubramanian, Vineeth N and Jawahar, CV},
  booktitle={BMVC},
  year={2018}
}

@inproceedings{rahman2016optimizing,
  title={Optimizing intersection-over-union in deep neural networks for image segmentation},
  author={Rahman, Md Atiqur and Wang, Yang},
  booktitle={ISVC},
  year={2016}
}

@inproceedings{hazan2010direct,
  title={Direct loss minimization for structured prediction},
  author={Hazan, Tamir and Keshet, Joseph and McAllester, David A},
  booktitle={NeurIPS},
  year={2010}
}

@inproceedings{song2016training,
  title={Training deep neural networks via direct loss minimization},
  author={Song, Yang and Schwing, Alexander and Urtasun, Raquel and others},
  booktitle={ICML},
  year={2016}
}

@article{grabocka2019learning,
  title={Learning Surrogate Losses},
  author={Grabocka, Josif and Scholz, Randolf and Schmidt-Thieme, Lars},
  journal={arXiv preprint arXiv:1905.10108},
  year={2019}
}

@inproceedings{gulrajani2017improved,
  title={Improved training of wasserstein gans},
  author={Gulrajani, Ishaan and Ahmed, Faruk and Arjovsky, Martin and Dumoulin, Vincent and Courville, Aaron C},
  booktitle={NeurIPS},
  year={2017}
}

@article{lee2018context,
  title={Context-adaptive entropy model for end-to-end optimized image compression},
  author={Lee, Jooyoung and Cho, Seunghyun and Beack, Seung-Kwon},
  journal={ICLR},
  year={2019}
}

@article{balle2018variational,
  title={Variational image compression with a scale hyperprior},
  author={Ball{\'e}, Johannes and Minnen, David and Singh, Saurabh and Hwang, Sung Jin and Johnston, Nick},
  journal={ICLR},
  year={2018}
}

@inproceedings{wang2003multiscale,
  title={Multiscale structural similarity for image quality assessment},
  author={Wang, Zhou and Simoncelli, Eero P and Bovik, Alan C},
  booktitle={ACSSC},
  year={2003}
}

@inproceedings{ledig2017photo,
  title={Photo-realistic single image super-resolution using a generative adversarial network},
  author={Ledig, Christian and Theis, Lucas and Husz{\'a}r, Ferenc and Caballero, Jose and Cunningham, Andrew and Acosta, Alejandro and Aitken, Andrew and Tejani, Alykhan and Totz, Johannes and Wang, Zehan and others},
  booktitle={CVPR},
  year={2017}
}

@article{dong2015image,
  title={Image super-resolution using deep convolutional networks},
  author={Dong, Chao and Loy, Chen Change and He, Kaiming and Tang, Xiaoou},
  journal={TPAMI},
  year={2015}
}

@article{ryoo2019assemblenet,
  title={Assemblenet: Searching for multi-stream neural connectivity in video architectures},
  author={Ryoo, Michael S and Piergiovanni, AJ and Tan, Mingxing and Angelova, Anelia},
  journal={NeurIPS},
  year={2019}
}

@article{zoph2016neural,
  title={Neural architecture search with reinforcement learning},
  author={Zoph, Barret and Le, Quoc V},
  journal={arXiv preprint arXiv:1611.01578},
  year={2016}
}

@inproceedings{DBLP:conf/cvpr/ShiWLYB16,
  author    = {Baoguang Shi and
               Xinggang Wang and
               Pengyuan Lyu and
               Cong Yao and
               Xiang Bai},
  title     = {Robust Scene Text Recognition with Automatic Rectification},
  booktitle = {CVPR},
  year      = {2016}
}

@inproceedings{DBLP:conf/nips/JaderbergSZK15,
  author    = {Max Jaderberg and
               Karen Simonyan and
               Andrew Zisserman and
               Koray Kavukcuoglu},
  title     = {Spatial Transformer Networks},
  booktitle = {NeurIPS},
  year      = {2015}
}

@inproceedings{DBLP:conf/bmvc/LiuCWSH16,
  author    = {Wei Liu and
               Chaofeng Chen and
               Kwan{-}Yee K. Wong and
               Zhizhong Su and
               Junyu Han},
  title     = {STAR-Net: {A} SpaTial Attention Residue Network for Scene Text Recognition},
  booktitle = {BMVC},
  year      = {2016}
}

@inproceedings{DBLP:conf/icdar/GomezSGNVMBK17,
  author    = {Raul Gomez and
               Baoguang Shi and
               Lluis Gomez{-}Bigorda and
               Lukas Neumann and
               Andreas Veit and
               Jiri Matas and
               Serge J. Belongie and
               Dimosthenis Karatzas},
  title     = {{ICDAR2017} Robust Reading Challenge on COCO-Text},
  booktitle = {ICDAR},
  year      = {2017}
}

@article{DBLP:journals/corr/JaderbergSVZ14,
  author    = {Max Jaderberg and
               Karen Simonyan and
               Andrea Vedaldi and
               Andrew Zisserman},
  title     = {Synthetic Data and Artificial Neural Networks for Natural Scene Text
               Recognition},
  journal   = {CoRR},
  year      = {2014}
}

@inproceedings{DBLP:conf/cvpr/GuptaVZ16,
  author    = {Ankush Gupta and
               Andrea Vedaldi and
               Andrew Zisserman},
  title     = {Synthetic Data for Text Localisation in Natural Images},
  booktitle = {CVPR},
  year      = {2016}
}

@article{DBLP:journals/corr/abs-1212-5701,
  author    = {Matthew D. Zeiler},
  title     = {{ADADELTA:} An Adaptive Learning Rate Method},
  journal   = {CoRR},
  year      = {2012}
}

@inproceedings{DBLP:conf/nips/ZhangZL15,
  author    = {Xiang Zhang and
               Junbo Jake Zhao and
               Yann LeCun},
  title     = {Character-level Convolutional Networks for Text Classification},
  booktitle = {NeurIPS},
  year      = {2015}
}

@article{DBLP:journals/corr/XuWCL15,
  author    = {Bing Xu and
               Naiyan Wang and
               Tianqi Chen and
               Mu Li},
  title     = {Empirical Evaluation of Rectified Activations in Convolutional Network},
  journal   = {CoRR},
  year      = {2015}
}

@inproceedings{DBLP:conf/cvpr/LiuLYCQY18,
  author    = {Xuebo Liu and
               Ding Liang and
               Shi Yan and
               Dagui Chen and
               Yu Qiao and
               Junjie Yan},
  title     = {{FOTS:} Fast Oriented Text Spotting With a Unified Network},
  booktitle = {CVPR},
  year      = {2018}
}

@inproceedings{DBLP:journals/jmlr/GlorotBB11,
  author    = {Xavier Glorot and
               Antoine Bordes and
               Yoshua Bengio},
  title     = {Deep Sparse Rectifier Neural Networks},
  booktitle = {AISTATS},
  year      = {2011}
}

@inproceedings{DBLP:conf/cvpr/EngilbergeCPC19,
  author    = {Martin Engilberge and
               Louis Chevallier and
               Patrick P{\'{e}}rez and
               Matthieu Cord},
  title     = {SoDeep: {A} Sorting Deep Net to Learn Ranking Loss Surrogates},
  booktitle = {CVPR},
  year      = {2019}
}

@inproceedings{prabhavalkar2018minimum,
  title={Minimum word error rate training for attention-based sequence-to-sequence models},
  author={Prabhavalkar, Rohit and Sainath, Tara N and Wu, Yonghui and Nguyen, Patrick and Chen, Zhifeng and Chiu, Chung-Cheng and Kannan, Anjuli},
  booktitle={ICASSP},
  year={2018}
}

@article{hochreiter1997long,
  title={Long short-term memory},
  author={Hochreiter, Sepp and Schmidhuber, J{\"u}rgen},
  journal={Neural computation},
  year={1997}
}

@article{patel2019deep,
  title={Deep perceptual compression},
  author={Patel, Yash and Appalaraju, Srikar and Manmatha, R},
  journal={arXiv preprint arXiv:1907.08310},
  year={2019}
}

@article{patel2020hierarchical,
  title={Hierarchical Auto-Regressive Model for Image Compression Incorporating Object Saliency and a Deep Perceptual Loss},
  author={Patel, Yash and Appalaraju, Srikar and Manmatha, R},
  journal={arXiv preprint arXiv:2002.04988},
  year={2020}
}

@InProceedings{bsc+14,
  title   = {Neural codes for image retrieval},
  author  = {Babenko, Artem and Slesarev, Anton and Chigorin, Alexandr
      and Lempitsky, Victor},
  booktitle = {ECCV},
  year    = {2014}
}

@InProceedings{dsl+09,
  author  = {Wei Dong and Richard Socher and Li Li-Jia and Kai Li and
      Li Fei-Fei},
  title   = {{ImageNet}: A large-scale hierarchical image database},
  booktitle = {CVPR},
  year    = 2009
}

@Article{gar+17,
  title   = {End-to-end learning of deep visual representations for
      image retrieval},
  author  = {Gordo, Albert and Almazan, Jon and Revaud, Jerome and
      Larlus, Diane},
  journal = {IJCV},
  year    = {2017}
}

@InProceedings{hcl06,
  title   = {Dimensionality reduction by learning an invariant
      mapping},
  author  = {Hadsell, Raia and Chopra, Sumit and LeCun, Yann},
  booktitle = {CVPR},
  year    = {2006}
}

@InProceedings{hzr+16,
  title   = {Deep residual learning for image recognition},
  author  = {He, Kaiming and Zhang, Xiangyu and Ren, Shaoqing and Sun,
      Jian},
  booktitle = {CVPR},
  year    = {2016}
}

@InProceedings{kb15,
  title   = {Adam: A method for stochastic optimization},
  author  = {Kingma, Diederik and Ba, Jimmy},
  booktitle = {ICLR},
  year    = {2015}
}

@Conference{nas+17,
  title   = {Large-Scale Image Retrieval with Attentive Deep Local
      Features},
  author  = {Noh, Hyeonwoo and Araujo, Andre and Sim, Jack and Weyand,
      Tobias and Han, Bohyung},
  booktitle = {ICCV},
  year    = {2017}
}

@InProceedings{ohb16,
  title   = {Siamese Network of Deep Fisher-Vector Descriptors for
      Image Retrieval},
  author  = {Ong, Eng-Jon and Husain, Sameed and Bober, Miroslaw},
  booktitle = {arXiv},
  year    = {2017}
}

@InProceedings{slj+15,
  title   = {Going deeper with convolutions},
  author  = {Szegedy, Christian and Liu, Wei and Jia, Yangqing and
      Sermanet, Pierre and Reed, Scott and Anguelov, Dragomir and
      Erhan, Dumitru and Vanhoucke, Vincent and Rabinovich,
      Andrew},
  booktitle = {CVPR},
  year    = {2015}
}

@Conference{sxj+15,
  title   = {Deep Metric Learning via Lifted Structured Feature
      Embedding},
  author  = {Song, Hyun Oh and Xiang, Yu and Jegelka, Stefanie and
      Savarese, Silvio},
  booktitle = {arXiv},
  year    = {2015}
}

@inproceedings{hfw+20,
  title={Momentum contrast for unsupervised visual representation learning},
  author={He, Kaiming and Fan, Haoqi and Wu, Yuxin and Xie, Saining and Girshick, Ross},
  booktitle={CVPR},
  year={2020}
}

@inproceedings{mbl20,
  title={A metric learning reality check},
  author={Musgrave, Kevin and Belongie, Serge and Lim, Ser-Nam},
  booktitle={ECCV},
  year={2020}
}

@inproceedings{tjc20,
  title={Learning and aggregating deep local descriptors for instance-level recognition},
  author={Tolias, Giorgos and Jenicek, Tomas and Chum, Ond{\v{r}}ej},
  booktitle={ECCV},
  year={2020}
}

@article{rtc19,
  title={Fine-tuning CNN image retrieval with no human annotation},
  author={Radenovi{\'c}, Filip and Tolias, Giorgos and Chum, Ond{\v{r}}ej},
  journal={PAMI},
  year={2019}
}

@inproceedings{mtl+17,
  title={No fuss distance metric learning using proxies},
  author={Movshovitz-Attias, Yair and Toshev, Alexander and Leung, Thomas K and Ioffe, Sergey and Singh, Saurabh},
  booktitle={ICCV},
  year={2017}
}

@inproceedings{wms+17,
  title={Sampling matters in deep embedding learning},
  author={Wu, Chao-Yuan and Manmatha, R and Smola, Alexander J and Krahenbuhl, Philipp},
  booktitle={ICCV},
  year={2017}
}

@inproceedings{whh+19,
  title={Multi-similarity loss with general pair weighting for deep metric learning},
  author={Wang, Xun and Han, Xintong and Huang, Weilin and Dong, Dengke and Scott, Matthew R},
  booktitle={CVPR},
  year={2019}
}

@article{zw18,
  title={Classification is a strong baseline for deep metric learning},
  author={Zhai, Andrew and Wu, Hao-Yu},
  journal={arXiv preprint arXiv:1811.12649},
  year={2018}
}

@inproceedings{tdt20,
  title={Proxynca++: Revisiting and revitalizing proxy neighborhood component analysis},
  author={Teh, Eu Wern and DeVries, Terrance and Taylor, Graham W},
  booktitle={ECCV},
  year={2020}
}

@inproceedings{skp+15,
  title={Facenet: A unified embedding for face recognition and clustering},
  author={Schroff, Florian and Kalenichenko, Dmitry and Philbin, James},
  booktitle={CVPR},
  year={2015}
}

@inproceedings{chx+19,
  title={Deep metric learning to rank},
  author={Cakir, Fatih and He, Kun and Xia, Xide and Kulis, Brian and Sclaroff, Stan},
  booktitle={CVPR},
  year={2019}
}

@inproceedings{rbo+19,
  title={Mic: Mining interclass characteristics for improved metric learning},
  author={Roth, Karsten and Brattoli, Biagio and Ommer, Bjorn},
  booktitle={ICCV},
  year={2019}
}

@inproceedings{rmp+20,
  title={Optimizing rank-based metrics with blackbox differentiation},
  author={Rol{\'\i}nek, Michal and Musil, V{\'\i}t and Paulus, Anselm and Vlastelica, Marin and Michaelis, Claudio and Martius, Georg},
  booktitle={CVPR},
  year={2020}
}

@inproceedings{jph+19,
  title={Metric learning with horde: High-order regularizer for deep embeddings},
  author={Jacob, Pierre and Picard, David and Histace, Aymeric and Klein, Edouard},
  booktitle={ICCV},
  year={2019}
}

@inproceedings{wzh+20,
  title={Cross-batch memory for embedding learning},
  author={Wang, Xun and Zhang, Haozhi and Huang, Weilin and Scott, Matthew R},
  booktitle={CVPR},
  year={2020}
}

@inproceedings{dgx+19,
  title={Arcface: Additive angular margin loss for deep face recognition},
  author={Deng, Jiankang and Guo, Jia and Xue, Niannan and Zafeiriou, Stefanos},
  booktitle={CVPR},
  year={2019}
}

@inproceedings{wwz+18,
  title={Cosface: Large margin cosine loss for deep face recognition},
  author={Wang, Hao and Wang, Yitong and Zhou, Zheng and Ji, Xing and Gong, Dihong and Zhou, Jingchao and Li, Zhifeng and Liu, Wei},
  booktitle={CVPR},
  year={2018}
}

@inproceedings{lwy+17,
  title={Sphereface: Deep hypersphere embedding for face recognition},
  author={Liu, Weiyang and Wen, Yandong and Yu, Zhiding and Li, Ming and Raj, Bhiksha and Song, Le},
  booktitle={CVPR},
  year={2017}
}

@inproceedings{qss+19,
  title={Softtriple loss: Deep metric learning without triplet sampling},
  author={Qian, Qi and Shang, Lei and Sun, Baigui and Hu, Juhua and Li, Hao and Jin, Rong},
  booktitle={ICCV},
  year={2019}
}

@inproceedings{evt+20,
  title={The group loss for deep metric learning},
  author={Elezi, Ismail and Vascon, Sebastiano and Torcinovich, Alessandro and Pelillo, Marcello and Leal-Taix{\'e}, Laura},
  booktitle={ECCV},
  year={2020}
}

@inproceedings{sel21,
  title={Learning Intra-Batch Connections for Deep Metric Learning},
  author={Seidenschwarz, Jenny and Elezi, Ismail and Leal-Taix{\'e}, Laura},
  booktitle={ICML},
  year={2021}
}

@inproceedings{brz+20,
  title={Metric learning: cross-entropy vs. pairwise losses},
  author={Boudiaf, Malik and Rony, J{\'e}r{\^o}me and Ziko, Imtiaz Masud and Granger, Eric and Pedersoli, Marco and Piantanida, Pablo and Ayed, Ismail Ben},
  booktitle={ECCV},
  year={2020}
}

@inproceedings{rms+20,
  title={Revisiting training strategies and generalization performance in deep metric learning},
  author={Roth, Karsten and Milbich, Timo and Sinha, Samarth and Gupta, Prateek and Ommer, Bjorn and Cohen, Joseph Paul},
  booktitle={ICML},
  year={2020}
}

@inproceedings{sohn16,
  title={Improved deep metric learning with multi-class n-pair loss objective},
  author={Sohn, Kihyuk},
  booktitle={NeurIPS},
  year={2016}
}

@inproceedings{wzw+17,
  title={Deep metric learning with angular loss},
  author={Wang, Jian and Zhou, Feng and Wen, Shilei and Liu, Xiao and Lin, Yuanqing},
  booktitle={ICCV},
  year={2017}
}

@inproceedings{ul16,
  title={Learning deep embeddings with histogram loss},
  author={Ustinova, Evgeniya and Lempitsky, Victor},
  booktitle={NeurIPS},
  year={2016}
}

@inproceedings{bxk+20,
  title={Smooth-AP: Smoothing the path towards large-scale image retrieval},
  author={Brown, Andrew and Xie, Weidi and Kalogeiton, Vicky and Zisserman, Andrew},
  booktitle={ECCV},
  year={2020}
}

@inproceedings{hls18,
  title={Local descriptors optimized for average precision},
  author={He, Kun and Lu, Yan and Sclaroff, Stan},
  booktitle={CVPR},
  year={2018}
}

@inproceedings{rar+19,
  title={Learning with average precision: Training image retrieval with a listwise loss},
  author={Revaud, Jerome and Almaz{\'a}n, Jon and Rezende, Rafael S and Souza, Cesar Roberto de},
  booktitle={ICCV},
  year={2019}
}

@inproceedings{ksd+13,
  title={3d object representations for fine-grained categorization},
  author={Krause, Jonathan and Stark, Michael and Deng, Jia and Fei-Fei, Li},
  booktitle={ICCV workshops},
  year={2013}
}

@inproceedings{vms+18,
  title={The inaturalist species classification and detection dataset},
  author={Van Horn, Grant and Mac Aodha, Oisin and Song, Yang and Cui, Yin and Sun, Chen and Shepard, Alex and Adam, Hartwig and Perona, Pietro and Belongie, Serge},
  booktitle={CVPR},
  year={2018}
}

@inproceedings{vlb+19,
  title={Manifold mixup: Better representations by interpolating hidden states},
  author={Verma, Vikas and Lamb, Alex and Beckham, Christopher and Najafi, Amir and Mitliagkas, Ioannis and Lopez-Paz, David and Bengio, Yoshua},
  booktitle={ICML},
  year={2019}
}

@article{zcd+17,
  title={mixup: Beyond empirical risk minimization},
  author={Zhang, Hongyi and Cisse, Moustapha and Dauphin, Yann N and Lopez-Paz, David},
  journal={arXiv preprint arXiv:1710.09412},
  year={2017}
}

@article{ksp+20,
  title={Hard negative mixing for contrastive learning},
  author={Kalantidis, Yannis and Sariyildiz, Mert Bulent and Pion, Noe and Weinzaepfel, Philippe and Larlus, Diane},
  journal={NeurIPS},
  year={2020}
}

@inproceedings{gk+20,
  title={Symmetrical synthesis for deep metric learning},
  author={Gu, Geonmo and Ko, Byungsoo},
  booktitle={AAAI},
  year={2020}
}

@inproceedings{vpa+21,
  title={It Takes Two to Tango: Mixup for Deep Metric Learning},
  author={Venkataramanan, Shashanka and Psomas, Bill and Avrithis, Yannis and Kijak, Ewa and Amsaleg, Laurent and Karantzalos, Konstantinos},
  booktitle={ICLR},
  year={2022}
}

@article{gkk+21,
  title={Proxy synthesis: Learning with synthetic classes for deep metric learning},
  author={Gu, Geonmo and Ko, Byungsoo and Kim, Han-Gyu},
  journal={AAAI},
  year={2021}
}

@inproceedings{yjw+16,
  title={Unitbox: An advanced object detection network},
  author={Yu, Jiahui and Jiang, Yuning and Wang, Zhangyang and Cao, Zhimin and Huang, Thomas},
  booktitle={ACM-MM},
  year={2016}
}

@inproceedings{rig+19,
  title={Generalized intersection over union: A metric and a loss for bounding box regression},
  author={Rezatofighi, Hamid and Tsoi, Nathan and Gwak, JunYoung and Sadeghian, Amir and Reid, Ian and Savarese, Silvio},
  booktitle={CVPR},
  year={2019}
}

@inproceedings{nsb+18,
  title={Neuro-IoU: Learning a Surrogate Loss for Semantic Segmentation.},
  author={Nagendar, Gattigorla and Singh, Digvijay and Balasubramanian, Vineeth N and Jawahar, CV},
  booktitle={BMVC},
  year={2018}
}

@inproceedings{bms+18,
  title={Variational image compression with a scale hyperprior},
  author={Ball{\'e}, Johannes and Minnen, David and Singh, Saurabh and Hwang, Sung Jin and Johnston, Nick},
  booktitle = {ICLR},
  year={2018}
}

@inproceedings{mae+18,
  title={Conditional probability models for deep image compression},
  author={Mentzer, Fabian and Agustsson, Eirikur and Tschannen, Michael and Timofte, Radu and Van Gool, Luc},
  booktitle={CVPR},
  year={2018}
}

@inproceedings{pam+21,
  title={Saliency Driven Perceptual Image Compression},
  author={Patel, Yash and Appalaraju, Srikar and Manmatha, R},
  booktitle={WACV},
  year={2021}
}

@inproceedings{bbx+17,
  title={An actor-critic algorithm for sequence prediction},
  author={Bahdanau, Dzmitry and Brakel, Philemon and Xu, Kelvin and Goyal, Anirudh and Lowe, Ryan and Pineau, Joelle and Courville, Aaron and Bengio, Yoshua},
  booktitle = {ICLR},
  year={2017}
}

@inproceedings{phm+20,
  title={Learning surrogates via deep embedding},
  author={Patel, Yash and Hoda{\v{n}}, Tom{\'a}{\v{s}} and Matas, Ji{\v{r}}{\'\i}},
  booktitle={ECCV},
  year={2020}
}

@inproceedings{pm+21,
  title={FEDS--Filtered Edit Distance Surrogate},
  author={Patel, Yash and Matas, Jiri},
  booktitle = {ICDAR},
  year={2021}
}

@inproceedings{ecp+19,
  title={Sodeep: a sorting deep net to learn ranking loss surrogates},
  author={Engilberge, Martin and Chevallier, Louis and P{\'e}rez, Patrick and Cord, Matthieu},
  booktitle={CVPR},
  year={2019}
}

@inproceedings{ltw+16,
  title={Deep Relative Distance Learning: Tell the Difference Between Similar Vehicles},
  author={Liu, Hongye and Tian, Yonghong and Wang, Yaowei and Pang, Lu and Huang, Tiejun},
  booktitle={CVPR},
  year={2016}
}

@inproceedings{rit+18,
  title={Revisiting oxford and paris: Large-scale image retrieval benchmarking},
  author={Radenovi{\'c}, Filip and Iscen, Ahmet and Tolias, Giorgos and Avrithis, Yannis and Chum, Ond{\v{r}}ej},
  booktitle={CVPR},
  year={2018}
}

@article{bkh+16,
  title={Layer normalization},
  author={Ba, Jimmy Lei and Kiros, Jamie Ryan and Hinton, Geoffrey E},
  journal={arXiv preprint arXiv:1607.06450},
  year={2016}
}

@article{ikm+15,
  title={On the approximation of the cut and step functions by logistic and Gompertz functions},
  author={Iliev, Anton Iliev and Kyurkchiev, Nikolay and Markov, Svetoslav},
  journal={Biomath},
  year={2015}
}

@article{ikm+17,
  title={On the Approximation of the step function by some sigmoid functions},
  author={Iliev, A and Kyurkchiev, Nikolay and Markov, S},
  journal={Mathematics and Computers in Simulation},
  year={2017}
}

@article{km+15,
  title={Sigmoid functions: some approximation and modelling aspects},
  author={Kyurkchiev, Nikolay and Markov, Svetoslav},
  journal={LAP LAMBERT Academic Publishing, Saarbrucken},
  year={2015}
}

@article{sh+09,
  title={Semantic hashing},
  author={Salakhutdinov, Ruslan and Hinton, Geoffrey},
  journal={International Journal of Approximate Reasoning},
  year={2009}
}

@inproceedings{mmt+17,
  title={The concrete distribution: A continuous relaxation of discrete random variables},
  author={Maddison, Chris J and Mnih, Andriy and Teh, Yee Whye},
  booktitle={ICLR},
  year={2017}
}

@inproceedings{gls+16,
  title={Muprop: Unbiased backpropagation for stochastic neural networks},
  author={Gu, Shixiang and Levine, Sergey and Sutskever, Ilya and Mnih, Andriy},
  booktitle={ICLR},
  year={2016}
}

@inproceedings{dbk+21,
  title={An image is worth 16x16 words: Transformers for image recognition at scale},
  author={Dosovitskiy, Alexey and Beyer, Lucas and Kolesnikov, Alexander and Weissenborn, Dirk and Zhai, Xiaohua and Unterthiner, Thomas and Dehghani, Mostafa and Minderer, Matthias and Heigold, Georg and Gelly, Sylvain and others},
  booktitle = {ICLR},
  year={2021}
}

@misc{rw2019timm,
  author = {Ross Wightman},
  title = {PyTorch Image Models},
  year = {2019},
  publisher = {GitHub},
  journal = {GitHub repository},
  doi = {10.5281/zenodo.4414861},
  howpublished = {\url{https://github.com/rwightman/pytorch-image-models}}
}

@inproceedings{lh+19,
  title={Decoupled weight decay regularization},
  author={Loshchilov, Ilya and Hutter, Frank},
  booktitle = {ICLR},
  year={2019}
}

@inproceedings{dzl+18,
  title={Deep adversarial metric learning},
  author={Duan, Yueqi and Zheng, Wenzhao and Lin, Xudong and Lu, Jiwen and Zhou, Jie},
  booktitle={CVPR},
  year={2018}
}

@inproceedings{zcl+19,
  title={Hardness-aware deep metric learning},
  author={Zheng, Wenzhao and Chen, Zhaodong and Lu, Jiwen and Zhou, Jie},
  booktitle={CVPR},
  year={2019}
}

@inproceedings{pgr+19,
  title={Self-supervised visual representations for cross-modal retrieval},
  author={Patel, Yash and Gomez, Lluis and Rusi{\~n}ol, Mar{\c{c}}al and Karatzas, Dimosthenis and Jawahar, CV},
  booktitle={ICMR},
  year={2019}
}

@inproceedings{zlx+21,
  title={Learning Better Visual Data Similarities via New Grouplet Non-Euclidean Embedding},
  author={Zhang, Yanfu and Luo, Lei and Xian, Wenhan and Huang, Heng},
  booktitle={ICCV},
  year={2021}
}

@inproceedings{lxz+19,
  title={Sampling wisely: Deep image embedding by top-k precision optimization},
  author={Lu, Jing and Xu, Chaofan and Zhang, Wei and Duan, Ling-Yu and Mei, Tao},
  booktitle={ICCV},
  year={2019}
}

@inproceedings{kpd+20,
  title={Rankmi: A mutual information maximizing ranking loss},
  author={Kemertas, Mete and Pishdad, Leila and Derpanis, Konstantinos G and Fazly, Afsaneh},
  booktitle={CVPR},
  year={2020}
}

@incollection{jaderberg2015spatial,
  title={Spatial transformer networks},
  author={Jaderberg, Max and Simonyan, Karen and Zisserman, Andrew and others},
  booktitle={NeurIPS},
  year={2015}
}

@incollection{jaderberg2016reading,
  title={Reading text in the wild with convolutional neural networks},
  author={Jaderberg, Max and Simonyan, Karen and Vedaldi, Andrea and Zisserman, Andrew},
  booktitle={IJCV},
  year={2016}
 }

@incollection{gupta2016synthetic,
  title={Synthetic data for text localisation in natural images},
  author={Gupta, Ankush and Vedaldi, Andrea and Zisserman, Andrew},
  booktitle={CVPR},
  year={2016}
}

@incollection{zhang2017uber,
  title={Uber-text: A large-scale dataset for optical character recognition from street-level imagery},
  author={Zhang, Ying and Gueguen, Lionel and Zharkov, Ilya and Zhang, Peter and Seifert, Keith and Kadlec, Ben},
  booktitle={CVPR workshop},
  year={2017}
}

@incollection{patel2020learning,
  title={Learning Surrogates via Deep Embedding},
  author={Patel, Yash and Hodan, Tomas and Matas, Jiri},
  booktitle={ECCV},
  year={2020}
}

@incollection{long2020unrealtext,
  title={UnrealText: Synthesizing realistic scene text images from the unreal world},
  author={Long, Shangbang and Yao, Cong},
  booktitle={CVPR},
  year={2020}
}

@incollection{litman2020scatter,
  title={SCATTER: selective context attentional scene text recognizer},
  author={Litman, Ron and Anschel, Oron and Tsiper, Shahar and Litman, Roee and Mazor, Shai and Manmatha, R},
  booktitle={CVPR},
  year={2020}
}

@incollection{liao2019mask,
  title={Mask textspotter: An end-to-end trainable neural network for spotting text with arbitrary shapes},
  author={Liao, Minghui and Lyu, Pengyuan and He, Minghang and Yao, Cong and Wu, Wenhao and Bai, Xiang},
  booktitle={TPAMI},
  year={2019}
}

@incollection{liu2018char,
  title={Char-net: A character-aware neural network for distorted scene text recognition},
  author={Liu, Wei and Chen, Chaofeng and Wong, Kwan-Yee K},
  booktitle={AAAI},
  year={2018}
}

@incollection{yang2019symmetry,
  title={Symmetry-constrained rectification network for scene text recognition},
  author={Yang, Mingkun and Guan, Yushuo and Liao, Minghui and He, Xin and Bian, Kaigui and Bai, Song and Yao, Cong and Bai, Xiang},
  booktitle={ICCV},
  year={2019}
}

@incollection{zhan2019esir,
  title={Esir: End-to-end scene text recognition via iterative image rectification},
  author={Zhan, Fangneng and Lu, Shijian},
  booktitle={CVPR},
  year={2019}
}

@incollection{li2019show,
  title={Show, attend and read: A simple and strong baseline for irregular text recognition},
  author={Li, Hui and Wang, Peng and Shen, Chunhua and Zhang, Guyu},
  booktitle={AAAI},
  year={2019}
}

@incollection{shi2018aster,
  title={Aster: An attentional scene text recognizer with flexible rectification},
  author={Shi, Baoguang and Yang, Mingkun and Wang, Xinggang and Lyu, Pengyuan and Yao, Cong and Bai, Xiang},
  booktitle={PAMI},
  year={2018}
}

@incollection{liao2019scene,
  title={Scene text recognition from two-dimensional perspective},
  author={Liao, Minghui and Zhang, Jian and Wan, Zhaoyi and Xie, Fengming and Liang, Jiajun and Lyu, Pengyuan and Yao, Cong and Bai, Xiang},
  booktitle={AAAI},
  year={2019}
}

@incollection{yao2014strokelets,
  title={Strokelets: A learned multi-scale representation for scene text recognition},
  author={Yao, Cong and Bai, Xiang and Shi, Baoguang and Liu, Wenyu},
  booktitle={CVPR},
  year={2014}
}

@incollection{jaderberg2014deep,
  title={Deep features for text spotting},
  author={Jaderberg, Max and Vedaldi, Andrea and Zisserman, Andrew},
  booktitle={ECCV},
  year={2014}
}

@incollection{klara,
  title={Text Recognition - Real World Data and Where to Find Them},
  author={Janou{\v{s}}kov{\'a}, Kl{\'a}ra and Matas, Jiri and Gomez, Lluis and Karatzas, Dimosthenis},
  booktitle={ICPR},
  year={2021}
}

@incollection{patel2016dynamic,
  title={Dynamic lexicon generation for natural scene images},
  author={Patel, Yash and Gomez, Lluis and Rusinol, Mar{\c{c}}al and Karatzas, Dimosthenis},
  booktitle={ECCV},
  year={2016}
}

@incollection{DBLP:booktitles/corr/abs-1212-5701,
  author    = {Matthew D. Zeiler},
  title     = {{ADADELTA:} An Adaptive Learning Rate Method},
  booktitle   = {CoRR},
  year      = {2012}
}

@incollection{DBLP:booktitles/corr/XuWCL15,
  author    = {Bing Xu and
               Naiyan Wang and
               Tianqi Chen and
               Mu Li},
  title     = {Empirical Evaluation of Rectified Activations in Convolutional Network},
  booktitle   = {CoRR},
  year      = {2015}
}

@incollection{rolinek2020optimizing,
  title={Optimizing Rank-based Metrics with Blackbox Differentiation},
  author={Rol{\'\i}nek, Michal and Musil, V{\'\i}t and Paulus, Anselm and Vlastelica, Marin and Michaelis, Claudio and Martius, Georg},
  booktitle={CVPR},
  year={2020}
}

@incollection{sutskeverSqeuence,
title = {Sequence to Sequence Learning with Neural Networks},
author = {Sutskever, Ilya and Vinyals, Oriol and Le, Quoc V},
booktitle = {NeurIPS},
year = {2014}
}

@incollection{linCOCO,
  author    = {Tsung{-}Yi Lin and
               Michael Maire and
               Serge J. Belongie and
               James Hays and
               Pietro Perona and
               Deva Ramanan and
               Piotr Doll{\'{a}}r and
               C. Lawrence Zitnick},
  title     = {Microsoft {COCO:} Common Objects in Context},
  booktitle = {ECCV},
  year      = {2014}
}

@inproceedings{ye2014text,
  title={Text detection and recognition in imagery: A survey},
  author={Ye, Qixiang and Doermann, David},
  booktitle={TPAMI},
  year={2014}
}

@inproceedings{long2020scene,
  title={Scene text detection and recognition: The deep learning era},
  author={Long, Shangbang and He, Xin and Yao, Cong},
  booktitle={IJCV},
  year={2020}
}

@inproceedings{wang2010word,
  title={Word spotting in the wild},
  author={Wang, Kai and Belongie, Serge},
  booktitle={ECCV},
  year={2010}
}

@inproceedings{lecun1998gradient,
  title={Gradient-based learning applied to document recognition},
  author={LeCun, Yann and Bottou, L{\'e}on and Bengio, Yoshua and Haffner, Patrick},
  booktitle={Proceedings of the IEEE},
  year={1998}
}

@inproceedings{busta2017deep,
  title={Deep textspotter: An end-to-end trainable scene text localization and recognition framework},
  author={Busta, Michal and Neumann, Lukas and Matas, Jiri},
  booktitle={ICCV},
  year={2017}
}

@inproceedings{gomez2019selective,
  title={Selective Style Transfer for Text},
  author={Gomez, Raul and Biten, Ali Furkan and Gomez, Lluis and Gibert, Jaume and Karatzas, Dimosthenis and Rusi{\~n}ol, Mar{\c{c}}al},
  booktitle={ICDAR},
  year={2019}
}

@inproceedings{he2016reading,
  title={Reading scene text in deep convolutional sequences},
  author={He, Pan and Huang, Weilin and Qiao, Yu and Loy, Chen and Tang, Xiaoou},
  booktitle={AAAI},
  year={2016}
}

@inproceedings{bookstein1989principal,
  title={Principal warps: Thin-plate splines and the decomposition of deformations},
  author={Bookstein, Fred L.},
  booktitle={TPAMI},
  year={1989}
}

@inproceedings{patel12018e2e,
  title={E2E-MLT-an unconstrained end-to-end method for multi-language scene text},
  author={Patel$^1$, Yash and Bu{\v{s}}ta$^1$, Michal and Matas$^1$, Jiri},
  year={2018}
}

@inproceedings{zhang2015character,
  title={Character-level convolutional networks for text classification},
  author={Zhang, Xiang and Zhao, Junbo and LeCun, Yann},
  booktitle={NeurIPS},
  year={2015}
}

@inproceedings{gomez2017lsde,
  title={Lsde: Levenshtein space deep embedding for query-by-string word spotting},
  author={G{\'o}mez, Llu{\'\i}s and Rusinol, Mar{\c{c}}al and Karatzas, Dimosthenis},
  booktitle={ICDAR},
  year={2017}
}

@inproceedings{wang2020decoupled,
  title={Decoupled attention network for text recognition},
  author={Wang, Tianwei and Zhu, Yuanzhi and Jin, Lianwen and Luo, Canjie and Chen, Xiaoxue and Wu, Yaqiang and Wang, Qianying and Cai, Mingxiang},
  booktitle={AAAI},
  year={2020}
}

@inproceedings{yu2020towards,
  title={Towards accurate scene text recognition with semantic reasoning networks},
  author={Yu, Deli and Li, Xuan and Zhang, Chengquan and Liu, Tao and Han, Junyu and Liu, Jingtuo and Ding, Errui},
  booktitle={CVPR},
  year={2020}
}

@inproceedings{qiao2020seed,
  title={Seed: Semantics enhanced encoder-decoder framework for scene text recognition},
  author={Qiao, Zhi and Zhou, Yu and Yang, Dongbao and Zhou, Yucan and Wang, Weiping},
  booktitle={CVPR},
  year={2020}
}

@inproceedings{yue2020robustscanner,
  title={RobustScanner: Dynamically Enhancing Positional Clues for Robust Text Recognition},
  author={Yue, Xiaoyu and Kuang, Zhanghui and Lin, Chenhao and Sun, Hongbin and Zhang, Wayne},
  booktitle={ECCV},
  year={2020}
}

@online{bpg,
	author = {Bellard, Fabrice},
	title = {BPG Image Format},
	year = 2014,
	url = {http://bellard. org/bpg/}
}

@article{skodras2001jpeg,
	title={The jpeg 2000 still image compression standard},
	author={Skodras, Athanassios and Christopoulos, Charilaos and Ebrahimi, Touradj},
	journal={IEEE Signal processing magazine},
	year={2001}
}

@inproceedings{mentzer2018conditional,
  title={Conditional probability models for deep image compression},
  author={Mentzer, Fabian and Agustsson, Eirikur and Tschannen, Michael and Timofte, Radu and Van Gool, Luc},
  booktitle={CVPR},
  year={2018}
}

@article{fasterrcnn,
	author = {Ren, Shaoqing and He, Kaiming and Girshick, Ross and Sun, Jian},
	title = {Faster {R-CNN}: Towards Real-Time Object Detection with Region Proposal Networks},
	journal = {TPAMI},
	year = {2017}
}

@inproceedings{chopra2005learning,
	title={Learning a similarity metric discriminatively, with application to face verification},
	author={Chopra, Sumit and Hadsell, Raia and LeCun, Yann},
	booktitle={CVPR},
	year={2005}
}

@article{srivastava2014dropout,
	title={Dropout: a simple way to prevent neural networks from overfitting},
	author={Srivastava, Nitish and Hinton, Geoffrey and Krizhevsky, Alex and Sutskever, Ilya and Salakhutdinov, Ruslan},
	journal=JMLR,
	year={2014},
}

@article{ioffe2015batch,
	title={Batch normalization: Accelerating deep network training by reducing internal covariate shift},
	author={Ioffe, Sergey and Szegedy, Christian},
	journal={arXiv preprint arXiv:1502.03167},
	year={2015}
}

@article{kingma2014adam,
	title={Adam: A method for stochastic optimization},
	author={Kingma, Diederik P and Ba, Jimmy},
	journal={arXiv preprint arXiv:1412.6980},
	year={2014}
}

@article{kodak,
	title={Kodak lossless true color image suite},
	author={Franzen, Rich},
	journal={source: http://r0k. us/graphics/kodak},
	year={1999}
}

@article{balle2016end,
	title={End-to-end optimized image compression},
	author={Ball{\'e}, Johannes and Laparra, Valero and Simoncelli, Eero P},
	journal={arXiv preprint arXiv:1611.01704},
	year={2016}
}

@article{patel2019human,
  title={Human Perceptual Evaluations for Image Compression},
  author={Patel, Yash and Appalaraju, Srikar and Manmatha, R},
  journal={arXiv preprint arXiv:1908.04187},
  year={2019}
}

@inproceedings{boudiaf2020unifying,
  title={A unifying mutual information view of metric learning: cross-entropy vs. pairwise losses},
  author={Boudiaf, Malik and Rony, J{\'e}r{\^o}me and Ziko, Imtiaz Masud and Granger, Eric and Pedersoli, Marco and Piantanida, Pablo and Ayed, Ismail Ben},
  booktitle={ECCV},
  year={2020}
}

@article{kalantidis2020hard,
  title={Hard negative mixing for contrastive learning},
  author={Kalantidis, Yannis and Sariyildiz, Mert Bulent and Pion, Noe and Weinzaepfel, Philippe and Larlus, Diane},
  journal={arXiv preprint arXiv:2010.01028},
  year={2020}
}

@inproceedings{chen2020simple,
  title={A simple framework for contrastive learning of visual representations},
  author={Chen, Ting and Kornblith, Simon and Norouzi, Mohammad and Hinton, Geoffrey},
  booktitle={ICML},
  year={2020},
}

@article{chen2020big,
  title={Big Self-Supervised Models are Strong Semi-Supervised Learners},
  author={Chen, Ting and Kornblith, Simon and Swersky, Kevin and Norouzi, Mohammad and Hinton, Geoffrey},
  journal={arXiv preprint arXiv:2006.10029},
  year={2020}
}

@Article{khosla2020supervised,
    title   = {Supervised Contrastive Learning},
    author  = {Prannay Khosla and Piotr Teterwak and Chen Wang and Aaron Sarna and Yonglong Tian and Phillip Isola and Aaron Maschinot and Ce Liu and Dilip Krishnan},
    journal = {arXiv preprint arXiv:2004.11362},
    year    = {2020},
}

@inproceedings{sulc2019improving,
  title={Improving cnn classifiers by estimating test-time priors},
  author={Sulc, Milan and Matas, Jiri},
  booktitle={ICCV Workshops},
  year={2019}
}

@article{sipka2021hitchhiker,
  title={The Hitchhiker's Guide to Prior-Shift Adaptation},
  author={Sipka, Tomas and Sulc, Milan and Matas, Jiri},
  journal={arXiv preprint arXiv:2106.11695},
  year={2021}
}

@inproceedings{alexandari2020maximum,
  title={Maximum likelihood with bias-corrected calibration is hard-to-beat at label shift adaptation},
  author={Alexandari, Amr and Kundaje, Anshul and Shrikumar, Avanti},
  booktitle={ICML},
  year={2020}
}

@article{saerens2002adjusting,
  title={Adjusting the outputs of a classifier to new a priori probabilities: a simple procedure},
  author={Saerens, Marco and Latinne, Patrice and Decaestecker, Christine},
  journal={Neural computation},
  year={2002}
}

@article{hinton2015distilling,
  title={Distilling the knowledge in a neural network},
  author={Hinton, Geoffrey and Vinyals, Oriol and Dean, Jeff},
  journal={arXiv preprint arXiv:1503.02531},
  year={2015}
}

@article{kittler1998combining,
  title={On combining classifiers},
  author={Kittler, Josef and Hatef, Mohamad and Duin, Robert PW and Matas, Jiri},
  journal={TPAMI},
  year={1998}
}

@article{breiman1996bagging,
  title={Bagging predictors},
  author={Breiman, Leo},
  journal={Machine learning},
  year={1996}
}

@article{ju2018relative,
  title={The relative performance of ensemble methods with deep convolutional neural networks for image classification},
  author={Ju, Cheng and Bibaut, Aur{\'e}lien and van der Laan, Mark},
  journal={Journal of Applied Statistics},
  year={2018}
}

@inproceedings{he2020momentum,
  title={Momentum contrast for unsupervised visual representation learning},
  author={He, Kaiming and Fan, Haoqi and Wu, Yuxin and Xie, Saining and Girshick, Ross},
  booktitle={CVPR},
  year={2020}
}

@article{grill2020bootstrap,
  title={Bootstrap your own latent: A new approach to self-supervised learning},
  author={Grill, Jean-Bastien and Strub, Florian and Altch{\'e}, Florent and Tallec, Corentin and Richemond, Pierre H and Buchatskaya, Elena and Doersch, Carl and Pires, Bernardo Avila and Guo, Zhaohan Daniel and Azar, Mohammad Gheshlaghi and others},
  journal={arXiv preprint arXiv:2006.07733},
  year={2020}
}

@article{caron2020unsupervised,
  title={Unsupervised learning of visual features by contrasting cluster assignments},
  author={Caron, Mathilde and Misra, Ishan and Mairal, Julien and Goyal, Priya and Bojanowski, Piotr and Joulin, Armand},
  journal={arXiv preprint arXiv:2006.09882},
  year={2020}
}

@inproceedings{graf2021dissecting,
  title={Dissecting supervised constrastive learning},
  author={Graf, Florian and Hofer, Christoph and Niethammer, Marc and Kwitt, Roland},
  booktitle={ICML},
  year={2021}
}

@article{krizhevsky2009learning,
  title={Learning multiple layers of features from tiny images},
  author={Krizhevsky, Alex and Hinton, Geoffrey and others},
  year={2009},
  publisher={Citeseer}
}

@misc{tinyimgnet,
author = {Stanford},
title = {{Tiny ImageNet Challenge, CS231N Course}},
url = {https://tiny-imagenet.herokuapp.com/}
}

@inproceedings{guo2017calibration,
  title={On calibration of modern neural networks},
  author={Guo, Chuan and Pleiss, Geoff and Sun, Yu and Weinberger, Kilian Q},
  booktitle={ICML},
  year={2017},
  organization={PMLR}
}

@inproceedings{wohlhart2013optimizing,
  title={Optimizing 1-nearest prototype classifiers},
  author={Wohlhart, Paul and Kostinger, Martin and Donoser, Michael and Roth, Peter M and Bischof, Horst},
  booktitle={CVPR},
  year={2013}
}

@inproceedings{patel2021feds,
  title={FEDS-Filtered Edit Distance Surrogate},
  author={Patel, Yash and Matas, Ji{\v{r}}{\'\i}},
  booktitle={ICDAR},
  year={2021}
}

@article{vsimsa2023docile_icdar,
  title={DocILE Benchmark for Document Information Localization and Extraction},
  author={{\v{S}}imsa, {\v{S}}t{\v{e}}p{\'a}n and {\v{S}}ulc, Milan and U{\v{r}}i{\v{c}}{\'a}{\v{r}}, Michal and Patel, Yash and Hamdi, Ahmed and Koci{\'a}n, Mat{\v{e}}j and Skalick{\`y}, Maty{\'a}{\v{s}} and Matas, Ji{\v{r}}{\'\i} and Doucet, Antoine and Coustaty, Micka{\"e}l and others},
  journal={arXiv preprint arXiv:2302.05658},
  year={2023}
}

@inproceedings{patel2021saliency,
  title={Saliency driven perceptual image compression},
  author={Patel, Yash and Appalaraju, Srikar and Manmatha, R},
  booktitle={WACV},
  year={2021}
}

@article{wei2023fully,
  title={Generalized Differentiable {RANSAC}},
  author={Wei, Tong and Patel, Yash and Matas, Jiri and Barath, Daniel},
  journal={arXiv preprint arXiv:2212.13185},
  year={2023}
}

@article{patel2023simcon,
  title={SimCon Loss with Multiple Views for Text Supervised Semantic Segmentation},
  author={Patel, Yash and Xie, Yusheng and Zhu, Yi and Appalaraju, Srikar and Manmatha, R},
  journal={arXiv preprint arXiv:2302.03432},
  year={2023}
}

@inproceedings{djukanovic2021neural,
  title={Neural network-based acoustic vehicle counting},
  author={Djukanovi{\'c}, Slobodan and Patel, Yash and Matas, Ji{\v{r}}{\'\i} and Virtanen, Tuomas},
  booktitle={European Signal Processing Conference (EUSIPCO)},
  year={2021}
}

@article{radenovic2023filtering,
  title={Filtering, Distillation, and Hard Negatives for Vision-Language Pre-Training},
  author={Radenovic, Filip and Dubey, Abhimanyu and Kadian, Abhishek and Mihaylov, Todor and Vandenhende, Simon and Patel, Yash and Wen, Yi and Ramanathan, Vignesh and Mahajan, Dhruv},
  booktitle={CVPR},
  year={2023}
}

@article{picek2022plant,
  title={Plant recognition by AI: Deep neural nets, transformers, and kNN in deep embeddings},
  author={Picek, Luk{\'a}{\v{s}} and {\v{S}}ulc, Milan and Patel, Yash and Matas, Ji{\v{r}}{\'\i}},
  journal={Frontiers in Plant Science},
  year={2022}
}

@article{aljundi2023contrastive,
  title={Contrastive Classification and Representation Learning with Probabilistic Interpretation},
  author={Aljundi, Rahaf and Patel, Yash and Sulc, Milan and Olmeda, Daniel and Chumerin, Nikolay},
  journal={AAAI},
  year={2023}
}

@inproceedings{patel2022recall,
  title={Recall@k surrogate loss with large batches and similarity mixup},
  author={Patel, Yash and Tolias, Giorgos and Matas, Ji{\v{r}}{\'\i}},
  booktitle={CVPR},
  year={2022}
}

@article{griffin2007caltech,
  title={Caltech-256 object category dataset},
  author={Griffin, Gregory and Holub, Alex and Perona, Pietro},
  year={2007},
  publisher={California Institute of Technology}
}

@article{hendrycks2016baseline,
  title={A baseline for detecting misclassified and out-of-distribution examples in neural networks},
  author={Hendrycks, Dan and Gimpel, Kevin},
  journal={arXiv preprint arXiv:1610.02136},
  year={2016}
}

@inproceedings{caron2021emerging,
  title={Emerging properties in self-supervised vision transformers},
  author={Caron, Mathilde and Touvron, Hugo and Misra, Ishan and J{\'e}gou, Herv{\'e} and Mairal, Julien and Bojanowski, Piotr and Joulin, Armand},
  booktitle={ICCV},
  year={2021}
}

@inproceedings{davari2022probing,
  title={Probing Representation Forgetting in Supervised and Unsupervised Continual Learning},
  author={Davari, MohammadReza and Asadi, Nader and Mudur, Sudhir and Aljundi, Rahaf and Belilovsky, Eugene},
  booktitle={CVPR},
  year={2022}
}

@article{winkens2020contrastive,
  title={Contrastive training for improved out-of-distribution detection},
  author={Winkens, Jim and Bunel, Rudy and Roy, Abhijit Guha and Stanforth, Robert and Natarajan, Vivek and Ledsam, Joseph R and MacWilliams, Patricia and Kohli, Pushmeet and Karthikesalingam, Alan and Kohl, Simon and others},
  journal={arXiv preprint arXiv:2007.05566},
  year={2020}
}

@inproceedings{chen2022contrastive,
  title={Contrastive Test-Time Adaptation},
  author={Chen, Dian and Wang, Dequan and Darrell, Trevor and Ebrahimi, Sayna},
  booktitle={CVPR},
  year={2022}
}

@inproceedings{movshovitz2017no,
  title={No fuss distance metric learning using proxies},
  author={Movshovitz-Attias, Yair and Toshev, Alexander and Leung, Thomas K and Ioffe, Sergey and Singh, Saurabh},
  booktitle={ICCV},
  year={2017}
}

@inproceedings{kim2020proxy,
  title={Proxy anchor loss for deep metric learning},
  author={Kim, Sungyeon and Kim, Dongwon and Cho, Minsu and Kwak, Suha},
  booktitle={CVPR},
  year={2020}
}

@inproceedings{sun2020circle,
  title={Circle loss: A unified perspective of pair similarity optimization},
  author={Sun, Yifan and Cheng, Changmao and Zhang, Yuhan and Zhang, Chi and Zheng, Liang and Wang, Zhongdao and Wei, Yichen},
  booktitle={CVPR},
  year={2020}
}

@article{li2020prototypical,
  title={Prototypical contrastive learning of unsupervised representations},
  author={Li, Junnan and Zhou, Pan and Xiong, Caiming and Hoi, Steven CH},
  journal={arXiv preprint arXiv:2005.04966},
  year={2020}
}

@inproceedings{chen2021exploring,
  title={Exploring simple siamese representation learning},
  author={Chen, Xinlei and He, Kaiming},
  booktitle={CVPR},
  year={2021}
}

@article{devlin2018bert,
  title={Bert: Pre-training of deep bidirectional transformers for language understanding},
  author={Devlin, Jacob and Chang, Ming-Wei and Lee, Kenton and Toutanova, Kristina},
  journal={arXiv preprint arXiv:1810.04805},
  year={2018}
}

@article{li2021align,
  title={Align before fuse: Vision and language representation learning with momentum distillation},
  author={Li, Junnan and Selvaraju, Ramprasaath and Gotmare, Akhilesh and Joty, Shafiq and Xiong, Caiming and Hoi, Steven Chu Hong},
  journal={NeurIPS},
  year={2021}
}

@article{zhou2019semantic,
  title={Semantic understanding of scenes through the ade20k dataset},
  author={Zhou, Bolei and Zhao, Hang and Puig, Xavier and Xiao, Tete and Fidler, Sanja and Barriuso, Adela and Torralba, Antonio},
  journal={IJCV},
  year={2019}
}

@inproceedings{jadon2020survey,
  title={A survey of loss functions for semantic segmentation},
  author={Jadon, Shruti},
  booktitle={CIBCB},
  year={2020}
}

@article{ruder2016overview,
  title={An overview of gradient descent optimization algorithms},
  author={Ruder, Sebastian},
  journal={arXiv preprint arXiv:1609.04747},
  year={2016}
}

@article{duchi2011adaptive,
  title={Adaptive subgradient methods for online learning and stochastic optimization.},
  author={Duchi, John and Hazan, Elad and Singer, Yoram},
  journal={Journal of machine learning research},
  year={2011}
}

@article{zeiler2012adadelta,
  title={Adadelta: an adaptive learning rate method},
  author={Zeiler, Matthew D},
  journal={arXiv preprint arXiv:1212.5701},
  year={2012}
}

@inproceedings{he2022masked,
  title={Masked autoencoders are scalable vision learners},
  author={He, Kaiming and Chen, Xinlei and Xie, Saining and Li, Yanghao and Doll{\'a}r, Piotr and Girshick, Ross},
  booktitle={CVPR},
  year={2022}
}

@inproceedings{ganin2015unsupervised,
  title={Unsupervised Domain Adaptation by Backpropagation},
  author={Ganin, Yaroslav and Ustinova, Evgeniya and Ajakan, Hana and Germain, Pascal and Larochelle, Hugo},
  booktitle={ICML},
  year={2015}
}

@article{ganin2016domain,
  title={Domain-Adversarial Training of Neural Networks},
  author={Ganin, Yaroslav and Ustinova, Evgeniya and Ajakan, Hana and Germain, Pascal and Larochelle, Hugo},
  journal={JMLR},
  year={2016}
}

@inproceedings{hussain2016deep,
  title={Deep Domain Confusion: Maximizing for Domain Invariance},
  author={Hussain, Sajid and Porikli, Fatih and Tsotsos, John K.},
  booktitle={ECCV},
  year={2016}
}

@inproceedings{long2015learning,
  title={Learning Transferable Features with Deep Adaptation Networks},
  author={Long, Mingsheng and Wang, Jianmin and Ding, Guiguang and Sun, Jiaguang and Yu, Philip S.},
  booktitle={ICML},
  year={2015}
}

@inproceedings{zheng2015cross,
  title={Cross-Domain Cascaded Deep Feature Learning},
  author={Zheng, Liang and Wang, Shengjin and Ogunbona, Philip O.},
  booktitle={CVPR},
  year={2015}
}

@inproceedings{liu2021self,
  title={Self-Supervised Domain Adaptation for Computer Vision Tasks},
  author={Liu, Yuxuan and Liu, Yunhui and Xia, Wei and Lin, Dahua},
  booktitle={CVPR},
  year={2021}
}

@inproceedings{liu2021rethinking,
  title={Rethinking the Value of Labels for Improving Class-Imbalanced Learning},
  author={Liu, Huidong and Wan, Mengting and Zheng, Liang and Yuan, Zehuan and Qin, Jing and Zhu, Zhanxing},
  booktitle={CVPR},
  year={2021}
}

@inproceedings{huang2020improving,
  title={Improving Unsupervised Domain Adaptation by Self-Training with Deep Reconstruction},
  author={Huang, Wei and Ling, Haibin and Lin, Weiyao},
  booktitle={ECCV},
  year={2020}
}

@inproceedings{li2020d,
  title={D-Sym: Diversified Synthesis for Domain Adaptive Object Detection},
  author={Li, Liang and Lu, Zhiwu and Huang, Yingjie and Tang, Jinhui},
  booktitle={CVPR},
  year={2020}
}

@inproceedings{tang2019multi,
  title={Multi-Source Domain Adaptation for Semantic Segmentation},
  author={Tang, Peng and Wang, Xinggang and Liu, Wenyu and Wang, Jingdong},
  booktitle={CVPR},
  year={2019}
}

@inproceedings{radford2021learning,
  title={Learning transferable visual models from natural language supervision},
  author={Radford, Alec and Kim, Jong Wook and Hallacy, Chris and Ramesh, Aditya and Goh, Gabriel and Agarwal, Sandhini and Sastry, Girish and Askell, Amanda and Mishkin, Pamela and Clark, Jack and others},
  booktitle={ICML},
  year={2021}
}

@inproceedings{mahajan2018exploring,
  title={Exploring the limits of weakly supervised pretraining},
  author={Mahajan, Dhruv and Girshick, Ross and Ramanathan, Vignesh and He, Kaiming and Paluri, Manohar and Li, Yixuan and Bharambe, Ashwin and Van Der Maaten, Laurens},
  booktitle={ECCV},
  year={2018}
}

@inproceedings{singh2022revisiting,
      title={{Revisiting Weakly Supervised Pre-Training of Visual Perception Models}}, 
      author={Singh, Mannat and Gustafson, Laura and Adcock, Aaron and Reis, Vinicius de Freitas and Gedik, Bugra and Kosaraju, Raj Prateek and Mahajan, Dhruv and Girshick, Ross and Doll{\'a}r, Piotr and van der Maaten, Laurens},
      booktitle={CVPR},
      year={2022}
}

@article{klein2019hyperparameter,
  title={Hyperparameter optimization in machine learning: A review},
  author={Klein, Aaron and Falkner, Stefan and Hutter, Frank},
  journal={JMLR},
  year={2019}
}

@article{law2020grid,
  title={Grid search hyperparameter tuning for deep learning models: A review},
  author={Law, Sean N and Wipf, David P and Gupta, Arjun},
  journal={Neural Computing and Applications},
  year={2020}
}

@article{smith2017learning,
  title={Learning rate schedules for faster convergence},
  author={Smith, Leslie N},
  journal={Deep Learning},
  year={2017}
}

@inproceedings{bengio2012practical,
  title={Practical recommendations for gradient-based training of deep architectures},
  author={Bengio, Yoshua and Courville, Aaron and Vincent, Pascal},
  booktitle={NeurIPS},
  year={2012}
}

@inproceedings{huang2021metricopt,
  title={Metricopt: Learning to optimize black-box evaluation metrics},
  author={Huang, Chen and Zhai, Shuangfei and Guo, Pengsheng and Susskind, Josh},
  booktitle={CVPR},
  year={2021}
}

@inproceedings{poganvcic2020differentiation,
  title={Differentiation of blackbox combinatorial solvers},
  author={Pogan{\v{c}}i{\'c}, Marin Vlastelica and Paulus, Anselm and Musil, Vit and Martius, Georg and Rolinek, Michal},
  booktitle={ICLR},
  year={2020}
}

@inproceedings{eban2017scalable,
  title={Scalable learning of non-decomposable objectives},
  author={Eban, Elad and Schain, Mariano and Mackey, Alan and Gordon, Ariel and Rifkin, Ryan and Elidan, Gal},
  booktitle={AISTAT},
  year={2017}
}

@inproceedings{rakotomamonjy2004optimizing,
  title={Optimizing Area Under Roc Curve with SVMs.},
  author={Rakotomamonjy, Alain},
  booktitle={ROCAI},
  year={2004}
}

@inproceedings{berman2018lovasz,
  title={The lov{\'a}sz-softmax loss: A tractable surrogate for the optimization of the intersection-over-union measure in neural networks},
  author={Berman, Maxim and Triki, Amal Rannen and Blaschko, Matthew B},
  booktitle={CVPR},
  year={2018}
}

@article{puthiya2014optimizing,
  title={Optimizing F-measures by cost-sensitive classification},
  author={Puthiya Parambath, Shameem and Usunier, Nicolas and Grandvalet, Yves},
  journal={NeurIPS},
  year={2014}
}

@article{xu2018autoloss,
  title={Autoloss: Learning discrete schedules for alternate optimization},
  author={Xu, Haowen and Zhang, Hao and Hu, Zhiting and Liang, Xiaodan and Salakhutdinov, Ruslan and Xing, Eric},
  journal={arXiv preprint arXiv:1810.02442},
  year={2018}
}

@article{kumar2010self,
  title={Self-paced learning for latent variable models},
  author={Kumar, M and Packer, Benjamin and Koller, Daphne},
  journal={NeurIPS},
  year={2010}
}

@article{ba2016layer,
  title={Layer normalization},
  author={Ba, Jimmy Lei and Kiros, Jamie Ryan and Hinton, Geoffrey E},
  journal={arXiv preprint arXiv:1607.06450},
  year={2016}
}

@inproceedings{goeau2018overview,
  title={Overview of expertlifeclef 2018: how far automated identification systems are from the best experts?},
  author={Go{\"e}au, Herv{\'e} and Bonnet, Pierre and Joly, Alexis},
  booktitle={CLEF},
  year={2018}
}

@inproceedings{goeau2017plant,
  title={Plant identification based on noisy web data: the amazing performance of deep learning (LifeCLEF 2017)},
  author={Goeau, Herve and Bonnet, Pierre and Joly, Alexis},
  booktitle={CLEF},
  year={2017}
}

@INPROCEEDINGS{8886046,
    author={D. {Zhou} and J. {Fang} and X. {Song} and C. {Guan} and J. {Yin} and Y. {Dai} and R. {Yang}},
    booktitle={2019 International Conference on 3D Vision (3DV)}, 
    title={IoU Loss for 2D/3D Object Detection}, 
    year={2019}
}
\addcontentsline{toc}{chapter}{Bibliography}
\end{document}